\documentclass[mnsc,nonblindrev]{informs3_hide}
\OneAndAHalfSpacedXI % default for MS

\usepackage[english]{babel}
\usepackage[autostyle, english = american]{csquotes}
\MakeOuterQuote{"}
\usepackage[colorlinks,linkcolor=black, citecolor=black]{hyperref}
\usepackage{cleveref}
\crefname{subsection}{section}{subsections}
\usepackage{enumitem}
\usepackage{eqnarray}
\usepackage{algorithm}
\usepackage{algorithmic}
\usepackage{comment}
 \usepackage{stmaryrd}
\SetSymbolFont{stmry}{bold}{U}{stmry}{m}{n}
\DeclareFontShape{OT1}{cmr}{bx}{sc}{<->ssub * cmr/bx/n}{}

\usepackage{bm,bbm}
\usepackage{amsfonts}
\usepackage{amsgen,amsmath,amstext,amsbsy,amsopn,amssymb}

\usepackage{natbib, multirow,multicol}
 \bibpunct[, ]{(}{)}{,}{a}{}{,}%
 \def\bibfont{\small}%

\newcommand{\abs}[1]{\left| #1 \right|}
\newcommand{\norm}[1]{\left\| #1 \right\|}

\DeclareMathAlphabet\mathbfcal{OMS}{cmsy}{b}{n} %define mathbfcal for tensor (if used)

\newcommand{\bI}{\mathbbm{1}}
\newcommand{\bE}{\mathbb{E}}
\newcommand{\bR}{\mathbb{R}}

\newcommand{\bp}{\mathbb{P}}

\newcommand{\VO}{V^{\mathrm{Off}}}
\newcommand{\VU}{V^{\mathrm{UB}}}

\newcommand{\by}{\bm{y}}

\newcommand{\ba}{\bm{a}}

\newcommand{\bb}{\bm{b}}
\newcommand{\bg}{\bm{g}}
\newcommand{\bh}{\bm{h}}

\newcommand{\bbeta}{\bm{\beta}}

\newcommand{\bmu}{\bm{\mu}}
\newcommand{\bpi}{\bm{\pi}}
\newcommand{\blambda}{\bm{\lambda}}

\newcommand{\bx}{\bm{x}}

\newcommand{\bA}{\bm{A}}

\newcommand{\bC}{\bm{C}}
\newcommand{\bM}{\bm{M}}
\newcommand{\bu}{\bm{u}}
\newcommand{\bv}{\bm{v}}
\newcommand{\bW}{\bm{W}}
\newcommand{\bSigma}{\bm{\Sigma}}
\newcommand{\bomega}{\bm{\omega} }
\newcommand{\bOmega}{\bm{\Omega} }
\newcommand{\brho}{\bm{\rho} }
\newcommand{\cB}{\mathcal{B} }
\newcommand{\cC}{\mathcal{C} }
\newcommand{\cD}{\mathcal{D} }

\newcommand{\cH}{\mathcal{H} }
\newcommand{\cL}{\mathcal{L} }

\newcommand{\cT}{\mathcal{T} }
\newcommand{\cP}{\mathcal{P} }
\newcommand{\tF}{{\rm F}}

\newcommand{\rhomin}{\underline{\rho} }
\newcommand{\rhomax}{ \overline{\rho}}

\newcommand{\est}{\mathsf{est}}
\newcommand{\sfm}{\mathsf{m}}

\newcommand{\sfB}{\mathsf{B}}
\newcommand{\NB}{\mathsf{NB}}
\usepackage[dvipsnames]{xcolor}

\newcommand{\commentWT}[1]{{\color{blue}[#1]}}

\NewDocumentEnvironment{myproof}{o}
\TheoremsNumberedThrough     % Preferred (Theorem 1, Lemma 1, Theorem 2)
\ECRepeatTheorems

\EquationsNumberedThrough    % Default: (1), (2), ...
\TITLE{High-dimensional Linear Bandits with Knapsacks}

\ARTICLEAUTHORS{%
\AUTHOR{$\text{Wanteng Ma}^1$ \quad $\text{Dong Xia}^2$ \quad $\text{Jiashuo Jiang}^3$}

\AFF{\  \\
$1~$ Department of Statistics and Data Science, University of Pennsylvania\\
$2~$ Department of Mathematics, The Hong Kong University of Science and Technology \\
$3~$ Department of Industrial Engineering \& Decision Analytics, The Hong Kong University of Science and Technology\\
}
}

\ABSTRACT{We investigate the contextual bandits with knapsack (CBwK) problem in a high-dimensional linear setting, where the feature dimension can be very large. Our goal is to harness sparsity to obtain sharper regret guarantees. To this end, we first develop an online variant of the hard thresholding algorithm that performs the sparse estimation in an online manner. We then embed this estimator in a primal–dual scheme: every knapsack constraint is paired with a dual variable, which is updated by an online learning rule to keep the cumulative resource consumption within budget. This integrated approach achieves a two-phase sub-linear regret that scales only logarithmically with the feature dimension, improving on the polynomial dependency reported in prior work. Furthermore, we show that either of the following structural assumptions is sufficient for a sharper regret bound of $\widetilde{O}(s_0\sqrt{T})$:
(i) a diverse-covariate condition; and
(ii) a margin condition. When both conditions hold simultaneously, we can further control the regret to $O\bigl(s_0^{2}\log(dT)\log T\bigr)$ by a dual resolving scheme. As a by-product, applying our framework to high-dimensional contextual bandits without knapsack constraints recovers the optimal regret rates in both the data-poor and data-rich regimes. Finally, numerical experiments confirm the empirical efficiency of our algorithms in high-dimensional settings.

}
\begin{document}
\maketitle
\section{Introduction}
Popularized by the seminal paper \cite{badanidiyuru2013bandits}, the bandit with knapsacks problem (BwK) is defined by solving an online \textit{knapsack} problem with global size constraints. This kind of problem is a special but important case of the online allocation problem, which imposes a reward-agnostic assumption on resource allocations. The bandit with knapsacks problem has been broadly applied to many scenarios, e.g., ad allocation, dynamic pricing, repeated auctions, etc. In fact, in several applications like online recommendation or online advertising, many contexts (or features, covariates) of rewards that we can observe are possibly high-dimensional, which significantly contribute to the decision-making and motivate us to consider a variant of the BwK problem, i.e., the contextual bandit with knapsacks problem \citep{badanidiyuru2014resourceful}. However, although the contextual bandit with knapsacks problem has been extensively studied under different settings \citep{agrawal2014bandits,agrawal2016linear,immorlica2022adversarial,liu2022non}, previous studies largely neglect the inherent high dimensionality of covariates, and in turn, incur regrets that depend polynomially on the large dimension $d$, making these methods less feasible in the high-dimensional setting. This motivates us to explore further approaches that can handle the BwK problem in the high-dimensional case, which is an emergent topic in online learning.

In this paper, we develop efficient algorithms for {linear contextual bandits with knapsacks} (CBwK) when the unknown parameter is high-dimensional and $s_0$-sparse. Our approach combines {primal estimation} with {dual-based allocation}. In the primal step, we introduce an {online hard-thresholding estimator} that achieves exact support recovery at the optimal statistical rate, matching the celebrated LASSO in accuracy while reducing computational cost. By pairing this estimator with various dual-update strategies that enforce the knapsack constraints, we obtain three regret bounds, depending on the structural assumptions on the covariates: $\widetilde{O}\left( \frac{\VU}{C_{\min}}\sqrt{T}+ \left(\frac{\VU}{C_{\min}}\right)^{\frac{1}{3}}T^{\frac{2}{3}} \right)$, $\widetilde{O}(s_0\sqrt{T})$, or $O\bigl(s_0^{2}\log(dT)\log T\bigr)$. 
All three bounds depend only logarithmically on the dimension $d$, where the bounds are formalized in Section~\ref{sec:formulation}. To derive these guarantees, we identify and generalize the {diverse covariate condition} and the {margin condition}—two key notions in contextual bandits \citep{zeevi2009woodroofe,perchet2013multi,bastani2021mostly,ren2023dynamic}—to the CBwK setting.

% In this paper we develop efficient algorithms for the linear contextual bandits with knapsacks (CBwK) when parameters are high-dimensional and $s_0$-sparse. Our method consists of two parts, primal estimation and dual-based allocation. On the primal side, we introduce an online hard-thresholding estimator that achieves exact sparse recovery at the optimal statistical error, matching the celebrated LASSO in accuracy while reducing computational cost. Coupled with different dual update strategies that enforce the knapsack constraints, this primal–dual scheme bounds the regret of sparse CBwK at three rates, depending on the structural assumptions placed on the covariates: $\widetilde{O}\left( \frac{\VU}{C_{\min}}\sqrt{T}+ \left(\frac{\VU}{C_{\min}}\right)^{\frac{1}{3}}T^{\frac{2}{3}} \right)$, $\widetilde{O}(s_0\sqrt{T})$, or $O\bigl(s_0^{2}\log(dT)\log T\bigr)$. All three bounds depend only logarithmically on the dimension $d$. The notations above are formalized in Section~\ref{sec:formulation}. To achieve these regret bounds, we identify and generalize the diverse covariate condition and margin condition, two important notions from contextual bandits, to our CBwK problem. 
% Moreover, we also show that the regret can be further improved to $\widetilde{O}\left( \frac{\VU}{C_{\min}}\sqrt{T} \right)$ with an additional diverse covariate or margin condition.

Our method also brings new insights into the general online sparse estimation and sparse bandit problem. For the sparse bandit problem, most of the existing literature heavily relies on LASSO (e.g, \cite{kim2019doubly,hao2020high,ren2023dynamic}), which explores sparsity by regularized sample average approximation (SAA). Although LASSO guarantees good theoretical results, it is hard to perform in an online fashion. In this paper, we solve the sparse recovery problem through a novel stochastic approximation approach with hard thresholding, which is more aligned with online learning and is also statistically optimal. This estimation algorithm leads to a by-product, i.e., a unified sparse bandit algorithm framework that reaches desired optimal regrets $\widetilde{O}(s_0^{2 / 3} T^{2 / 3})$  and $\widetilde{O}(\sqrt{s_0 T})$, in both data-poor and data-rich regimes respectively, which satisfies the so-called ``the best of two worlds'' \citep{hao2020high}.
% In our work, we propose to recover exact sparse structures through a hard thresholding method

\subsection{Main Results and Contributions}
Our main results and contributions can be summarized as follows.

First, we develop a new online sparse estimation algorithm, named Online HT, that performs the sparse estimation in an online manner. Note that previous methods for sparse estimation, like LASSO (e.g. \cite{hao2020high,li2022simple,ren2023dynamic}) and iterative hard thresholding \citep{blumensath2009iterative,nguyen2017linear}, perform the estimation in an offline manner and thus require us to store the entire historical data set, on the size of $O(d\cdot T)$, which can be costly when both the dimension and time epoch are large. In contrast, our algorithm is an online variant of the hard thresholding method and features a gradient-averaging technique that only requires us to store the average of the previous estimations, on the size of $O(d^2)$, instead of the entire data set. Moreover, the computation complexity of the sparse estimation step can be reduced by our approach. To be specific, the computational complexity of Online HT is $O(d^2)$ per iteration and $O(d^2T)$ in total, while the computational complexity of classical LASSO solution is $O(d^3+d^2 t)$ per iteration \citep{efron2004least}, and $O(d^3T+d^2 T^2)$ in total if we require constant updates of the estimation, e.g., \cite{kim2019doubly,ren2023dynamic}. In this way, our online estimator enjoys a greater computational benefit than the offline estimator established in the previous literature. 

Second, we demonstrate that the online updates of the Online HT algorithm integrate seamlessly with a primal--dual framework for solving the high-dimensional CBwK problem. Specifically, we associate a dual variable with each resource constraint. Unlike prior work, we treat the current output of Online HT as the primal estimate used in the dual update. For the dual step, we consider two schemes: online mirror descent for the general setting and dual resolving for the non-degenerate setting. Our analysis shows that, in the general case, the primal--dual method combined with mirror descent exploits sparsity to attain regret $\widetilde{O}\left( \frac{\VU}{C_{\min}}\sqrt{T}+ \left(\frac{\VU}{C_{\min}}\right)^{\frac{1}{3}}T^{\frac{2}{3}} \right)$ under no covariate assumptions, or $\widetilde{O}(s_0\sqrt{T})$ under either the diverse covariate or margin condition, by appropriately tuning the exploration rate in the epsilon-greedy algorithm. Furthermore, when the dual problem is non-degenerate, coupling the primal Online HT with dual resolving yields the sharper bound $O\bigl(s_0^{2}\log(dT)\log T\bigr)$ under both conditions. All of these guarantees depend only logarithmically on the dimension $d$, underscoring the efficiency of our approach in extremely high-dimensional regimes. To the best of our knowledge, this is the first work to achieve both $\widetilde{O}(s_0\sqrt{T})$ and $O\bigl(s_0^{2}\log(dT)\log T\bigr)$ regret in the high-dimensional CBwK problem. We also conduct numerical experiments to further illustrate the superiority of the empirical performances of our algorithm under the high-dimensional setting.

%We then construct a primal-dual framework based on Online HT to solve the high-dimensional linear contextual BwK problem. By keeping updating primal variables with Online HT and dual variables with mirror descent, we can efficiently perform the online allocation of CBwK problem and enjoys a regret that depends logarithmically on the dimension $d$.

Apart from the knapsack problem, our Online HT algorithm can be broadly applied to many other high-dimensional problems to achieve statistically optimal results. For instance, we apply the Online HT to the high-dimensional contextual bandit problem, which can be regarded as a special case of the high-dimensional contextual CBwK problem where the resource constraints are absent. We show that our algorithm reaches the desired optimal regrets $\widetilde{O}(s_0^{2 / 3} T^{2 / 3})$ for the data-poor regime and $\widetilde{O}(\sqrt{s_0 T})$ for the general data-rich regimes under the extra diverse covariate condition. In this way, we achieve the so-called ``the best of two worlds'' \citep{hao2020high} without additional phase splitting and signal requirements \citep{hao2020high,jang2022popart}. 

Compared with our preliminary conference version \cite{ma2024high},  this paper makes several significant advances.
(i) We demonstrate that either the diverse covariate condition or the margin condition alone suffices to achieve the $\widetilde{O}(s_0\sqrt{T})$ regret bound.
(ii) We propose a novel dual‑resolving algorithm for CBwK and develop a theory that establishes its $O\bigl(s_0^{2}\log(dT)\log T\bigr)$ regret under stronger assumptions.
(iii) For logarithmic regret, our refined analysis reveals an $O(m^2)$ dependence on the number of constraints—matching the full‑information allocation results \citep{li2022online,ma2024optimal} and can be further improved to $O(m)$ when knapsack sizes are known. We have also updated the numerical experiments to highlight the empirical performance of our methods.

% regularity of the underlying distribution.

%can be broadly applied to many online high-dimensional problems. Our algorithm framework features online hard thresholding and gradient averaging and runs \textit{fully online}. With averaged gradient descent, our method can consecutively update the sparse estimation with low computational cost and statistical optimal estimation rate. Also, when applied to high-dimensional bandit problem, our algorithm reaches desired optimal regrets $\widetilde{O}(s_0^{2 / 3} T^{2 / 3})$ for data-poor regime and $\widetilde{O}(\sqrt{s_0 T})$ for general data-rich regimes under extra diverse covariate condition, achieving the so-called ``the best of two worlds'' \citep{hao2020high} without additional phase splitting and signal requirements \citep{hao2020high,jang2022popart}. %that relies on restricted phase elimination

%We then construct a primal-dual framework based on Online HT to solve the high-dimensional linear contextual BwK problem. By keeping updating primal variables with Online HT and dual variables with mirror descent, we can efficiently perform the online allocation of CBwK problem and enjoys a regret that depends logarithmically on the dimension $d$. 

%With the consecutive sparse estimation of Online HT

\subsection{Related Literature}
Bandit with knapsacks problem \citep{badanidiyuru2013bandits,agrawal2014bandits} can be viewed as a special case of online allocation problem,   where reward functions are unknown for decision-makers. Unlike other resource allocation problems \citep{jiang2020online,balseiro2023best,ma2024optimal},  BwK problem poses strong demands on balancing exploration and exploitation. In the face of uncertainty, this trade-off is mainly handled by, e.g., elimination-based algorithms \citep{badanidiyuru2013bandits,badanidiyuru2018bandits}, or UCB \citep{agrawal2014bandits, liu2022non}, or primal-dual algorithms \citep{badanidiyuru2013bandits,li2021symmetry}, which are all guaranteed to be optimal for problem independent settings. In the contextual BwK problem (CBwK), some well-established methods have been proposed, including policy elimination \citep{badanidiyuru2014resourceful} and UCB-type algorithm \citep{agrawal2016linear}, which both originated from contextual bandit problem. Recently, \cite{slivkins2023contextual} summarized a general primal-dual framework for contextual BwK with a regression-based primal algorithm. However, the currently well-known CBwK methods \citep{badanidiyuru2014resourceful,agrawal2016linear,slivkins2023contextual} all suffer from at least $O(\sqrt{d})$ dependence on the dimension in the regret, which hugely confines their applicants to the low-dimensional case. For example, the $O(d\cdot\sqrt{T})$ regret bound in \cite{agrawal2016linear} and the $O(\sqrt{d\cdot T})$ regret bound in \cite{han2023optimal}. The failure of classic CBwK methods for large $d$ strongly motivates us to explore the CBwK problem with high-dimensional contexts, which is frequently encountered in the real world like user-specific recommendations and personalized treatments \citep{bastani2020online}.

%non-degenerate problem:

The high-dimensional contextual bandit problems have been extensively studied in the previous literature, which serves as the precursor of the CBwK problems.
Based on the LASSO method, many sampling strategies have been devised.  Noticeable force-sampling strategy in \cite{bastani2020online} achieves a regret $O\left(s_0^2 \cdot\left(\log d+\log T\right)^2 \right)$  under the margin condition, and has been improved by \cite{wang2018minimax} to a sharper minimax rate $O\left(s_0^2 \cdot\left(\log d+s_0\right) \cdot \log T\right)$ with concave penalized LASSO. \cite{kim2019doubly} has constructed a doubly-robust $\varepsilon$-greedy sampling strategy by re-solving LASSO, yielding a regret of order $\widetilde{O}(s_0 \sqrt{T})$ under vanishing noise size. \cite{hao2020high} introduced an Explore-then-Commit LASSO bandit framework with the regret $\widetilde{O}(s_0^{2 / 3} T^{2 / 3})$, and this framework has been followed up by, e.g., \cite{li2022simple,jang2022popart}. As is shown in \cite{jang2022popart}, the regret lower bound of sparse bandit problem is $\Omega\left(\phi_{\min}^{-2 / 3} s_0^{2 / 3} T^{2 / 3}\right)$ in the data-poor regime $d\gtrsim T^{\frac{1}{3}} s_0^{\frac{4}{3}}$. However, another stream of work showed that, for the general data-rich regime,
% optimal sparse linear bandits \cite{jang2022popart} with given population covariance matrix. 
the optimal regret is of order $\Omega(\sqrt{s_0 T})$  \citep{chu2011contextual,ren2023dynamic} and can be obtained with additional covariate conditions,   for example, diverse covariate condition \citep{ren2023dynamic}, and balanced covariance condition,  \citep{oh2021sparsity,ariu2022thresholded}, etc. The two-phase optimal regret of the sparse bandit problem leads to an open question, i.e., can we achieve ``the best of two worlds'' of sparse bandit problem in both data-poor and data-rich regimes with a unified framework \citep{hao2020high}? In our paper, we will answer this question affirmatively by providing our Online HT algorithm in the sparse bandit setting.
%Generally, to achieve a regret lower than $\widetilde{O}(s_0^{2 / 3} T^{2 / 3})$,

The idea of hard thresholding is applied in our methodology for the consecutive online estimation. Hard thresholding finds its application in sparse recovery primarily for the iterative hard thresholding methods \citep{blumensath2009iterative}. One of the most intriguing properties of hard thresholding is that it can return an exact sparse estimation given any sparsity level. Nonetheless, the poor smoothness behavior inhered in the hard thresholding projector \citep{shen2017tight} makes it difficult to analyze the error for iterative methods, especially for stochastic gradient descent methods with large variances. Therefore, current applications of hard thresholding mainly focus on batch learning \citep{nguyen2017linear,yuan2021stability} or hybrid learning \citep{zhou2018efficient}, while hard thresholding methods for online learning are still largely unexplored. 

\section{Notations}
Throughout the paper, we use $\widetilde{O}(\cdot)$ to denote the big-O rate that omits the logarithmic terms. We write $f\lesssim g$ if $f\le Cg$ for some universal numerical constant $C>0$. The value of $C$ in the paper may change from line to line and is independent of key quantities in the paper, such as $T,d,K,m,s_0$ and the problem-dependent parameters unless otherwise specified. We write $[K]$ as the set of positive integers from $1$ to $K$, i.e., $\{1,2,\dots,K\}$. We shall denote the scalas by normal symbols and vectors/matrices by bold symbols. For matrix $\bA$, we use $\norm{\bA}_p$ to represent the operator norm induced by $\ell_p$ vector norm. Moreover, we use $\norm{\cdot}_{\max}$ to represent the maximum absolute value of entries, and use $\norm{\cdot}_{2,\max}$ to represent the maximum $\ell_2$ norm within all the rows, i.e., $\norm{\bM}_{2,\max}=\max_{i}\norm{\bm{e}_i^\top \bM}_2$. The $K$ dimensional unit simplex is denoted by $\Delta^K=\{\bx\in\bR^K:x_i\ge0,\forall i\in [r],\ \norm{\bx}_1=\sum_{i=1}^r x_i\le 1\}$. For vectors $\ba$ and $\bb$ of the same length, the notation $\ba\preceq\bb$ denotes the entrywise inequality $\ba_i \le \bb_i$ for every index $i$.

\section{ High-dimensional Contextual BwK}\label{sec:formulation}
We consider the high-dimensional contextual bandit with the knapsack problem over a finite horizon of $T$ periods. There are $m$ resources and each resource $i\in[m]$ has an initial capacity $C_i>0$. The capacity vector is denoted by $\bC\in \bR^m$. We normalize the vector $\bC$ such that $C_i/T\in[0,1]$ for each $i\in[m]$. We denote $C_{\min}=\min_{i\in[m]}\{C_i\}$, and $C_{\max}=\max_{i\in[m]}\{C_i\}$. There are $K$ arms and a null arm that generate no reward and consume no resources to perform void action. At each period $t\in[T]$, one query arrives, denoted by query $t$, and is associated with a feature $\bx_t\in\mathbb{R}^d$. We assume that the feature $\bx_t$ is drawn from a distribution $F(\cdot)$ independently at each period $t$. For each arm $a\in[K]$, query $t$ is associated with an \textit{unknown} reward $r_t(a, \bx_t)$ and an \textit{unknown} size $\bb_t(a,\bx_t)=\left[b_{1,t}(a, \bx_t),\dots, b_{m,t}(a, \bx_t)\right]^\top\in\mathbb{R}^m_{\ge0}$. Note that the reward $r_t(a, \bx_t)$ and the size $\bb_t(a,\bx_t)$ depends on the feature $\bx_t$ and the arm $a$. For each arm $a\in[K]$, we assume that the size $\bb_t(a,\bx_t)$ follows the following relationship
\begin{equation}\label{eqn:sizecovariate}
\bb_t(a, \bx_t)=\bm{W}^{\star}_a \bx_t +\bomega_t,  \ a\in[K] %+\bxi_{t}^{c}
\end{equation}
where $\bm{W}_a^{\star} \in\mathbb{R}^{m\times d}$ is a weight matrix and is assumed to be \textit{unknown}, specified for each arm $a\in[K]$. $\bomega_{t}\in\bR^{m}$ is a $m$-dimensional random noise vector independently for each $t$, and each entry $\omega_{t,i}$ is with mean $0$.  %Note that all our results can be directly generalized to the setting where the weight matrix $\bm{W}_a^*$ is unknown for each $a\in[K]$, as described in \Cref{sec:Conclusion}. 
%The noise $\bxi_t^c$ is zero-mean independent sub-Gaussian with parameter $\sigma$. 
The reward $r_t(a, \bx_t)$ is stochastic and is assumed to follow the relationship
\begin{equation}\label{eqn:rewardcovariate}
r_t(a, \bx_t)= \left\langle \bmu^{\star}_{a},\bx_t \right\rangle +\xi_{t}, \ a\in[K] 
\end{equation}
where $\bmu^\star_a\in\mathbb{R}^d$ is an \textit{unknown} weight vector, specified for each arm $a\in[m]$, and $\xi_{t} $ is a $0$-mean random noise. We further assume that the noises $\xi_t$ and $\bomega_t$ are sub-Gaussian with parameter $\sigma$, i.e., $\bE\exp(\xi_t^2/\sigma^2)\le 2$, $\bE \exp(\omega_{t,i}^2/\sigma^2)\le 2$. In the following discussion, we will sometimes write the reward $r(a, \bx_t)$ as $r_t$ for simplicity. We also make bounded assumption on the size of the expected resource consumption and reward, i.e.,
there exists constants $R_{\max}$ and $B_{\max}$ such that for any arm $a$ and covariate $\bx_t\sim F$, it holds that  $\abs{\left\langle \bmu^{\star}_{a},\bx_t \right\rangle} \le R_{\max}$, $\|\bm{W}^\star_a\bx_t\|_{\infty}\le B_{\max}$. %$\norm{\bb(a,\bx_t) }_{\infty} \leq B_{\max}$. %\|\bm{W}^*_a\bm{x}\|_{\infty} 
We also assume that there is a null arm representing ``void-action'', i.e., consuming no resource and gain no reward. We denote the null arm as $\bmu^{\star}_0=\bm{0}$, $\bW^{\star}_0=\bm{0}$ with
\begin{equation*}
    \bb_t(0,\bx_t)=\bm{0}, \quad r_t(0,\bx_t)=0.
\end{equation*}
This assumption is standard in online allocation literature \citep{agrawal2016linear,agrawal2016efficient,balseiro2023survey} as the decision maker can always decide not to enter the market.

After seeing the feature $\bx_t$, a decision maker must decide online which arm to pull. If arm $a_t$ is pulled for query $t$, then each resource $i\in[m]$ will be consumed by $b_{i,t}(a_t, \bx_t)$ units and a reward $r_t(a_t, \bx_t)$ will be collected. The realized value of $r_t(a_t, \bx_t)$ is also observed. Note that query $t$ is only feasible to be served if the remaining capacities exceed $\bb_t(a_t, \bx_t)$ component-wise. The decision maker's goal is to maximize the total collected reward subject to the resource capacity constraint. 

% To represent the relaxed benchmark, we use decision rules
% $y_{a,t}(\bx_t)\in[0,1]$, where $y_{a,t}(\cdot)$ denotes a possibly
% time-dependent measurable mapping from the current covariate to the probability
% of selecting arm $a$ at period $t$. 

To accommodate the possibility of taking no action, we equivalently introduce
\textit{relaxed} decision rules $y_{a,t}(\cdot)$, where
$y_{a,t}(\bx_t)\in[0,1]$ denotes the probability of selecting arm $a$ at time
$t$ after observing the covariate $\bx_t$. These decision rules satisfy
$\sum_{a\in[K]}y_{a,t}(\bx_t)\le 1$, where the remaining probability corresponds
to the void action. The realized reward at time $t$ can then be written as $  r_t
    =
    \sum_{a\in[K]}
    \left((\bmu_a^\star)^\top\bx_t+\xi_{a,t}\right)
    y_{a,t}(\bx_t).$ This notation is used mainly for the relaxed
fluid formulation. For an online policy, the corresponding realized action at time $t$: $y_{a,t}\in\{0,1\}$ is
adapted to the filtration generated by the past history and the current covariate.

\begin{comment}
    
\commentWT{We need to address the ``void-action'' issue: we set the decision region as $\sum_{a\in[K]} y_{a,t}(\bx_t) \le 1$, , with the reward as $\sum_{t=1}^T\sum_{a\in[K]} ((\bmu_a^\star)^\top \bx_t+\xi_{t}) \cdot y_{a,t} $}

\commentWT{Also, please use $\bb_t$ and $r_t$ with the subscript $t$!}
\end{comment}

% The benchmark is the offline decision maker that is aware of the value of $\bmu_a^\star$ and $\bx_t$ for all $a\in[K]$, $t\in[T]$ and always makes the optimal decision in hindsight. We denote by $\{y^{\text{off}}_{a,t}, \forall a\in[K]\}_{t=1}^T$ the offline decision of the offline optimum, which is an optimal solution to the following offline problem:
% \begin{equation*}
%     \begin{gathered}
%         \VO(I)=\max_{y_{a,t}}  \sum_{t=1}^T\sum_{a\in[K]} ((\bmu_a^\star)^\top \bx_t +\xi_{t}) \cdot y_{a,t}\label{lp:offlineoptimum}\\
% \mathrm{s.t.\ }  \sum_{t=1}^T\left(\sum_{a\in[K]} \bm{W}_a^{\star} \bx_t +\bomega_t\right)\cdot y_{a,t} \preceq \bC \nonumber\\
%  y_{a,t}\in\{0,1\},\  \sum_{a\in[K]} y_{a,t} \le 1, \quad \quad \forall t\in[T] \nonumber
%     \end{gathered}
% \end{equation*}

The benchmark is an offline decision maker that knows the true parameters
$\{\bmu_a^\star,\bm{W}_a^\star\}_{a\in[K]}$ and observes the full covariate
sequence $\{\bx_t\}_{t=1}^T$ in advance. Following the standard fluid-benchmark
convention in stochastic BwK and CBwK models
\citep{badanidiyuru2013bandits,agrawal2016linear}, this benchmark evaluates
rewards and resource consumptions through their conditional means. We denote by
$\{y^{\mathrm{off}}_{a,t},a\in[K]\}_{t=1}^T$ the decisions of the offline
optimum, which solve
\begin{equation*}
    \begin{gathered}
        \VO(I)=
        \max_{y_{a,t}}
        \sum_{t=1}^T\sum_{a\in[K]}
        (\bmu_a^\star)^\top \bx_t \cdot y_{a,t} \\
        \mathrm{s.t.}\quad
        \sum_{t=1}^T
        \sum_{a\in[K]}
        \bm{W}_a^{\star}\bx_t \cdot y_{a,t}
        \preceq \bC, \\
        y_{a,t}\in\{0,1\},\quad
        \sum_{a\in[K]} y_{a,t}\le 1,
        \quad \forall t\in[T].
    \end{gathered}
\end{equation*}
Here $I=\{(\bx_t,\xi_t,\bomega_t)\}_{t=1}^T$ denotes the realized random sequence. The reward
and consumption noises are realized only after an action is selected and are
therefore not available to the offline benchmark when making decisions.

 For any feasible online policy $\pi$, we use \textit{regret} to measure its performance, which is defined as follows:
\begin{equation}\label{eqn:regret}
    \text{Regret}(\pi):= \mathbb{E}_{I\sim F}[\VO(I)]-\mathbb{E}_{ I\sim F}[V^{\pi}(I)]
\end{equation}
where $I=\{(\bx_t,\xi_t,\bomega_t)\}_{t=1}^T\sim F$ denotes that $\bx_t$ follows distribution $F(\cdot)$ independently for each $t\in[T]$, and $V^{\pi}(I)$ denotes the total value collected under the policy $\pi$.  A common upper bound of  $\mathbb{E}_{I\sim F}[V^{\text{off}}(I)]$ can be written as the following fluid relaxation:
\begin{align*}
V^{\mathrm{UB}}
=
\max_{\{y_{a,t}(\cdot)\}}
&\quad
\sum_{t=1}^T\sum_{a\in[K]}
\bE_{\bx_t\sim F}
\left[
(\bmu_a^\star)^\top \bx_t \cdot y_{a,t}(\bx_t)
\right] \\
\mathrm{s.t.}
&\quad
\sum_{t=1}^T\sum_{a\in[K]}
\bE_{\bx_t\sim F}
\left[
\bW_a^\star \bx_t \cdot y_{a,t}(\bx_t)
\right]
\le \bC, \\
&\quad
y_{a,t}(\bx_t)\in[0,1],
\qquad
\sum_{a\in[K]}y_{a,t}(\bx_t)\le 1,
\qquad \forall t\in[T].
\end{align*}
Here, the resource constraint is imposed in expectation over the covariate
distribution. The last line specifies that, for any realized covariate, the
relaxed action probabilities lie in the unit simplex with a possible void action.

The following result is standard in the literature, which formally establishes the fact that $\VU$ can be used to upper bound the regret of any policy $\pi$.
\begin{lemma}[folklore]\label{lem:Upperbound}
We have $\mathbb{E}_{I\sim F}[\VO(I)]\leq\VU$.
\end{lemma}
Therefore, in what follows, we benchmark against $\VU$ and we exploit the structures of $\VU$ to derive our online policy and bound the regret.

\subsection{High-dimensional features and sparsity structures}
We consider the case where the dimension of the feature $d$ is very large, and a sparsity structure exists for the weight vector $\bmu_a^\star$. Specifically, we assume that there exists $s_0$ such that the uniform sparsity level can be bounded: $\|\bmu_a^\star \|_0\le s_0$ for each $a\in[K]$, given $s_0\ll d$, and a bound on the general range of arms: $\norm{\bmu_a^\star}_\infty \le 1$. Accordingly, we assume that $\bm{W}_a^{\star}$ are also row-wise sparse with each row satisfying $\max_{a\in[K],i\in [m] }\norm{\bm{W}_{a,i\cdot}^{\star} }_{0}\le s_0$, with maximum entry satisfying $\max_{a\in[K] }\norm{\bm{W}_{a}^{\star} }_{\max}\le 1$. The entrywise bound on $\bW_a^\star$ is imposed as a normalization condition for
the estimation problem. This normalization places the initial estimator within a constant-radius neighborhood of $\bW_a^\star$, a standard condition in the analysis of iterative gradient-based and thresholded-gradient procedures \citep{loh2017statistical,loh2017support}.
The matrix-valued parameter $\bm{W}_a^{\star}$ captures the dependence of resource consumption on high-dimensional covariates. We impose a row-wise sparsity structure on $\bm{W}_a^{\star}$ to make the model tractable in high dimensions. Related matrix-structured models, including sparse and low-rank formulations, have been studied in the operations and revenue-management literature; see, e.g., \cite{kallus2020dynamic,liu2022learning,cai2025doubly}. To establish the theory of online learning, one must ensure that the information of each $\bmu_a^\star$ and $\bm{W}_{a}^{\star} $ can be retrieved statistically based on the observation. The following basic assumptions are necessary for such sparse learning.

\begin{assumption}\label{assump:general}
We make the following assumptions on the covariate throughout the paper.
\begin{enumerate}[label=1.\arabic*,ref= 1.\arabic*]
    \item There exists a constant $D$ such that the covariate $\bx_t$ is uniformly bounded: $\norm{\bx_t}_\infty \le  D $.\label{asm:size-of-prob}
    \item For any $s$, the covariance matrix  $\bSigma:=\bE \bx_t \bx_t^\top$ has the $2s$-sparse minimal eigenvalue $\phi_{\min}(s)$ and $2s$-sparse maximal eigenvalue $\phi_{\max}(s)$ \citep{meinshausen2008lasso}, where $\phi_{\min }(s)$ is defined as: 
    \begin{equation*}\label{asm:bounded-ftr}
        \phi_{\min }(s)=\min _{\bbeta:\|\bbeta\|_{0} \leq\lceil 2s\rceil} \frac{\bbeta^\top \bSigma \bbeta}{\bbeta^\top \bbeta}. % \text{ and } \  \phi_{\max}(s)=\max _{\bbeta:\|\bbeta\|_{0} \leq\lceil 2s\rceil } \frac{\bbeta^\top \bSigma \bbeta}{\bbeta^\top \bbeta}.
    \end{equation*}
     $\phi_{\max }(s)$ is also correspondingly defined. Then the condition number can be denoted by $\kappa = \frac{\phi_{\max}(s)}{\phi_{\min }(s)}$.
\end{enumerate}
\end{assumption}
% \begin{assumption}\label{asm:size-of-prob}
%     The covariate $\bx_t$ is uniformly bounded: $\norm{\bx_t}_\infty \le  D $.  %with the true parameter confined in a region $\norm{\bmu_\star}_\infty\lesssim D\sqrt{\log d} $  0-mean and
% \end{assumption}
% \begin{assumption}
%  There exists a constant $B_{\max}$ such that for any covariate $\bx$, it holds that $\|\bm{W}^*_a\bm{x}\|_{\infty}\leq B_{\max}$.   
% \end{assumption}
% \begin{assumption}\label{asm:bounded-ftr}
%     The covariance matrix  $\bSigma:=\bE \bx_t \bx_t^\top$ has the $2s$-sparse minimal eigenvalue $\phi_{\min}(s)$ and $2s$-sparse maximal eigenvalue $\phi_{\max}(s)$ \citep{meinshausen2008lasso}:
%     \begin{equation*}
%         \phi_{\min }(s)=\min _{\bbeta:\|\bbeta\|_{0} \leq\lceil 2s\rceil} \frac{\bbeta^\top \bSigma \bbeta}{\bbeta^\top \bbeta}, \text{ and } \  \phi_{\max}(s)=\max _{\bbeta:\|\bbeta\|_{0} \leq\lceil 2s\rceil } \frac{\bbeta^\top \bSigma \bbeta}{\bbeta^\top \bbeta}.
%     \end{equation*}
%     We further define the condition number of our problem as $\kappa = \frac{\phi_{\max}(s)}{\phi_{\min }(s)}$.
% \end{assumption}
The sparse minimal eigenvalue condition essentially shares the same idea as the restrict eigenvalue conditions that have been broadly used in the high-dimensional sparse bandit problem \citep{bastani2020online,hao2020high,oh2021sparsity,li2022simple}. It ensures that the sparse structure can be detected from the sampling. In the following discussion, we assume that $\kappa=O(1)$ is bounded.
%%%%% better take 2s or 3s for math derivation.
\begin{comment}
    \commentWT{Dependence on $\kappa$ is about $\kappa^4$}
\end{comment}

\section{Optimal Online Sparse Estimation}
The primal task for our online learning problem is to estimate the high-dimensional arms during the exploration, which serves as the foundation of our learning strategies. To this end, we focus on estimating one specific arm in this section, say, estimating $\bmu_a^\star$ for one $a\in[K]$ with the observation $\bx_t$ and $r_t$. Estimating $\bm{W}_{a}^{\star} $ can be similarily conducted by treating $\bb_i(a,\bx_t)$ as the response of each row $\bW_{a,i\cdot}^{\star}$.
Since $\|\bmu_a^\star\|_0\le s_0$ for $s_0\ll d$, for the linear problem, recovering $\bmu_a^\star$ is equivalent to the following $\ell_0$-constrained optimization problem:
\begin{equation}\label{eq:sparse-recov-det}
   \min_{\norm{\bmu}_0 \le s_0 } f(\bmu) := \bE (r_t- \bmu^\top \bx_t )^2 = \norm{\bmu-\bmu_a^\star}_{\bSigma}^2+\sigma^2.
\end{equation}
To solve $\eqref{eq:sparse-recov-det}$, LASSO is widely used in the literature. Despite its statistical optimality, this method relies heavily on accumulated data for the $\ell_1$-regularized optimization, making it difficult to adapt to an online setting—particularly for sequential estimations. Consequently, in high-dimensional online learning, there is a pressing need for a sparse estimation algorithm that runs \textit{fully online} while still achieving the optimal statistical rate. Within the context of the $\epsilon$-greedy sampling strategy, we present our proposed optimal online sparse estimation algorithm in Algorithm~\ref{alg:online-iht}. To simplify notation, we define the sparse projection $\cH_s(x)$ as the hard-thresholding operator that zeros out all entries in $x$ except the largest (in absolute value) $s$ entries. Here, we denote $\varrho = s_0/s$ as the relative sparsity level.

%stochastic gradient $\bg_t=2 \bx_t (\widetilde{\bmu}_{t-1}^\top \bx_t-r_t)= 2 \bx_t \bx_t^\top(\widetilde{\bmu}_{t-1}-\bmu_\star)-2\xi_t \cdot \bx_t$
\begin{algorithm}[ht]
   \caption{Online Hard Thresholding with Averaged Gradient (Online HT)}
   \label{alg:online-iht}
\begin{algorithmic}[1]
   \STATE {\bfseries Input:} $T$, step size $\beta_t$, sparsity level $s$, $s_0$, arm $a$, $ \bmu_{a,0}=\bm{0}$
    \FOR{$t = 1,\dots,T$} 
   % \FOR{$i=1$ {\bfseries to} $m-1$}
\STATE{For the fixed arm $a$, observe $y_{a,t}=\mathbbm{1}\{y_t=a\}$ with marginal selection probability $p_{a,t}=\mathbb{P}(y_t=a\mid \cH_{t-1},\bx_t)$} % p_t=(1-\varepsilon_t)\widetilde{y}_t+ \frac{\varepsilon_t}{2} \textbf{from} 1  \textbf{to}
        \IF{$p_{a,t}=0$}
         \STATE{Treat $y_{a,t}/p_{a,t}=0$}
        \ENDIF
         \STATE {Compute the covariance matrix \\ $\widehat{\bSigma}_{a,t} = 1/t\cdot \left( (t-1)\widehat{\bSigma}_{a,t-1}+ y_{a,t} \bx_t \bx_t^\top/p_{a,t}  \right)$ }
         \STATE{Get averaged stochastic gradient: \\ $\bg_{a,t} = 2  \widehat{\bSigma}_{a,t} \bmu_{a,t-1} - \frac{2}{t}\sum_{j=1}^{t} y_{a,j}\bx_j r_j/p_{a,j}  $ }
         \STATE{Gradient descent with hard thresholding: \\ ${\bmu}_{a,t}= \cH_s({\bmu}_{a,t-1}-\beta_t \bg_{a,t} ) $}
        % \State {Averaging:  $\bmu_t= 1/t \cdot \left(  (t-1){\bmu}_{t-1}+ \widetilde{\bmu}_t \right) $}
         \STATE{Exact $s_0$-sparse estimation: ${\bmu}_{a,t}^{ \mathsf{s} } = \cH_{s_0}({\bmu}_{a,t} )  $ }
    \ENDFOR
%    \OUTPUT
\STATE {\bfseries Output:} $\left\{\bmu_t^{ \mathsf{s} }\right\}$, $t\in [T]$
   %    \REPEAT
   % \UNTIL{$noChange$ is $true$}
\end{algorithmic}
\end{algorithm}

\begin{theorem}\label{thm:sparse-est}
   Define $ \underline{p_j} =\inf {p_{a,j}  } $ as the lower bound of each $p_{a,j}$ and suppose $\underline{p_j} > \Omega(j^{-1})$. If we take the relative sparsity level satisfying $\varrho:=s_0/s = \frac{1}{36\kappa^4}$, and $\beta_t= \frac{1}{4\kappa\phi_{\max}(s) } $, then under \Cref{assump:general}, the output of Algorithm \ref{alg:online-iht} satisfies 
    \begin{equation*}
        \bE \norm{\bmu_{a,t}^{\mathsf{s} } -\bmu_a^{\star} }_2^2 \lesssim \frac{\sigma^2 D^2 s_0 }{\phi_{\min}^2(s)  } \frac{\log d }{t^2} \left(\sum_{j=1}^{t} \frac{1}{
        \underline{p_j} } \right),
    \end{equation*}   
    for all $t\ge t_0+C\kappa^2\log s_0$, and the high-probability bound
        \begin{equation*}
        \norm{\bmu_{a,t}^{\mathsf{s} } -\bmu_a^\star }_2^2 \lesssim \frac{\sigma^2 D^2 s_0 }{\phi_{\min}^2(s)  } \frac{\log (dT/\varepsilon) }{t^2} \left(\sum_{j=1}^{t} \frac{1}{ \underline{p_j} } \right),
    \end{equation*}
    which holds for all $t\ge t_0+C\kappa^2\log s_0$ with probability at least $1-\varepsilon$. Here the $t_0$ is the time such that $t_0/\sqrt{ \sum_{j=1}^{ t_0}1/\underline{p_j}}\ge C\kappa^2 \frac{s_0 D\sqrt{\log(dT)}}{ \phi_{\min}(s) } $.
    %These two bounds also hold for exact $s_0$-sparse estimation $\bmu_t^{ }$.
\end{theorem}
Algorithm \ref{alg:online-iht} serves as an online counterpart of the classic LASSO method. It achieves the statistically optimal rate of sparse estimation in the sense that, if we force $p_{a,j}=\frac{1}{K}$ for each $j$ (which corresponds to the uniform sampling case), then we obtain the estimation error $O\left(\frac{s_0\sigma^2 \log d }{\phi_{\min}^2(s) t }\right)$, which matches the well-known optimal sparse estimation error rate \citep{ye2010rate,tsybakov2011exponential}. Algorithm \ref{alg:online-iht} needs to continuously maintain an empirical covariance matrix $\widehat{\bSigma}_{a,t}$, which takes up $O(d^2)$ storage space; however, all the updates of $\widehat{\bSigma}_{a,t}$ and stochastic gradients $\bg_{a,t}$ can be computed linearly, which leads to the fast $O(d^2T)$ total computational complexity. Moreover, our bound can be easily extended to the uniform bound over all arms $\bE \max_{a\in [K]} \norm{\bmu_{a,t}^{\mathsf{s} } -\bmu_a^{\star} }_2^2 $, with only an additional $\log K$ term on the error rate. See the supplementary materials for details. 
The $ \underline{p_j} $ here is used to adapt our algorithm to the $\epsilon$-greedy exploration strategy. If for each $j$, the arm $a$ can be sampled with minimum probability $\epsilon_j$, then we have $p_{a,j}=1-(K-1)\epsilon_j$ or $p_{a,j} = \epsilon_j$ for arm $a$, implying that $ \underline{p_j}=\epsilon_j$. The inverse probability weight $1/p_{a,j}$ we use in Algorithm \ref{alg:online-iht} serves to correct the empirical covariance matrix and the gradients of each iteration by importance sampling\citep{chen2021statistical}, making the gradient estimation consistent. Actually, the error rate in Theorem \ref{thm:sparse-est} also applies to $s$-sparse estimator ${\bmu}_{a,t}$. Thus, when $s_0$ is unknown, we can just use ${\bmu}_{a,t}$ instead.
%de-bias sampling distribution correct

For the hard-thresholding type method, the major challenge for the online algorithm design is the gradient information loss caused by truncation. In the online update,  the hard thresholding operator will zero out all the small signals, which contain valuable gradient information for the next update \citep{murata2018sample,zhou2018efficient}. Moreover, the missing information will accumulate during the online iteration, rendering it difficult for previous methods to recover a sparse structure \citep{nguyen2017linear,murata2018sample,zhou2018efficient}. To tackle this issue, we choose a slightly larger sparsity level that allows us to preserve more information on the gradient. We show that a larger sparsity level (which depends on the condition number $\kappa$) allows us to keep enough information so that the truncation effect is negligible.

%The notion of averaging is also used in the sparse estimation in, e.g., \cite{han2023online} but is with different objectives. \citep{han2023online} does averaging on estimators, which is used for online inference, and the soft thresholding in \cite{han2023online} can not guarantee exact $s_0$ sparse recovery.  , and use the gradient averaging in each step to obtain a more accurate characterization of the gradient.

\begin{remark}
    The rationale behind the gradient averaging in Algorithm \ref{alg:online-iht} is actually the poor smoothness property of the hard thresholding operator, i.e., projection onto $\ell_0$-constraint space. Unlike the convex projection or higher-order low-rank projection, the projection onto the $\ell_0$-constraint space exhibits an inflating smoothness behavior. To be specific, the projection onto the convex space shares the nice property $ \norm{\cP(\bx+\Delta) - \bx }_2\le  \norm{\Delta}_2$, with no inflation on the error. The projection onto the low-rank space (e.g., SVD or HOSVD on low-rank matrix or tensor) also satisfies $ \norm{\cP(\bx+\Delta) - \bx }_\tF \le \norm{\Delta}_\tF+C\norm{\Delta}_\tF^2$ if $\Delta$ is in the tangent space of the manifold \citep{kressner2014low,cai2022generalized}, which leads to tiny inflation in online low-rank learning \citep{cai2023online}. However, the projection onto $\ell_0$-constraint space can only ensure $\norm{\cP(\bx+\Delta) - \bx }_2\le (1+\delta_s)\norm{\Delta}_2 $, where $\delta_s$ is a non-zero parameter depending on the relative sparsity level and is unimprovable \citep{shen2017tight}, which causes trouble for online sparse recovery. To mitigate the inevitable inflation, gradient averaging is employed to reduce the variance, thereby enabling us to achieve the optimal convergence rate. Thus, the two hard-thresholding levels play different roles: $s$ is a working
sparsity level to stabilize the online iterative update and prevent the
loss of useful gradient information, while $s_0$ is the target sparsity level
used to output the final exact sparse estimator. In theory, we take
$s=36\kappa^4s_0$ to guarantee contraction; in practice, $s_0$ can be tuned by
standard validation or stability-based procedures, and $s$ can be chosen as a
moderate multiple of $s_0$.
\end{remark}

For the CBwK problem, since $\bb_t(a,\bx_t)$ are also unknown for decision-makers, we need to consecutively estimate the size, or equivalently, $\bW_a^{\star}$. To this end, we can treat each row $\bW_{a,i\cdot}^{\star}$ as a sparse vector (substituting $\bmu_{a}^{\star}$) with $\bb_{t,i}(a,\bx_t)$ as the response (substituting $r_t$), and estimate them using Algorithm \ref{alg:online-iht}. The error of estimating ${\bW}_{a,i\cdot}^{\star}$ shares the same order as estimating $\bmu_a^\star$. See the supplementary materials for the exact error bound of the estimation $\widehat{\bW}_{a,t}$.
% The error bound of estimating $\bW_a^{\star}$ follows the \Cref{thm:sparse-est}. 

% \subsection{Estimate Cost}

\section{Online Allocation: CBwK Problem}\label{sec:CBwK-regret}
\subsection{The dual problem}

In this section, we handle the CBwK problem described in \Cref{sec:formulation}. Our algorithm adopts a primal-dual framework, where we introduce a dual variable to reflect the capacity consumption of each resource. The dual variable can be interpreted as the Lagrangian dual variable for $\VU$, with the dual function:
\begin{equation*}
    \begin{gathered}
              L(\blambda)= \sum_{t=1}^T \mathbb{E}_{\bx_t\sim F} \big[\max_{\bm{y}_t(\bx_t)\in\Delta^K}\big\{  \sum_{a\in[K]}(\bmu_a^\star)^\top \bx_t\cdot y_{a,t}(\bx_t) -   (\bm{W}_a^{\star} \bx_t)^{\top}\blambda\cdot y_{a,t}(\bx_t)\big\} \big] + \bC^\top \blambda, 
    \end{gathered}
\end{equation*}
% $$
%     \begin{aligned}
%         L(\bm{\eta}) & = \sum_{t=1}^T \mathbb{E}_{\bx_t\sim F} \big[\max_{\bm{y}_t(\bx_t)\in\Delta^K}\big\{  \sum_{a\in[K]}(\bmu_a^\star)^\top \bx_t\cdot y_{a,t}(\bx_t)   \\
%        & -   Z\cdot(\bm{W}_a^{\star} \bx_t)^{\top}\bm{\eta}\cdot y_{a,t}(\bx_t)\big\} \big] +\bC^\top \bm{\eta}, 
%     \end{aligned}
% $$
where $\Delta^K$ denotes the unit simplex in $\bR^K$: $\Delta^K=\{\bm{y}\in\mathbb{R}^K: y_a\geq0, \forall a\in[K], \text{~and~}\sum_{a\in[K]}y_a\le 1\}$. Note that if the weight vector $\bmu^\star_a$, cost $\bW_a^{\star}$ are given for each arm $a\in[K]$ and the distribution $F(\cdot)$ is known, one can directly solve the dual problem $\min_{\blambda \succeq 0}L(\blambda)$ to obtain the optimal dual variable $\blambda^*$ and then the primal variable $y_{a,t}(\bx_t)$ can be decided by solving the inner maximization problem in the definition of the dual function $L(\blambda)$. However, since they are all unknown, one cannot directly solve the dual problem. Instead, we will employ an online learning algorithm (mirror descent) and use the information we obtained at each period as the feedback to the online algorithm to update the dual variable $\bm{\eta}_t$.

\subsection{Rescaling and online mirror descent}
It has been broadly recognized in the online optimization  that selecting an appropriate reference function in mirror descent can substantially reduce the regret’s dependence on the dimensionality \citep{agrawal2016efficient,balseiro2023best}. Therefore, to achieve a regret with minimal dependence on the number of resources $m$, we apply the Hedge algorithm (with the negative entropy function as the reference function) by rescaling the dual variable $\blambda$ to the unit simplex. To this end, we define $\bm{\eta}=\blambda/Z $ and $\bm{\eta}^*=\blambda^*/Z\in \Delta^{m}$. We can then write
\begin{equation*}
    \begin{aligned}
              L(\blambda) & = L(Z\cdot\bm{\eta}) \\
              & = \sum_{t=1}^T \mathbb{E}_{\bx_t\sim F} \big[\max_{\bm{y}_t(\bx_t)\in\Delta^K}\big\{  \sum_{a\in[K]}(\bmu_a^\star)^\top \bx_t\cdot y_{a,t}(\bx_t)   
        -   Z\cdot(\bm{W}_a^{\star} \bx_t)^{\top}\bm{\eta}\cdot y_{a,t}(\bx_t)\big\} \big] + Z\cdot\bC^\top \bm{\eta}, 
    \end{aligned}
\end{equation*}
and $Z$ is a scaling parameter that we will specify later.

 Then, we plug in the dual variable $\bm{\eta}_t$, as well as estimates of $\bmu_a^\star$ and $\bW_a^{\star}$ for each $a\in[K]$, to solve the inner maximization problem in the definition of the dual function $L(\cdot)$ to obtain the primal variable $y_{a,t}(\bx_t)$. Note that this primal-dual framework has been developed previously in the literature (e.g. \cite{badanidiyuru2013bandits, agrawal2016linear}) of bandits with knapsacks for UCB algorithms. The innovation of our algorithm is that, instead of using UCB in the primal selection, we use $\epsilon$-greedy for exploration with a finer estimate of $\bmu^\star_a$ and $\bW_a^{\star}$ via \Cref{alg:online-iht}, which enables us to exploit the sparsity structure of the problem and obtain improved regret bound. Our formal algorithm is presented in \Cref{alg:LagrangianBwK}. In the algorithm, the vector $\bm{\alpha}_t$ in the dual update is the unnormalized
multiplicative-weight vector. Each coordinate of the previous dual variable
$\bm{\eta}_{t-1}$ is multiplied by an exponential factor depending on the
realized resource-consumption violation
$b_i(y_t,\bx_t)-C_i/T$. The vector is then normalized, or projected, back to the
unit simplex to obtain $\bm{\eta}_t$. This is the standard Hedge update for
online linear optimization with the negative-entropy mirror map.

\begin{algorithm}[ht!]
\caption{Primal-Dual High-dimensional CBwK Algorithm}
\label{alg:LagrangianBwK}
\begin{algorithmic}[1]
\STATE {\bfseries Input:} $Z$, $\epsilon$-greedy probability $\epsilon_t$ for each $t$, $\delta$.
\STATE {In the first $m$ rounds, pull each arm once and initialize $\bm{\eta}_m=\frac{1}{m}\bm{1}_m$. Set $ \bmu_{a,m}^{\mathsf{s}}=\bm{0}$, $\widehat{\bW}_{a,m}=\bm{0}$ }
\FOR {$t=m+1,..., T$}

\STATE {Observe the feature $\bx_t$.}
%construct UCB estimate for the expected reward $u_{a,t}\in[0,1]$ and the LCB estimate for the resource consumption vector, $\bm{L}_{a,t}\in[0,1]^m$.
\STATE  Estimate $\operatorname{EstCost}(a)=\bx_t^\top \widehat{\bW}_{a,t-1}^\top\bm{\eta}_{t-1}$ for each arm $a\in[K]$.
\STATE  Sample a random variable $\nu_t\sim \operatorname{Ber}(K\epsilon_t) $,
\IF{$\nu_t=0$}
\STATE{$y_t=\widetilde{y}_t:=\displaystyle\arg\max_{a\in[K]} \{(\bmu_{a,t-1}^{\mathsf{s}} )^\top\bx_t-Z\cdot\operatorname{EstCost}(a)\}$}
\ELSE
\STATE{$y_t$ is uniformly selected from $[K]$}
\ENDIF

% \STATE and pull the arm $y_t$ defined as follows: % where  $ p_t=K\epsilon_t$
% \[
% y_t=\left\{\begin{aligned}
%    & ,  \text{ if } \nu_t=0\\
%    &a, ~~~~\text{w.p.~} 1/K \text{~for~each~arm~}a\in[K]  \text{ if }\nu_t=1. 
% \end{aligned}
% \right.
% \]

\STATE  Receive $r_t$ and $\bb(y_t,\bx_t)$. If one of the constraints is violated, then EXIT.
\STATE  Update for each resource $i\in[m]$,
\[
\alpha_{t}(i)=\bm{\eta}_{t-1}(i)\cdot\exp(\delta\cdot{(b_i(y_t,\bx_t)-\frac{C_i}{T})\cdot(1-\nu_t)})
\]
and project $\bm{\alpha}_{t}$ into the unit simplex $\{ \bm{\eta}: \|\bm{\eta}\|_1\leq1, \bm{\eta}\geq0 \}$ to obtain $\bm{\eta}_{t}$ as follows:
\[
\bm{\eta}_{t}(i)= \left\{ \begin{array}{ccl}
          \alpha_{t}(i) & \mbox{if}
         & \norm{\bm{\alpha}_{t} }_1\le 1 \\ 
           \frac{\alpha_{t}(i)}{\sum_{i'\in[m]}\alpha_{t}(i')}  & \mbox{if} & \norm{\bm{\alpha}_{t} }_1> 1 \\
                \end{array}\right.
               , ~~~\forall i\in[m].
\]

\STATE  For each arm $a\in[K]$, update the estimate $\bmu_{a,t}^{\mathsf{s} }$ and $\widehat{\bW}_{a,t}$ from \Cref{alg:online-iht} by letting 
\begin{equation*}
    p_{a,t}=\left\{ \begin{array}{ccl}
          1-(K-1)\epsilon_t & \mbox{for} & a= \widetilde{y}_t\\
        \epsilon_t & \mbox{for} & a\neq \widetilde{y}_t
                \end{array}\right. , \quad \text{and }  y_{a,t}= \mathbbm{1}\{y_t=a\}, \quad \forall a\in[K].
\end{equation*}
\ENDFOR
\end{algorithmic}
\end{algorithm}

\section{Regret of Primal-Dual CBwK}\label{sec:regret-sqrt}
\subsection{Regret analysis}
In this section, we conduct regret analysis of \Cref{alg:LagrangianBwK}, where Algorithm~\ref{alg:online-iht} inside  is run with the working sparsity level and
step size specified in Theorem~\ref{thm:sparse-est}. We first show how regret depends on the choice of $\epsilon_t$, for each $t\in[T]$, as well as the estimation error of our estimator of $\bmu^\star_a$, $\bW_a^\star$ for each $a\in[K]$. We then specify the exact value of $\epsilon_t$ and utilize the estimation error characterized in \Cref{thm:sparse-est} to derive our final regret bound.   

\begin{comment}
    \commentWT{According to the current version, the regret of Hedge should be 
\begin{equation*}
    \mathbb{E}\left[\sup_{\bm{\eta} }R(\cT,\bm{\eta}) \right] \leq \sqrt{\left(B_{\max}+\frac{C_{\max}}{T}+\sigma\sqrt{\log d}\right)\cdot T\cdot \log m}
\end{equation*}
We can denote $\widetilde{B}_{\max}=B_{\max}+\frac{C_{\max}}{T}+\sigma \sqrt{\log d}$ for simplicity.
}
\end{comment}

%Let $R_{\max}=\sup \abs{\left\langle \bx_t, \bmu_{\star}^{a} \right\rangle} $. We first prove the following theorem on the regret control of CBwK.
\begin{theorem}\label{thm:Primaldual}
%Denote by $\bm{y}_t=\bm{e}_{a_t}\in\mathbb{R}^{K}$, which indicates the action taken at period $t$. 
Denote by $\pi$ the process of our \Cref{alg:LagrangianBwK}, and $\tau$ the stopping time of \Cref{alg:LagrangianBwK}. Write  $\widetilde B_{\max}:=
\left(B_{\max}+\frac{C_{\max}}{T}\right)^2+\sigma^2\log m $.
If $Z$ satisfies $Z\ge \frac{\VU}{C_{\min}}$, then, under \Cref{assump:general}, the regret of the policy $\pi$ can be upper bounded as follows
\begin{align*}
        & \operatorname{Regret}(\pi) \leq Z C\sqrt{ T\widetilde B_{\max}\cdot \log m} \\
        & +Z\cdot\mathbb{E} \left[\sum_{t=1}^{\tau}\max_{a}\abs{\left\langle 
\bx_t, \bmu_{a}^\star-\bmu_{a,t-1}^{\mathsf{s} } \right \rangle} +D\norm{\widehat{\bW}_{a,t}-{\bW}_{a}^{\star} }_{\infty}\right] \\
& + Z\cdot(4 R_{\max} + 2 B_{\max}Z)\cdot\sum_{t=1}^T K \epsilon_t,
\end{align*}
by setting $\delta= c_{\delta}\sqrt{\frac{\log m}{T \widetilde B_{\max}}}$ for some sufficiently small constant $c_{\delta}>0$.
%, where $R_{\max}=\sup_{\bx, a\in[K]} \abs{\left\langle \bx, \bmu^{\star}_{a} \right\rangle} $ and $B_{\max}$ denotes an upper bound of $b_i(y_t, \bx_t)$ as specified in \Cref{assump:general}.
\end{theorem}
% Ds_0 = R_{\max}
The three terms in \Cref{thm:Primaldual} exhibit distinct components of \Cref{alg:LagrangianBwK} that contribute to the final regret bound. The first term represents the effect of the dual update using the Hedge algorithm \citep{freund1997decision}. While the last two terms arise from online sparse estimation and $\epsilon$-greedy exploration, both of which can be categorized as consequences of the primal update. To make the estimation error in \Cref{thm:sparse-est} and \Cref{cor:uniform-est} meaningful for all $[T]$, we can imagine that the primal estimation process runs from $t=1$ to $T$ such that $\{(\bmu^{\mathsf{s}}_{a,t},\widehat{\bW}_{a,t})\}_{a\in[K]}$ are well defined for each $t\in[T]$ (however, we only have access to $\{(\bmu^{\mathsf{s}}_{a,t},\widehat{\bW}_{a,t})\}_{a\in[K]}$ before the dual allocation algorithm stops). 
Given that the estimation error is confined by \Cref{cor:uniform-est} and \Cref{prop:sparse-est-W}, we can establish the following regret bound:

\begin{theorem}\label{thm:BwK-regret-1}
Under \Cref{assump:general}, suppose that
$\frac{\VU}{C_{\min}}\le Z \le C_Z(\frac{\VU}{C_{\min}}+1)$ and $    T \ge
    C_{\mathrm b}
    \frac{
        K
        \left(
            R_{\max}
            +
            B_{\max}\left(\frac{\VU}{C_{\min}}+1\right)
        \right)
        s_0^2D
        \log^{3/2}(mdKT)
    }{
        \sigma\phi_{\min}^2(s)
    }$ for some numerical constant $C_Z, C_{\mathrm b} >0$. Then the regret of
Algorithm~\ref{alg:LagrangianBwK} satisfies
\begin{align*}
\begin{aligned}
    \operatorname{Regret}(\pi)
    \lesssim \;&
    \left(\frac{\VU}{C_{\min}}+1\right)
    \sqrt{\widetilde B_{\max}T\log m} \\
    &+
    \left(\frac{\VU}{C_{\min}}+1\right)
    \frac{
        \left(
            R_{\max}
            +
            B_{\max}\left(\frac{\VU}{C_{\min}}+1\right)
        \right)^{1/3}
        K^{1/3}
        \sigma^{2/3}D^{4/3}
        s_0^{2/3}
        T^{2/3}
        \log^{1/3}(mdK)
    }{
        \phi_{\min}^{2/3}(s)
    } .
\end{aligned}
\end{align*}
by setting
$\delta=c_{\delta}\sqrt{\frac{\log m}{T\widetilde B_{\max}}}$
and $\epsilon_t=\epsilon_0t^{-1/3}\wedge 1/K$ for some sufficiently small constant $c_{\delta}>0$ and some $T$-independent $\epsilon_0$.
\end{theorem}

The result shows that the regret has two components, $ \widetilde O\left(
    \frac{V^{\mathrm{UB}}}{C_{\min}}\sqrt T
    +
    \left(\frac{V^{\mathrm{UB}}}{C_{\min}}\right)^{1/3}T^{2/3}
    \right).$
Since rewards are bounded, $V^{\mathrm{UB}}$ is at most of order $T$ and is
typically linear in $T$ in non-degenerate instances. Thus the ratio
$V^{\mathrm{UB}}/C_{\min}$ measures the effective scarcity of the resources.
For example, when the capacities scale linearly with $T$, this ratio is often
$O(1)$; when the capacities are smaller, the ratio becomes larger.

The two terms in the regret bound correspond to different sources of error. The
first term is due to the dual update for managing the knapsack constraints,
which is implemented through the Hedge or multiplicative-weights update
\citep{freund1997decision,arora2012multiplicative}. Hence, this term increases
with the scarcity ratio $V^{\mathrm{UB}}/C_{\min}$. The second term is due to
the primal sparse estimation and explicit exploration. This is analogous to the
estimation--exploration trade-off in high-dimensional sparse bandits
\citep{hao2020high,li2022simple,jang2022popart}. The two terms are balanced when
$V^{\mathrm{UB}}/C_{\min}\asymp T^{1/4}$, in which case both are of order
$T^{3/4}$ up to logarithmic factors. If
$V^{\mathrm{UB}}/C_{\min}=O(T^{1/4})$, then the primal-estimation term dominates;
in particular, when $V^{\mathrm{UB}}/C_{\min}=O(1)$, the bound becomes
$\widetilde O(T^{2/3})$. If
$V^{\mathrm{UB}}/C_{\min}\gg T^{1/4}$, then the dual-update term dominates and
the regret is
$\widetilde O((V^{\mathrm{UB}}/C_{\min})\sqrt T)$, which can be larger because
the scarcity ratio itself is large. Therefore, the scarce-resource regime does
not lead to a smaller regret bound; rather, it shifts the bottleneck from primal
estimation to dual learning. Most notably, the regret depends only
logarithmically on the feature dimension $d$, improving the polynomial
dependence on $d$ in previous CBwK results \citep{agrawal2016linear}.

%\subsection{Estimating reward-constraint ratio}
\begin{remark}
The \Cref{alg:LagrangianBwK} require an estimation of reward-constraint ratio $Z$. Such an estimation can be obtained from linear programming similar to that in \cite{agrawal2016linear}. However, different from \cite{agrawal2016linear}, we will use the estimators obtained in \Cref{alg:online-iht} to construct a relaxed linear programming. To be specific, we choose a parameter $T_0$ and use the first $ T_0$ periods to obtain an approximation of $\VU$ , i.e., $\hat{V}$, by uniform sampling. We show that as long as $T_0=\widetilde O\left(s_0^2 \cdot \frac{T^2}{C^2_{\min}}\right)$, we will have $Z=O(\frac{\VU}{C_{\min}}+1)$ with high probability. If further the constraints grow linearly, i.e., $C_{\min}=\Omega(T)$, we only require $T_0=\widetilde O\left(1\right)$ in general. See the \Cref{sec:ParameterEstimation} in Supplement for details.
\end{remark}

\subsection{Improved regret with diverse covariate}
In Theorem \ref{thm:BwK-regret-1}, it is shown that the primal update may become the bottleneck of the regret. This happens because we have to compromise between exploration and exploitation. However, in some cases, when the covariates are diverse enough, our dual allocation algorithm will naturally explore sufficient arms, leading to significant improvement in the exploitation. We now describe such a case with the notion of diverse covariate condition \citep{ren2023dynamic}.

\begin{assumption}[Diverse covariate]
\label{asm:diverse-cov}
    There are (possibly $K$-dependent) positive constants $\gamma(K)$ and $\zeta(K)$, such that for  any unit vector $\bv \in \mathbb{R}^d$, $\norm{\bv}_2=1$ and any $a \in[K]$, conditional on the history $\cH_{t-1}$, there is 
$$\mathbb{P}\left( \bv^{\top} \left( \bx_t \bx_t^{\top} \right) \bv\cdot \bI \left\{y_t=a \right\} \geq\gamma(K)\middle| \cH_{t-1} \right) \geq \zeta(K),$$
where $y_t= \arg\max_{a\in[K]\cup\{0\}}\left\langle \bmu_{a,t-1}^{\mathsf{s}}  - Z\cdot\widehat{\bW}_{a,t-1}^\top\bm{\eta}_{t-1} , \bx_t\right\rangle$ is the greedy selection.
\end{assumption}
% \jnote{which one is better?}
% \begin{assumption}[\jnote{Diverse covariate 2?}]
% \label{asm:diverse-cov}
%     There are (possibly $K$-dependent) positive constants $\gamma(K)$ and $\zeta(K)$, such that for any $\bm{\eta}\in\Delta^K$, any possible estimate $\{\hat{\bmu}_a\}_{\forall a\in[K]}$, any unit vector $\bv \in \mathbb{R}^d$, $\norm{\bv}_2=1$ and any $a \in[K]$, it holds 
% $$\mathbb{P}\left( \bv^{\top} \bx \bx^{\top} \bv\cdot \bI\left\{y(\bx)=a \right\} \geq\gamma(K) \right) \geq \zeta(K),$$
% where $\bx\sim F$ and $y(\bx)=\operatorname{argmax}_{a\in[K]} \{(\hat{\bmu}_{a} )^\top\bx-Z\cdot\bm{b}(a,\bx)^\top\bm{\eta} \}$.
% \end{assumption}
Here, we assume that $\zeta(K)\lesssim 1/K$ without loss of generality. Such a diverse covariate condition states that when we perform the online allocation task, our dual-based algorithm can ensure sufficient exploration. This can be viewed as a primal-dual version of the diverse covariate condition for greedy algorithms \citep{han2020sequential,bastani2021mostly,ren2023dynamic}. When $K=2$, this assumption can be met by a stronger covariate diversity condition in \cite{bastani2021mostly}, i.e.,
\begin{equation}\label{eq:div-cov-K2}
    \lambda_{\min }\left(\mathbb{E}\left[\bx \bx^{\top} \bI\left\{\bx^{\top} \bv \geq 0\right\}\right]\right) \geq \lambda_0, \text{ for any } \bv\in \bR^d.
\end{equation}
For more examples of distributions that satisfy \eqref{eq:div-cov-K2} (and consequently, Assumptions \ref{asm:diverse-cov}), please refer to \cite{bastani2021mostly}.
If our covariate is diverse enough, we can just set $\epsilon_t=0$ in Algorithm \ref{alg:LagrangianBwK} to obtain a good performance of primal exploration. We can show that the primal estimation under diverse covariate condition satisfies $ \bE \norm{\bmu_{a,t}^{\mathsf{s}}-\bmu_{a}^{\star} }_2^2\lesssim \frac{s_0\log d}{t}$. We defer the detail of the $O\left(\frac{s_0\log d}{t}\right)$ sparse estimation  to \Cref{thm:sparse-est-divcov} in the Appendix. 
This rate suggests that under the diverse covariate condition, our primal estimation algorithm can recover the sparse arms with a statistical error rate that is optimal for $t$. This greatly improves the primal performance of our algorithm and thus leads to a sharper regret bound for the BwK problem. We describe this improved regret in \Cref{thm:BwK-regret-2}.

% \begin{algorithm}[H]
% \caption{Averaged Online Hard Thresholding with Diverse Covariate}
% \label{alg:online-iht-divcov}
% \begin{algorithmic}[1]
% \Require $T$, step size $\beta_t$, sparsity level $s$, ${\bmu}^{ \mathsf{s} }_{a,0}= \bmu_{a,0}=0$
%     \For{$t = 1,\dots,T$} 
%     \State {Receive covariates $\bx_t$}
%         \State {Sample the reward $r_t$ according to the decision variable $y_t\in \sigma(\cH_{t-1},\bx_t)  $}.
%         % p_t=(1-\varepsilon_t)\widetilde{y}_t+ \frac{\varepsilon_t}{2}
%     %\textbf{from} 1  \textbf{to}
%         \State {Compute the covariance matrix for each $a\in [K]$:
%         $$\widehat{\bSigma}_{a,t} = 1/t\cdot \left( (t-1)\widehat{\bSigma}_{a,t-1}+ \bI\left\{y_t= a \right\} \bx_t \bx_t^\top  \right)$$ }
%         \State {Get averaged stochastic gradient for each $a\in [K]$: $\bg_{a,t} = 2  \widehat{\bSigma}_{a,t} \bmu_{a,t-1} - \frac{2}{t}\sum_{j=1}^{t} \bI\left\{y_t= a \right\}\bx_j r_j $}
%         \State {Gradient descent with hard thresholding for each $a\in [K]$: ${\bmu}_{a,t}= \cH_s({\bmu}_{a,t-1}-\beta_t \bg_{a,t}) $}
%         % \State {Averaging:  $\bmu_t= 1/t \cdot \left(  (t-1){\bmu}_{t-1}+ \widetilde{\bmu}_t \right) $}
%         \State{Exact $s_0$-sparse estimation ${\bmu}_{a,t}^{ \mathsf{s} } = \cH_{s_0}({\bmu}_{a,t} )  $ }
%     \EndFor
% \Ensure{$\left\{\bmu_{a,t}^{ \mathsf{s} }\right\}$, $a\in [K]$, $t\in [T]$}
% \end{algorithmic}

% \end{algorithm}

\begin{theorem}\label{thm:BwK-regret-2}
Under Assumption \ref{assump:general}, \ref{asm:diverse-cov},  if $Z$ satisfies $\frac{\VU}{C_{\min}}\le Z \le C_Z(\frac{\VU}{C_{\min}}+1)$,  then the regret of the Algorithm \ref{alg:LagrangianBwK}  can be upper bounded by:
\begin{equation*}
\begin{aligned}
    & \operatorname{Regret}(\pi) \lesssim \left(\frac{\VU}{C_{\min}}+1\right)\sqrt{ T\widetilde B_{\max}\log m} + \frac{    s_0\sqrt{ T \log K\log (mdK) }  }{  \gamma(K)\zeta(K)   } 
\end{aligned}
\end{equation*}
by setting $\delta=c_{\delta} \sqrt{\frac{\log m}{T\cdot \widetilde B_{\max} }}$, and $\epsilon_t=0$ for each $t\in[T]$.
\end{theorem}

The rationale behind setting $\epsilon_t=0$ in Algorithm \ref{alg:LagrangianBwK} is that, when our covariate vectors exhibit sufficient diversity, our strategy will automatically explore enough arms while simultaneously optimizing regret. This condition is typically met in the online allocation problem where the optimal strategy is often a distribution within arms, rather than a single arm \citep{badanidiyuru2018bandits}. This starkly contrasts with the classical multi-armed bandit problem, where the optimal solution is typically confined to a single arm. 
Theorem \ref{thm:BwK-regret-2} significantly reduces the impact of primal update on the regret from $\widetilde{O}\left(\left(\frac{\VU}{C_{\min}}\right)^{\frac{1}{3}}T^{\frac{2}{3}} \right) $ to a sharper $\widetilde{O}\left( s_0\sqrt{T} \right) $, which makes the impact of the dual update the dominating factor of regret, giving the bound $ \operatorname{Regret}(\pi)=\widetilde{O}\left( \frac{\VU}{C_{\min}}\sqrt{T} \right)$. 

\subsection{Improved regret with margin condition}
Apart from the diverse covariate condition, we identify another covariate condition, namely--margin condition for CBwK problem that can ensure fast exploration and sharp regret.
Margin condition was initially proposed in the study of classification \citep{mammen1999smooth,tsybakov2004optimal}, and it has now been widely perceived as a key concept in contextual multi-armed bandits \citep{zeevi2009woodroofe,perchet2013multi,bastani2021mostly}. We now describe a new version of margin condition for CBwK problem in terms of Lagrangian:
\begin{assumption}[Static dual margin condition]\label{asm:margin-static} Denote $a^*_t\in[K]\cup\{0\}$ as the optimal arm such that 
\begin{equation*}\begin{aligned}
 a^*_t =  \arg\max_{a\in[K]\cup\{0\}} \left\langle \bmu_{a}^\star - (\bm{W}_{a}^{\star})^{\top}\blambda^*,\bx_t \right\rangle.
\end{aligned}
\end{equation*}
Then, for any $0<\delta\le R_{\max}$, there exists a constant $M_{\sfm }>0$ such that
\begin{equation*}
       \bp\left(\left\langle \bmu_{a^*_t}^\star - (\bm{W}_{a^*_t}^{\star})^{\top}\blambda^*,\bx_t \right\rangle - \max_{a\neq a^*_t}\left\langle \bmu_{a}^\star - (\bm{W}_{a}^{\star})^{\top}\blambda^*,\bx_t \right\rangle \le \delta\right)\le M_{\sfm } \delta.
\end{equation*}
\end{assumption}
Notice that, Assumption \ref{asm:margin-static} only requires the dual value to be well separated from the sub-optimal Lagrangian for a fixed $\blambda=\blambda^*$ (that is why we call it ``static''). This is a simple generalization of the classic margin condition in contextual bandits \citep{zeevi2009woodroofe,perchet2013multi,bastani2021mostly}, where we replace the maximum expected reward $\max_{a\in[K]\cup\{0\}} \left\langle \bmu_{a}^\star,\bx_t \right\rangle$ by its dual value $\max_{a\in[K]\cup\{0\}} \left\langle \bmu_{a}^\star - (\bm{W}_{a}^{\star})^{\top}\blambda^*,\bx_t \right\rangle$. We now show that, if the dual problem exhibits margin condition, then we can control the regret of CBwK at order $\widetilde{O}(s_0\sqrt{T})$:
\begin{theorem}\label{thm:sqrt-margin}
Under Assumption \ref{assump:general}, \ref{asm:margin-static}, if
    $\frac{\VU}{C_{\min}}\le Z \le C_Z(\frac{\VU}{C_{\min}}+1) $ then the regret of the Algorithm \ref{alg:LagrangianBwK}  can be upper bounded by:
\begin{equation*}
\begin{aligned}
    & \operatorname{Regret}(\pi) \lesssim   (\frac{\VU}{C_{\min}}+1)\left( \frac{\sigma D^2 s_0 }{\phi_{\min}(s)  }\sqrt{M_{\sfm} K R_{\max} T\log (mdKT)}+ \sqrt{\widetilde{B}_{\max} T \log m}\right) 
\end{aligned}
\end{equation*}
by setting $\delta=c_{\delta}\sqrt{\frac{\log m}{T\cdot \widetilde B_{\max}}}$, and $\epsilon_t=\epsilon_0 t^{-\gamma} \wedge 1/K $ some sufficiently small constant $c_{\delta}>0$ and some $T$-independent $\epsilon_0$, with $\gamma=\frac{1}{2}+o(1/\sqrt{\log T})$ for each $t\in[T]$.
\end{theorem}
The exact value of $\gamma$ can be found in the supplement. 
\begin{remark}
    We remark that the diverse covariate condition for Theorem \ref{thm:BwK-regret-2} and the margin condition for Theorem \ref{thm:sqrt-margin} improve the regret bound from different perspectives. Specifically, the diverse covariate condition removes the need for explicit exploration, allowing every sample to update the primal parameters; this accelerates the estimation error rate and consequently tightens the regret. In contrast, although the margin condition does not accelerate online estimation, it guarantees that indistinguishable contexts appear only rarely, which likewise yields an improved regret bound.
\end{remark}

\begin{comment}
    \commentWT{Check the $\log(mdK)$ term}
\end{comment}

%one possible condition: where $a^*=\underset{a \in[K]}{\operatorname{argmax} }  \ x_t^{\top} \bmu_a$.

\section{Application of Online HT to High-dimensional Bandit Problem}
An important application of our Algorithm \ref{alg:online-iht} is the high-dimensional bandit problem \citep{carpentier2012bandit,hao2020high}, where we do not consider the knapsacks but only focus on reward maximization (or, we can treat the bandit problem as a special CBwK problem where the constraints are always met). Here we associate our algorithm with $\epsilon$-greedy strategy and show that our high-dimensional bandit algorithm by Online HT can achieve both the $\widetilde{O}( s_0^{\frac{2}{3}}  T^{\frac{2}{3}} ) $ regret which is optimal in the data-poor regime, and the $\widetilde{O}( \sqrt{s_0 T} ) $  regret which is optimal in the data-rich regime, and thus enjoys the so-called ``the best of two worlds''.

\begin{algorithm}[ht]
\caption{High Dimensional Bandit by Online HT}
\label{alg:Bandit}
\begin{algorithmic}[1]
\STATE $\epsilon$-greedy sampling probability $\epsilon_t$ for each $t$. $\bmu^{\mathsf{s} }_{a,0}=0$, step size $\beta_t$.
\FOR {$t=1,..., T$}

\STATE Observe the feature $\bx_t$.
%construct UCB estimate for the expected reward $u_{a,t}\in[0,1]$ and the LCB estimate for the resource consumption vector, $\bm{L}_{a,t}\in[0,1]^m$.
\STATE Sample a random variable $\nu_t\sim \operatorname{Ber}(K \epsilon_t) $.
\STATE {Pull the arm $y_t$ with $\epsilon_t$-greedy strategy defined as follows:
\[
y_t=\left\{\begin{aligned}
   & \arg\max_{a\in[K] }  \left\langle \bx_t,\bmu^{\mathsf{s} }_{a,t-1} \right\rangle, & \text{if~} \nu_t=0\\
   &a, ~\text{w.p.~} 1/K \text{~for~each~}a\in[K]  &\text{~if~}\nu_t=1
\end{aligned}
\right.
\]

 and receive a reward $r_t$.}
\STATE {For each $a\in[K]$, update the sparse estimate $\bmu_{a,t}^{\mathsf{s}}$ by \Cref{alg:online-iht} with each $p_{a,t}=(1-K\epsilon_t )y_{a,t}+\epsilon_t $}
\ENDFOR
\end{algorithmic}
\end{algorithm}
\begin{theorem}\label{thm:bandit-poor}
Let $R_{\max}=\sup \abs{\left\langle \bx_t, \bmu_{a}^{\star} \right\rangle} $. Choosing $\epsilon_t=\sigma^{\frac{2}{3}} D^{\frac{4}{3}} s_0^{\frac{2}{3}} (\log (dK))^{\frac{1}{3}}t^{-\frac{1}{3}} / \left(R_{\max} K\right)^{\frac{2}{3}} \wedge 1/K$, our   Algorithm \ref{alg:Bandit} incurs the regret 
\begin{equation*}
\begin{aligned}
        \operatorname{Regret}(\pi)
        & \lesssim \frac{ R_{\max}^{\frac{1}{3}} K^{\frac{1}{3}} \sigma^{\frac{2}{3}} D^{\frac{4}{3}} s_0^{\frac{2}{3}}  T^{\frac{2}{3}} (\log (dK))^{\frac{1}{3}}   }{\phi_{\min}(s)^{\frac{2}{3}} }
\end{aligned}
\end{equation*}
    %  & = \bE\left[\sum_{t=1}^{T} \left\langle \bx_t, \bmu_{\mathsf{opt} }(\bx_t) \right\rangle - \sum_{t=1}^{T} \left\langle \bx_t, \bmu_{y_t }^\star \right\rangle \right] \\
\end{theorem}
Theorem \ref{thm:bandit-poor} states the optimality of our high-dimensional bandit algorithm under minimal assumptions, which matches the $\Omega\left(\phi_{\min}^{-2 / 3} s_0^{2 / 3} T^{2 / 3}\right)$ lower bound \citep{jang2022popart} in the data-poor regime $d\ge T^{\frac{1}{3}}s_0^{\frac{4}{3}}$. We further show that, we can use the same algorithm framework to achieve better regret given the diverse covariate condition, which will match the regret lower bound for data-rich regimes. We present our result in \Cref{thm:bandit-rich}.

\begin{theorem}\label{thm:bandit-rich}
    
Suppose $\bx_t$ is further sparse marginal sub-Gaussian: 
$$\bE\exp ( \bu^\top \bx_t )\le \exp(c \phi_{\max}(s_0) \norm{\bu}_2^2 /2), $$
for  any $2s_0$-sparse vector $\bu$. Assume the following diverse covariate condition \citep{ren2023dynamic} holds: There are positive constants $\gamma(K)$ and $\zeta(K)$, such that for  any unit vector $\bv \in \mathbb{R}^d$,  and any $a \in[K]$, there is 
$$\mathbb{P}\left( \bv^{\top} x_t x_t^{\top} \bv\cdot \bI\left\{a^\star_t=a \right\} \geq\gamma(K)\middle| \cH_{t-1} \right) \geq \zeta(K),$$
where $a^\star_t = \max_{a\in[K] } \left\langle \bx_t , {\bmu}_{a,t-1}^{ \mathsf{s} }  \right\rangle$ is selected greedily. Denote $\kappa_1 = \frac{\phi_{\max}(s) }{\gamma(K)\zeta(K)}$. Setting $\epsilon_t=0$, we have the following regret bound for Algorithm \ref{alg:Bandit}:
\begin{equation*}
    \operatorname{Regret}(\pi)
    \lesssim
    \frac{
    \left(
    \kappa_1\log K
    \wedge
    \frac{s_0D^2}{\gamma(K)\zeta(K)}
    \right)^{1/2}
    \sigma D\sqrt{s_0\log(dK)T}
    }
    {\sqrt{\gamma(K)\zeta(K)}}.
\end{equation*}

\end{theorem} 

The regret of our bandit algorithm indeed matches the known lower bound of high-dimensional bandit problems $\Omega(\sqrt{s_0 T})$  \citep{chu2011contextual,ren2023dynamic}.  Compared with previous LASSO-based frameworks, no additional assumption on the range of arms (e.g., $\ell_2$-norm bound of $\bmu_a^\star$ \citep{ren2023dynamic}) or the minimum signal strength \citep{hao2020high,jang2022popart} is needed for our algorithm to achieve the optimal regret in the data-rich regime, as long as the diverse covariate condition holds. The sparse marginal sub-Gaussian assumption here is used to yield a more precise characterization of errors w.r.t $s_0$. If without this assumption, there will be no $\kappa_1$ term in the regret bound.

\section{Logarithmic Regret of CBwK by Resolving}\label{sec:resolving}
 Although the results of $\widetilde{O}(s_0^{\frac{2}{3}} T^{\frac{2}{3}} )$ and $\widetilde{O}(\sqrt{s_0 T})$ regrets in  Section \ref{sec:CBwK-regret} and \ref{sec:regret-sqrt} exhibit logarithmic dependence on the dimension of features $d$,  they still suffer from the polynomial dependence on the total time $T$. However, as is pointed out by a stream of studies in online allocation problems \citep{li2022online,balseiro2023survey,jiang2022degeneracy,ma2024optimal,bray2024logarithmic}, one can expect $O(\log T)$ regret under various conditions by using the re-solving heuristic. Thus, it is natural to ask whether it is possible to achieve logarithmic regret for both $d$ and $T$ in CBwK problem.

To achieve this strong logarithmic regret, a key challenge is tackling the unknown reward under bandit feedback. Unlike the settings in \citep{li2022online,balseiro2023survey, jiang2022degeneracy, ma2024optimal}, where the expected reward for each action is known prior to decision-making, the reward-agnostic bandit framework requires learning the rewards/parameters, thereby introducing additional errors in both decision-making and cumulative regret. Although \cite{bray2024logarithmic} included a brief study of online allocation with unknown rewards, its analysis is strictly confined to a simple multi-secretary problem without any contextual information. In the CBwK problem, however, the impact of these learning-induced errors remains largely unexplored, particularly in the presence of high-dimensional features.

Two studies are particularly relevant for developing resolving heuristics for CBwK problems. \cite{vera2021online} analyze CBwK through Bellman inequalities and, via resolving, obtain an $O(\log T)$ regret bound; however, their analysis relies on an arm gap condition and assumes a one-dimensional context (arrival types). More recently, \cite{chen2024contextual} address CBwK with a resolving scheme based on distribution estimation, yet the randomness in their primal problem is driven solely by an external factor, which simplifies learning. Furthermore, the density estimator used in \cite{chen2024contextual} is sub-optimal for linear bandit problems with multiple dimensions.

% To carefully handle such errors and control the regret, a better dual performance in the primal-dual algorithm is necessary. To this end, we re-cast the dual problem: 
% to the unknown exact primal information

To study the resolving heuristic in our high-dimensional CBwK problem, we denote the initial average capacity as $\brho_0={\bC}/{T}$. In this section, we assume that the capacity $\bC$ grows linearly with $T$ such that  the initial average capacity satisfies $\rhomin\le\rho_{0,i} \le\rhomax $, $\forall i\in[m]$ for some $\rhomin,\rhomax>0$. Denote a  neighborhood of $\brho_0$ as $\cB(\brho_0,\delta_\rho)=\bigotimes_{i=1}^m [\rho_{0,i}-\delta_{\rho},\rho_{0,i}+\delta_\rho]$, where $\delta_\rho< 0.5\rhomin $.

Denote the population version of the dual problem as
\begin{equation*}
     D(\blambda,\brho) =  \bE\left[\max_{\bm{y}_t(\bx_t)\in\Delta^K}\big\{  \sum_{a\in[K]}(\bmu_a^\star)^\top \bx_t\cdot y_{a,t}(\bx_t) -   (\bm{W}_a^{\star} \bx_t)^{\top}\blambda\cdot y_{a,t}(\bx_t)\big\} \right]+ \brho^\top \blambda.
\end{equation*}
We have $L(\blambda)=T  \cdot D(\blambda,\brho_0)$. 
We write the optimal $\blambda$ that minimizes $ D(\blambda,\brho)$ for each $\brho$ as $\blambda^*(\brho)=\arg\min_{\blambda\succeq \bm{0}} D(\blambda,\brho)$. The optimal dual variable $\blambda^*$ in previous sections is thus $\blambda^*=\blambda^*(\brho_0)$. Denote $a^*_t(\blambda)\in[K]\cup\{0\}$ as the optimal arm such that 
\begin{equation*}\begin{aligned}
 a^*_t(\blambda) =  \arg\max_{a\in[K]\cup\{0\}} \left\langle \bmu_{a}^\star - (\bm{W}_{a}^{\star})^{\top}\blambda,\bx_t \right\rangle.
\end{aligned}
\end{equation*}

We now assume the following local regularity conditions of the dual problems
when $\brho$ changes within a small neighborhood of the initial average
capacities $\brho_0$.

\begin{assumption}[Local dual diversity and margin conditions]
\label{asm:local-dual-conditions}
The following two conditions hold.

\begin{enumerate}[label=(\roman*)]
    \item \textit{Local dual diverse covariate condition.}
    There are (possibly $K$-dependent) positive constants $\gamma(K)$ and
    $\zeta(K)$, such that for any unit vector $\bv \in \mathbb{R}^d$,
    $\norm{\bv}_2=1$, and any $a \in[K]$, there is
    $$
    \mathbb{P}\left(
    \bv^{\top} \left( \bx_t \bx_t^{\top} \right) \bv
    \cdot
    \bI \left\{ a  = a^*_t(\blambda^*(\brho) )  \right\}
    \geq\gamma(K)
    \right) \geq \zeta(K)
    $$
    for any $\brho\in \cB(\brho_0,\delta_\rho)$, where
    $a^*_t(\blambda^*(\brho)) =
    \arg\max_{a\in[K]\cup\{0\}}
    \left\langle
    \bmu_{a}^\star - (\bm{W}_{a}^{\star})^{\top}\blambda^*(\brho),
    \bx_t
    \right\rangle$.

    \item \textit{Local dual margin condition.}
    For any $0<\delta\le R_{\max}$, there exists a constant $M_{\sfm }>0$
    such that
    \begin{equation}\label{eq:local-margin}
       \bp\left(
       \left\langle
       \bmu_{a^*_t(\blambda) }^\star
       -(\bm{W}_{a^*_t(\blambda)}^{\star})^{\top}\blambda,
       \bx_t
       \right\rangle
       -
       \max_{a\neq a^*_t(\blambda)}
       \left\langle
       \bmu_{a}^\star
       -(\bm{W}_{a}^{\star})^{\top}\blambda,
       \bx_t
       \right\rangle
       \le \delta
       \right)
       \le M_{\sfm } \delta,
    \end{equation}
    for any
    $\blambda\in
    \{\blambda\succeq \bm 0:
    \norm{\blambda-\blambda^*(\brho)}_{1}\le\delta_{\lambda},
    \text{ where }\brho\in \cB(\brho_0,\delta_\rho)\}$.
\end{enumerate}
\end{assumption}

\begin{remark}
Assumption~\ref{asm:local-dual-conditions} consists of local variants of the
diverse covariate condition (Assumption~\ref{asm:diverse-cov}) and the margin
condition (Assumption~\ref{asm:margin-static}) mentioned in
Section~\ref{sec:CBwK-regret}. The local dual diverse covariate condition is
weaker than Assumption~\ref{asm:diverse-cov} in the sense that the maximum arm
is taken with respect to a group of fixed dual functions
$\left\langle \bmu_{a}^\star -
(\bm{W}_{a}^{\star})^{\top}\blambda^*(\brho),\bx_t \right\rangle$, rather than
dual functions with stochastic estimations. It can also be satisfied by
\eqref{eq:div-cov-K2} and the conditions listed in
\cite{bastani2021mostly}. The local dual margin condition is a local version of
Assumption~\ref{asm:margin-static} in the sense that it requires the dual value
to be well separated from the sub-optimal Lagrangian for $\blambda$ that is in
a local region around optimal $\blambda^*(\brho)$, for a group of $\brho$ close
to $\brho_0$. Under the second-order growth condition (which will be described
in Assumption~\ref{asm:dual-regularity-log}), we can show that the dual
optimal variable $\blambda^*(\brho)$ changes smoothly with respect to $\brho$.
Thus, a sufficient condition for the local dual margin condition is to assume
that \eqref{eq:local-margin} holds for a group of $\blambda$ that are close to
$\blambda^*$.
\end{remark}

Now, we denote the partial derivative of a function with respect to the argument
$\blambda$ as $\nabla_{\blambda}$ (if the function is not differentiable at some
point, we can use its sub-gradient instead). We reveal that the local dual
margin condition actually implies the smoothness condition of the expected dual
problem, which we detail in Lemma~\ref{lemma:smoothness}.
 \begin{lemma}[Margin condition implies smoothness]\label{lemma:smoothness}
 Under Assumption~\ref{assump:general} and
 Assumption~\ref{asm:local-dual-conditions}, we have
    \begin{equation}
        \norm{\nabla_{\blambda} D(\blambda,\brho)
        - \nabla_{\blambda} D(\blambda^*(\brho),\brho)}_{\infty}
        \le4B_{\max}^2 M_{\sfm}
        \norm{\blambda^*-\blambda}_1,
    \end{equation}
    for any
    $\blambda\in\{\blambda\succeq \bm 0:
    \norm{\blambda-\blambda^*(\brho)}_{1}\le\delta_{\lambda},
    \text{ where }\brho\in \cB(\brho_0,\delta_\rho)\}$.
    \end{lemma}
    
To establish $O(\log T)$ regret for online allocation, we impose the regularity conditions, including both locally second-order growth condition and non-degeneracy condition on the dual problem following \cite{li2022online,balseiro2023survey,ma2024optimal,bray2024logarithmic}.
    
% \cap 2\frac{\VU}{C_{\min}} \Delta^m
\begin{assumption}[Regularity conditions on dual problem]\label{asm:dual-regularity-log}
We assume the following conditions on the population version of dual problem $ D(\blambda,\brho)$:
\begin{enumerate}[label=5.\arabic*,ref= 5.\arabic*]
    \item (locally second order growth)   Given any $\brho\in \cB(\brho_0,\delta_\rho)$, the dual problem $D(\blambda,\brho)$ shares the second order growth for the argument $\blambda$ around $\blambda^*(\brho)$: 
    \begin{equation*}
        \left\langle \nabla_{\blambda} D(\blambda,\brho) - \nabla_{\blambda} D(\blambda^*(\brho),\brho), \blambda - \blambda^*(\brho) \right\rangle \ge  \cL \norm{\blambda - \blambda^*(\brho)}_1^2,
    \end{equation*}
    for $\blambda\in\{\blambda \succeq \bm 0:\norm{\blambda-\blambda^*(\brho)}_{1}\le\delta_{\lambda}\}$. \label{asm:dual-regularity-log-1}
    
    \item (non-degenerate) For $i\in[m]$,  the optimal dual variable $\blambda^*(\brho_0)$ satisfies  $\bE\big(\brho_{0}-\bW_{a^*_t(\blambda^*(\brho_0))}^{\star}\bx_t\big)_i =0$ if and only if $\lambda^*_i(\brho_0)>0$. \label{asm:dual-regularity-log-2} 
\end{enumerate}
    % \bE \left\langle  (\bm{W}_{a^*_t(\brho_0)}^{\star} -\bm{W}_{a^*_t(\brho)}^{\star}) \bx_t, \brho - \brho_0 \right\rangle \ge  \cL \norm{\blambda^ - \brho_0}_2
\end{assumption}

Define the binding/non-binding dimensions with respect to the initial $\brho_0$ as 
$$I_{\sfB}=\left\{i\in[m]:\bE\big(\brho_{0}-\bW_{a^*_t(\blambda^*(\brho_0))}^{\star}\bx_t\big)_i=0\right\}, \quad I_{\NB}=\left\{i\in[m]:\bE\big(\brho_{0}-\bW_{ a^*_t(\blambda^*(\brho_0)) }^{\star}\bx_t\big)_i>0\right\}.$$

Then, Assumption \ref{asm:dual-regularity-log-2} imposes strong complementary slackness on the dual problem under initial average capacity.  Denote the gaps in binding and non-binding dimensions as $\delta_{\sfB}=\min_{i\in I_{\sfB}}\lambda^*_i(\brho_0)>0$, $\delta_{\NB}=\min_{i\in I_{\NB}}\bE\big(\brho_{0}-\bW_{a^*_t(\blambda^*(\brho_0))}^{\star}\bx_t\big)_i>0$.
% gap between the fluid benchmark and offline maximum to be well controlled.
The following lemma shows that when $\brho$ changes within a certain region around $\brho_0$, the binding and non-binding dimensions remain unchanged.
\begin{lemma}[\cite{ma2024optimal}]\label{lemma:region-B+} Under Assumptions \ref{asm:dual-regularity-log}, 
    there exists a constant $\delta_{\rho}'= \frac{\cL\delta_{\sfB }}{2}\wedge\frac{\cL \delta_{\NB}}{16 B_{\max}^2 M_{\sfm}} \wedge \delta_{\rho}$ such that for any $\brho$ in
$$
  \cB^+(\brho_0,\delta_\rho') := \left\{\brho:\rho_i\in [\rho_{0,i}-\delta_{\rho}',\rho_{0,i}+\delta_\rho'] \text{ if } i\in I_{\sfB}, \  \rho_i\ge\rho_{0,i}-\delta_{\rho}' \text{ if } i\in I_{\NB}\right\}, 
$$
 the dual problems $D(\blambda^*, \brho_0)$ and $D(\blambda^*(\brho), \brho)$ share the same binding and non-binding dimensions. 
\end{lemma}

Notice that, in the non-binding dimensions, enlarging the resource constraints will not change the optimal dual solution $\blambda^*(\brho)$. Therefore, it is convenient to consider the region $\cB^+(\brho_0,\delta_\rho')$ for further studies as elements in $\cB(\brho_0,\delta_\rho')$ and $\cB^+(\brho_0,\delta_\rho')$ share the same optimal dual solution region.
% =\{\brho:\rho_i\in [\rho_{0,i}-\delta_{\rho}',\rho_{0,i}+\delta_\rho'] \text{ if } i\in I_{\sfB}, \  \rho_i\ge\rho_{0,i}-\delta_{\rho}' \text{ if } i\in I_{\NB} \}

The non-degenerate condition is imposed as we compare the algorithm performance with $\VU$. Indeed, \cite{bumpensanti2020re,vera2021bayesian} shows that if the problem is degenerate, then there exists a hard instance that incurs substantial $\Omega(\sqrt{T})$ gap between the fluid relaxation $\VU$ and the optimal hindsight reward, making the $O(\log T)$ regret unattainable.
We now describe our primal-dual algorithm with resolving in Algorithm \ref{alg:LagrangianBwK-resolve}. Then, we develop theories for dual convergence and regret control with the presence of online HT sparse estimation.

\begin{algorithm}[ht!]
\caption{Primal-Dual High-dimensional CBwK by Resolving}
\label{alg:LagrangianBwK-resolve}
\begin{algorithmic}[1]
\STATE {\bfseries Input:} reward-constraint ratio $Z$, initial period $T_0$, $\brho_0=\bC/T$. $ \bmu_{a,0}^{\mathsf{s}}=\bm{0}$, $\widehat{\bW}_{a,0}=\bm{0}$, $\widehat{\blambda}_{t-1}=\bm{0}$.
\FOR {$t=1,\dots, T$}

\STATE {Observe the feature $\bx_t$.}
\STATE  Estimate $\operatorname{EstCost}(a)=\bx_t^\top \widehat{\bW}_{a,t-1}^\top\widehat{\blambda}_{t-1}$ for each arm $a\in[K]$.

\IF{$1\le t \le T_0$}
\STATE{   Initialization by pulling each arm uniformly with $\bp( y_t=a)=\frac{1}{K}$,  }
\ELSIF{$t \ge T_0+1$}
\STATE{$y_t=\widetilde{y}_t:=\displaystyle\arg\max_{a\in[K]} \{(\bmu_{a,t-1}^{\mathsf{s}} )^\top\bx_t-\operatorname{EstCost}(a)\} = \displaystyle\arg\max_{a\in[K]} \left\{ \left\langle\bmu_{a,t-1}^{\mathsf{s}} -\widehat{\bW}_{a,t-1}^\top\widehat{\blambda}_{t-1},\bx_t \right\rangle \right\} $}
\ENDIF
% \STATE {and use Algorithm \ref{alg:online-iht} to update primal estimation} % and 

%construct UCB estimate for the expected reward $u_{a,t}\in[0,1]$ and the LCB estimate for the resource consumption vector, $\bm{L}_{a,t}\in[0,1]^m$.

% \STATE and pull the arm $y_t$ defined as follows: % where  $ p_t=K\epsilon_t$
% \[
% y_t=\left\{\begin{aligned}
%    & ,  \text{ if } \nu_t=0\\
%    &a, ~~~~\text{w.p.~} 1/K \text{~for~each~arm~}a\in[K]  \text{ if }\nu_t=1. 
% \end{aligned}
% \right.
% \]

\STATE  Receive $r_t$ and $\bb_t(y_t,\bx_t)$. If one of the constraints is violated, then EXIT.
\STATE {Update average remaining resource: 
\begin{equation*}
    \brho_{t}=\left((T-t+1)\brho_{t-1}- \bb_t(y_t,\bx_t)\right)/(T-t)
\end{equation*}
}

\STATE  Update $\widehat{\blambda }_t$ by solving the programming:
\begin{equation}\label{eq:dual-resolve}
    \widehat{\blambda }_t = \arg\min_{\blambda \succeq \bm{0}, \norm{\blambda}_1\le Z} \bar{D}_t(\bmu_{a,t}^{\mathsf{s} },\widehat{\bm{W}}_{a},\blambda,\brho_{t}) : = \frac{1}{t}\sum_{j=1}^t \max_{a\in[K]\cup\{0\}}\left\langle \bmu_{a,t}^{\mathsf{s} } - (\widehat{\bm{W}}_{a})^{\top}\blambda,\bx_j \right\rangle + \brho_t^\top \blambda
\end{equation}

\STATE  For each arm $a\in[K]$, update the estimate $\bmu_{a,t}^{\mathsf{s} } $ and $\widehat{\bW}_{a,t}$ from \Cref{alg:online-iht} by letting 
\begin{equation*}
    p_{a,t}= y_{a,t}= \mathbbm{1}\{y_t=a\}, \quad \forall a\in[K].
\end{equation*}
\ENDFOR
\end{algorithmic}
\end{algorithm}

Our approach to establishing dual convergence and regret control departs greatly
from previous work
\citep{li2022online,balseiro2023survey,jiang2022degeneracy,ma2024optimal,bray2024logarithmic}
in the sense that the primal estimation and dual optimization are integrated
organically, making the dual convergence theories in the prior work fail. In our
framework, these two components are intrinsically linked: the primal update
relies on the accuracy of the dual variables, while the dual convergence obtained
through resolving, in turn, depends on the current primal solution. To address
this mutual dependence, we establish in the supplement a dual convergence result
that holds for any primal estimator derived from historical data, and then invoke
it inductively to obtain the primal estimation error rate under resolving. In
particular, when $\{\brho_j\}_{j=1}^{t}$ remain within
$\cB^+(\brho_0,\delta_\rho')$, the primal estimator satisfies
\begin{equation}\label{eq:primal-log}
    \max_{a}\norm{\bmu_{a,t}^{ \mathsf{s} }-\bmu_a^\star}_2^2
\vee
\norm{\widehat{\bW}_{a,t}-\bW_a^\star}_{2,\max}^2
\lesssim
\frac{\sigma^2D^2s_0\log(mKdT)}
{\gamma^2(K)\zeta^2(K)t},
\end{equation}
with high probability; see Theorem~\ref{thm:est-dual-convergence}, \ref{thm:est-induction} in the supplement for more details.

Now, exploiting the dual convergence and primal estimation rates established in
the supplement, we can prove that our resolving algorithm achieves
$O({\log(mKdT)}\log T)$ regret.

% Notice that the primal estimation is updated following an online fashion, meaning that its newly collected gradient is always independent from the old $\brho_{t-1}$. Therefor, the rate in always valid as long as $\{\brho_{j}\}_{j=1}^{t-1}$ is within $\cB^+(\brho_0,\delta_{\rho}')$ and the algorithm does not stop at $t$.

\begin{theorem}\label{thm:resolving-regret}
Under Assumptions~\ref{assump:general},
\ref{asm:local-dual-conditions}, and
\ref{asm:dual-regularity-log}, our
Algorithm~\ref{alg:LagrangianBwK-resolve} with $T\gtrsim Z^2\log(mKdT)$ achieves the regret:
\begin{equation*}
    \operatorname{Regret}(\pi)  \lesssim
    \frac{
    B_{\max}^6 \widetilde{B}_{\max} M_{\sfm }^4 D^4Z^2
    \sigma^2 s_0^2 m^2
    {\log(mKdT)} \log T}
    {\cL^2 \delta'^4_{\rho} \gamma^2(K)\zeta^2(K)}
    =
    O(s_0^2m^2{\log(mKdT)}\log T).
\end{equation*}
\end{theorem}

\begin{remark}[Dependence on $m$]
Our algorithm exhibits an $O(m^2)$ dependence on the number of resources, which
matches previous resolving-based online allocation results
\citep{li2022online,ma2024optimal}, even in the presence of unknown reward and
resource consumptions. The two factors of $m$ come from different sources. One
factor is due to the large parameter space of the unknown resource-consumption
matrices $\{\bW_a^\star\}_{a\in[K]}$: under row-wise sparsity, their total degree
of freedom is of order $s_0mK$. If $\{\bW_a^\star\}_{a\in[K]}$ are known, this
part of the parameter space is removed, and the regret bound can be reduced to
$O(s_0^2m\log(KdT)\log T)$.

The remaining factor of $m$ comes from the dual convergence required by the
resolving step. Unlike the Hedge-based update in
Theorem~\ref{thm:Primaldual}, which incurs only logarithmic dependence on $m$,
resolving requires the empirical dual optimizer $\widehat{\blambda}_t$ to
converge to the population dual optimizer $\blambda^*$. Since
$\blambda^*$ is an $m$-dimensional parameter, this dual-convergence step
typically incurs dimension dependence $\|\widehat{\blambda}_t-\blambda^*\|_2^2
    \lesssim \frac{m}{t}$,
up to logarithmic and problem-dependent factors. This type of dependence also
appears in prior resolving-based analyses
\citep{li2022online,jiang2022degeneracy,ma2024optimal} and is difficult to
remove under a dual-convergence-based proof.
\end{remark}

\begin{remark}[Inexact resolving] Our analysis on the non-degenerate dual problem grants us the feasibility to handle inexact resolving when optimizing \eqref{eq:dual-resolve}. Even when the programming \eqref{eq:dual-resolve} is not exactly solved, we can still control the regret as long as the optimization error is small. Similar to \cite{ma2024optimal}, we can show that a free $O\big(\frac{1}{t}\big)$ optimization error on the dual objective function is admitted at each step to still achieve $\widetilde{O}(\log T)$ regret.
    
\end{remark}

\begin{remark} We establish new conditions for CBwK problems that yield regret below $O(\sqrt{T})$ rate. In contrast, earlier studies that achieve $\widetilde{O}(\log T)$ regret rely on different assumptions. For instance, \cite{sankararaman2021bandits} requires a single resource constraint together with a unique best‐arm condition, while \cite{vera2021online} adopts a similar optimal‐arm‐gap assumption. \cite{chen2024contextual} assumes a unique, non‐degenerate primal problem, which is stricter than our dual non-degenerate condition. Their studies focus on a simplified model in which the randomness of an arrival is determined by an observable external factor. Therefore, our conditions and analyses provide whole new insights into logarithmic regret control of CBwK. 
%It is worth mentioning that for the contextual bandit problem without knapsack, diverse covariate and margin conditions can also lead to a $O(\log T)$ regret, as is shown in \cite{bastani2021mostly} for low-dimensional case and \cite{duan2024regret} for high-dimensional case.

\end{remark}

\section{Numerical Experiments}\label{sce:num-exp}

\subsection{Sparse recovery}\label{sec:sparse-rec}

\begin{figure}[!ht]
\centering
\includegraphics[width=\textwidth]{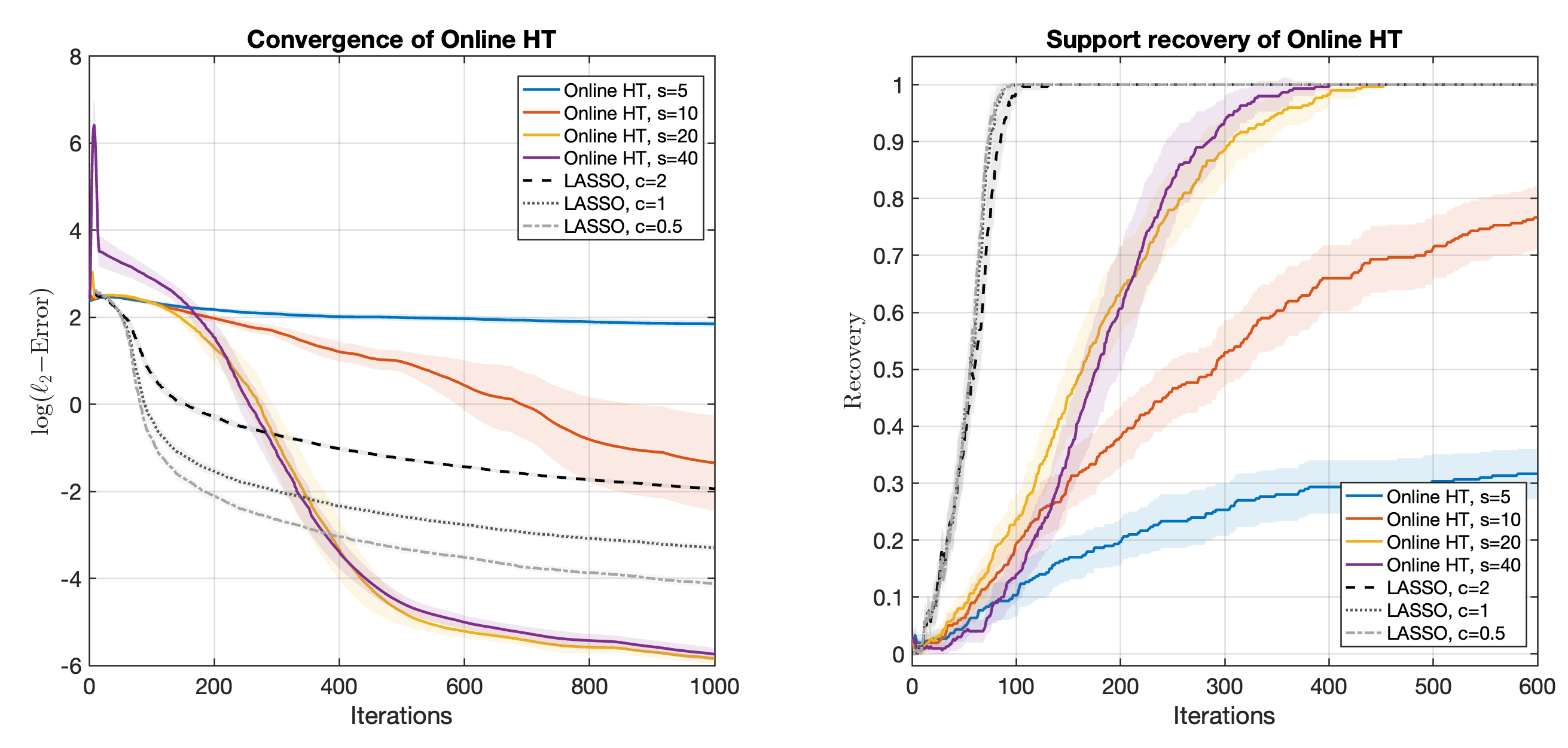}
\caption{Primal performance of Online HT vs LASSO. The true sparsity level is
fixed at $s_0=10$, while $s$ denotes the working sparsity level used by Online
HT. Shaded regions represent approximate $95\%$ confidence bands.}
\label{fig:primal}
\end{figure}

We first examine the feasibility of our primal algorithm in the sparse recovery
problem. To check the performance of Algorithm~\ref{alg:online-iht}, suppose
now we only consider one arm $\bmu_{\star}$, and we want to estimate it in an
online process. To this end, we always choose $y_t=1$ and thus $p_t=1$. At each
$t$, we measure the sparse estimation error
$\norm{\bmu_t^{\mathsf{s}}-\bmu_{\star}}_2^2$, and the support recovery rate
$\abs{\operatorname{supp}(\bmu_t^{\mathsf{s}})\cap \Omega_{\star}}/s_0$, which
indicates the ratio of the support set we have detected. The result is presented
in Figure~\ref{fig:primal}. Here we set $d=1000$, $s_0=10$, $\sigma=0.5$, and
$\bSigma$ to be the power decaying covariance matrix
$\bSigma_{ij}=\alpha^{\abs{i-j}}$, where $\alpha=0.5$.

We compare Online HT with LASSO, a prevalent method used in online
high-dimensional bandit problems
\citep{kim2019doubly,hao2020high,ren2023dynamic}. In this experiment, the true
sparsity level is fixed at $s_0=10$, while the working sparsity level $s$ used
by Online HT varies over $s\in\{5,10,20,40\}$. This design illustrates the
information-preserving role of the working sparsity level. When $s<s_0$, the
hard-thresholding step necessarily discards part of the true support, leading to
information loss and poor support recovery. When $s\ge s_0$, the true support
can be retained by the thresholding operator, and the estimator achieves fast
error decay and accurate support recovery. For LASSO, we select the
regularization level $\lambda=c\cdot \sqrt{\log(dt)/t}$ with
$c\in\{2,1,0.5\}$. The shaded regions in Figure~\ref{fig:primal} represent
approximate $95\%$ confidence bands over independent trials.

The results show that Online HT is robust to moderate overspecification of the
working sparsity level: the choices $s=20$ and $s=40$ achieve fast
decrease in estimation error and nearly exact support recovery. In contrast,
underspecifying the sparsity level, e.g., $s=5$, leads to inferior support
recovery because the algorithm is forced to keep fewer coordinates than the true
support size. Compared with LASSO, Online HT achieves competitive or better
estimation error while maintaining exact sparsity through hard thresholding.
Moreover, Online HT has efficient computational cost: its update costs
$O(d^2)$ per iteration and $O(d^2T)$ in total, while classical LASSO solutions
cost $O(d^3+d^2t)$ per iteration \citep{efron2004least}, and
$O(d^3T+d^2T^2)$ in total if the estimator is updated at every time step, as in
\cite{kim2019doubly,ren2023dynamic}. This computational advantage is important
in online decision-making, where the estimator must be updated sequentially.

\subsection{Online bandit problem}

\begin{figure}[!ht]
\centering
\includegraphics[width=\textwidth]{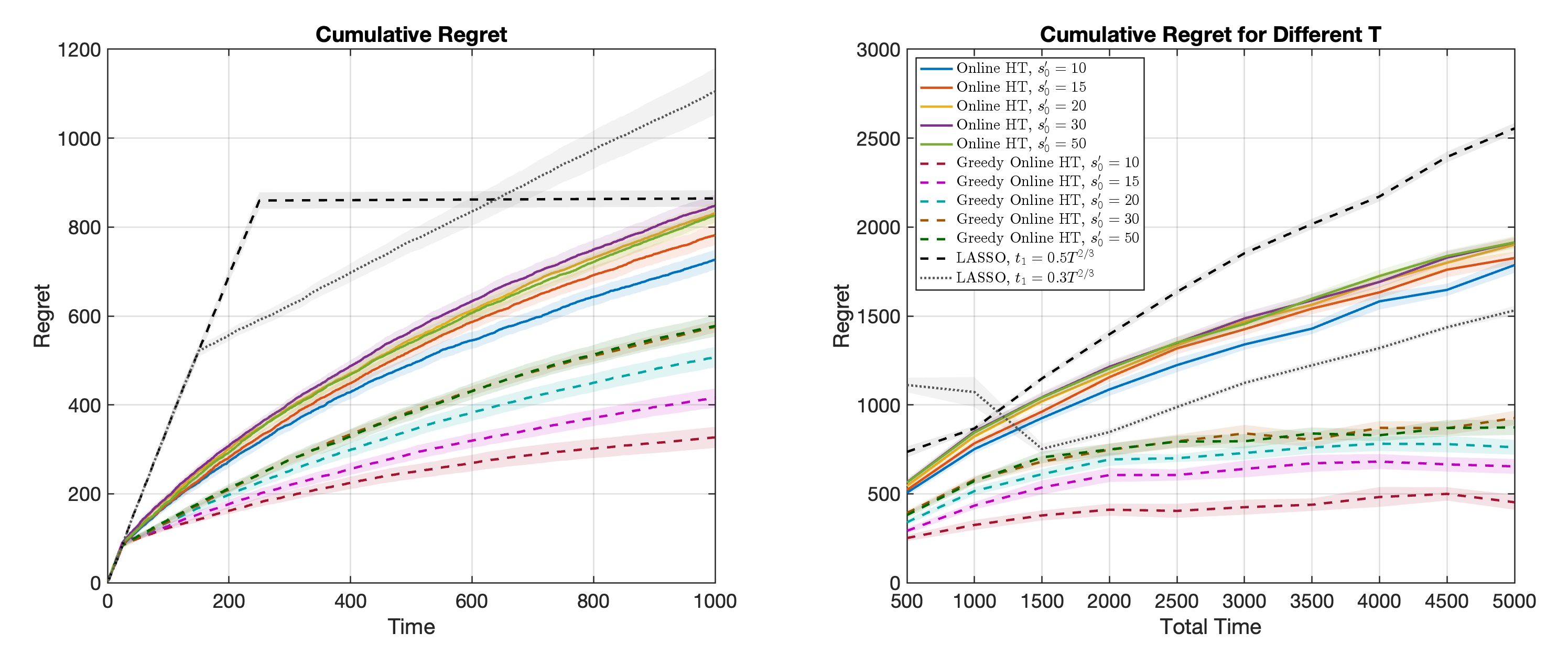}
\caption{Regret of Online HT, greedy Online HT, and LASSO bandit. The label
$s_0'$ denotes the working sparsity level used by Online HT, while the true
sparsity level is $s_0=10$. Shaded regions represent approximate $95\%$
confidence bands.}
\label{fig:bandit-reg}
\end{figure}

We then apply Algorithm~\ref{alg:Bandit} to the high-dimensional linear bandit
problem. We choose $d=100$, $s_0=10$, and $K=5$. The covariates are generated
as in Section~\ref{sec:sparse-rec}. We study both the regret accumulation for a
fixed horizon and the regret growth with respect to different horizons. The
result is presented in Figure~\ref{fig:bandit-reg}.

We compare the $\epsilon$-greedy Online HT method, the greedy Online HT method,
and LASSO bandit algorithms based on Explore-Then-Commit, as in
\cite{hao2020high,li2022simple,jang2022popart}. For Online HT and greedy
Online HT, we consider misspecified working sparsity levels
$s_0'\in\{10,15,20,30,50\}$, where $s_0=10$ is the true sparsity level. For the
LASSO bandit baselines, we use exploration lengths
$t_1=0.3T^{2/3}$ and $t_1=0.5T^{2/3}$. The shaded regions represent approximate
$95\%$ confidence bands over independent trials.

The results show that greedy Online HT achieves the smallest regret in this
simulation, consistent with the favorable diverse-covariate setting where
explicit exploration is less necessary. The $\epsilon$-greedy Online HT method
also substantially improves over the LASSO bandit baselines. The comparison
across $s_0'$ further shows that Online HT is stable under moderate
overspecification of the sparsity level, while the correct or mildly
overspecified choices tend to perform best. Overall, the experiment illustrates
that hard-thresholding-based online estimation can effectively exploit sparsity
in high-dimensional linear bandits.

\subsection{High-dimensional BwK}

We now focus on the linear BwK problem with high-dimensional sparse reward and
resource-consumption parameters. We compare Online HT, the resolving-based
Online HT algorithm, and the classic UCB-based linear BwK algorithm, i.e.,
linCBwK~\citep{agrawal2016linear}. Notice that, in
\cite{agrawal2016linear}, linCBwK is designed for the {\sffamily Model-C}
bandit problem, but it can be generalized to our {\sffamily Model-P} setting by
maintaining upper confidence regions for multiple arms simultaneously. The exact
formulas of the upper confidence regions can be found in
\cite{agrawal2016linear}.

\begin{figure}[!ht]
\centering
\includegraphics[width=\textwidth]{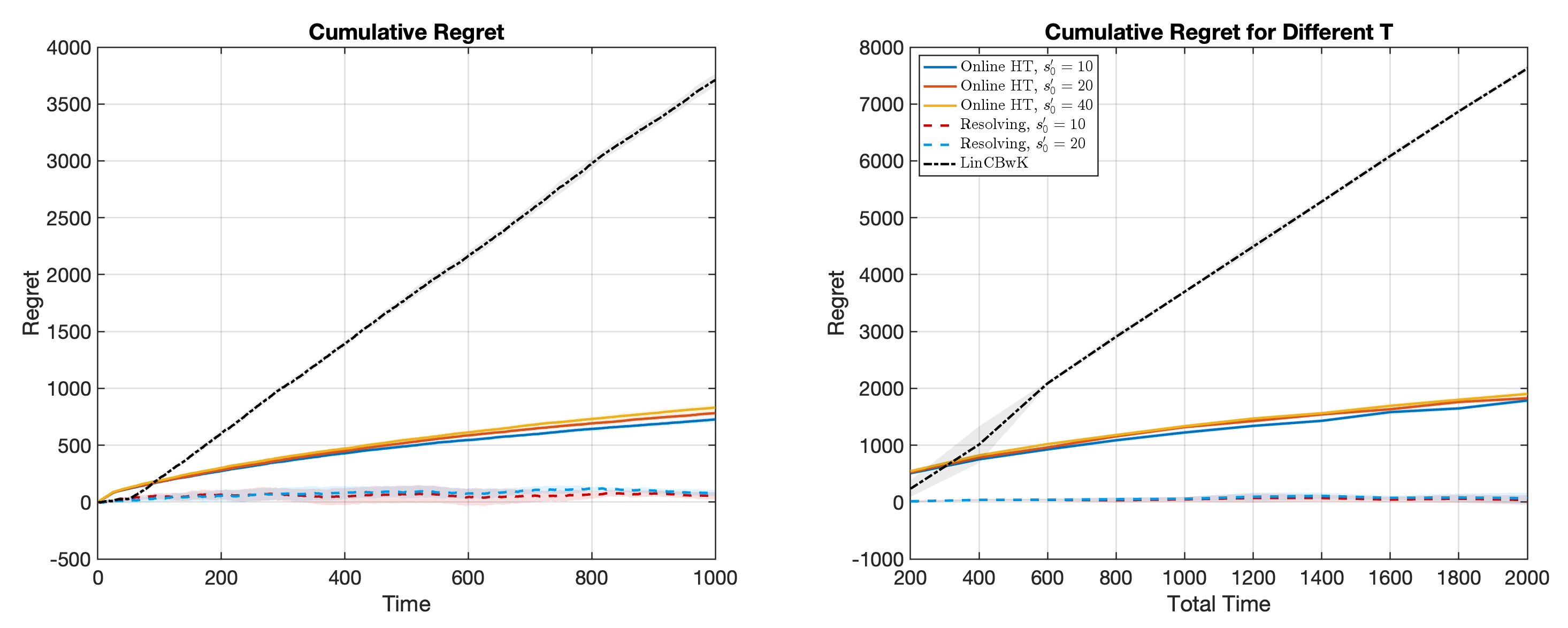}
\caption{Regret of Online HT, resolving-based Online HT, and linCBwK for the
high-dimensional CBwK problem. The label $s_0'$ denotes the working sparsity
level used by Online HT, while the true sparsity level is $s_0=10$. Shaded
regions represent approximate $95\%$ confidence bands.}
\label{fig:bwk-reg}
\end{figure}

We set $d=200$, $s_0=10$, $K=5$, and $m=10$. The covariates are generated
using the same power-decaying covariance matrix as in
Section~\ref{sec:sparse-rec}. The resource-consumption matrices
$\{\bW_a^\star\}_{a\in[K]}$ are generated row-wise sparse, with each row sharing
the same support as the corresponding reward parameter $\bmu_a^\star$. The
average capacities are generated uniformly. We report regret directly in
Figure~\ref{fig:bwk-reg}, rather than relative reward, to make the comparison
with the theoretical regret guarantees more transparent.

For Online HT, we examine the misspecified working sparsity levels
$s_0'\in\{10,20,40\}$. For resolving, which is computationally more expensive,
we use $s_0'\in\{10,20\}$. The shaded regions represent approximate $95\%$
confidence bands. The results show that linCBwK suffers from large regret in
this high-dimensional setting, which is consistent with its
$\widetilde O(d\sqrt{T})$ dependence. In contrast, the Online HT methods
substantially reduce regret by exploiting sparsity. Most notably, the resolving
method has much slower regret growth than the non-resolving Online HT baseline,
corroborating the logarithmic-regret theory $\widetilde{O}(\log T)$ developed in
Section~\ref{sec:resolving}. The comparison across $s_0'$ also shows that the
resolving method remains stable under moderate sparsity overspecification.

All experiments were run on an Apple M3 Max CPU. The sparse recovery experiment
uses $d=1000$, $s_0=10$, $T=1000$, and 50 independent trials. The online bandit
experiment uses $d=100$, $K=5$, $s_0=10$, $T=1000$, and 50 independent trials.
The high-dimensional BwK experiment uses $d=200$, $m=10$, $K=5$, $s_0=10$,
$T=1000$, and 50 independent trials with different working sparsity
levels. All experiments finished within minutes. This runtime is consistent
with the computational advantage of Online HT: each online update only requires
a gradient step followed by hard thresholding, avoiding the repeated solution of
a high-dimensional LASSO problem at every time step.

\section{Discussion on Assumption Validation}
\label{sec:discussion}

The improved regret bounds in Sections~\ref{sec:CBwK-regret}--\ref{sec:resolving}
rely on additional structural conditions, such as diverse covariates, margin
conditions, and local dual regularity. These assumptions should be viewed as
interpretable sufficient conditions for sharper regret guarantees rather than
conditions that must be exactly known before deployment. In particular, the
diverse-covariate condition is a distributional condition on the covariates and
can be implied by more primitive assumptions, such as Gaussian covariates or
covariates that are sufficiently spread over an isotropic ball
\citep{bastani2021mostly}. Hence, when historical or initial exploration data
are available, one can empirically examine whether the covariates are reasonably
well-spread, for example through the empirical covariance matrix or sparse
restricted eigenvalue diagnostics.

The margin condition is less directly verifiable before deployment because it
concerns the separation between the best and sub-optimal arms. However, the
corresponding hyperparameter dependence is mild: the margin condition mainly
affects logarithmic-level choices in the exploration schedule, and moderate
misspecification does not change the qualitative form of the algorithm. Moreover,
the resolving-based algorithm in Section~\ref{sec:resolving} further reduces
the tuning burden, since it does not require a pre-specified exploration
probability depending on whether the diverse-covariate or margin condition
holds; apart from the reward-constraint ratio $Z$, it is driven by the remaining
resource vector and the empirical dual resolving step. In practice, one may use
initial data to check covariate diversity and then choose the
diverse-covariate or resolving-based variant when the diagnostic is favorable,
or otherwise use the more conservative $\epsilon$-greedy version.

% Although in this paper we mainly focus on the case when the consumption $\bm{W}_a^\star$ for each arm is known, it is direct to generalize our results to the unknown $\bm{W}_a^\star$ by estimating them with Algorithm \ref{alg:online-iht}. Substituting $\bm{W}_a^\star$ with an estimated version $\widehat{\bm{W}}_a$ may incur additional estimation error, but this error can be generally controlled in a similar fashion to \Cref{thm:sparse-est}. As we proceed to discuss the consumption-agnostic instance, we will also posit that $\bm{W}_a^\star$ is row-wise sparse, a necessary assumption to render the problem tractable. The exploration of this particular aspect is earmarked for future work.

\newpage

\bibliographystyle{apalike}
\bibliography{myreferences}

@inproceedings{han2023optimal,
  title={Optimal Contextual Bandits with Knapsacks under Realizability via Regression Oracles},
  author={Han, Yuxuan and Zeng, Jialin and Wang, Yang and Xiang, Yang and Zhang, Jiheng},
  booktitle={International Conference on Artificial Intelligence and Statistics},
  pages={5011--5035},
  year={2023},
  organization={PMLR}
}

@article{freund1997decision,
  title={A decision-theoretic generalization of on-line learning and an application to boosting},
  author={Freund, Yoav and Schapire, Robert E},
  journal={Journal of computer and system sciences},
  volume={55},
  number={1},
  pages={119--139},
  year={1997},
  publisher={Elsevier}
}

@inproceedings{badanidiyuru2013bandits,
  title={Bandits with knapsacks},
  author={Badanidiyuru, Ashwinkumar and Kleinberg, Robert and Slivkins, Aleksandrs},
  booktitle={2013 IEEE 54th Annual Symposium on Foundations of Computer Science},
  pages={207--216},
  year={2013},
  organization={IEEE}
}

@article{jiang2020online,
  title={Online stochastic optimization with wasserstein-based nonstationarity},
  author={Jiang, Jiashuo and Li, Xiaocheng and Zhang, Jiawei},
  journal={Management Science},
  year={2025},
  publisher={INFORMS}
}

@article{agrawal2016linear,
  title={Linear contextual bandits with knapsacks},
  author={Agrawal, Shipra and Devanur, Nikhil},
  journal={Advances in Neural Information Processing Systems},
  volume={29},
  year={2016}
}

@article{hao2020high,
  title={High-dimensional sparse linear bandits},
  author={Hao, Botao and Lattimore, Tor and Wang, Mengdi},
  journal={Advances in Neural Information Processing Systems},
  volume={33},
  pages={10753--10763},
  year={2020}
}

@article{bastani2020online,
  title={Online decision making with high-dimensional covariates},
  author={Bastani, Hamsa and Bayati, Mohsen},
  journal={Operations Research},
  volume={68},
  number={1},
  pages={276--294},
  year={2020},
  publisher={INFORMS}
}

@article{kim2019doubly,
  title={Doubly-robust lasso bandit},
  author={Kim, Gi-Soo and Paik, Myunghee Cho},
  journal={Advances in Neural Information Processing Systems},
  volume={32},
  year={2019}
}

@inproceedings{li2022simple,
  title={A simple unified framework for high dimensional bandit problems},
  author={Li, Wenjie and Barik, Adarsh and Honorio, Jean},
  booktitle={International Conference on Machine Learning},
  pages={12619--12655},
  year={2022},
  organization={PMLR}
}

@article{jang2022popart,
  title={Popart: Efficient sparse regression and experimental design for optimal sparse linear bandits},
  author={Jang, Kyoungseok and Zhang, Chicheng and Jun, Kwang-Sung},
  journal={Advances in Neural Information Processing Systems},
  volume={35},
  pages={2102--2114},
  year={2022}
}

@article{blumensath2009iterative,
  title={Iterative hard thresholding for compressed sensing},
  author={Blumensath, Thomas and Davies, Mike E},
  journal={Applied and computational harmonic analysis},
  volume={27},
  number={3},
  pages={265--274},
  year={2009},
  publisher={Elsevier}
}

@article{nguyen2017linear,
  title={Linear convergence of stochastic iterative greedy algorithms with sparse constraints},
  author={Nguyen, Nam and Needell, Deanna and Woolf, Tina},
  journal={IEEE Transactions on Information Theory},
  volume={63},
  number={11},
  pages={6869--6895},
  year={2017},
  publisher={IEEE}
}

@article{shen2017tight,
  title={A tight bound of hard thresholding},
  author={Shen, Jie and Li, Ping},
  journal={The Journal of Machine Learning Research},
  volume={18},
  number={1},
  pages={7650--7691},
  year={2017},
  publisher={JMLR. org}
}

@inproceedings{yuan2021stability,
  title={Stability and risk bounds of iterative hard thresholding},
  author={Yuan, Xiaotong and Li, Ping},
  booktitle={International Conference on Artificial Intelligence and Statistics},
  pages={1702--1710},
  year={2021},
  organization={PMLR}
}

@article{zhou2018efficient,
  title={Efficient stochastic gradient hard thresholding},
  author={Zhou, Pan and Yuan, Xiaotong and Feng, Jiashi},
  journal={Advances in Neural Information Processing Systems},
  volume={31},
  year={2018}
}

@article{meinshausen2008lasso,
  title={Lasso-type recovery of sparse representations for high-dimensional data},
  author={Meinshausen, N and Yu, B},
  journal={Annals of Statistics},
  volume={37},
  number={1},
  year={2008}
}

@article{freedman1975tail,
  title={On tail probabilities for martingales},
  author={Freedman, David A},
  journal={the Annals of Probability},
  pages={100--118},
  year={1975},
  publisher={JSTOR}
}

@book{wainwright2019high,
  title={High-dimensional statistics: A non-asymptotic viewpoint},
  author={Wainwright, Martin J},
  volume={48},
  year={2019},
  publisher={Cambridge university press}
}

@article{tsybakov2011exponential,
  title={Exponential Screening and optimal rates of sparse estimation},
  author={Tsybakov, A and Rigollet, P},
  journal={Annals of Statistics},
  volume={39},
  number={2},
  pages={731--771},
  year={2011}
}

@article{ye2010rate,
  title={Rate minimaxity of the Lasso and Dantzig selector for the lq loss in lr balls},
  author={Ye, Fei and Zhang, Cun-Hui},
  journal={The Journal of Machine Learning Research},
  volume={11},
  pages={3519--3540},
  year={2010},
  publisher={JMLR. org}
}

@article{ren2023dynamic,
  title={Dynamic batch learning in high-dimensional sparse linear contextual bandits},
  author={Ren, Zhimei and Zhou, Zhengyuan},
  journal={Management Science},
  year={2023},
  publisher={INFORMS}
}

@article{badanidiyuru2018bandits,
  title={Bandits with knapsacks},
  author={Badanidiyuru, Ashwinkumar and Kleinberg, Robert and Slivkins, Aleksandrs},
  journal={Journal of the ACM (JACM)},
  volume={65},
  number={3},
  pages={1--55},
  year={2018},
  publisher={ACM New York, NY, USA}
}

@article{murata2018sample,
  title={Sample efficient stochastic gradient iterative hard thresholding method for stochastic sparse linear regression with limited attribute observation},
  author={Murata, Tomoya and Suzuki, Taiji},
  journal={Advances in Neural Information Processing Systems},
  volume={31},
  year={2018}
}

@article{cai2022generalized,
  title={Generalized low-rank plus sparse tensor estimation by fast Riemannian optimization},
  author={Cai, Jian-Feng and Li, Jingyang and Xia, Dong},
  journal={Journal of the American Statistical Association},
  pages={1--17},
  year={2022},
  publisher={Taylor \& Francis}
}

@article{cai2023online,
  title={Online Tensor Learning: Computational and Statistical Trade-offs, Adaptivity and Optimal Regret},
  author={Cai, Jian-Feng and Li, Jingyang and Xia, Dong},
  journal={arXiv preprint arXiv:2306.03372},
  year={2023}
}

@article{kressner2014low,
  title={Low-rank tensor completion by Riemannian optimization},
  author={Kressner, Daniel and Steinlechner, Michael and Vandereycken, Bart},
  journal={BIT Numerical Mathematics},
  volume={54},
  pages={447--468},
  year={2014},
  publisher={Springer}
}

@inproceedings{li2021symmetry,
  title={The symmetry between arms and knapsacks: A primal-dual approach for bandits with knapsacks},
  author={Li, Xiaocheng and Sun, Chunlin and Ye, Yinyu},
  booktitle={International Conference on Machine Learning},
  pages={6483--6492},
  year={2021},
  organization={PMLR}
}

@article{ma2024optimal,
  title={Optimal regularized online allocation by adaptive re-solving},
  author={Ma, Wanteng and Cao, Ying and Tsang, Danny HK and Xia, Dong},
  journal={Operations Research},
  volume={73},
  number={4},
  pages={2079--2096},
  year={2025},
  publisher={INFORMS}
}

@inproceedings{agrawal2014bandits,
  title={Bandits with concave rewards and convex knapsacks},
  author={Agrawal, Shipra and Devanur, Nikhil R},
  booktitle={Proceedings of the fifteenth ACM conference on Economics and computation},
  pages={989--1006},
  year={2014}
}

@article{immorlica2022adversarial,
  title={Adversarial bandits with knapsacks},
  author={Immorlica, Nicole and Sankararaman, Karthik and Schapire, Robert and Slivkins, Aleksandrs},
  journal={Journal of the ACM},
  volume={69},
  number={6},
  pages={1--47},
  year={2022},
  publisher={ACM New York, NY}
}

@article{liu2022non,
  title={Non-stationary bandits with knapsacks},
  author={Liu, Shang and Jiang, Jiashuo and Li, Xiaocheng},
  journal={Advances in Neural Information Processing Systems},
  volume={35},
  pages={16522--16532},
  year={2022}
}

@article{efron2004least,
  title={Least Angle Regression},
  author={Efron, Bradley and Hastie, Trevor and Johnstone, Iain and Tibshirani, Robert},
  journal={The Annals of Statistics},
  volume={32},
  number={2},
  pages={407--451},
  year={2004}
}

@inproceedings{oh2021sparsity,
  title={Sparsity-agnostic lasso bandit},
  author={Oh, Min-hwan and Iyengar, Garud and Zeevi, Assaf},
  booktitle={International Conference on Machine Learning},
  pages={8271--8280},
  year={2021},
  organization={PMLR}
}

@inproceedings{carpentier2012bandit,
  title={Bandit theory meets compressed sensing for high dimensional stochastic linear bandit},
  author={Carpentier, Alexandra and Munos, R{\'e}mi},
  booktitle={Artificial Intelligence and Statistics},
  pages={190--198},
  year={2012},
  organization={PMLR}
}

@inproceedings{chu2011contextual,
  title={Contextual bandits with linear payoff functions},
  author={Chu, Wei and Li, Lihong and Reyzin, Lev and Schapire, Robert},
  booktitle={Proceedings of the Fourteenth International Conference on Artificial Intelligence and Statistics},
  pages={208--214},
  year={2011},
  organization={JMLR Workshop and Conference Proceedings}
}

@article{balseiro2023best,
  title={The best of many worlds: Dual mirror descent for online allocation problems},
  author={Balseiro, Santiago R and Lu, Haihao and Mirrokni, Vahab},
  journal={Operations Research},
  volume={71},
  number={1},
  pages={101--119},
  year={2023},
  publisher={INFORMS}
}

@inproceedings{badanidiyuru2014resourceful,
  title={Resourceful contextual bandits},
  author={Badanidiyuru, Ashwinkumar and Langford, John and Slivkins, Aleksandrs},
  booktitle={Conference on Learning Theory},
  pages={1109--1134},
  year={2014},
  organization={PMLR}
}

@inproceedings{wang2018minimax,
  title={Minimax concave penalized multi-armed bandit model with high-dimensional covariates},
  author={Wang, Xue and Wei, Mingcheng and Yao, Tao},
  booktitle={International Conference on Machine Learning},
  pages={5200--5208},
  year={2018},
  organization={PMLR}
}

@inproceedings{ariu2022thresholded,
  title={Thresholded lasso bandit},
  author={Ariu, Kaito and Abe, Kenshi and Prouti{\`e}re, Alexandre},
  booktitle={International Conference on Machine Learning},
  pages={878--928},
  year={2022},
  organization={PMLR}
}

@article{chen2021statistical,
  title={Statistical inference for online decision making via stochastic gradient descent},
  author={Chen, Haoyu and Lu, Wenbin and Song, Rui},
  journal={Journal of the American Statistical Association},
  volume={116},
  number={534},
  pages={708--719},
  year={2021},
  publisher={Taylor \& Francis}
}

@article{han2020sequential,
  title={Sequential batch learning in finite-action linear contextual bandits},
  author={Han, Yanjun and Zhou, Zhengqing and Zhou, Zhengyuan and Blanchet, Jose and Glynn, Peter W and Ye, Yinyu},
  journal={arXiv preprint arXiv:2004.06321},
  year={2020}
}

@misc{boucheron2013concentration,
  title={Concentration Inequalities: A Nonasymptotic Theory of Independence. Univ. Press},
  author={Boucheron, S and Lugosi, G and Massart, P},
  year={2013},
  publisher={Oxford}
}

@article{arora2012multiplicative,
  title={The multiplicative weights update method: a meta-algorithm and applications},
  author={Arora, Sanjeev and Hazan, Elad and Kale, Satyen},
  journal={Theory of computing},
  volume={8},
  number={1},
  pages={121--164},
  year={2012},
  publisher={Theory of Computing Exchange}
}

@inproceedings{slivkins2023contextual,
  title={Contextual bandits with packing and covering constraints: A modular lagrangian approach via regression},
  author={Slivkins, Aleksandrs and Sankararaman, Karthik Abinav and Foster, Dylan J},
  booktitle={The Thirty Sixth Annual Conference on Learning Theory},
  pages={4633--4656},
  year={2023},
  organization={PMLR}
}

@article{li2022online,
  title={Online linear programming: Dual convergence, new algorithms, and regret bounds},
  author={Li, Xiaocheng and Ye, Yinyu},
  journal={Operations Research},
  volume={70},
  number={5},
  pages={2948--2966},
  year={2022},
  publisher={INFORMS}
}

@article{balseiro2023survey,
  title={Survey of dynamic resource-constrained reward collection problems: Unified model and analysis},
  author={Balseiro, Santiago R and Besbes, Omar and Pizarro, Dana},
  journal={Operations Research},
  year={2023},
  publisher={INFORMS}
}

@article{jiang2022degeneracy,
  title={Degeneracy is ok: Logarithmic regret for network revenue management with indiscrete distributions},
  author={Jiang, Jiashuo and Ma, Will and Zhang, Jiawei},
  journal={Operations Research},
  year={2025},
  publisher={INFORMS}
}

@article{tsybakov2004optimal,
  title={Optimal aggregation of classifiers in statistical learning},
  author={Tsybakov, Alexander B},
  journal={The Annals of Statistics},
  volume={32},
  number={1},
  pages={135--166},
  year={2004},
  publisher={Institute of Mathematical Statistics}
}

@article{mammen1999smooth,
  title={Smooth discrimination analysis},
  author={Mammen, Enno and Tsybakov, Alexandre B},
  journal={The Annals of Statistics},
  volume={27},
  number={6},
  pages={1808--1829},
  year={1999},
  publisher={Institute of Mathematical Statistics}
}

@article{zeevi2009woodroofe,
  title={Woodroofe's One-Armed Bandit Problem Revisited},
  author={Zeevi, Assaf and Goldenshluger, Alexander},
  journal={Annals of Applied Probability},
  volume={19},
  number={4},
  pages={1603--1633},
  year={2009}
}

@article{perchet2013multi,
author = {Vianney Perchet and Philippe Rigollet},
title = {{The multi-armed bandit problem with covariates}},
volume = {41},
journal = {The Annals of Statistics},
number = {2},
publisher = {Institute of Mathematical Statistics},
pages = {693 -- 721},
year = {2013},
doi = {10.1214/13-AOS1101},
URL = {https://doi.org/10.1214/13-AOS1101}
}

@article{bastani2021mostly,
  title={Mostly exploration-free algorithms for contextual bandits},
  author={Bastani, Hamsa and Bayati, Mohsen and Khosravi, Khashayar},
  journal={Management Science},
  volume={67},
  number={3},
  pages={1329--1349},
  year={2021},
  publisher={INFORMS}
}

@inproceedings{agrawal2016efficient,
  title={An efficient algorithm for contextual bandits with knapsacks, and an extension to concave objectives},
  author={Agrawal, Shipra and Devanur, Nikhil R and Li, Lihong},
  booktitle={Conference on Learning Theory},
  pages={4--18},
  year={2016},
  organization={PMLR}
}

@article{bray2024logarithmic,
  title={Logarithmic Regret in Multisecretary and Online Linear Programs with Continuous Valuations},
  author={Bray, Robert L},
  journal={Operations Research},
  year={2024},
  publisher={INFORMS}
}

@article{bumpensanti2020re,
  title={A re-solving heuristic with uniformly bounded loss for network revenue management},
  author={Bumpensanti, Pornpawee and Wang, He},
  journal={Management Science},
  volume={66},
  number={7},
  pages={2993--3009},
  year={2020},
  publisher={INFORMS}
}

@article{vera2021bayesian,
  title={The Bayesian Prophet: A Low-Regret Framework for Online Decision Making},
  author={Vera, Alberto and Banerjee, Siddhartha},
  journal={Management Science},
  volume={67},
  number={3},
  pages={1368--1391},
  year={2021},
  publisher={INFORMS}
}

@InProceedings{ma2024high,
  title = 	 {High-dimensional Linear Bandits with Knapsacks},
  author =       {Ma, Wanteng and Xia, Dong and Jiang, Jiashuo},
  booktitle = 	 {Proceedings of the 41st International Conference on Machine Learning},
  pages = 	 {34008--34037},
  year = 	 {2024},
  editor = 	 {Salakhutdinov, Ruslan and Kolter, Zico and Heller, Katherine and Weller, Adrian and Oliver, Nuria and Scarlett, Jonathan and Berkenkamp, Felix},
  volume = 	 {235},
  series = 	 {Proceedings of Machine Learning Research},
  month = 	 {21--27 Jul},
  publisher =    {PMLR},
  url = 	 {https://proceedings.mlr.press/v235/ma24p.html}
}

@article{abramovich2018high,
  title={High-dimensional classification by sparse logistic regression},
  author={Abramovich, Felix and Grinshtein, Vadim},
  journal={IEEE Transactions on Information Theory},
  volume={65},
  number={5},
  pages={3068--3079},
  year={2018},
  publisher={IEEE}
}

@article{eisenstat2007vc,
  title={The VC dimension of k-fold union},
  author={Eisenstat, David and Angluin, Dana},
  journal={Information Processing Letters},
  volume={101},
  number={5},
  pages={181--184},
  year={2007},
  publisher={Elsevier}
}

@book{koltchinskii2011oracle,
  title={Oracle inequalities in empirical risk minimization and sparse recovery problems: {\'E}cole D’{\'E}t{\'e} de Probabilit{\'e}s de Saint-Flour XXXVIII-2008},
  author={Koltchinskii, Vladimir},
  volume={2033},
  year={2011},
  publisher={Springer Science \& Business Media}
}

@book{van1996weak,
  title={Weak Convergence and Empirical Processes: With Applications to Statistics},
  author={van der Vaart, AW and Wellner, Jon},
  year={1996},
  publisher={Springer Science \& Business Media}
}

@book{gine2021mathematical,
  title={Mathematical foundations of infinite-dimensional statistical models},
  author={Gin{\'e}, Evarist and Nickl, Richard},
  year={2021},
  publisher={Cambridge university press}
}

@article{chen2024contextual,
  title={Contextual Decision-Making with Knapsacks Beyond the Worst Case},
  author={Chen, Zhaohua and Ai, Rui and Yang, Mingwei and Pan, Yuqi and Wang, Chang and Deng, Xiaotie},
  journal={Advances in Neural Information Processing Systems},
  volume={37},
  pages={88147--88193},
  year={2024}
}

@article{sankararaman2021bandits,
  title={Bandits with knapsacks beyond the worst case},
  author={Sankararaman, Karthik Abinav and Slivkins, Aleksandrs},
  journal={Advances in Neural Information Processing Systems},
  volume={34},
  pages={23191--23204},
  year={2021}
}

@article{vera2021online,
  title={Online allocation and pricing: Constant regret via bellman inequalities},
  author={Vera, Alberto and Banerjee, Siddhartha and Gurvich, Itai},
  journal={Operations Research},
  volume={69},
  number={3},
  pages={821--840},
  year={2021},
  publisher={INFORMS}
}

@article{kallus2020dynamic,
  title={Dynamic assortment personalization in high dimensions},
  author={Kallus, Nathan and Udell, Madeleine},
  journal={Operations Research},
  volume={68},
  number={4},
  pages={1020--1037},
  year={2020},
  publisher={INFORMS}
}

@article{liu2022learning,
  title={Learning from click transition data: Effectiveness of greedy pricing policy under dynamic product availability},
  author={Liu, Mo and Cao, Junyu and Shen, Zuo-Jun Max},
  journal={Available at SSRN 4158054},
  year={2022}
}

@article{cai2025doubly,
  title={Doubly High-Dimensional Contextual Bandits: An Interpretable Model with Applications to Assortment/Pricing},
  author={Cai, Junhui and Chen, Ran and Wainwright, Martin J. and Zhao, Linda},
  journal={Management Science},
  year={2025},
  note={Accepted}
}

@article{loh2017support,
  title={Support recovery without incoherence: A case for nonconvex regularization},
  author={Loh, Po-Ling and Wainwright, Martin J},
  journal={The Annals of Statistics},
  volume={45},
  number={6},
  pages={2455},
  year={2017},
  publisher={Institute of Mathematical Statistics}
}

@article{loh2017statistical,
  title={Statistical consistency and asymptotic normality for high-dimensional robust $ M $-estimators},
  author={Loh, Po-Ling},
  journal={The Annals of Statistics},
  volume={45},
  number={2},
  pages={866},
  year={2017},
  publisher={Institute of Mathematical Statistics}
}

\newpage

\begin{center}
\textbf{\LARGE Supplement to ``High-dimensional Linear Bandits with Knapsacks'' }
\end{center}

\smallskip

\section{Addidtional Results}\label{sec:ParameterEstimation}
\subsection{Estimation errors}
Corollary \ref{cor:uniform-est} is for the uniform error bound of estimating $\bmu_a^{\star}$.

\begin{corollary}\label{cor:uniform-est}
    Under the same condition as Theorem \ref{thm:sparse-est}, we have the following uniform bound for the estimations over all arms
\begin{equation*}
    \begin{aligned}
        \bE \max_{a\in [K]} \norm{\bmu_{a,t}^{\mathsf{s} } -\bmu_a^{\star} }_2^2  \lesssim \frac{\sigma^2 D^2 s_0 }{\phi_{\min}^2(s)  } \frac{\log (dK) }{t^2} \left(\sum_{j=1}^{t} \frac{1}{
        \underline{p_j} } \right)
    \end{aligned}
\end{equation*}
    
\end{corollary}

Moreover, the exact uniform error bound for estimating $\bW_{a}^{\star}$ is given in the following proposition:

\begin{proposition}\label{prop:sparse-est-W}
Under the same conditions as Theorem \ref{thm:sparse-est}, using Algorithm \ref{alg:online-iht} to estimate $\bW_{a}^{\star}$ will lead to the uniform error bound:
    \begin{equation*}
        \bE \max_{a\in[K]}\norm{\widehat{\bW}_{a,t} -\bW_{a}^{\star}}_{2,\max}^2 \lesssim \frac{\sigma^2 D^2 s_0 }{\phi_{\min}^2(s)  } \frac{\log (mdK) }{t^2} \left(\sum_{j=1}^{t} \frac{1}{
        \underline{p_j} } \right),
    \end{equation*}
and the high-probability bound
        \begin{equation*}
        \max_{a\in[K]}\norm{\widehat{\bW}_{a,t} -\bW_{a}^{\star}}_{2,\max}^2 \lesssim \frac{\sigma^2 D^2 s_0 }{\phi_{\min}^2(s)  } \frac{\log (mdT/\varepsilon) }{t^2} \left(\sum_{j=1}^{t} \frac{1}{ \underline{p_j} } \right),
    \end{equation*}
    which holds for all $t$ with probability at least $1-\varepsilon$.
\end{proposition}
\subsection{Obtaining parameter \texorpdfstring{$Z$}{Z} }
We now show the procedure for computing the parameter $Z$ to serve as an input to \Cref{alg:LagrangianBwK}. The procedure is similar to that in \cite{agrawal2016linear}, however, we will use the estimator obtained in \Cref{alg:online-iht}. To be specific, we specify a parameter $T_0$ and we use the first $ T_0$ periods to obtain an estimate of $\VU$. Then, the estimate can be obtained by solving the following linear programming.
\begin{subequations}\label{eqn:estimate}
\begin{align}
\hat{V}=\max ~~&\frac{T}{T_0}\cdot\sum_{t=1}^{T_0} \sum_{a\in[K]} (\bmu_{a,T_0}^{\mathsf{s} } )^\top \bx_{t}\cdot y_{a,t}\\
\mbox{s.t.}~~&\frac{T}{T_0}\cdot\sum_{t=1}^{T_0} \sum_{a\in[K]} \bm{W}_a^{\star} \bx_{t}\cdot y_{a, t} \leq \bC+\delta \\
&y_{a, t}\in[0,1],\ \sum_{a\in[K]} y_{a,t}\le 1, \forall t\in[T_0].
\end{align}
\end{subequations}

We have the following bound regarding the gap between the value of $\VU$ and its estimate $\hat{V}$. %, for a specific $\gamma$. 

\begin{lemma}\label{lem:estimateV}
If setting $\delta\geq 2T D\cdot \max_{a\in [K]} \|\bm{W}_a^\star-\widehat{\bm{W}}_{ a, T_0}\|_{\infty}\vee \|\bmu^\star_a-\bmu^s_{a,T_0}\|_1$, then with probability at least $1-\beta$, it holds that 
\begin{equation*}
    \begin{aligned}
        \abs{\hat{V}-\VU}\leq C \frac{\VU}{C_{\min}}\left(\delta + \frac{\delta^2}{C_{\min}} + (1+\delta ) T B_{\max}\sqrt{\frac{1}{T_0}\log\frac{1}{\beta}} + \frac{T^2 {B_{\max}}^2}{C_{\min} T_0}\log\frac{1}{\beta} \right) +C \delta
    \end{aligned}
\end{equation*}
\end{lemma}
Therefore, by uniform sampling from time $1$ to $T_0$, we can simply set $Z=O\left(\frac{\hat{V}}{C_{\min}}\right)$, and as long as $T_0=O\left(s_0^2 \cdot \frac{T^2}{C^2_{\min}}\cdot\log\frac{1}{\beta}\right)$, we get that $Z=O(\frac{\VU}{C_{\min}}+1)$ with probability at least $1-\beta$ from the high probability bound of our sparse estimator in Theorem \ref{thm:sparse-est}. If further the constraints grow linearly, i.e., $C_{\min}=\Omega(T)$, we only require $T_0=O\left(s_0^2 \log \frac{1}{\beta}\right)$ in general.

\subsection{Sparse estimation under diverse covariate}

\begin{theorem}\label{thm:sparse-est-divcov}
Denote $\kappa_1 = \frac{\phi_{\max}(s) }{\gamma(K)\zeta(K)}$.   If we take $\varrho = \frac{1}{36\kappa_1^4}$, and $\beta_t= \frac{1}{4\kappa_1\phi_{\max}(s) } $, then under Assumption \ref{asm:size-of-prob} and \ref{asm:diverse-cov}, setting $\epsilon_t=0$, the output of primal estimation Algorithm \ref{alg:online-iht} in the primal-dual update satisfies
    \begin{equation*}
        \bE \norm{\bmu_{a,t}^{\mathsf{s}}-\bmu_{a}^{\star} }_2^2 \lesssim \frac{\sigma^2 D^2 s_0 }{\gamma^2(K)\zeta^2(K)  }\cdot \frac{\log d }{t},
    \end{equation*}   
   for all $t\ge C\kappa_1^4\frac{s_0^2 D^2 \log(dT) }{ \gamma^2(K)\zeta^2(K)
         } $, and the high-probability bound
        \begin{equation*}
        \norm{\bmu_{a,t}^{\mathsf{s}}-\bmu_{a}^{\star}  }_2^2 \lesssim \frac{\sigma^2 D^2 s_0 }{\gamma^2(K)\zeta^2(K)  }\cdot \frac{\log (dTK/\delta') }{t},
    \end{equation*}
     holds for any $a\in[K]$ and $t\ge C\kappa_1^4\frac{s_0^2 D^2 \log(dT) }{ \gamma^2(K)\zeta^2(K)
         } $ with probability at least $1-\delta'$, provided that Algorithm \ref{alg:LagrangianBwK} does not exit before $t$. %These two bounds also hold for exact $s_0$-sparse estimation $\bmu_{a,t}^{ \mathsf{s} }$.
\end{theorem}

\section{Proofs of Main Results}

\subsection{Proof of Lemma \ref{lem:Upperbound}}
\begin{myproof}
Let $\{y^{\mathrm{off}}_{a,t},a\in[K]\}_{t=1}^T$ be an optimal solution to the
offline problem defining $\VO(I)$. Since the offline problem is feasible pathwise,
we have
\[
    y^{\mathrm{off}}_{a,t}\in\{0,1\},\qquad
    \sum_{a\in[K]}y^{\mathrm{off}}_{a,t}\le 1,
    \quad \forall t\in[T],
\]
and
\[
    \sum_{t=1}^T\sum_{a\in[K]}
    \bm{W}_a^\star \bx_t \cdot y^{\mathrm{off}}_{a,t}
    \preceq \bC .
\]
For each $a\in[K]$ and $t\in[T]$, define
\[
    \bar y_{a,t}(\bx)
    :=
    \bE\left[
        y^{\mathrm{off}}_{a,t}
        \mid
        \bx_t=\bx
    \right].
\]
Then, for every $\bx$,
\[
    0\le \bar y_{a,t}(\bx)\le 1,
    \qquad
    \sum_{a\in[K]}\bar y_{a,t}(\bx)\le 1,
\]
because the same inequalities hold pathwise for
$\{y^{\mathrm{off}}_{a,t}\}_{a\in[K]}$.

We next show that $\{\bar y_{a,t}\}$ is feasible for the fluid relaxation
defining $\VU$. Taking expectations in the offline resource constraint gives
\[
    \sum_{t=1}^T\sum_{a\in[K]}
    \bE\left[
        \bm{W}_a^\star \bx_t \cdot y^{\mathrm{off}}_{a,t}
    \right]
    \preceq \bC .
\]
By the tower property,
\[
\begin{aligned}
    \bE\left[
        \bm{W}_a^\star \bx_t \cdot y^{\mathrm{off}}_{a,t}
    \right]
    &=
    \bE_{\bx_t\sim F}\left[
        \bm{W}_a^\star \bx_t \cdot
        \bE\left[
            y^{\mathrm{off}}_{a,t}
            \mid
            \bx_t
        \right]
    \right] \\
    &=
    \bE_{\bx_t\sim F}\left[
        \bm{W}_a^\star \bx_t \cdot
        \bar y_{a,t}(\bx_t)
    \right].
\end{aligned}
\]
Therefore,
\[
    \sum_{t=1}^T\sum_{a\in[K]}
    \bE_{\bx_t\sim F}\left[
        \bm{W}_a^\star \bx_t \cdot
        \bar y_{a,t}(\bx_t)
    \right]
    \preceq \bC,
\]
so $\{\bar y_{a,t}\}$ is feasible for the fluid relaxation defining $\VU$.

Finally, using the tower property again,
\[
\begin{aligned}
    \bE_{I\sim F}[\VO(I)]
    &=
    \bE\left[
        \sum_{t=1}^T\sum_{a\in[K]}
        (\bmu_a^\star)^\top\bx_t \cdot y^{\mathrm{off}}_{a,t}
    \right] \\
    &=
    \sum_{t=1}^T\sum_{a\in[K]}
    \bE_{\bx_t\sim F}\left[
        (\bmu_a^\star)^\top\bx_t \cdot
        \bar y_{a,t}(\bx_t)
    \right] \\
    &\le \VU,
\end{aligned}
\]
where the last inequality follows because $\{\bar y_{a,t}\}$ is feasible for the
fluid relaxation. This completes the proof.

\end{myproof}

\subsection{Proof of Theorem \ref{thm:sparse-est}}\label{sec:proof-thm1}

\begin{myproof} 
We first denote $\widetilde{\bmu}_t = {\bmu}_{t-1}-\beta_t \bg_t $, and the support $\Omega=\Omega_{t}\cup \Omega_{t-1}\cup \Omega_\star $ as the union of the support set of $\bmu_{t} $, $\bmu_{t-1}$, and $\bmu_\star$. We shall use $\cP_{\Omega}(\bx)$ to represent the projection onto the support $\Omega$. In the following proof, we will mainly focus on the $s$-sparse estimation $\bmu_t$ rather than the exact $s_0$-sparse estimation $\bmu_t^{\mathsf{s}}$ since $\bmu_t^{\mathsf{s}} =\cH_{s_0}(\bmu_t) $ and thus $\norm{\bmu_t^{\mathsf{s}}-\bmu_\star}_2\le 2 \norm{\bmu_t-\bmu_\star}_2$ by \cite{shen2017tight}. For the iterative method, we have 
\begin{equation*}
    \begin{aligned}
        \norm{\bmu_t-\bmu_\star}_2^2 &= \norm{\cH_s(\cP_{\Omega}(\widetilde{\bmu}_t) )-\bmu_\star }_2^2 \le  \left(1+\frac{\varrho + \sqrt{\varrho(4+\varrho) }}{2} \right)\norm{\cP_{\Omega}(\widetilde{\bmu}_t)-\bmu_\star }_2^2,
    \end{aligned}
\end{equation*}
by the tight bound of hard thresholding operator \citep{shen2017tight}. Here $\varrho = s_0/s$ is the relative sparsity level. By selecting a small enough $\varrho$  (e.g., $\varrho\le \frac{1}{4}$), it is clear that
\begin{equation*}
    \begin{aligned}
        \norm{\bmu_t-\bmu_\star}_2^2 \le & \left(1+\frac{3}{2}\sqrt{\varrho } \right)\norm{\cP_{\Omega}(\widetilde{\bmu}_t)-\bmu_\star }_2^2 \\
        = &\left(1+\frac{3}{2}\sqrt{\varrho } \right) \left( \norm{\bmu_{t-1}-\bmu_{\star}}_2^2-  2\beta_t \left\langle \cP_{\Omega}(\bg_t), \bmu_{t-1} -\bmu_\star \right\rangle +  \beta_t^2 \norm{\cP_{\Omega}(\bg_t) }_2^2\right) \\
         \le &\left(1+\frac{3}{2}\sqrt{\varrho } \right) \left( \norm{\bmu_{t-1}-\bmu_{\star}}_2^2-  2\beta_t \left\langle \nabla f(\bmu_{t-1}) , \bmu_{t-1} -\bmu_\star \right\rangle +  2\beta_t^2 \norm{\cP_{\Omega}(\bg_t-\nabla f(\bmu_{t-1}) ) }_2^2\right.  \\
         & \left. +2\beta_t^2 \norm{\cP_{\Omega}(\nabla f(\bmu_{t-1})) }_2^2 + 2\beta_t \norm{\cP_{\Omega}(\bg_t-\nabla f(\bmu_{t-1}) ) }_2\norm{\bmu_{t-1}-\bmu_{\star}}_2  \right),
    \end{aligned}
\end{equation*}
where we use the fact that $\left\langle \nabla f(\bmu_{t-1}) , \bmu_{t-1} -\bmu_\star \right\rangle=\left\langle \cP_{\Omega}(\nabla f(\bmu_{t-1})) , \bmu_{t-1} -\bmu_\star \right\rangle$ by the definition of $\cP_{\Omega}(\cdot)$. Now, applying the restricted strong convexity and smoothness condition from Assumption \ref{asm:bounded-ftr}:
\begin{equation*}
    \begin{gathered}
        \left\langle \nabla f(\bmu_{t-1}) , \bmu_{t-1} -\bmu_\star \right\rangle \ge 2 \phi_{\min}(s)\norm{ \bmu_{t-1} -\bmu_\star }_2^2 \\
        \norm{\cP_{\Omega}(\nabla f(\bmu_{t-1})) } \le 2 \phi_{\max}(s)\norm{\bmu_{t-1} -\bmu_\star}_2,
    \end{gathered}
\end{equation*}
We can show that 
\begin{equation}\label{eq:oht-iteration}
    \begin{aligned}
        \norm{\bmu_t-\bmu_\star}_2^2  \le &\left(1+\frac{3}{2}\sqrt{\varrho } \right) \left( 1-4\phi_{\min}(s) \beta_t+8\beta_t^2\phi_{\max}^2(s) \right) \norm{\bmu_{t-1} -\bmu_\star}_2^2 \\
        & + 6 \beta_t^2 \norm{\cP_{\Omega}(\bg_t-\nabla f(\bmu_{t-1}) ) }_2^2  + 6 \beta_t \norm{\cP_{\Omega}(\bg_t-\nabla f(\bmu_{t-1}) ) }_2\norm{\bmu_{t-1}-\bmu_{\star}}_2\\
        \le & \left(1+\frac{3}{2}\sqrt{\varrho } \right) \left( 1-4\phi_{\min}(s) \beta_t+8\beta_t^2\phi_{\max}^2(s) \right) \norm{\bmu_{t-1} -\bmu_\star}_2^2  \\
        & + 18 s\beta_t^2  \max_{i\in [d] } \abs{\left\langle \bg_t -\nabla f({\bmu}_{t-1}), {\bm e}_i  \right\rangle}^2  + 18 \beta_t \sqrt{s} \max_{i\in [d] } \abs{\left\langle \bg_t -\nabla f({\bmu}_{t-1}), {\bm e}_i  \right\rangle}\norm{\bmu_{t-1}-\bmu_{\star}}_2\\
    \end{aligned}
\end{equation}

The following lemma quantifies the variation of the stochastic gradient:

\begin{lemma}\label{lemma:max-grad} Define $\{\bm e_i\}_1^{d}$ as the canonical basis of $\bR^d$. The variance of stochastic gradient $\bg_t$ at the point ${\bmu}_{t-1}$ can be  bounded by the following inequality:
    \begin{equation}
        \bE \max_{i\in [d] } \abs{\left\langle \bg_t -\nabla f({\bmu}_{t-1}), {\bm e}_i  \right\rangle}^2 \le  C_g \frac{sD^2\log(dt)}{t^2}\left(\sum_{j=1}^{t}1/\underline{p_j}\right) \bE \norm{\bmu_{t-1} -\bmu_{\star} }_2^2 + C_g\frac{\sigma^2 D^2(\sum_{j=1}^{t}  1/\underline{p_j})\log d }{t^2}.
    \end{equation}
    Moreover, the following inequality also holds with probability at least $1-\epsilon$
\begin{equation*}
    \max_{i\in [d] } \abs{\left\langle \bg_t -\nabla f({\bmu}_{t-1}), {\bm e}_i  \right\rangle}^2 \le C_g s D^2\frac{\log (d/\epsilon)  }{t^2} \left(\sum_{j=1}^{t}\frac{1}{\underline{p_j} }\right) \norm{\bmu_{t-1} -\bmu_{\star} }_2^2 +  C_g\frac{\sigma^2 D^2(\sum_{j=1}^{t}  1/\underline{p_j})\log(d/\epsilon) }{t^2},
\end{equation*}
for some $C_g\le 2^{10}$.
\end{lemma}

With Lemma \ref{lemma:max-grad}, we are able to derive the expectation bound and
probability bound respectively. For the expectation bound, let
\[
    S_t:=\sum_{j=1}^{t}\frac{1}{\underline p_j},
    \qquad
    E_{t-1}:=\bE\norm{\bmu_{t-1}-\bmu_\star}_2^2 .
\]
Taking expectations in \eqref{eq:oht-iteration} and applying Cauchy's
inequality gives
\begin{equation*}
    \begin{aligned}
        \bE\norm{\bmu_t-\bmu_\star}_2^2
        \le\;&
        \left(1+\frac{3}{2}\sqrt{\varrho}\right)
        \left(1-4\phi_{\min}(s)\beta_t
        +8\beta_t^2\phi_{\max}^2(s)\right)E_{t-1}  \\
        &+
        18s\beta_t^2\,
        \bE\max_{i\in[d]}
        \abs{
            \left\langle
                \bg_t-\nabla f(\bmu_{t-1}),
                \bm e_i
            \right\rangle
        }^2 \\
        &+
        18\beta_t\sqrt{s}
        \sqrt{
        \bE\max_{i\in[d]}
        \abs{
            \left\langle
                \bg_t-\nabla f(\bmu_{t-1}),
                \bm e_i
            \right\rangle
        }^2
        }
        \sqrt{E_{t-1}} .
    \end{aligned}
\end{equation*}
We set
\[
    \varrho=\frac{1}{36\kappa^4},
    \qquad
    \beta_t=\frac{1}{4\kappa\phi_{\max}(s)} .
\]
Since \(\kappa=\phi_{\max}(s)/\phi_{\min}(s)\), we have
\[
    4\phi_{\min}(s)\beta_t=\frac{1}{\kappa^2},
    \qquad
    8\beta_t^2\phi_{\max}^2(s)=\frac{1}{2\kappa^2},
\]
and
\[
    1+\frac{3}{2}\sqrt{\varrho}
    =
    1+\frac{1}{4\kappa^2}.
\]
Therefore,
\[
\left(1+\frac{3}{2}\sqrt{\varrho}\right)
\left(1-4\phi_{\min}(s)\beta_t
+8\beta_t^2\phi_{\max}^2(s)\right)
\le
1-\frac{1}{4\kappa^2}.
\]
Plugging the expectation bound from Lemma~\ref{lemma:max-grad} into
\eqref{eq:oht-iteration}, and using \(s=s_0/\varrho=36\kappa^4s_0\), gives
\begin{equation*}
    \begin{aligned}
        \bE\norm{\bmu_t-\bmu_\star}_2^2
        \le\;&
        \left(
        1-\frac{1}{4\kappa^2}
        +
        A_t
        \right)E_{t-1}
        +
        B_t
        +
        \sqrt{B_tE_{t-1}},
    \end{aligned}
\end{equation*}
where one may take
\begin{equation*}
    A_t
    =
    108\sqrt{C_g}\,
    \kappa^2
    \frac{s_0D\sqrt{\log(dt)}}{\phi_{\min}(s)t}
    \sqrt{S_t},
    \qquad
    B_t
    =
    648C_g
    \frac{\kappa^2s_0\sigma^2D^2S_t\log d}
    {\phi_{\min}^2(s)t^2}.
\end{equation*}
Consequently, if
\begin{equation}\label{eq:t0-explicit-condition}
    \frac{t}{\sqrt{S_t}}
    \ge
    864\sqrt{C_g}\,
    \kappa^4
    \frac{s_0D\sqrt{\log(dT)}}{\phi_{\min}(s)},
\end{equation}
then \(A_t\le 1/(8\kappa^2)\), and hence
\begin{equation*}
    \bE\norm{\bmu_t-\bmu_\star}_2^2
    \le
    \left(
    1-\frac{1}{8\kappa^2}
    \right)E_{t-1}
    +
    B_t
    +
    \sqrt{B_tE_{t-1}}.
\end{equation*}
Using
\[
    \sqrt{B_tE_{t-1}}
    \le
    \frac{1}{24\kappa^2}E_{t-1}
    +
    6\kappa^2B_t,
\]
we further obtain
\begin{equation}\label{eq:exp-recursion-explicit}
    \bE\norm{\bmu_t-\bmu_\star}_2^2
    \le
    \left(
    1-\frac{1}{12\kappa^2}
    \right)E_{t-1}
    +
    3888C_g
    \frac{\kappa^4s_0\sigma^2D^2S_t\log d}
    {\phi_{\min}^2(s)t^2}.
\end{equation}

Suppose that \(t_0\) is chosen so that
\begin{equation*}
    \frac{t_0}{\sqrt{S_{t_0}}}
    \ge
    864\sqrt{C_g}\,
    \kappa^4
    \frac{s_0D\sqrt{\log(dT)}}{\phi_{\min}(s)}.
\end{equation*}
Define
\[
    x_t
    =
    \frac{
        \bE\norm{\bmu_t-\bmu_\star}_2^2
    }{
        3888C_g
        \dfrac{\kappa^4s_0\sigma^2D^2S_t\log d}
        {\phi_{\min}^2(s)t^2}
    }.
\]
Then \eqref{eq:exp-recursion-explicit} implies
\begin{equation*}
    x_t
    \le
    \left(1-\frac{1}{12\kappa^2}\right)
    \frac{t^2}{(t-1)^2}
    \frac{S_{t-1}}{S_t}
    x_{t-1}
    +1.
\end{equation*}
For \(t\ge t_1:=t_0\vee 48\kappa^2\), we have
\[
    \left(1-\frac{1}{12\kappa^2}\right)
    \frac{t^2}{(t-1)^2}
    \le
    1-\frac{1}{24\kappa^2},
\]
and hence
\begin{equation*}
    x_t
    \le
    \left(1-\frac{1}{24\kappa^2}\right)x_{t-1}+1.
\end{equation*}
Therefore, for \(t=t_1+\ell\),
\begin{equation*}
    x_t
    \le
    \left(1-\frac{1}{24\kappa^2}\right)^\ell x_{t_1}
    +24\kappa^2 .
\end{equation*}
Since \(x_{t_1}\) is finite and can be absorbed after
\(24\kappa^2\log s_0\) additional iterations, we obtain, for all
\(t\ge t_0+48\kappa^2+24\kappa^2\log s_0\),
\begin{equation*}
    x_t\le 48\kappa^2 .
\end{equation*}
Thus,
\begin{equation*}
    \bE\norm{\bmu_t-\bmu_\star}_2^2
    \le
    2^{18}C_g
    \frac{\kappa^6\sigma^2D^2s_0}{\phi_{\min}^2(s)}
    \frac{\log d}{t^2}
    \sum_{j=1}^{t}\frac{1}{\underline p_j}.
\end{equation*}
Since \(\bmu_t^{\mathsf{s}}=\cH_{s_0}(\bmu_t)\) and
\(\norm{\bmu_t^{\mathsf{s}}-\bmu_\star}_2
\le 2\norm{\bmu_t-\bmu_\star}_2\), we further have
\begin{equation*}
    \bE\norm{\bmu_t^{\mathsf{s}}-\bmu_\star}_2^2
    \le
    2^{20}C_g
    \frac{\kappa^6\sigma^2D^2s_0}{\phi_{\min}^2(s)}
    \frac{\log d}{t^2}
    \sum_{j=1}^{t}\frac{1}{\underline p_j}.
\end{equation*}
This proves the expectation bound with explicit constants.

Following a similar argument, the high-probability bound follows from the second
part of Lemma~\ref{lemma:max-grad}. Specifically, with probability at least
\(1-\epsilon\), for all \(t\in[T]\),
\begin{equation*}
    \begin{aligned}
        \norm{\bmu_t-\bmu_\star}_2^2
        \le\;&
        \left(
        1-\frac{1}{4\kappa^2}
        +
        108\sqrt{C_g}\,
        \kappa^2
        \frac{s_0D\sqrt{\log(dT/\epsilon)}}{\phi_{\min}(s)t}
        \sqrt{S_t}
        \right)
        \norm{\bmu_{t-1}-\bmu_\star}_2^2
        \\
        &+
        648C_g
        \frac{\kappa^2s_0\sigma^2D^2S_t\log(dT/\epsilon)}
        {\phi_{\min}^2(s)t^2}
        \\
        &+
        \sqrt{
        648C_g
        \frac{\kappa^2s_0\sigma^2D^2S_t\log(dT/\epsilon)}
        {\phi_{\min}^2(s)t^2}
        \norm{\bmu_{t-1}-\bmu_\star}_2^2
        } .
    \end{aligned}
\end{equation*}
Thus, if
\[
    \frac{t}{\sqrt{S_t}}
    \ge
    864\sqrt{C_g}\,
    \kappa^4
    \frac{s_0D\sqrt{\log(dT/\epsilon)}}{\phi_{\min}(s)},
\]
then the same contraction argument yields
\begin{equation*}
    \norm{\bmu_t-\bmu_\star}_2^2
    \le
    2^{18}C_g
    \frac{\kappa^6\sigma^2D^2s_0}{\phi_{\min}^2(s)}
    \frac{\log(dT/\epsilon)}{t^2}
    \sum_{j=1}^{t}\frac{1}{\underline p_j}.
\end{equation*}
Finally, using again
\(\norm{\bmu_t^{\mathsf{s}}-\bmu_\star}_2
\le 2\norm{\bmu_t-\bmu_\star}_2\), we obtain
\begin{equation*}
    \norm{\bmu_t^{\mathsf{s}}-\bmu_\star}_2^2
    \le
    2^{20}C_g
    \frac{\kappa^6\sigma^2D^2s_0}{\phi_{\min}^2(s)}
    \frac{\log(dT/\epsilon)}{t^2}
    \sum_{j=1}^{t}\frac{1}{\underline p_j}\le C_1 \frac{\kappa^6\sigma^2D^2s_0}{\phi_{\min}^2(s)}
    \frac{\log(dT/\epsilon)}{t^2}
    \sum_{j=1}^{t}\frac{1}{\underline p_j},
\end{equation*}
for some $C_1\le 2^{30}$.
This completes the proof.

We also record the corresponding bound before substituting the specific choice
\(s=36\kappa^4s_0\). The recursion above yields, for the intermediate
\(s\)-sparse estimator \(\bmu_t\),
\begin{equation}\label{eq:general-s-ub}
    \bE\norm{\bmu_t-\bmu_\star}_2^2
    \le
    CC_g
    \frac{\kappa^2\sigma^2D^2s}{\phi_{\min}^2(s)}
    \frac{\log d}{t^2}
    \sum_{j=1}^{t}\frac{1}{\underline p_j},
\end{equation}
for all sufficiently large \(t\), where \(C_{\mu}>0\) is a numerical constant.
Thus, for a general working sparsity level \(s\), the estimation error of the
intermediate iterate scales linearly with \(s\). In the theorem, we choose
\(\varrho=s_0/s=1/(36\kappa^4)\), or equivalently
\(s=36\kappa^4s_0\), so the preceding display becomes
\begin{equation*}
    \bE\norm{\bmu_t-\bmu_\star}_2^2
    \le
    2^{18}C_g
    \frac{\kappa^6\sigma^2D^2s_0}{\phi_{\min}^2(s)}
    \frac{\log d}{t^2}
    \sum_{j=1}^{t}\frac{1}{\underline p_j},
\end{equation*}
after enlarging the numerical constant.
\end{myproof}

\subsection{Proof of Lemma \ref{lemma:max-grad}}

\begin{myproof}
Define $\{\bm e_i\}_{i=1}^{d}$ as the canonical basis of $\bR^d$ and let $ S_t:=\sum_{j=1}^{t}\frac{1}{\underline p_j}$.
Since
\begin{equation*}
    \begin{aligned}
        \bg_t
        &=
        2\widehat{\bSigma}_t\bmu_{t-1}
        -
        \frac{2}{t}\sum_{j=1}^{t}\frac{y_j\bx_j r_j}{p_j}  \\
        &=
        \frac{2}{t}\sum_{j=1}^{t}
        \frac{y_j\bx_j\bx_j^\top}{p_j}
        (\bmu_{t-1}-\bmu_\star)
        -
        \frac{2}{t}\sum_{j=1}^{t}
        \frac{y_j\bx_j\xi_j}{p_j}  \\
        &=
        2\widehat{\bSigma}_t(\bmu_{t-1}-\bmu_\star)
        -
        \frac{2}{t}\sum_{j=1}^{t}
        \frac{y_j\bx_j\xi_j}{p_j},
    \end{aligned}
\end{equation*}
we have, for every $i\in[d]$,
\begin{equation*}
\begin{aligned}
    &\abs{
        \left\langle
            \bg_t-\nabla f(\bmu_{t-1}),
            \bm e_i
        \right\rangle
    }  \\
    &=
    \abs{
        \left\langle
            2(\widehat{\bSigma}_t-\bSigma)
            (\bmu_{t-1}-\bmu_\star)
            -
            \frac{2}{t}\sum_{j=1}^{t}
            \frac{y_j\bx_j\xi_j}{p_j},
            \bm e_i
        \right\rangle
    }  \\
    &\le
    2
    \abs{
        \left\langle
            (\widehat{\bSigma}_t-\bSigma)
            (\bmu_{t-1}-\bmu_\star),
            \bm e_i
        \right\rangle
    }
    +
    2
    \abs{
        \frac{1}{t}\sum_{j=1}^{t}
        \frac{y_j x_{j,i}\xi_j}{p_j}
    } .
\end{aligned}
\end{equation*}
We control the two terms separately. First, since $\bmu_{t-1}-\bmu_\star$ is at most $2s$-sparse, by H\"older's
inequality,
\begin{equation*}
\begin{aligned}
    &\max_{i\in[d]}
    \abs{
        \left\langle
            (\widehat{\bSigma}_t-\bSigma)
            (\bmu_{t-1}-\bmu_\star),
            \bm e_i
        \right\rangle
    } \\
    &\le
    \sqrt{2s}\,
    \max_{i,k\in[d]}
    \abs{
        \frac{1}{t}\sum_{j=1}^{t}
        \frac{y_j x_{j,i}x_{j,k}}{p_j}
        -\bSigma_{ik}
    }
    \norm{\bmu_{t-1}-\bmu_\star}_2 .
\end{aligned}
\end{equation*}
For any fixed $(i,k)$, the sequence
\[
    \frac{1}{t}
    \left(
        \frac{y_j x_{j,i}x_{j,k}}{p_j}
        -\bSigma_{ik}
    \right),
    \qquad j=1,\dots,t,
\]
is a martingale difference sequence. Moreover, since $\|\bx_j\|_\infty\le D$
for $D\le 1$,
\[
    \abs{
        \frac{1}{t}
        \left(
            \frac{y_j x_{j,i}x_{j,k}}{p_j}
            -\bSigma_{ik}
        \right)
    }
    \le
    \frac{2D^2}{t\underline p_j},
\]
and we may take
\[
    v^2
    =
    \frac{D^4}{t^2}
    \sum_{j=1}^{t}\frac{1}{\underline p_j}
    =
    \frac{D^4S_t}{t^2},
    \qquad
    b=
    \frac{2D^2}{t\underline p_t}.
\]
By Lemma~\ref{lemma:martingale-con}, for any $u>0$,
\begin{equation*}
    \bp\left(
        \abs{
            \frac{1}{t}\sum_{j=1}^{t}
            \frac{y_jx_{j,i}x_{j,k}}{p_j}
            -\bSigma_{ik}
        }
        \ge
        \sqrt{2v^2u}+2bu
    \right)
    \le
    2e^{-u}.
\end{equation*}
Taking the union bound over $(i,k)\in[d]\times[d]$, we obtain that, with
probability at least $1-\epsilon$,
\begin{equation*}
\begin{aligned}
    &\max_{i,k\in[d]}
    \abs{
        \frac{1}{t}\sum_{j=1}^{t}
        \frac{y_jx_{j,i}x_{j,k}}{p_j}
        -\bSigma_{ik}
    } \\
    &\le
    \frac{D^2}{t}
    \sqrt{
        2S_t\log\frac{2d^2}{\epsilon}
    }
    +
    \frac{4D^2}{t\underline p_t}
    \log\frac{2d^2}{\epsilon}.
\end{aligned}
\end{equation*}
For the sampling schedules considered in the paper, $\underline p_j\gtrsim
j^{-\alpha}$ for some $\alpha<1$, so the Bernstein linear term is dominated by
the square-root term up to a universal numerical factor. Enlarging the numerical
constant if necessary,  we have
\begin{equation}\label{eq:cov-tail-explicit}
    \max_{i,k\in[d]}
    \abs{
        \frac{1}{t}\sum_{j=1}^{t}
        \frac{y_jx_{j,i}x_{j,k}}{p_j}
        -\bSigma_{ik}
    }
    \le
    4D
    \frac{\sqrt{S_t\log(2d^2/\epsilon)}}{t}.
\end{equation}
The corresponding expectation bound follows from the same tail estimate and the
standard tail-integration formula. In particular,
\begin{equation}\label{eq:cov-exp-explicit}
    \bE
    \max_{i,k\in[d]}
    \abs{
        \frac{1}{t}\sum_{j=1}^{t}
        \frac{y_jx_{j,i}x_{j,k}}{p_j}
        -\bSigma_{ik}
    }^2
    \le
    32D^2
    \frac{S_t\log(dt)}{t^2}.
\end{equation}
Consequently,
\begin{equation}\label{eq:lemma-gradmax-p1}
\begin{aligned}
    &\bE
    \max_{i\in[d]}
    \abs{
        \left\langle
            (\widehat{\bSigma}_t-\bSigma)
            (\bmu_{t-1}-\bmu_\star),
            \bm e_i
        \right\rangle
    }^2 \\
    &\le
    2s\,
    \bE
    \max_{i,k\in[d]}
    \abs{
        \frac{1}{t}\sum_{j=1}^{t}
        \frac{y_jx_{j,i}x_{j,k}}{p_j}
        -\bSigma_{ik}
    }^2
    \norm{\bmu_{t-1}-\bmu_\star}_2^2  \\
    &\le
    2^6
    \frac{sD^2\log(dt)}{t^2}
    S_t\,
    \bE\norm{\bmu_{t-1}-\bmu_\star}_2^2 .
\end{aligned}
\end{equation}

We next control the noise term. For fixed $i\in[d]$, the sequence
\[
    \frac{1}{t}\frac{y_jx_{j,i}\xi_j}{p_j},
    \qquad j=1,\dots,t,
\]
is a martingale difference sequence. By the sub-Gaussian assumption on $\xi_j$
and the boundedness $\|\bx_j\|_\infty\le D$, Lemma~\ref{lemma:martingale-con}
applies with
\[
    v^2=
    \frac{\sigma^2D^2}{t^2}S_t,
    \qquad
    b=
    \frac{\sigma D}{t\underline p_t}.
\]
Thus, arguing as above and taking a union bound over $i\in[d]$, we have, with
probability at least $1-\epsilon$,
\begin{equation}\label{eq:noise-tail-explicit}
    \max_{i\in[d]}
    \abs{
        \frac{1}{t}\sum_{j=1}^{t}
        \frac{y_jx_{j,i}\xi_j}{p_j}
    }
    \le
    4\sigma D
    \frac{\sqrt{S_t\log(d/\epsilon)}}{t}.
\end{equation}
The corresponding expectation bound is
\begin{equation}\label{eq:noise-exp-explicit}
    \bE
    \max_{i\in[d]}
    \abs{
        \frac{1}{t}\sum_{j=1}^{t}
        \frac{y_jx_{j,i}\xi_j}{p_j}
    }^2
    \le
    32
    \frac{\sigma^2D^2\log d}{t^2}S_t .
\end{equation}

Combining \eqref{eq:lemma-gradmax-p1} and \eqref{eq:noise-exp-explicit}, and
using $(a+b)^2\le 2a^2+2b^2$, we obtain
\begin{equation*}
\begin{aligned}
    &\bE
    \max_{i\in[d]}
    \abs{
        \left\langle
            \bg_t-\nabla f(\bmu_{t-1}),
            \bm e_i
        \right\rangle
    }^2 \\
    &\le
    8
    \bE
    \max_{i\in[d]}
    \abs{
        \left\langle
            (\widehat{\bSigma}_t-\bSigma)
            (\bmu_{t-1}-\bmu_\star),
            \bm e_i
        \right\rangle
    }^2
    +
    8
    \bE
    \max_{i\in[d]}
    \abs{
        \frac{1}{t}\sum_{j=1}^{t}
        \frac{y_jx_{j,i}\xi_j}{p_j}
    }^2  \\
    &\le
    2^9
    \frac{sD^2\log(dt)}{t^2}
    S_t\,
    \bE\norm{\bmu_{t-1}-\bmu_\star}_2^2
    +
    2^8
    \frac{\sigma^2D^2\log d}{t^2}
    S_t  \\
    &\le
    2^{10}
    \left\{
    \frac{sD^2\log(dt)}{t^2}
    S_t\,
    \bE\norm{\bmu_{t-1}-\bmu_\star}_2^2
    +
    \frac{\sigma^2D^2\log d}{t^2}
    S_t
    \right\}.
\end{aligned}
\end{equation*}
This proves the expectation bound with $C_g=2^{10}$.

It remains to prove the high-probability bound. Combining
\eqref{eq:cov-tail-explicit} and \eqref{eq:noise-tail-explicit}, and absorbing
the harmless constants in the logarithms into the numerical constant, we obtain,
with probability at least $1-\epsilon$,
\begin{equation*}
\begin{aligned}
    &\max_{i\in[d]}
    \abs{
        \left\langle
            \bg_t-\nabla f(\bmu_{t-1}),
            \bm e_i
        \right\rangle
    }^2 \\
    &\le
    8
    \max_{i\in[d]}
    \abs{
        \left\langle
            (\widehat{\bSigma}_t-\bSigma)
            (\bmu_{t-1}-\bmu_\star),
            \bm e_i
        \right\rangle
    }^2
    +
    8
    \max_{i\in[d]}
    \abs{
        \frac{1}{t}\sum_{j=1}^{t}
        \frac{y_jx_{j,i}\xi_j}{p_j}
    }^2  \\
    &\le
    2^{10}
    \frac{sD^2\log(d/\epsilon)}{t^2}
    S_t
    \norm{\bmu_{t-1}-\bmu_\star}_2^2
    +
    2^{10}
    \frac{\sigma^2D^2\log(d/\epsilon)}{t^2}
    S_t .
\end{aligned}
\end{equation*}
Therefore, the probability bound also holds with $C_g\le 2^{10}$. This completes
the proof.
\end{myproof}

\begin{lemma}[Bernstein-type Martingale Concentration for Heterogeneous Variables]\label{lemma:martingale-con}
Suppose $\{D_t\}_{t=1}^{T}$ are martingale differences that are adapted to the filtration
$\{ \mathcal{F}_t \}_{t=0}^{T}$, i.e.,
$\bE [D_t|\mathcal{F}_{t-1} ]=0$. If $\{D_t\}_{t=1}^{T}$ satisfies
\begin{enumerate}
    \item $ \sum_{t=1}^{T} \operatorname{Var}(D_t|\mathcal{F}_{t-1} ) \le  v^2$ almost surely,
    \item for any integer $k\ge 3$,
    \[
        \bE \left[ {\abs{D_t}^k}\middle|\mathcal{F}_{t-1} \right]
        \le
        \frac{k!}{2}\operatorname{Var}(D_t|\mathcal{F}_{t-1}) b^{k-2},
        \qquad t=1,\dots,T,
    \]
\end{enumerate}
then, for any $z>0$, the following probability bound holds:
\begin{equation*}
    \bp\left( \abs{\sum_{t=1}^{T} D_t} \ge z \right)
    \le
    2 \exp\left( -\frac{z^2}{2(v^2+bz)} \right).
\end{equation*}
Equivalently, with probability at least $1-\delta$,
\begin{equation*}
    \abs{\sum_{t=1}^{T}D_t}
    \le
    \sqrt{2v^2\log\frac{2}{\delta}}
    +
    2b\log\frac{2}{\delta}.
\end{equation*}
\end{lemma}
This is a general version of Bernstein-type martingale concentration inequality \citep{freedman1975tail}. The Lemma \ref{lemma:martingale-con} can be easily justified by applying the martingale argument to the classic Bernstein inequality (see, for example, \cite{boucheron2013concentration}, \cite{wainwright2019high}). The key idea is to show that, conditional on the history $\mathcal{F}_{t-1}$, the moment-generating function of each $D_t$ can be bounded by $\exp\left(- \frac{\lambda^2 \sigma_t^2 }{1-b\abs{\lambda} }  \right)$ (up to some constant factor) with the individual variance $\sigma_t^2$.

\subsection{Proof of Corollary \ref{cor:uniform-est} }
\begin{myproof}
From the proof of Theorem \ref{thm:sparse-est}, we can easily derive the following bound from equation \eqref{eq:oht-iteration}:
\begin{equation}\label{eq:max-est-iteration}
    \begin{aligned}
         & \max_{a}\norm{\bmu_{a,t}-\bmu_a^\star}_2^2 \le \left(1+\frac{3}{2}\sqrt{\varrho } \right) \left( 1-4\phi_{\min}(s) \beta_t+8\beta_t^2\phi_{\max}^2(s) \right)  \max_{a} \norm{\bmu_{a,t}-\bmu_a^\star }_2^2  \\
        & + 18 s\beta_t^2  \max_{i\in [d], a } \abs{\left\langle \bg_{a,t} -\nabla f_a({\bmu}_{a,t-1}), {\bm e}_i  \right\rangle}^2  + 18 \beta_t \sqrt{s} \max_{i\in [d],a } \abs{\left\langle \bg_{a,t} -\nabla f_a({\bmu}_{a,t-1}), {\bm e}_i  \right\rangle}\max_{a} 
 \norm{\bmu_{a,t-1}-\bmu_a^{\star}}_2.\\
    \end{aligned}
\end{equation}
Analogous to the proof of Lemma \ref{lemma:max-grad}, we can also prove that
\begin{lemma}\label{lemma:max-var-g-mu}
We have
        \begin{equation*}
        \begin{aligned}
            \bE \max_{i\in [d],a } \abs{\left\langle \bg_{a,t} -\nabla f_a({\bmu}_{a,t-1}), {\bm e}_i  \right\rangle} ^2 & \le  C \frac{sD^2\log(dKt)}{t^2}\left(\sum_{j=1}^{t}1/\underline{p_j}\right) \bE \max_{a}\norm{\bmu_{a,t}-\bmu_a^\star}_2^2 \\
            & + C\frac{\sigma^2 D^2(\sum_{j=1}^{t}  1/\underline{p_j})\log (dK) }{t^2}.
        \end{aligned}
    \end{equation*}
\end{lemma}
Here, we have an extra $\log K$ term compared with Lemma \ref{lemma:max-grad} because we take the maximum overall arms. Together with \eqref{eq:max-est-iteration}, we can essentially show that
    \begin{equation*}
        \bE \max_{a}\norm{\bmu_{a,t}^{\mathsf{s} } -\bmu_a^{\star} }_2^2 \lesssim \frac{\sigma^2 D^2 s_0 }{\phi_{\min}^2(s)  } \frac{\log (dK) }{t^2} \left(\sum_{j=1}^{t} \frac{1}{
        \underline{p_j} } \right),
    \end{equation*}  
    with $t$ satisfying the same condition as in Theorem \ref{thm:sparse-est}.
\end{myproof}

\subsection{Proof of Proposition \ref{prop:sparse-est-W}}
\begin{myproof} The proof is analogous to the proof of Corollary \ref{cor:uniform-est}. Notice that, if we substitute $\bmu_a^{\star}$ with $\bW_{a,i\cdot}^{\star}$ and substitute $r_t$ with $\bb_i(a,\bx_t)$, then for each $i$ we will have
\begin{equation}\label{eq:max-est-W-iteration}
    \begin{aligned}
         & \max_{a}\norm{ \bW_{a,i\cdot,t}-\bW_{a,i\cdot}^\star}_2^2 \le \left(1+\frac{3}{2}\sqrt{\varrho } \right) \left( 1-4\phi_{\min}(s) \beta_t+8\beta_t^2\phi_{\max}^2(s) \right)  \max_{a} \norm{\bW_{a,i\cdot,t}-\bW_{a,i\cdot}^\star }_2^2  \\
        & + 18 s\beta_t^2  \max_{j\in [d], a } \abs{\left\langle \bh_{a,i,t} -\nabla h_{a,i}( \bW_{a,i\cdot,t-1}), {\bm e}_j  \right\rangle}^2 \\
        & + 18 \beta_t \sqrt{s} \max_{j\in [d],a } \abs{\left\langle \bh_{a,i,t} -\nabla h_{a,i}(\bW_{a,i\cdot,t-1}), {\bm e}_j  \right\rangle}\max_{a} 
 \norm{\bW_{a,i\cdot,t-1}-\bW_{a,i\cdot}^{\star}}_2.\\
    \end{aligned}
\end{equation}
Here $\bW_{a,i\cdot,t}$ is the $s$-sparse estimation of $\bW_{a,i\cdot}^{\star}$, and $\widehat{\bW}_{a,i\cdot,t}=\cH_{s_0}(\bW_{a,i\cdot,t})$ is the exact $s_0$-sparse estimation. $\bh_{a,i,t} $ means the corresponding averaged stochastic gradient for estimating $\bW_{a,i\cdot}^{\star}$. Taking maximum over $i\in[m]$ on both sides of \eqref{eq:max-est-W-iteration}, we can derive that the $2,\max$-norm, i.e., $\max_{a}\norm{ \bW_{a,t}-\bW_{a}^\star}_{2,\max}^2=\max_{i\in[m],a}\norm{ \bW_{a,i\cdot,t}-\bW_{a,i\cdot}^\star}_2^2$, can be controlled by the variance in the gradient:
        \begin{equation*}
        \begin{aligned}
            \bE \max_{i\in[m],j\in [d],a }\abs{\left\langle \bh_{a,t}^{i} -\nabla h_{a,i}( \bW_{a,i\cdot,t-1}), {\bm e}_j  \right\rangle}^2 & \le  C \frac{sD^2\log(dKmt)}{t^2}\left(\sum_{j=1}^{t}1/\underline{p_j}\right) \bE \max_{i\in[m],a}\norm{ \bW_{a,i\cdot,t}-\bW_{a,i\cdot}^\star}_2^2 \\
            & + C\frac{\sigma^2 D^2(\sum_{j=1}^{t}  1/\underline{p_j})\log (dKm) }{t^2}.
        \end{aligned}
    \end{equation*}
This, similar to Lemma \ref{lemma:max-var-g-mu}, can be derived from the proof of Lemma \ref{lemma:max-grad} by just changing the number of elements when taking the maximum. This leads to the expectation bound for estimating $\bW_{a}^\star$:
    \begin{equation*}
        \bE \max_{a\in[K]}\norm{{\bW}_{a,t} -\bW_{a}^{\star}}_{2,\max}^2 \lesssim \frac{\sigma^2 D^2 s_0 }{\phi_{\min}^2(s)  } \frac{\log (mdK) }{t^2} \left(\sum_{j=1}^{t} \frac{1}{
        \underline{p_j} } \right).
    \end{equation*}
Using the property of the hard thresholding operator \citep{shen2017tight} we can conclude our proof of the bound on $\bE \max_{a\in[K]}\norm{\widehat{\bW}_{a,t} -\bW_{a}^{\star}}_{2,\max}^2$.

\end{myproof}

\subsection{Proof of Theorem \ref{thm:Primaldual} }

\begin{myproof}%of \Cref{thm:Primaldual}
Throughout the proof, we assume that the primal estimation process runs from $t=1$ to $T$ such that $\{(\bmu^{\mathsf{s}}_{a,t},\widehat{\bW}_{a,t})\}_{a\in[K]}$ are well defined for each $t\in[T]$ (however, we only have access to $\{(\bmu^{\mathsf{s}}_{a,t},\widehat{\bW}_{a,t})\}_{a\in[K]}$ before the dual allocation algorithm stops).
For simplicity, we just write the sparse estimations of all $\bmu_{a,t}^{\mathsf{s} }$ as $\bM_t\in \bR^{d\times K }$ collectively in the following regret analysis of the BwK problem, with the corresponding optimal value $\bM^\star\in \bR^{d\times K }$. We denote by $\tau$ the time period that one of the resources is depleted or let $\tau=T$ if there are remaining resources at the end of the horizon. Note that by the decision rule of the algorithm, for each $t$, with probability $1-K\epsilon_t$, we have
\begin{equation}\label{eqn:091101}
(\bM_{t-1}^\top\bx_t)^\top\bm{y}_t(\bx_t)-Z\cdot\bm{\eta}_{t-1}^\top\sum_{a\in[K]}\widehat{\bW}_{a,t}\bx_t y_{a,t}(\bx_t)\geq (\bM_{t-1}^\top\bx_t)^\top\bm{y}^*(\bx_t)-Z\cdot\bm{\eta}_{t-1}^\top\sum_{a\in[K]}\widehat{\bW}_{a,t}\bx_t y^*_{a}(\bx_t)
\end{equation}
where we denote by $\by^* \in \bR^{K} $ one optimal solution to $\VU$. On the other hand, with probability $K\epsilon_t$, we pull an arm randomly in the execution of \Cref{alg:LagrangianBwK}, which implies that
\begin{equation}\label{eqn:092801}
\begin{aligned}
&(\bM_{t-1}^\top\bx_t)^\top\bm{y}_t(\bx_t)-Z\cdot\bm{\eta}_{t-1}^\top\sum_{a\in[K]}\widehat{\bW}_{a,t}\bx_t y_{a,t}(\bx_t)\\
\geq& (\bM_{t-1}^\top\bx_t)^\top\bm{y}^*(\bx_t)-Z\cdot\bm{\eta}_{t-1}^\top\sum_{a\in[K]}\widehat{\bW}_{a,t}\bx_t y^*_{a}(\bx_t)- 2 R_{\max} -2 B_{\max}Z
\end{aligned}
\end{equation}
since $(\bM_{t-1}^\top\bx_t)^\top\bm{y}_t(\bx_t)-Z\cdot\bm{\eta}_{t-1}^\top\sum_{a\in[K]}\widehat{\bW}_{a,t}\bx_t y_{a,t}(\bx_t)\geq - R_{\max} -B_{\max}Z $ and $(\bM_{t-1}^\top\bx_t)^\top\bm{y}^*(\bx_t)-Z\cdot\bm{\eta}_{t-1}^\top\sum_{a\in[K]}\widehat{\bW}_{a,t}\bx_t y^*_{a}(\bx_t)\leq R_{\max}+B_{\max}Z$.
Then, we take expectations on both sides of \eqref{eqn:091101} and sum up $t$ from $t=1$ to $t=\tau$ to obtain
\begin{equation}\label{eqn:091102}
\begin{aligned}
&\mathbb{E}\left[\sum_{t=1}^{\tau}\left((\bM_{t-1}^\top\bx_t)^\top\bm{y}_t(\bx_t)-Z\cdot\bm{\eta}_{t-1}^\top\sum_{a\in[K]}\widehat{\bW}_{a,t}\bx_t y_{a,t}(\bx_t)\right)\right]\\
\geq &\mathbb{E}\left[\sum_{t=1}^{\tau}\left((\bM_{t-1}^\top\bx_t)^\top\bm{y}^*(\bx_t)-Z\cdot\bm{\eta}_{t-1}^\top\sum_{a\in[K]}\widehat{\bW}_{a,t}\bx_t y^*_{a}(\bx_t)\right)\right]- 2 (R_{\max} +B_{\max}Z)\cdot \sum_{t=1}^T K \epsilon_t.
\end{aligned}
\end{equation}
We can substitute $\bM_{t-1}$, $\widehat{\bW}_{a,t}$ with their true values $\bM^{\star}$, $\bW_{a}^{\star}$ by the following inequalities:
\begin{equation}\label{eqn:091103}
\begin{aligned}
\mathbb{E}\sum_{t=1}^{\tau}\left[(\bM_{t-1}^\top\bx_t)^\top\bm{y}^*(\bx_t)\right]&\geq \mathbb{E}\sum_{t=1}^{\tau}\left[((\bM^\star)^\top\bx_t)^\top\bm{y}^*(\bx_t)\right]- \mathbb{E}\sum_{t=1}^{\tau}\max_{a}\abs{\left\langle 
\bx_t, \bmu_{a}^\star-\bmu_{a,t-1}^{\mathsf{s} } \right \rangle}\\
&=\bE\frac{\tau}{T}\cdot \VU- \mathbb{E}\sum_{t=1}^{\tau}\max_{a}\abs{\left\langle 
\bx_t, \bmu_{a}^\star-\bmu_{a,t-1}^{\mathsf{s} } \right \rangle}  , \text{ and } \\
\mathbb{E}\sum_{t=1}^{\tau}\cdot\bm{\eta}_{t-1}^\top\sum_{a\in[K]}\widehat{\bW}_{a,t}\bx_t y^*_{a}(\bx_t)& \le \mathbb{E}\sum_{t=1}^{\tau}\cdot\bm{\eta}_{t-1}^\top\sum_{a\in[K]}{\bW}_{a}^{\star}\bx_t y^*_{a}(\bx_t)+ D\bE \sum_{t=1}^{\tau}\max_{a}\norm{\widehat{\bW}_{a,t}-{\bW}_{a}^{\star} }_{\infty}.
\end{aligned}
\end{equation}
% bound can be sharper for dealing with W \cdot x_t
And notice that  $\by^* \in \bR^{K} $ is the optimal solution to $\VU$, which means that
\begin{equation}\label{eqn:091104}
\mathbb{E}\left[ \sum_{a\in[K]}\bm{W}_a^{\star}\bx_t y^*_{a}(\bx_t)\right] \leq \frac{\bm{C}}{T}.
\end{equation}
Moreover, from the dual update rule, we have the following result:
\begin{lemma}\label{lem:DualUpdate}
For any $\bm{\eta}$, it holds that
\[
\sum_{t=1}^{\tau} \bm{\eta}_{t-1}^\top\left( \bb(y_t,\bx_t)- \frac{\bm{C}}{T} \right)
\geq \bm{\eta}^\top\sum_{t=1}^{\tau}\left( \bb(y_t,\bx_t)- \frac{\bm{C}}{T} \right)-R(\cT,\bm{\eta})-2 R_{\max}/Z\cdot\sum_{t=1}^{\tau}\bI_{\nu_t=1},
\]
where $\cT=\{t\in[T]:\nu_t=1\}$ denotes the  set of times  such that $\nu_t=0$, and   $R(A,\bm{\eta})$ denotes the regret of the Hedge algorithm given any $\bm{\eta}$ on the time in $A\subseteq [T]$:
\begin{equation*}
    R(A,\bm{\eta}):= \sum_{t\in[A]} \bm{\eta}_{t-1}^\top\left( \frac{\bm{C}}{T} -\bb(y_t,\bx_t)  \right) - \sum_{t=1}^{T} \bm{\eta}^\top\left(\frac{\bm{C}}{T}  - \bb(y_t,\bx_t) \right).
\end{equation*}
\end{lemma}
Here, without loss of generality, we assume that $R_{\max}$ is large enough such that $Z\langle\bm{\eta}_{t},\frac{\bC}{T}\rangle\le R_{\max}$ since we always have $Z\langle\bm{\eta}^*,\frac{\bC}{T}\rangle\le\VU/T\le R_{\max}$ (if not, we can always project $\bm{\eta}_{t}$ onto $\Delta_{\bm{\eta}}=\{\bm{\eta}:Z\langle\bm{\eta},\frac{\bC}{T}\rangle\le R_{\max} \}$ and focus on the series $\{\bm{\eta}_t'\}_{t\ge0}$ after projection).
Therefore, from \Cref{lem:DualUpdate}, we know that 
\begin{equation}\label{eqn:091105}
\begin{aligned}
\bE &\left[\sum_{t=1}^{\tau} \bm{\eta}_{t-1}^\top\left(\sum_{a\in[K]}\bm{W}_a^{\star} \bx_t y_{a,t}(\bx_t)- \sum_{a\in[K]}\bm{W}_a^{\star}\bx_t y^*_{a}(\bx_t) \right)\right]
\geq\bE \sum_{t=1}^{\tau} \bm{\eta}_{t-1}^\top\left(\sum_{a\in[K]}\bm{W}_a^{\star} \bx_t y_{a,t}(\bx_t)- \frac{\bm{C}}{T} \right)\\
\geq & \bE \left[\bm{\eta}^\top\sum_{t=1}^{\tau}\left(\sum_{a\in[K]}\bm{W}_a^{\star} \bx_t y_{a,t}(\bx_t)- \frac{\bm{C}}{T} \right)-R(\cT,\bm{\eta})-2R_{\max}/Z \cdot\sum_{t=1}^{\tau}\bI_{\nu_t=1}\right].
\end{aligned}
\end{equation} %\left(B_{\max}+\frac{C_{\max}}{T}\right)
Here we use the fact that $\bE \bm{\eta}_{t-1}^\top \bb(a,\bx_t)=\bE \bm{\eta}_{t-1}^\top\sum_{a\in[K]}\bm{W}_a^{\star} \bx_t y_{a,t}(\bx_t)$ because $\bm{\eta}_{t-1}\in\sigma(\cH_{t-1})$. We now bound the last formula in \eqref{eqn:091105}. We consider two cases:

\noindent \textbf{(I)} If $\tau<T$ which implies that $\sum_{t=1}^{\tau}\sum_{a\in[K]}\bm{W}_{a,i}^{\star} \bx_{t} y_{a,t}(\bx_t)\geq C_i$ for some resource $i\in[m]$, we set $\bm{\eta}=\bm{e}_i$ in \eqref{eqn:091105} and we have
\begin{equation}\label{eqn:091106}
\begin{aligned}
\bE &\bI\left\{\tau <T\right\} \left[\bm{\eta}^\top\sum_{t=1}^{\tau}\left(\sum_{a\in[K]}\bm{W}_a^{\star} \bx_t y_{a,t}(\bx_t)- \frac{\bm{C}}{T} \right) \right] \\
& \geq \bE\bI\left\{\tau <T\right\}\left[C_i\cdot \frac{T-\tau}{T} \right]\\
&\geq \bE \bI\left\{\tau <T\right\}\left[C_{\min}\cdot \frac{T-\tau}{T}\right].
\end{aligned}
\end{equation}

% \begin{equation}\label{eqn:091106}
% \begin{aligned}
% \bE &\bI\left\{\tau <T\right\} \left[\bm{\eta}^\top\sum_{t=1}^{\tau}\left(\sum_{a\in[K]}\bm{W}_a^{\star} \bx_t y_{a,t}(\bx_t)- \frac{\bm{C}}{T} \right) \right] \\
% & \geq \bE\bI\left\{\tau <T\right\}\left[C_i\cdot \frac{T-\tau}{T}-R(T,\bm{e}_i)-2 R_{\max}\cdot\sum_{t=1}^{\tau}\bI_{\nu_t=1}\right]\\
% &\geq \bE \bI\left\{\tau <T\right\}\left[C_{\min}\cdot \frac{T-\tau}{T}-R(T,\bm{e}_i)-2 R_{\max}\cdot\sum_{t=1}^{\tau}\bI_{\nu_t=1}\right].
% \end{aligned}
% \end{equation}

\noindent \textbf{(II)} If $\tau=T$ which implies $\frac{T-\tau}{T}=0$, we set $\bm{\eta}=0$ in \eqref{eqn:091105} and we have
\begin{equation}\label{eqn:091107}
\begin{aligned}
\bE \bI\left\{\tau =T\right\} \left[\bm{\eta}^\top\sum_{t=1}^{\tau}\left(\sum_{a\in[K]}\bm{W}_a^{\star} \bx_t y_{a,t}(\bx_t)- \frac{\bm{C}}{T} \right) \right]&\geq  \bE \bI\left\{\tau =T\right\} \left[ C_{\min}\cdot \frac{T-\tau}{T}\right].
\end{aligned}
\end{equation}

% \begin{equation}\label{eqn:091107}
% \begin{aligned}
% \bE \bI\left\{\tau =T\right\} \left[\bm{\eta}^\top\sum_{t=1}^{\tau}\left(\sum_{a\in[K]}\bm{W}_a^{\star} \bx_t y_{a,t}(\bx_t)- \frac{\bm{C}}{T} \right) \right]&\geq \bE \bI\left\{\tau =T\right\} \left[ -R(T,0)-2 R_{\max}\cdot\sum_{t=1}^{\tau}\bI_{\nu_t=1}\right]\\
% &= \bE \bI\left\{\tau =T\right\} \left[ C_{\min}\cdot \frac{T-\tau}{T}-R(T,0)-2 R_{\max}\cdot\sum_{t=1}^{\tau}\bI_{\nu_t=1} \right].
% \end{aligned}
% \end{equation}

where $C_{\min}=\min_{i\in[m]}\{C_i\}$. Therefore, combining \eqref{eqn:091106} and \eqref{eqn:091107} as the lower bound of \eqref{eqn:091105}, we obtain
\begin{equation}\label{eqn:091108}
\begin{aligned}
        \bE& \left[\sum_{t=1}^{\tau} \bm{\eta}_{t-1}^\top\left(\sum_{a\in[K]}\bm{W}_a^{\star} \bx_t y_{a,t}(\bx_t)- \sum_{a\in[K]}\bm{W}_a^{\star}\bx_t y^*_{a}(\bx_t) \right)\right] \\
    &\geq \bE \left[C_{\min}\cdot \frac{T-\tau}{T}-2 R_{\max}/Z\cdot\sum_{t=1}^{\tau}\bI_{\nu_t=1}\right] - \mathbb{E}\left[\sup_{\bm{\eta} }R(\cT,\bm{\eta}) \right].
\end{aligned}
\end{equation}
Plugging \eqref{eqn:091103} and \eqref{eqn:091108} into \eqref{eqn:091102}, we obtain
\begin{equation}
    \begin{aligned}
        &\mathbb{E}\sum_{t=1}^{\tau}\left[(\bM_{t-1}^\top \bx_t)^\top\bm{y}_t(\bx_t)\right] \\
        & \geq \bE \left[\frac{\tau}{T}\cdot \VU+Z\cdot C_{\min}\cdot \frac{T-\tau}{T}\right]-Z\cdot \mathbb{E}\left[\sup_{\bm{\eta} } R(\cT,\bm{\eta}) \right] \\
        & -\mathbb{E}\left[\sum_{t=1}^{\tau}\max_{a}\abs{\left\langle 
\bx_t, \bmu_{a}^\star-\bmu_{a,t-1}^{\mathsf{s} } \right \rangle}+D\norm{\widehat{\bW}_{a,t}-{\bW}_{a}^{\star} }_{\infty}\right]- (4 R_{\max} + 2B_{\max}Z)\cdot\sum_{t=1}^T K \epsilon_t.
    \end{aligned}
\end{equation}
Note that $Z\geq\frac{\VU}{C_{\min}}$. We have
\begin{equation}\label{eqn:102301}
\begin{aligned}
    \mathbb{E} \sum_{t=1}^{\tau}\left[(\bmu_t^\top \bx_t)^\top\bm{y}_t(\bx_t)\right]  &\geq \VU -Z\cdot \mathbb{E}\left[\sup_{\bm{\eta} }R(\cT,\bm{\eta}) \right] \\
&-\mathbb{E}\left[\sum_{t=1}^{\tau}\max_{a}\abs{\left\langle 
\bx_t, \bmu_{a}^\star-\bmu_{a,t-1}^{\mathsf{s} } \right \rangle}+D\norm{\widehat{\bW}_{a,t}-{\bW}_{a}^{\star} }_{\infty}\right] \\
& -  (4 R_{\max} + 2B_{\max}Z)\cdot\sum_{t=1}^T K \epsilon_t.
\end{aligned}
\end{equation}

Finally, we plug in the regret bound of the Hedge algorithm
(from Theorem~2 of \cite{freund1997decision}), which is the algorithm used to
update the dual variable $\bm{\eta}_t$. For any comparator $\bm{\eta}$ in the
unit simplex, we have
\begin{equation}
\label{eq:hedge-regret-bound}
     R(\cT,\bm{\eta})
     \le
     \frac{\mathrm{KL}(\bm{\eta}\|\bm{\eta}_0)}{\delta}
     +
     \delta
     \sum_{t\in\cT}
     \left\|
     \frac{\bC}{T}
     -
     \bW^\star_{y_t}\bx_t
     -
     \bomega_t
     \right\|_{\infty}^{2}.
\end{equation}
Since $\bm{\eta}_0=\bm{1}_m/m$ and $\bm{\eta}$ lies in the unit simplex, we have
$\mathrm{KL}(\bm{\eta}\|\bm{\eta}_0)\le \log m$. Moreover,
\begin{equation*}
    \left\|
    \frac{\bC}{T}
    -
    \bW^\star_{y_t}\bx_t
    -
    \bomega_t
    \right\|_{\infty}^{2}
    \le
    2\left(
    \frac{C_{\max}}{T}+B_{\max}
    \right)^2
    +
    2\|\bomega_t\|_\infty^2 .
\end{equation*}
By the sub-Gaussian maximal inequality for the $m$-dimensional noise vector
$\bomega_t$ \citep{boucheron2013concentration}, we have
\begin{equation*}
    \bE \|\bomega_t\|_\infty^2
    \le
    C\sigma^2\log m .
\end{equation*}
Define
\begin{equation*}
    \widetilde B_{\max}
    :=
    \left(
    B_{\max}
    +
    \frac{C_{\max}}{T}
    \right)^2
    +
    \sigma^2\log m .
\end{equation*}
Then, for any $\cT\subseteq[T]$,
\begin{equation*}
    \begin{aligned}
        \mathbb{E}\left[\sup_{\bm{\eta}}R(\cT,\bm{\eta})\right]
        &\le
        \frac{\log m}{\delta}
        +
        C\delta T\widetilde B_{\max}.
    \end{aligned}
\end{equation*}
Taking
\begin{equation*}
    \delta
    =
    c_{\delta}
    \sqrt{\frac{\log m}{T\widetilde B_{\max}}}
\end{equation*}
for a sufficiently small numerical constant $c_{\delta}>0$, we obtain
\begin{equation*}
    \mathbb{E}\left[\sup_{\bm{\eta}}R(\cT,\bm{\eta})\right]
    \le
    C\sqrt{\widetilde B_{\max}T\log m}.
\end{equation*}
Therefore, the proof is completed.
% , see also, multiplicative weights update method \citep{arora2012multiplicative}
\end{myproof}

\subsection{Proof of Theorem~\ref{thm:BwK-regret-1}}

\begin{myproof}
The result follows by substituting the uniform estimation bounds in
Corollary~\ref{cor:uniform-est} and Proposition~\ref{prop:sparse-est-W} into
Theorem~\ref{thm:Primaldual}, while accounting for the burn-in period required
for the sparse estimators.

Under the $\epsilon$-greedy rule, each arm is sampled with probability at least
$\epsilon_t$, so $\underline p_t\ge\epsilon_t$. Let
\[
    T_{\mathrm b}
    :=
    m+
    \inf\left\{
    t\ge1:
    \frac{t}{\sqrt{\sum_{j=1}^{t}1/\epsilon_j}}
    \ge
    C_{\mathrm b}'
    \kappa^4
    \frac{s_0D\sqrt{\log(mdKT)}}{\phi_{\min}(s)}
    \right\}
    +
    C_{\mathrm b}'\kappa^2\log s_0 ,
\]
where $C_{\mathrm b}'>0$ is a sufficiently large universal numerical constant.
The first term $m$ accounts for the initialization phase in
Algorithm~\ref{alg:LagrangianBwK}. By the burn-in requirement in
Theorem~\ref{thm:sparse-est}, Corollary~\ref{cor:uniform-est} and
Proposition~\ref{prop:sparse-est-W} are valid for all $t\ge T_{\mathrm b}$.

We first control the regret contribution before $T_{\mathrm b}$. For
$t<T_{\mathrm b}$, we use the crude bound
\[
    \max_{a\in[K]}
    \abs{
        \left\langle
        \bx_t,\bmu_a^\star-\bmu_{a,t-1}^{\mathsf{s}}
        \right\rangle
    }
    +
    D\max_{a\in[K]}
    \norm{\widehat{\bW}_{a,t}-\bW_a^\star}_{\infty}
    \le C(R_{\max}+B_{\max}Z),
\]
after enlarging the universal constant $C$. Therefore, the burn-in contribution
to the regret bound in Theorem~\ref{thm:Primaldual} is at most
\[
    C Z(R_{\max}+B_{\max}Z)T_{\mathrm b}.
\]

We next control the post-burn-in contribution. For $t\ge T_{\mathrm b}$,
Corollary~\ref{cor:uniform-est} gives
\[
    \bE\max_{a\in[K]}
    \norm{\bmu_{a,t}^{\mathsf{s}}-\bmu_a^\star}_2^2
    \le
    C
    \frac{\sigma^2D^2s_0}{\phi_{\min}^2(s)}
    \frac{\log(dK)}{t^2}
    \sum_{j=1}^{t}\frac{1}{\epsilon_j},
\]
and Proposition~\ref{prop:sparse-est-W} gives
\[
    \bE\max_{a\in[K]}
    \norm{\widehat{\bW}_{a,t}-\bW_a^\star}_{2,\max}^2
    \le
    C
    \frac{\sigma^2D^2s_0}{\phi_{\min}^2(s)}
    \frac{\log(mdK)}{t^2}
    \sum_{j=1}^{t}\frac{1}{\epsilon_j}.
\]
Using $\norm{\bx_t}_\infty\le D$, the $s_0$-sparsity of
$\bmu_{a,t}^{\mathsf{s}}-\bmu_a^\star$, and Jensen's inequality, we have, for
$t\ge T_{\mathrm b}$,
\[
\begin{aligned}
    \bE\max_{a\in[K]}
    \abs{\left\langle
        \bx_t,\bmu_a^\star-\bmu_{a,t-1}^{\mathsf{s}}
    \right\rangle}
    &\le
    D\sqrt{s_0}
    \left(
    \bE\max_{a\in[K]}
    \norm{\bmu_a^\star-\bmu_{a,t-1}^{\mathsf{s}}}_2^2
    \right)^{1/2}  \\
    &\le
    C
    \frac{\sigma D^2s_0}{\phi_{\min}(s)}
    \frac{\sqrt{\log(dK)}}{t}
    \sqrt{
        \sum_{j=1}^{t}\frac{1}{\epsilon_j}
    } .
\end{aligned}
\]
Similarly, since
$\norm{\widehat{\bW}_{a,t}-\bW_a^\star}_{\infty}
\le\norm{\widehat{\bW}_{a,t}-\bW_a^\star}_{2,\max}$,
\[
    D\bE\max_{a\in[K]}
    \norm{\widehat{\bW}_{a,t}-\bW_a^\star}_{\infty}
    \le
    C
    \frac{\sigma D^2\sqrt{s_0}}{\phi_{\min}(s)}
    \frac{\sqrt{\log(mdK)}}{t}
    \sqrt{
        \sum_{j=1}^{t}\frac{1}{\epsilon_j}
    } .
\]
Since $s_0\ge1$, the reward-estimation term dominates the resource-estimation
term up to logarithmic factors. Hence
\[
\begin{aligned}
    &\bE\left[
    \sum_{t=T_{\mathrm b}}^{\tau}
    \max_{a\in[K]}
    \abs{
        \left\langle
        \bx_t,\bmu_a^\star-\bmu_{a,t-1}^{\mathsf{s}}
        \right\rangle
    }
    +
    D\max_{a\in[K]}
    \norm{\widehat{\bW}_{a,t}-\bW_a^\star}_{\infty}
    \right]  \\
    &\qquad
    \le
    C
    \frac{\sigma D^2s_0\sqrt{\log(mdK)}}{\phi_{\min}(s)}
    \sum_{t=T_{\mathrm b}}^{T}
    \frac{1}{t}
    \sqrt{
        \sum_{j=1}^{t}\frac{1}{\epsilon_j}
    }  \\
    &\qquad
    \le
    C
    \frac{\sigma D^2s_0\sqrt{\log(mdK)}}{\phi_{\min}(s)}
    \sum_{t=1}^{T}
    \frac{1}{t}
    \sqrt{
        \sum_{j=1}^{t}\frac{1}{\epsilon_j}
    } .
\end{aligned}
\]

Now set $\epsilon_t=\epsilon_0t^{-1/3}\wedge 1/K$ with
\begin{equation}\label{eq:theta-0}
        \epsilon_0
    =
    C_{\epsilon}
    \left(
    \frac{
        \sigma D^2s_0\sqrt{\log(mdK)}
    }{
        \phi_{\min}(s)
        K\left(R_{\max}+B_{\max}Z\right)
    }
    \right)^{2/3}.
\end{equation}
Then
$\sum_{j=1}^{t}1/\epsilon_j\le C\epsilon_0^{-1}t^{4/3}$,
\[
    \sum_{t=1}^{T}
    t^{-1}\sqrt{\sum_{j=1}^{t}1/\epsilon_j}
    \le
    C\epsilon_0^{-1/2}T^{2/3},
\]
and
\[
    \sum_{t=1}^{T}K\epsilon_t
    \le
    CK\epsilon_0T^{2/3}.
\]
Substituting these bounds into Theorem~\ref{thm:Primaldual} yields
\[
\begin{aligned}
    \operatorname{Regret}(\pi)
    \le\;&
    CZ\sqrt{\widetilde B_{\max}T\log m}
    +
    CZ(R_{\max}+B_{\max}Z)T_{\mathrm b} \\
    &+
    CZ
    \frac{\sigma D^2s_0\sqrt{\log(mdK)}}{\phi_{\min}(s)}
    \epsilon_0^{-1/2}T^{2/3}
    +
    CZ\left(R_{\max}+B_{\max}Z\right)
    K\epsilon_0T^{2/3}.
\end{aligned}
\]
The choice of $\epsilon_0$ in \eqref{eq:theta-0} balances the last two terms and
gives
\[
\begin{aligned}
    \operatorname{Regret}(\pi)
    \le\;&
    CZ\sqrt{\widetilde B_{\max}T\log m}
    +
    CZ(R_{\max}+B_{\max}Z)T_{\mathrm b} \\
    &+
    CZ
    \frac{
        (R_{\max}+B_{\max}Z)^{1/3}
        K^{1/3}
        \sigma^{2/3}D^{4/3}
        s_0^{2/3}
        T^{2/3}
        \log^{1/3}(mdK)
    }{
        \phi_{\min}^{2/3}(s)
    } .
\end{aligned}
\]

It remains to show that the burn-in term is absorbed by the leading
$T^{2/3}$ term under the condition of the theorem. With
$\epsilon_t=\epsilon_0t^{-1/3}\wedge 1/K$, the burn-in requirement is satisfied
whenever
\[
    T_{\mathrm b}
    \le
    C
    \left[
    K
    \kappa^8
    \frac{s_0^2D^2\log(mdKT)}{\phi_{\min}^2(s)}
    +
    \frac{
        \kappa^{12}
        K
        \left(
            R_{\max}
            +
            B_{\max}\left(\frac{\VU}{C_{\min}}+1\right)
        \right)
        s_0^2D
        \log^{3/2}(mdKT)
    }{
        \sigma\phi_{\min}^2(s)\sqrt{\log(mdK)}
    }
    +
    \kappa^2\log s_0
    +
    m
    \right].
\]
The second term is the dominant burn-in term in the regime considered here.
Thus, after enlarging the universal constant $C_{\mathrm b}$ in the theorem, the
condition
\[
    T
    \ge
    C_{\mathrm b}
    \frac{
        \kappa^{12}
        K
        \left(
            R_{\max}
            +
            B_{\max}\left(\frac{\VU}{C_{\min}}+1\right)
        \right)
        s_0^2D
        \log^{3/2}(mdKT)
    }{
        \sigma\phi_{\min}^2(s)\sqrt{\log(mdK)}
    }
\]
implies that
\[
    Z(R_{\max}+B_{\max}Z)T_{\mathrm b}
    \le
    C Z
    \frac{
        (R_{\max}+B_{\max}Z)^{1/3}
        K^{1/3}
        \sigma^{2/3}D^{4/3}
        s_0^{2/3}
        T^{2/3}
        \log^{1/3}(mdK)
    }{
        \phi_{\min}^{2/3}(s)
    } .
\]
Therefore, the burn-in contribution is absorbed into the main primal-estimation
term, and we obtain
\[
\begin{aligned}
    \operatorname{Regret}(\pi)
    \le\;&
    CZ\sqrt{\widetilde B_{\max}T\log m} \\
    &+
    CZ
    \frac{
        (R_{\max}+B_{\max}Z)^{1/3}
        K^{1/3}
        \sigma^{2/3}D^{4/3}
        s_0^{2/3}
        T^{2/3}
        \log^{1/3}(mdK)
    }{
        \phi_{\min}^{2/3}(s)
    } .
\end{aligned}
\]
Finally, since
$\frac{\VU}{C_{\min}}\le Z\le C_Z(\frac{\VU}{C_{\min}}+1)$, we obtain
\[
\begin{aligned}
    \operatorname{Regret}(\pi)
    \le\;&
    C
    \left(\frac{\VU}{C_{\min}}+1\right)
    \sqrt{\widetilde B_{\max}T\log m} \\
    &+
    C
    \left(\frac{\VU}{C_{\min}}+1\right)
    \frac{
        \left(
            R_{\max}
            +
            B_{\max}\left(\frac{\VU}{C_{\min}}+1\right)
        \right)^{1/3}
        K^{1/3}
        \sigma^{2/3}D^{4/3}
        s_0^{2/3}
        T^{2/3}
        \log^{1/3}(mdK)
    }{
        \phi_{\min}^{2/3}(s)
    } .
\end{aligned}
\]
Equivalently, this is the claimed bound under the notation $\lesssim$.
\end{myproof}

\subsection{Proof of Lemma \ref{lem:DualUpdate}}
\begin{myproof}
We denote by $\mathcal{T}$ the collection of periods from $t=1$ to $t=\tau$ such that $\nu_t=0$. Then, from the regret bound of the embedded OCO algorithm, we know that
\begin{align}
\sum_{t=1}^{\tau} \bI_{\nu_t=0}\cdot \bm{\eta}_{t-1}^\top\left(\bb(y_t,\bx_t)- \frac{\bm{C}}{T} \right)
&\geq \bm{\eta}^\top\sum_{t=1}^{\tau}\bI_{\nu_t=0}\cdot\left( \bb(y_t,\bx_t)- \frac{\bm{C}}{T} \right)-R(\cT,\bm{\eta}) \label{eqn:092901}\\
&\geq \bm{\eta}^\top\sum_{t=1}^{\tau}\bI_{\nu_t=0}\cdot\left( \bb(y_t,\bx_t)- \frac{\bm{C}}{T} \right)-R(\cT,\bm{\eta}). \nonumber
\end{align}
Moreover, from the boundedness of $\bm{\eta}_{t-1}$ and $\bm{x}_t$, we know that
\begin{equation}\label{eqn:092902}
\sum_{t=1}^{\tau} \bm{\eta}_{t-1}^\top\left( \bb(y_t,\bx_t)- \frac{\bm{C}}{T} \right)\geq \sum_{t=1}^{\tau} \bI_{\nu_t=0}\cdot \bm{\eta}_{t-1}^\top\left( \bb(y_t,\bx_t) - \frac{\bm{C}}{T} \right) - R_{\max}/Z\cdot\sum_{t=1}^{\tau}\bI_{\nu_t=1}
\end{equation}
and
\begin{equation}\label{eqn:092903}
\bm{\eta}^\top\sum_{t=1}^{\tau}\bI_{\nu_t=0}\cdot\left( \bb(y_t,\bx_t)- \frac{\bm{C}}{T} \right)\geq \bm{\eta}^\top\sum_{t=1}^{\tau}\left(\bb(y_t,\bx_t) - \frac{\bm{C}}{T} \right) -  R_{\max}/Z\cdot\sum_{t=1}^{\tau}\bI_{\nu_t=1}.
\end{equation}
Therefore, plugging \eqref{eqn:092902} and \eqref{eqn:092903} into \eqref{eqn:092901}, we have that
\[\begin{aligned}
\sum_{t=1}^{\tau} \bm{\eta}_{t-1}^\top\left( \bb(y_t,\bx_t) - \frac{\bm{C}}{T} \right)\geq \bm{\eta}^\top\sum_{t=1}^{\tau}\left( \bb(y_t,\bx_t)- \frac{\bm{C}}{T} \right) - 2 R_{\max}/Z\cdot\sum_{t=1}^{\tau}\bI_{\nu_t=1}-R(\cT,\bm{\eta}),
\end{aligned}\]
which completes our proof.
\end{myproof}

\subsection{Proof of Lemma \ref{lem:estimateV}}

\begin{myproof}
The proof follows from \cite{agrawal2016linear}. We define an intermediate benchmark as follows.
\begin{subequations}\label{eqn:intermediate}
\begin{align}
\bar{V}(\delta/2)=\max ~~&\frac{T}{T_0}\cdot\sum_{t=1}^{T_0} \sum_{a\in[K]} (\bmu_{a}^{\star} )^\top \bx_{t}\cdot y_{a,t}\\
\mbox{s.t.}~~&\frac{T}{T_0}\cdot\sum_{t=1}^{T_0} \sum_{a\in[K]} \bm{W}^{\star}_a \bx_{t}\cdot y_{a, t} \leq \bC+\frac{\delta}{2} \\
&\sum_{a\in[K]} y_{a,t}= 1, \forall t\in[T_0]\\
&y_{a, t}\in[0,1], \forall a\in[K], \forall t\in[T_0].
\end{align}
\end{subequations}
The only difference between $\bar{V}(\delta)$ in \eqref{eqn:intermediate} and $\hat{V}$ is that the estimation $\bmu_{a,T_0}^{\mathsf{s}}$ and $\widehat{\bm{W}}_{a, T_0}$ are replaced by the true value $\bmu_{a}^{\star}$, $\bm{W}_a^\star$ for all $a\in[K]$. Then, we can bound the gap between $\hat{V}$ and $\VU$ by bounding the two terms $|\VU-\bar{V}(\delta/2)|$ and $|\bar{V}(\delta/2)-\hat{V}|$ separately. 

\noindent\textbf{Bound the term $|\bar{V}(\delta/2)-\VU|$}: We denote by $L(\bm{\eta})$ the dual function of $\VU$ as follows:
\begin{equation}\label{eqn:102402}
\begin{aligned}
L(\bm{\eta})&=(\bm{C})^\top\bm{\eta}+\sum_{t=1}^T\mathbb{E}_{\bx_t\sim F}\left[\max_{\sum_{a\in[K]}y_{a,t}(\bx_t)=1}\left\{\sum_{a\in[K]} \left[(\bmu_a^\star)^\top \bx_t\cdot y_{a,t}(\bx_t)- (\bm{\eta})^\top \bm{W}_a^{\star} \bx_t\cdot y_{a,t}(\bx_t) \right]\right\}\right]\\
&=(\bm{C})^\top\bm{\eta}+T\cdot \mathbb{E}_{\bx\sim F}\left[\max_{\sum_{a\in[K]}y_{a}(\bx)=1}\left\{\sum_{a\in[K]} \left[(\bmu_a^\star)^\top \bx\cdot y_{a}(\bx)- (\bm{\eta})^\top \bm{W}_a^{\star} \bx\cdot y_{a}(\bx) \right]\right\}\right].
\end{aligned}
\end{equation}

We also denote by $\bar{L}(\bm{\eta})$ the dual function of $\bar{V}(\delta/2)$ as follows:
\begin{equation}\label{eqn:102403}
\bar{L}(\bm{\eta})=\left(\bm{C}+\frac{\delta}{2}\right)^\top\bm{\eta}+\frac{T}{T_0}\cdot\sum_{t=1}^{T_0}\max_{\sum_{a\in[K]}y_{a,t}=1}\left\{\sum_{a\in[K]} \left[(\bmu_a^\star)^\top \bx_{t}\cdot y_{a,t}\right]- \sum_{a\in[K]} (\bm{\eta})^\top\left[ \bm{W}_a^{\star} \bx_{t}\cdot y_{a,t} \right]\right\}.
\end{equation}
Then, the function $\bar{L}(\bm{\eta})$ can be regarded as a sample average approximation of $L(\bm{\eta})$. We then proceed to bound the range of the optimal dual variable for $\VU$ and $\hat{V}$. Denote by $\bm{\eta}^*$ an optimal dual variable for $\VU$. Then, it holds that
\[
(\bm{C})^\top\bm{\eta}^* \leq \VU.
\]
Notice that this is valid only when we have the null-action. This
also implies that
\[
\bm{\eta}^*\in\Omega^*:=\left\{ \bm{\eta}\geq0:  \|\bm{\eta}\|_{1} \leq \frac{\VU}{C_{\min}}\right\}.
\]
Similarly, denote by $\bar{\bm{\eta}}^*$ an optimal dual variable for $\bar{V}(\delta/2)$ and we can obtain that
\[
\bar{\bm{\eta}}^*\in\bar{\Omega}^*:=\left\{ \bm{\eta}\geq0:  \|\bm{\eta}\|_{1} \leq \frac{\bar{V}(\delta/2)}{C_{\min}+\delta/2}\right\}.
\]
Note that 
\begin{equation}\label{eqn:102404}
\VU=L(\bm{\eta}^*) \geq \bar{L}(\bm{\eta}^*)-|L(\bm{\eta}^*)-\bar{L}(\bm{\eta}^*)|\geq \bar{L}(\bar{\bm{\eta}^*})- |L(\bm{\eta}^*)-\bar{L}(\bm{\eta}^*)|= \bar{V}(\delta/2)- |L(\bm{\eta}^*)-\bar{L}(\bm{\eta}^*)|
\end{equation}
and
\begin{equation}\label{eqn:102405}
\bar{V}(\delta/2)=\bar{L}(\bar{\bm{\eta}}^*)\geq L(\bar{\bm{\eta}}^*)-|\bar{L}(\bar{\bm{\eta}}^*)-L(\bar{\bm{\eta}}^*)|\geq L(\bm{\eta}^*)-|\bar{L}(\bar{\bm{\eta}}^*)-L(\bar{\bm{\eta}}^*)|=\VU-|\bar{L}(\bar{\bm{\eta}}^*)-L(\bar{\bm{\eta}}^*)|.
\end{equation}
%Therefore, we have
%\begin{equation}\label{eqn:102406}
%|\bar{V}(\delta/2)-\VU| \leq \max\left\{ |\bar{L}(\bar{\bm{\eta}}^*)-L(\bar{\bm{\eta}}^*)|, |\bar{L}(\bm{\eta}^*)-L(\bm{\eta}^*)| \right\}. 
%\end{equation}
Define a random variable $H(\bx)=\max_{\sum_{a\in[K]}y_{a}(\bx)=1}\left\{ \left[(\bmu_a^\star)^\top \bx\cdot y_{a}(\bx)- (\bm{\eta}^*)^\top \bm{W}_a^{\star} \bx\cdot y_{a}(\bx) \right]\right\}$ where $\bx\sim F$. It is clear to see that $|H(\bx)|\leq ( R_{\max} + \frac{\VU}{C_{\min}}\cdot B_{\max} )$ where $B_{\max}$ denotes an upper bound on $\bm{W}^\star_a \bx$ for every $a\in[K]$ and $\bx$. Then, we have %\sum_{a\in[K]}
\begin{equation}\label{eqn:102407}
\begin{aligned}
|\bar{L}(\bm{\eta}^*)-L(\bm{\eta}^*)|&=\frac{\delta}{2}\cdot\|\bm{\eta}^*\|_1+ \left|\mathbb{E}_{\bx\sim F}[H(\bx)]-\frac{T}{T_0}\cdot\sum_{t=1}^{T_0}H(\bx_{t})\right|\\
&\leq \frac{\delta}{2}\cdot\frac{\VU}{C_{\min}}+ T\cdot(R_{\max}+\frac{\VU}{C_{\min}}\cdot B_{\max})\cdot\sqrt{\frac{1}{2T_0}\cdot\log\frac{4}{\beta}}
%&\leq \frac{\delta}{2}\cdot\frac{\bar{V}(\delta/2)}{C_{\min}}+ T\cdot(R_{\max}+\frac{\VU}{C_{\min}}\cdot B_{\max})\cdot\sqrt{\frac{1}{2T_0}\cdot\log\frac{4}{\beta}}
\end{aligned}
\end{equation}
holds with probability at least $1-\frac{\beta}{2}$, where the inequality follows from the standard Hoeffding's inequality. Similarly, we have
\begin{equation}\label{eqn:102408}
\begin{aligned}
|\bar{L}(\bar{\bm{\eta}}^*)-L(\bar{\bm{\eta}}^*)|&\leq \frac{\delta}{2}\cdot\|\bar{\bm{\eta}}^*\|_1+T\cdot(R_{\max}+\frac{\bar{V}(\delta/2)}{C_{\min}+\delta/2}\cdot B_{\max})\cdot\sqrt{\frac{1}{2T_0}\cdot\log\frac{4}{\beta}}\\
&\leq \frac{\delta}{2}\cdot\frac{\bar{V}(\delta/2)}{C_{\min}+\delta/2}+ T\cdot(R_{\max}+\frac{\bar{V}(\delta/2)}{C_{\min}}\cdot B_{\max})\cdot\sqrt{\frac{1}{2T_0}\cdot\log\frac{4}{\beta}}\\
&\leq \frac{\delta}{2}\cdot\frac{\bar{V}(\delta/2)}{C_{\min}}+ T\cdot(R_{\max}+\frac{\bar{V}(\delta/2)}{C_{\min}}\cdot B_{\max})\cdot\sqrt{\frac{1}{2T_0}\cdot\log\frac{4}{\beta}}
\end{aligned}
\end{equation}
holds with probability at least $1-\frac{\beta}{2}$. From the union bound, we know that with probability at least $1-\beta$, both \eqref{eqn:102407} and \eqref{eqn:102408} hold. Therefore, from \eqref{eqn:102404} and \eqref{eqn:102405}, we have the following two inequalities
\begin{equation}\label{eqn:102409}
\bar{V}(\delta/2)-\VU\leq \frac{\delta}{2}\cdot\frac{\VU}{C_{\min}}+ T\cdot(R_{\max}+\frac{\VU}{C_{\min}}\cdot B_{\max})\cdot\sqrt{\frac{1}{2T_0}\cdot\log\frac{4}{\beta}}
\end{equation}
and
\begin{equation}\label{eqn:010801}
\VU-\bar{V}(\delta/2)\leq \frac{\delta}{2}\cdot\frac{\bar{V}(\delta/2)}{C_{\min}}+ T\cdot(R_{\max}+\frac{\bar{V}(\delta/2)}{C_{\min}}\cdot B_{\max})\cdot\sqrt{\frac{1}{2T_0}\cdot\log\frac{4}{\beta}}
\end{equation}
holds with probability at least $1-\beta$.

\noindent\textbf{Bound the term $|\bar{V}(\delta/2)-\hat{V}|$}: We first denote by $\bar{\by}$ an optimal solution to $\bar{V}(\delta/2)$. Note that
\begin{equation}\label{eqn:0104}
\norm{\frac{T}{T_0}\cdot\sum_{t=1}^{T_0} \sum_{a\in[K]} (\bm{W}_{a}^\star )\bx_{t}\cdot \bar{y}_{a,t}-\frac{T}{T_0}\cdot\sum_{t=1}^{T_0} \sum_{a\in[K]} (\widehat{\bm{W}}_{a, T_0} ) \bx_{t}\cdot \bar{y}_{a,t} }_\infty \leq T\cdot D\cdot \max_{a\in [K]} \|\bm{W}_a^\star-\widehat{\bm{W}}_{ a, T_0}\|_{\infty}.
\end{equation}
Since $\frac{\delta}{2}\geq T\cdot D\cdot \max_{a\in [K]} \|\bm{W}_a^\star-\widehat{\bm{W}}_{ a, T_0}\|_{\infty}$, we know that 
$\bar{\by}$ is a feasible solution to $\hat{V}$. Also, note that
\begin{equation}\label{eqn:102401}
\left|\frac{T}{T_0}\cdot\sum_{t=1}^{T_0} \sum_{a\in[K]} (\bmu_{a}^\star )^\top \bx_{t}\cdot \bar{y}_{a,t}-\frac{T}{T_0}\cdot\sum_{t=1}^{T_0} \sum_{a\in[K]} (\bmu_{a, T_0}^{\mathsf{s}} )^\top \bx_{t}\cdot \bar{y}_{a,t} \right|\leq T\cdot D\cdot \max_{a\in [K]} \|\bmu_a^\star-\bmu_{ a, T_0}^{\mathsf{s}}\|_{1}.
\end{equation}
Therefore, we know that
\begin{equation}\label{eqn:010802}
\bar{V}(\delta/2)\leq \frac{T}{T_0}\cdot\sum_{t=1}^{T_0} \sum_{a\in[K]} (\bmu_{a, T_0}^{\mathsf{s}} )^\top \bx_{t}\cdot \bar{y}_{a,t}+T\cdot D\cdot \max_{a\in [K]} \|\bmu_a^\star-\bmu_{ a, T_0}^{\mathsf{s}}\|_{1} \leq \hat{V}+T\cdot D\cdot \max_{a\in [K]} \|\bmu_a^\star-\bmu_{ a, T_0}^{\mathsf{s}}\|_{1}.
\end{equation}
On the other hand, we denote by $\hat{\by}$ an optimal solution to $\hat{V}$. Then, note that
\begin{equation}\label{eqn:0105}
\norm{\frac{T}{T_0}\cdot\sum_{t=1}^{T_0} \sum_{a\in[K]} (\bm{W}_{a}^\star ) \bx_{t}\cdot \hat{y}_{a,t}-\frac{T}{T_0}\cdot\sum_{t=1}^{T_0} \sum_{a\in[K]} (\widehat{\bm{W}}_{a, T_0} ) \bx_{t}\cdot \hat{y}_{a,t} }_{\infty}\leq T\cdot D\cdot \max_{a\in [K]} \|\bm{W}_a^\star-\widehat{\bm{W}}_{ a, T_0}\|_{\infty}.
\end{equation}
We have
\[
\frac{T}{T_0}\cdot\sum_{t=1}^{T_0} \sum_{a\in[K]} (\bm{W}_{a}^\star ) \bx_{t}\cdot \hat{y}_{a,t}\leq \frac{T}{T_0}\cdot\sum_{t=1}^{T_0} \sum_{a\in[K]} (\widehat{\bm{W}}_{a, T_0} ) \bx_{t}\cdot \hat{y}_{a,t}+ T\cdot D\cdot \max_{a\in [K]} \|\bm{W}_a^\star-\widehat{\bm{W}}_{ a, T_0}\|_{\infty} \leq \bC+\delta.
\]
Thus, we know that
$\hat{\by}$ is a feasible solution to $\bar{V}(\frac{3}{2}\delta)$ and again, from \eqref{eqn:102401}, it holds that
\begin{equation}\label{eqn:010803}
\hat{V}\leq \frac{T}{T_0}\cdot\sum_{t=1}^{T_0} \sum_{a\in[K]} (\bmu_{a, T_0}^{\mathsf{s}} )^\top \bx_{t}\cdot \hat{y}_{a,t}+T\cdot D\cdot \max_{a\in [K]} \|\bmu_a^\star-\bmu_{ a, T_0}^{\mathsf{s}}\|_{1} \leq \bar{V}(\frac{3}{2}\delta)+T\cdot D\cdot \max_{a\in [K]} \|\bmu_a^\star-\bmu_{ a, T_0}^{\mathsf{s}}\|_{1}.
\end{equation}
Therefore, combining \eqref{eqn:102409} and \eqref{eqn:010803}, we have
\begin{equation}\label{eqn:010804}
\hat{V}\leq \VU+ \frac{3}{2}\delta\cdot\frac{\VU}{C_{\min}}+ T\cdot(R_{\max}+\frac{\VU}{C_{\min}}\cdot B_{\max})\cdot\sqrt{\frac{1}{2T_0}\cdot\log\frac{4}{\beta}}+T\cdot D\cdot \max_{a\in [K]} \|\bmu_a^\star-\bmu_{ a, T_0}^{\mathsf{s}}\|_{1}.
\end{equation}
Also, combining \eqref{eqn:010801} and \eqref{eqn:010802}, we have
\begin{equation}\label{eqn:010805}
\VU \leq \hat{V}+\frac{\delta}{2}\cdot\frac{\bar{V}(\delta/2)}{C_{\min}}+ T\cdot(R_{\max}+\frac{\bar{V}(\delta/2)}{C_{\min}}\cdot B_{\max})\cdot\sqrt{\frac{1}{2T_0}\cdot\log\frac{4}{\beta}}+T\cdot D\cdot \max_{a\in [K]} \|\bmu_a^\star-\bmu_{ a, T_0}^{\mathsf{s}}\|_{1}.
\end{equation}
We further use the bound on $\bar{V}(\delta/2)-\VU$ in \eqref{eqn:102409} to plug in \eqref{eqn:010805}, and we obtain that
% \[\begin{aligned}
% |\hat{V}-\VU|\leq& \frac{3}{2}\delta\cdot\frac{\VU}{C_{\min}}+ (1+\delta)\cdot T\cdot(R_{\max}+\frac{\VU}{C_{\min}}\cdot B_{\max})\cdot\sqrt{\frac{1}{2T_0}\cdot\log\frac{4}{\beta}}+T\cdot D\cdot \max_{a\in [K]} \|\bmu_a^\star-\bmu_{ a, T_0}^{\mathsf{s}}\|_{1} \\
% &+\frac{\delta^2\cdot\VU}{2\cdot C_{\min}} + \frac{\delta T B_{\max}}{C_{\min}}\cdot\sqrt{\frac{1}{2T_0}\cdot\log\frac{4}{\beta}}\cdot\left( \frac{\VU}{C_{\min}} +T\cdot(R_{\max}+\frac{\VU}{C_{\min}}\cdot B_{\max})\cdot\sqrt{\frac{1}{2T_0}\cdot\log\frac{4}{\beta}} \right),
% \end{aligned}\]
\begin{equation*}
    \begin{aligned}
        \abs{\hat{V}-\VU}\leq C \frac{\VU}{C_{\min}}\left(\delta + \frac{\delta^2}{C_{\min}} + (1+\delta ) T B_{\max}\sqrt{\frac{1}{2T_0}\log\frac{4}{\beta}} + T^2 {B_{\max}}^2\frac{1}{2T_0}\log\frac{4}{\beta} \frac{1}{C_{\min}}\right) + \delta
    \end{aligned}
\end{equation*}
which completes our proof.
\end{myproof}

\subsection{Proof of Theorem \ref{thm:sqrt-margin}}

\begin{myproof}
For each $t$, denote the corresponding dual of reward as 
\begin{equation}
    D_t(\blambda,\brho) =  \max_{\bm{y}_t(\bx_t)\in\Delta^K}\big\{  \sum_{a\in[K]}(\bmu_a^\star)^\top \bx_t\cdot y_{a,t}(\bx_t) -   (\bm{W}_a^{\star} \bx_t)^{\top}\blambda\cdot y_{a,t}(\bx_t)\big\} + \brho^\top \blambda,
\end{equation}
with its expectation:
\begin{equation}
    D(\blambda,\brho) =  \bE\left[\max_{\bm{y}_t(\bx_t)\in\Delta^K}\big\{  \sum_{a\in[K]}(\bmu_a^\star)^\top \bx_t\cdot y_{a,t}(\bx_t) -   (\bm{W}_a^{\star} \bx_t)^{\top}\blambda\cdot y_{a,t}(\bx_t)\big\} \right]+ \brho^\top \blambda.
\end{equation}
By strong duality, we have $\VU = T\cdot D(\blambda^*,\frac{\bC}{T})$, and $\bE\left\langle\blambda^*,\frac{\bC}{T} - \bW^{\star}_{a^*_t} \bx_t\right\rangle = 0$. Still, we denote by $\tau$ the time period that one of the resources is depleted or $\tau=T$ if there are remaining resources at $T$. We then have:
\begin{equation}\label{eq:reg-decomp-static}
    \begin{aligned}
        &\operatorname{Regret}(\pi) = \VU  - \bE\sum_{t=1}^\tau (\bmu_{y_t}^\star)^\top \bx_t 
        \\
        & = \bE\sum_{t=1}^\tau\left[  D(\blambda^*,\frac{\bC}{T}) - (\bmu_{y_t}^\star)^\top \bx_t\right]\mathbbm{1}\{\nu_t=0\}  + \bE\sum_{t=1}^\tau\left[  D(\blambda^*,\frac{\bC}{T}) - (\bmu_{y_t}^\star)^\top \bx_t\right]\mathbbm{1}\{\nu_t=1\}+ \bE \frac{T-\tau}{T}\VU 
        \\
        & \le \underbrace{\bE \sum_{t\in[\tau]\cap\cT} \left[\left\langle \bmu_{a^*_t}^\star - (\bm{W}_{a^*_t}^{\star})^{\top}\blambda^*,\bx_t \right\rangle - 
        \left\langle \bmu_{y_t}^\star - (\bm{W}_{y_t}^{\star})^{\top}\blambda^*,\bx_t \right\rangle \right]}_{\mathfrak{A}_1} + \underbrace{\bE\sum_{t\in[\tau]\cap\cT} \left\langle\blambda^*,\frac{\bC}{T}- \bm{W}_{y_t}^{\star} \bx_t\right\rangle }_{ \mathfrak{A}_2} 
        \\
        & \quad \quad +\bE \frac{T-\tau}{T}\VU + 2R_{\max}K\sum_{t=1}^{T}\epsilon_t
    \end{aligned}
\end{equation}
We now turn to handle $\mathfrak{A}_1$ and $\mathfrak{A}_2$ in \eqref{eq:reg-decomp-static}. For $\mathfrak{A}_1$ in \eqref{eq:reg-decomp-static}, we have
\begin{equation}\label{eq:reg-decomp-static-A1}
    \begin{aligned}
      \mathfrak{A}_1  & = \bE\sum_{t\in[\tau]\cap\cT } \left[\left\langle \bmu_{a^*_t}^\star - (\bm{W}_{a^*_t}^{\star})^{\top}\blambda^*,\bx_t \right\rangle - 
        \left\langle \bmu_{y_t}^\star - (\bm{W}_{y_t}^{\star})^{\top}\blambda^*,\bx_t \right\rangle \right]\mathbbm{1}\left\{y_t\neq a^*_t \right\}
        \\
    & =  \bE\sum_{t\in[\tau]\cap\cT } \left[\left\langle \bmu_{a^*_t}^\star - (\bm{W}_{a^*_t}^{\star})^{\top}Z\cdot\bm{\eta}_{t-1},\bx_t \right\rangle - 
        \left\langle \bmu_{y_t}^\star - (\bm{W}_{y_t}^{\star})^{\top}Z\cdot\bm{\eta}_{t-1} ,\bx_t \right\rangle \right]\mathbbm{1}\left\{y_t\neq a^*_t \right\} 
        \\
        & \quad \quad + \bE\sum_{t\in[\tau]\cap\cT }\left\langle Z\cdot (\bm{\eta}_{t-1}-\bm{\eta}^*) ,(\bm{W}_{a^*_t}^{\star} - \bm{W}_{y_t}^\star)\bx_t \right\rangle
        \\
      % & = \bE \sum_{t=1}^\tau \left[\left\langle \bmu_{a^*_t}^\star - (\bm{W}_{a^*_t}^{\star})^{\top}\blambda^*,\bx_t \right\rangle - \max_{a\neq a^*_t}\left\langle \bmu_{a}^\star - (\bm{W}_{a}^{\star})^{\top}\blambda^*,\bx_t \right\rangle \right] \mathbbm{1}\left\{\right\}
    \end{aligned}
\end{equation}
By the optimality of $\blambda^*=Z\cdot\bm{\eta}^*$, we have 
\begin{equation}\label{eq:opt-lambda}
   \bE \left\langle \blambda-\blambda^* ,(\frac{\bC}{T}- \bm{W}_{a^*_t}^{\star})\bx_t \right\rangle =  Z\cdot\bE \left\langle \bm{\eta}-\bm{\eta}^* ,(\frac{\bC}{T}- \bm{W}_{a^*_t}^{\star})\bx_t \right\rangle \ge 0
\end{equation}
We also have 
\begin{equation}\label{eq:yt-at-diff}
    \begin{aligned}
        & \mathbbm{1}\left\{y_t\neq a^*_t \right\} 
        = \mathbbm{1}\left\{ \left\langle \bmu_{y_t,t-1}^{ \mathsf{s} } - (\widehat{\bm{W}}_{y_t,t-1})^{\top}Z\cdot\bm{\eta}_{t-1},\bx_t \right\rangle > \left\langle \bmu_{a^*_t,t-1}^{ \mathsf{s} } - (\widehat{\bm{W}}_{a^*_t,t-1})^{\top}Z\cdot\bm{\eta}_{t-1},\bx_t \right\rangle \right\} 
        \\
        &\le \mathbbm{1}\left\{ \left\langle \bmu_{y_t}^\star - (\bm{W}_{y_t}^{\star})^{\top} Z\cdot\bm{\eta}_{t-1},\bx_t \right\rangle > \left\langle \bmu_{a^*_t}^\star - (\bm{W}_{a^*_t}^{\star})^{\top} Z\cdot\bm{\eta}_{t-1},\bx_t \right\rangle \right.
        \\
        & \quad \quad \left.- 2DZ\sqrt{s_0}\max_{a}\norm{ \bmu_{a}^\star -\bmu_{a,t-1}^{ \mathsf{s} } }_2\vee \norm{\widehat{\bW}_{a,t-1} -\bW_{a}^{\star}}_{2,\max} \right\} \\
        &   \le \mathbbm{1}\left\{ \left\langle \bmu_{y_t}^\star - (\bm{W}_{y_t}^{\star})^{\top} Z\cdot\bm{\eta}_{t-1},\bx_t \right\rangle > \left\langle \bmu_{a^*_t}^\star - (\bm{W}_{a^*_t}^{\star})^{\top} Z\cdot\bm{\eta}_{t-1},\bx_t \right\rangle -CZ\cdot \frac{\sigma D^2 s_0^{} }{\phi_{\min}(s)  } \sqrt{\frac{ \log (dT) }{t^{1-\gamma}} }\right\} 
        \\
        & \quad +  \mathbbm{1}\left\{  \max_{a}\norm{ \bmu_{a}^\star -\bmu_{a,t-1}^{ \mathsf{s} } }_2\vee \norm{\widehat{\bW}_{a,t-1} -\bW_{a}^{\star}}_{2,\max} > C \frac{\sigma D s_0^{0.5} }{\phi_{\min}(s)  } \sqrt{\frac{ \log (dT) }{t^{1-\gamma}} } \right\}
    \end{aligned}
\end{equation}
From Theorem \ref{thm:sparse-est}, we have that
\begin{equation}\label{eq:est-high-prob-margin}
   \bp\left( \max_{a}\norm{ \bmu_{a}^\star -\bmu_{a,t-1}^{ \mathsf{s} } }_2\vee \norm{\widehat{\bW}_{a,t-1} -\bW_{a}^{\star}}_{2,\max} \le C \frac{\sigma D s_0^{0.5} }{\phi_{\min}(s)  } \sqrt{\frac{ \log (dT) }{t^{1-\gamma}} }\right) \ge 1-\frac{1}{T^2}.
\end{equation}
Plugging \eqref{eq:opt-lambda}, \eqref{eq:yt-at-diff}, \eqref{eq:est-high-prob-margin} into \eqref{eq:reg-decomp-static-A1}, we have:
\begin{equation}\label{eq:control-A1}
    \begin{aligned}
         \mathfrak{A}_1  &\le  \bE  \sum_{t\in[\tau]\cap\cT }  CZ\cdot \frac{\sigma D^2 s_0 }{\phi_{\min}(s)  } \sqrt{\frac{ \log (dT) }{t^{1-\gamma}} } \mathbbm{1}\left\{ \left\langle \bmu_{y_t}^\star - (\bm{W}_{y_t}^{\star})^{\top} Z\cdot\bm{\eta}_{t-1},\bx_t \right\rangle > \left\langle \bmu_{a^*_t}^\star - (\bm{W}_{a^*_t}^{\star})^{\top} Z\cdot\bm{\eta}_{t-1},\bx_t \right\rangle \right.
         \\
         & \quad \quad \quad \quad \quad \quad \quad \quad \quad \quad \quad \quad \quad \quad \quad \quad \quad \quad \quad \quad \quad \quad \left.  -CZ\cdot \frac{\sigma D^2 s_0 }{\phi_{\min}(s)  } \sqrt{\frac{ \log (dT) }{t^{1-\gamma}} }\right\} 
         \\
         & \quad+ T\cdot(R_{\max}+B_{\max}\frac{\VU}{C_{\min}})\frac{1}{T^2} +   \bE\sum_{t\in[\tau]\cap\cT }\left\langle Z\cdot (\bm{\eta}_{t-1}-\bm{\eta}^*) ,(\frac{\bC}{T} - \bm{W}_{y_t}^\star)\bx_t \right\rangle 
         \\
         & \le \bE  \sum_{t\in[\tau]\cap\cT }    CZ\cdot \frac{\sigma D^2 s_0^{1.5} }{\phi_{\min}(s)  } \sqrt{\frac{ \log (dT) }{t^{1-\gamma}} } \mathbbm{1}\left\{ \left\langle \bmu_{y_t}^\star - (\bm{W}_{y_t}^{\star})^{\top} Z\cdot\bm{\eta}_{t-1},\bx_t \right\rangle > \left\langle \bmu_{a^*_t}^\star - (\bm{W}_{a^*_t}^{\star})^{\top} Z\cdot\bm{\eta}_{t-1},\bx_t \right\rangle \right.
         \\
         & \quad \quad \quad \quad \quad \quad \quad \quad \quad \quad \quad \quad \quad \quad \quad \quad \quad \quad \quad \quad \quad \quad \left.  -CZ\cdot \frac{\sigma D^2 s_0 }{\phi_{\min}(s)  } \sqrt{\frac{ \log (dT) }{t^{1-\gamma}} }\right\} 
         \\
         & \quad + \bE R(\cT,\bm{\eta}^*) + (R_{\max}+B_{\max}\frac{\VU}{C_{\min}})\frac{1}{T} 
         \\
    \end{aligned}
\end{equation}

According to \eqref{eq:yt-at-diff}, we note that $\mathbbm{1}\left\{y_t\neq a^*_t \right\} $ can be well approximated by
\begin{equation*}
   \mathbbm{1}\left\{y_t\neq a^*_t \right\}  =\mathbbm{1}\left\{ \left\langle \bmu_{y_t}^\star - (\bm{W}_{y_t}^{\star})^{\top} Z\cdot\bm{\eta}_{t-1},\bx_t \right\rangle > \left\langle \bmu_{a^*_t}^\star - (\bm{W}_{a^*_t}^{\star})^{\top} Z\cdot\bm{\eta}_{t-1},\bx_t \right\rangle - \delta_{\est}\right\} +o_p(1),
\end{equation*}
where $\delta_{\est}(t) =  CZ\cdot \frac{\sigma D^2 s_0 }{\phi_{\min}(s)  } \sqrt{\frac{ \log (dT) }{t^{1-\gamma}} } $ is the high-probability bound of the estimation error, and we write is as $\delta_{\est}$ when no ambiguity. Notice that given $\nu_t=0$, the approximation of $\mathbbm{1}\left\{y_t\neq a^*_t \right\} $  can be bounded by
\begin{equation}\label{eq:exp-approx-ytat}
    \begin{aligned}
        &\bE\mathbbm{1} \left\{ \left\langle \bmu_{y_t}^\star - (\bm{W}_{y_t}^{\star})^{\top} Z\cdot\bm{\eta}_{t-1},\bx_t \right\rangle > \left\langle \bmu_{a^*_t}^\star - (\bm{W}_{a^*_t}^{\star})^{\top} Z\cdot\bm{\eta}_{t-1},\bx_t \right\rangle - \delta_{\est}\right\} 
        \\
        & = \bE \mathbbm{1} \left\{   \underbrace{\left\langle \bmu_{a^*_t}^\star - (\bm{W}_{a^*_t}^{\star})^{\top} Z\cdot\bm{\eta}^*,\bx_t \right\rangle - \left\langle \bmu_{y_t}^\star - (\bm{W}_{y_t}^{\star})^{\top} Z\cdot\bm{\eta}^*,\bx_t \right\rangle}_{\mathfrak{B}_1} < \underbrace{ Z\left\langle\bm{\eta}_{t-1}-\bm{\eta}^*,(\bm{W}_{a_t^*}^{\star} - \bm{W}_{y_t}^{\star}) \bx_t\right\rangle  }_{\mathfrak{B}_2}+    \delta_{\est} \right\}
        \\
        & \le \bE \mathbbm{1}  \left\{ \mathfrak{B}_1 < \mathfrak{B}_2 + \delta_{\est}, \mathfrak{B}_2< \delta_{\est}\right\} +  \bE \mathbbm{1}  \left\{ \mathfrak{B}_2\ge \delta_{\est}\right\}
        \\
        & \le \bp\left( \mathfrak{B}_1 <2 \delta_{\est}\right) +  \bE \mathfrak{B}_2/\delta_{\est} 
        \\
        & = \bp\left(  \left\langle \bmu_{a^*_t}^\star - (\bm{W}_{a^*_t}^{\star})^{\top} Z\cdot\bm{\eta}_{t-1},\bx_t \right\rangle - \left\langle \bmu_{y_t}^\star - (\bm{W}_{y_t}^{\star})^{\top} Z\cdot\bm{\eta}_{t-1},\bx_t \right\rangle < 2\delta_{\est}\right) 
        \\
        & \quad \quad + \bE Z\left\langle\bm{\eta}_{t-1}-\bm{\eta}^*,(\bm{W}_{a_t^*}^{\star} - \bm{W}_{y_t}^{\star}) \bx_t\right\rangle/\delta_{\est} 
        \\
        & \le 2M_{\sfm}\delta_{\est} + \bE Z\left\langle\bm{\eta}_{t-1}-\bm{\eta}^*,(\bm{W}_{a_t^*}^{\star} - \bm{W}_{y_t}^{\star}) \bx_t\right\rangle/\delta_{\est} 
        \\
        & \le  2M_{\sfm}\delta_{\est} + \bE Z\left\langle\bm{\eta}_{t-1}-\bm{\eta}^*,\frac{\bC}{T} - \bm{W}_{y_t}^{\star} \bx_t\right\rangle/\delta_{\est}.
    \end{aligned}
\end{equation}
Here in the last inequality, we use the optimality of $\blambda^*=Z\bm{\eta}^*$.
Plugging \eqref{eq:exp-approx-ytat} into \eqref{eq:control-A1} and invoking the  optional stopping theorem, we have 
\begin{equation}\label{eq:control-A1-final}
\begin{aligned}
        \mathfrak{A}_1 & \le \sum_{t=1}^T 2 M_{\sfm} \delta_{\est}^2(t) + 2Z \bE R(\cT,\bm{\eta}^*) + (R_{\max}+B_{\max}\frac{\VU}{C_{\min}})\frac{1}{T} 
    \\
    &  \le C  M_{\sfm} Z^2 \frac{\sigma^2 D^4 s_0^{2} }{\phi_{\min}^2(s)  } T^{\gamma}\log (dT) + 2Z \bE R(\cT,\bm{\eta}^*)  +1
    \\
    & \le C  M_{\sfm} Z^2 \frac{\sigma^2 D^4 s_0^{2} }{\phi_{\min}^2(s)  } T^{\gamma}\log (dT)+CZ\sqrt{\widetilde{B}_{\max} T \log m} +1
\end{aligned}
\end{equation}
provided that $T\ge R_{\max}+B_{\max}\frac{\VU}{C_{\min}}$.

We now control $\mathfrak{A}_2$ in \eqref{eq:reg-decomp-static} by
\begin{equation}\label{eq:control-A2}
\begin{aligned}
       \mathfrak{A}_2 & = \bE\sum_{t\in[\tau]\cap\cT} \left\langle\blambda^*,\frac{\bC}{T}- \bm{W}_{y_t}^{\star} \bx_t\right\rangle  =  \bE\sum_{t\in[\tau]\cap\cT} Z\left\langle\bm{\eta}^*-\bm{\eta}_{t-1},\frac{\bC}{T}- \bm{W}_{y_t}^{\star} \bx_t\right\rangle + Z\left\langle\bm{\eta}_{t-1},\frac{\bC}{T}- \bm{W}_{y_t}^{\star} \bx_t\right\rangle 
      \\ 
    & \le \bE\sum_{t\in[\tau]\cap\cT} Z\left\langle\bm{\eta}^*-\bm{\eta}_{t-1},(\bm{W}_{a_t^*}^{\star} - \bm{W}_{y_t}^{\star}) \bx_t\right\rangle + Z\left\langle\bm{\eta}_{t-1},\frac{\bC}{T}- \bm{W}_{y_t}^{\star} \bx_t\right\rangle  
    \\
    & \le \bE\sum_{t\in[\tau]\cap\cT} Z\left\langle\bm{\eta}^*-\bm{\eta}_{t-1},(\bm{W}_{a_t^*}^{\star} - \bm{W}_{y_t}^{\star}) \bx_t\right\rangle + \bE\sum_{t\in[\tau]\cap\cT} Z\left\langle\bm{\eta}_{t-1},\frac{\bC}{T}- \bm{W}_{y_t}^{\star} \bx_t\right\rangle 
    \\
    % & \quad \quad + \sum_{t=1}^T \frac{\VU}{C_{\min}}K(\frac{C_{\max}}{T}+B_{\max})\epsilon_t.
\end{aligned}
\end{equation}
Here the first inequality is because of the optimality \eqref{eq:opt-lambda}. Plugging  \eqref{eqn:091105}, \eqref{eqn:091106}, \eqref{eqn:091107} in \eqref{eq:control-A2}, with $\epsilon_t=\Theta(t^{-\gamma})$, we have:
\begin{equation}\label{eq:control-A2-2}
\begin{aligned}
       \mathfrak{A}_2 
    & \le \bE\sum_{t\in[\tau]\cap\cT} Z\left\langle\bm{\eta}^*-\bm{\eta}_{t-1},(\bm{W}_{a_t^*}^{\star} - \bm{W}_{y_t}^{\star}) \bx_t\right\rangle \mathbbm{1}\left\{y_t\neq a^*_t \right\}  -Z \bE C_{\min}\frac{T-\tau}{T} 
    \\
    & \quad \quad + Z\mathbb{E}\left[\sup_{\bm{\eta} } R(\cT,\bm{\eta}) \right]   + C\cdot\sum_{t=1}^T K R_{\max}\epsilon_t
    \\
    &\leq \bE\sum_{t\in[\tau]\cap\cT} Z\left\langle\bm{\eta}^*-\bm{\eta}_{t-1},(\bm{W}_{a_t^*}^{\star} - \bm{W}_{y_t}^{\star}) \bx_t\right\rangle\mathbbm{1}\left\{y_t\neq a^*_t \right\}  -Z \bE C_{\min}\frac{T-\tau}{T}
    \\
    & \quad \quad +Z\sqrt{\widetilde{B}_{\max} T \log m}  +CK R_{\max} T^{1-\gamma}
    \\
\end{aligned}
\end{equation}
By the selection rule, when $\nu_t=0$ we have:
\begin{equation}\label{eq:select-diff}
    \begin{gathered}
        \left\langle \bmu_{y_t,t-1}^{ \mathsf{s} } - (\widehat{\bm{W}}_{y_t,t-1})^{\top}Z\cdot\bm{\eta}_{t-1},\bx_t \right\rangle > \left\langle \bmu_{a^*_t,t-1}^{ \mathsf{s} } - (\widehat{\bm{W}}_{a^*_t,t-1})^{\top}Z\cdot\bm{\eta}_{t-1},\bx_t \right\rangle \\
        \left\langle \bmu_{a^*_t}^\star - (\bm{W}_{a^*_t}^{\star})^{\top} Z\cdot\bm{\eta}^*,\bx_t \right\rangle   >  \left\langle \bmu_{y_t}^\star - (\bm{W}_{y_t}^{\star})^{\top} Z\cdot\bm{\eta}^*,\bx_t \right\rangle
    \end{gathered}
\end{equation}
From \eqref{eq:select-diff}, we can derive:
\begin{equation}\label{eq:diff-by-selection}
    \begin{aligned}
        \left\langle  (\widehat{\bm{W}}_{a^*_t,t-1}-\widehat{\bm{W}}_{y_t,t-1})^{\top}Z\cdot\bm{\eta}_{t-1},\bx_t \right\rangle & > \left\langle (\bm{W}_{a^*_t}^{\star}-\bm{W}_{y_t}^{\star})^{\top} Z\cdot\bm{\eta}^*,\bx_t \right\rangle  
        \\
        & \quad + \left\langle \bmu_{y_t}^\star - \bmu_{y_t,t-1}^{ \mathsf{s} } ,\bx_t \right\rangle + \left\langle \bmu_{a^*_t,t-1}^{ \mathsf{s} } - \bmu_{a^*_t}^\star,\bx_t \right\rangle 
        \\
       \left\langle (\bm{W}_{a^*_t}^{\star}-\bm{W}_{y_t}^{\star})^{\top} Z\cdot(\bm{\eta}_{t-1}-\bm{\eta}^*),\bx_t \right\rangle   & > - 4DZ\sqrt{s_0}\max_{a}\norm{ \bmu_{a}^\star -\bmu_{a,t-1}^{ \mathsf{s} } }_2\vee \norm{\widehat{\bW}_{a,t} -\bW_{a}^{\star}}_{2,\max} 
    \end{aligned}
\end{equation}
Plugging \eqref{eq:diff-by-selection} into \eqref{eq:control-A2-2}, and use the high-probability trunction \eqref{eq:est-high-prob-margin} with the bound \eqref{eq:exp-approx-ytat}, we have the bound for $\mathfrak{A}_2$:
\begin{equation}\label{eq:control-A2-final}
\begin{aligned}
         \mathfrak{A}_2 
    & \le   \sum_{t=1}^T 2M_{\sfm}\delta_{\est}^2(t)  + \bE \sum_{t=1}^\tau  Z\left\langle\bm{\eta}_{t-1}-\bm{\eta}^*,\frac{\bC}{T} - \bm{W}_{y_t}^{\star} \bx_t\right\rangle + \frac{\widetilde{B}_{\max}}{T}
    \\
    & \quad \quad -Z \bE C_{\min}\frac{T-\tau}{T} +Z\sqrt{\widetilde{B}_{\max} T \log m}  +CK R_{\max} T^{1-\gamma}
    \\ 
    & \le   C   M_{\sfm} Z^2 \frac{\sigma^2 D^4 s_0^{2} }{\phi_{\min}^2(s)  } T^{\gamma}\log (dT)+CZ\sqrt{\widetilde{B}_{\max} T \log m} +1 
    \\
    &\quad \quad -Z \bE C_{\min}\frac{T-\tau}{T} +Z\sqrt{\widetilde{B}_{\max} T \log m}  +CK R_{\max} T^{1-\gamma}
\end{aligned}
\end{equation}

Combining the bounds for $\mathfrak{A}_1$,  $\mathfrak{A}_2$ in \eqref{eq:control-A1-final}, \eqref{eq:control-A2-final}, we yield the following control for \eqref{eq:reg-decomp-static}:
\begin{equation*}
    \begin{aligned}
        &\operatorname{Regret}(\pi) = \VU  - \bE\sum_{t=1}^\tau (\bmu_{y_t}^\star)^\top \bx_t 
        \\
        & \le {\mathfrak{A}_1} + { \mathfrak{A}_2} 
         +\bE \frac{T-\tau}{T}\VU + 2R_{\max}K\sum_{t=1}^{T}\epsilon_t
         \\
         & \le  C  M_{\sfm} Z^2 \frac{\sigma^2 D^4 s_0^{2} }{\phi_{\min}^2(s)  } T^{\gamma}\log (dT)+CZ\sqrt{\widetilde{B}_{\max} T \log m} 
         \\
         & \quad \quad +\bE \frac{T-\tau}{T}\VU -Z \bE C_{\min}\frac{T-\tau}{T}  +CK R_{\max} T^{1-\gamma}+1
    \end{aligned}
\end{equation*}
Setting 
$$\gamma= \frac{1}{2}\left[1+\log\left(\frac{KR_{\max}}{M_{\sfm}Z^2 \sigma^2 D^4 s_0^{2}\log (dT)/\phi_{\min}^2(s)   }\right)/\log T\right] , $$ 
and assuming that $\frac{\VU}{C_{\min}}\le Z\le C(\frac{\VU}{C_{\min}}+1)$, we have

\begin{equation*}
    \begin{aligned}
        \operatorname{Regret}(\pi)&\le C Z \frac{\sigma D^2 s_0 }{\phi_{\min}(s)  } \sqrt{M_{\sfm} K R_{\max} T\log (dT)}+CZ\sqrt{\widetilde{B}_{\max} T \log m} +1 \\
        & = O\left( (\frac{\VU}{C_{\min}}+1)\left( \frac{\sigma D^2 s_0 }{\phi_{\min}(s)  }\sqrt{M_{\sfm} K R_{\max} T\log (dT)}+ \sqrt{\widetilde{B}_{\max} T \log m}\right) \right)
    \end{aligned}
\end{equation*}
\end{myproof}

\subsection{Proof of Theorem \ref{thm:bandit-poor} and \ref{thm:bandit-rich}}
\begin{myproof}

Our proof essentially follows the basic ideas of regret analysis for $\epsilon$-greedy algorithms, with a fine-grained process on the estimation error. For the $\epsilon$-greedy algorithm, we have
\begin{equation*}
    \begin{aligned}
        &\operatorname{Regret} = \bE\left[\sum_{t=1}^{T} \left\langle \bx_t, \bmu_{\mathsf{opt} }(\bx_t) \right\rangle - \sum_{t=1}^{T} \left\langle \bx_t, \bmu_{y_t }^\star \right\rangle \right] \\
        &= \bE\left[\sum_{t=1}^{T} \left\langle \bx_t, \bmu_{\mathsf{opt} }(\bx_t)  - \bmu_{\mathsf{opt},t-1 }^{\mathsf{s} }(\bx_t)\right\rangle - \sum_{t=1}^{T} \left\langle \bx_t, \bmu_{y_t^*,t-1 }^{ \mathsf{s}}  - \bmu_{\mathsf{opt},t-1 }^{ \mathsf{s} }(\bx_t)\right\rangle \right. \\
        & \ \ +  \left. \left\langle \bx_t, \bmu_{y_t^*,t-1 }^{ \mathsf{s}}  - \bmu_{y_t^*}^\star\right\rangle +  \left\langle \bx_t, \bmu_{y_t^* }^\star  - \bmu_{y_t}^\star\right\rangle \right]\\
        & \le \sum_{t=1}^{T} \bE \norm{\bx_t}_\infty \left( \norm{ \bmu_{\mathsf{opt} }(\bx_t)  - \bmu_{\mathsf{opt},t-1 }^{\mathsf{s} }(\bx_t) }_1  + \norm{ \bmu_{y_t^* }^{ \mathsf{s}}  - \bmu_{y_t^*,t-1}^\star }_1 \right) +   2\sum_{t=1}^{T} K \epsilon_t R_{\max}
    \end{aligned}
\end{equation*}
where $y_t^*$ means the greedy action $y_t^*= \arg\max_{a\in[K] }  \left\langle \bx_t,\bmu^{\mathsf{s} }_{a,t-1} \right\rangle $, and $\bmu_{\mathsf{opt},t-1 }^{\mathsf{s} }(\bx_t)$ indicates the estimation of the optimal arm $\bmu_{\mathsf{opt} }(\bx_t)$. The inequality uses the fact of greedy action, and the uniform risk bound. 
This leads to the regret-bound 
\begin{equation*}
\begin{aligned}
        \operatorname{Regret} & \le 2D \sum_{t=1}^{T} \bE \sqrt{s_0} \max_{a} \norm{\bmu_{a,t}^{\mathsf{s} } -\bmu_a^{\star} }_2  +   2\sum_{t=1}^{T} K\epsilon_t R_{\max} \\
        & \lesssim \frac{\sigma D^2 s_0 \sqrt{\log (dK)  } }{\phi_{\min}(s)  } \sum_{t=1}^T  \left( \frac{1}{t}\sqrt{ \sum_{j=1}^{t} \frac{1}{
        \epsilon_j } } \right)  + \sum_{t=1}^T K \epsilon_t R_{\max}  .
\end{aligned}
\end{equation*}
Choosing $\epsilon_t=\sigma^{\frac{2}{3}} D^{\frac{4}{3}} s_0^{\frac{2}{3}} (\log (dK))^{\frac{1}{3}}t^{-\frac{1}{3}} / \left(K R_{\max} \right)^{\frac{2}{3}} \wedge 1/K$, the statement in Theorem \ref{thm:bandit-poor} can be justified. For the Theorem \ref{thm:bandit-rich}, since it can be viewed as a special case of $\epsilon$-greedy strategy (with $\epsilon=0$), we have
    \begin{equation*}
          \operatorname{Regret}  \le 2D \sum_{t=1}^{T} \bE \max_a \abs{\left\langle \bx_t, \bmu_{a,t-1 }^{ \mathsf{s}}  - \bmu_{a}^\star\right\rangle},
    \end{equation*}  
    where the estimation error can be guaranteed by
    \begin{equation}\label{eq:max-est-divcov}
        \bE \max_a\norm{\bmu_{a,t}^{ \mathsf{s}}-\bmu_{a}^{\star} }_2^2 \lesssim \frac{\sigma^2 D^2 s_0 }{\gamma^2(K)\zeta^2(K)  } \frac{\log (dK) }{t}.
    \end{equation}
This error bound can be easily derived from the proof of Theorem \ref{thm:sparse-est-divcov}. Here each term $\max_a \abs{\left\langle \bx_t, \bmu_{a,t-1 }^{ \mathsf{s}} - \bmu_{a}^\star\right\rangle}$ in the regret can be controlled by two ways:
\begin{equation}\label{eq:bandit-l1bound}
    \bE \max_a \abs{\left\langle \bx_t, \bmu_{a,t-1 }^{ \mathsf{s}} - \bmu_{a}^\star\right\rangle} \le D \bE \max_a \norm{\bmu_{a,t-1}-\bmu_{a}^{\star} }_1,
\end{equation}
and 
\begin{equation}\label{eq:bandit-l2bound}
\begin{aligned}
       & \bE \left[\max_a \abs{\left\langle \bx_t, \bmu_{a,t-1 }^{ \mathsf{s}} - \bmu_{a}^\star \right\rangle}-\bE \abs{\left\langle \bx_t, \bmu_{a,t-1 }^{ \mathsf{s}} - \bmu_{a}^\star \right\rangle} \right] \\
        &\le \int_0^{\infty} \bp\left( \max_a \abs{\left\langle \bx_t, \bmu_{a,t-1 }^{ \mathsf{s}} - \bmu_{a}^\star \right\rangle}-\bE \abs{\left\langle \bx_t, \bmu_{a,t-1 }^{ \mathsf{s}} - \bmu_{a}^\star \right\rangle} \ge z \right) dz
\end{aligned}
\end{equation}
Combining \eqref{eq:max-est-divcov} with \eqref{eq:bandit-l1bound}, it is easy to show that the regret bound:
\begin{equation*}
     \operatorname{Regret}  \le 2 D \sum_{t=1}^{T} \bE \max_a \abs{\left\langle \bx_t, \bmu_{a,t-1 }^{ \mathsf{s}}  - \bmu_{a}^\star\right\rangle} \lesssim \frac{\sigma D^2 s_0 \sqrt{\log (dK)  T} }{\gamma(K)\zeta(K)  }.
\end{equation*}
We use \eqref{eq:bandit-l2bound} to give another bound. Notice that $\bx_t$ is independent of the history $\cH_{t-1}$, which implies that, conditional on the history $\cH_{t-1}$,
\begin{equation*}
    \begin{aligned}
        \bE\abs{\left\langle \bx_t, \bmu_{a,t-1 }^{ \mathsf{s}}  - \bmu_{a}^\star\right\rangle} & \le\sqrt{ \bE\left(\bmu_{a,t-1}^{ \mathsf{s}}-\bmu_{a}^{\star}\right)^\top \bx_t \bx_t^\top \left(\bmu_{a,t-1}^{ \mathsf{s}}-\bmu_{a}^{\star}\right)  }\le \sqrt{ \norm{\bmu_{a,t-1}^{ \mathsf{s}}-\bmu_{a}^{\star}}_{\bSigma}^2 } . \\
        & \le \sqrt{\phi_{\max}(s_0)} \norm{\bmu_{a,t-1}^{ \mathsf{s}}-\bmu_{a}^{\star}}_2.
    \end{aligned}
\end{equation*}
Since $\bx_t$ is marginal sub-Gaussian, the $\abs{\left\langle \bx_t, \bmu_{a,t-1 }^{ \mathsf{s}}  - \bmu_{a}^\star\right\rangle}$ has a tail behavior by Chernoff bound:
\begin{equation*}
    \bp\left( \abs{\left\langle \bx_t, \bmu_{a,t-1 }^{ \mathsf{s}}  - \bmu_{a}^\star\right\rangle}-\bE  \abs{\left\langle \bx_t, \bmu_{a,t-1 }^{ \mathsf{s}}  - \bmu_{a}^\star\right\rangle} \ge z \right)\le \exp\left(-\frac{cz^2 }{ \phi_{\max}(s_0) \norm{\bmu_{a,t-1 }^{ \mathsf{s}}  - \bmu_{a}^\star}_2^2 }\right),
\end{equation*}
and also
\begin{equation*}
\begin{aligned}
         &\bp\left( \max_a \abs{\left\langle \bx_t, \bmu_{a,t-1 }^{ \mathsf{s}}  - \bmu_{a}^\star\right\rangle}-\bE  \abs{\left\langle \bx_t, \bmu_{a,t-1 }^{ \mathsf{s}}  - \bmu_{a}^\star\right\rangle} \ge z \right)\\
         &\le  1 \wedge \exp\left(\log K-\frac{cz^2 }{ \phi_{\max}(s_0) \max_a \norm{\bmu_{a,t-1 }^{ \mathsf{s}}  - \bmu_{a}^\star}_2^2 }\right).
\end{aligned}
\end{equation*}
% $\norm{\bmu_{a,t-1}^{ \mathsf{s}}-\bmu_{a}^{\star}}_1$, 
% and second-order moment bound $\phi_{\max}(s)\norm{\bmu_{a,t-1}^{ \mathsf{s}}-\bmu_{a}^{\star}}_2^2 $, by Bernstein tail inequality, we have 
This instantly gives rise to the maxima inequality by \eqref{eq:bandit-l2bound} 
\begin{equation*}
\begin{aligned}
       & \bE \left[\max_a \abs{\left\langle \bx_t, \bmu_{a,t-1 }^{ \mathsf{s}} - \bmu_{a}^\star \right\rangle}-\bE \abs{\left\langle \bx_t, \bmu_{a,t-1 }^{ \mathsf{s}} - \bmu_{a}^\star \right\rangle} \right] \\
        &\le \int_0^{\infty}  1 \wedge \exp\left(\log K-\frac{cz^2 }{ \phi_{\max}(s_0) \max_a \norm{\bmu_{a,t-1 }^{ \mathsf{s}}  - \bmu_{a}^\star}_2^2 }\right) dz \\
        & \lesssim \sqrt{\log K \phi_{\max}(s_0)}  \max_a \norm{\bmu_{a,t-1 }^{ \mathsf{s}}  - \bmu_{a}^\star}_2
\end{aligned}
\end{equation*}
We thus have 
\begin{equation*}
\begin{aligned}
       &\bE\max_a \abs{\left\langle \bx_t, \bmu_{a,t-1 }^{ \mathsf{s}} - \bmu_{a}^\star \right\rangle} \\
       & \le   \bE \left[\max_a \abs{\left\langle \bx_t, \bmu_{a,t-1 }^{ \mathsf{s}} - \bmu_{a}^\star \right\rangle}-\bE \abs{\left\langle \bx_t, \bmu_{a,t-1 }^{ \mathsf{s}} - \bmu_{a}^\star \right\rangle} \right] +    \max_a
    \bE\abs{\left\langle \bx_t, \bmu_{a,t-1 }^{ \mathsf{s}}  - \bmu_{a}^\star\right\rangle} \\
    & \lesssim \sqrt{\log K \phi_{\max}(s_0)}  \max_a \norm{\bmu_{a,t-1 }^{ \mathsf{s}}  - \bmu_{a}^\star}_2,
\end{aligned}
\end{equation*}
conditional on the history $\cH_{t-1}$. Together with the estimation error \eqref{eq:max-est-divcov}, we can derive another regret bound:

    \begin{equation*}
\begin{aligned}
              \operatorname{Regret}  &\le 2D \sum_{t=1}^{T} \bE \max_a \abs{\left\langle \bx_t, \bmu_{a,t-1 }^{ \mathsf{s}}  - \bmu_{a}^\star\right\rangle} \lesssim \sqrt{\log K \phi_{\max}(s_0)}  \frac{\sigma D  \sqrt{s_0 \log (dK)  T} }{\gamma(K)\zeta(K)  } \\
              & \lesssim   \frac{\sqrt{\kappa_1} \sigma D  \sqrt{s_0 \log K\log (dK)  T} }{ \sqrt{\gamma(K)\zeta(K)}  } 
\end{aligned}
    \end{equation*}  
Associate these two regret bounds, we finish the proof.
    
\end{myproof}
\subsection{Proof of Theorem \ref{thm:sparse-est-divcov}}\label{sec:proof-sparse-est-divcov}

\begin{myproof}
    The proof shares a similar fashion with the proof of Theorem \ref{thm:sparse-est}. The key difference is that, instead of focusing on the concentration of the gradient $\bg_{a,t}$ to the population version $\nabla f^{a}(\bmu_{a,t-1} )$, we consider a series of new objective functions $\{f_t^{a}\}$ that is changing over time, and derive the concentration of  $\bg_{a,t}$ to $\nabla f^a_t(\bmu_{t-1})$. To this end, we defined the history-dependent covariance matrices $\bE \left[\bx_t \bx_t^{\top}\cdot \bI\left\{y_t=a \right\} \middle| \cH_{t-1}\right]$, and their average: $\Bar{\bSigma}_{a,t} = \sum_{j=1}^{t}\bE \left[\bx_j \bx_j^{\top}\cdot \bI\left\{y_j=a \right\} \middle| \cH_{j-1}\right]/t$. We write the corresponding objective function that $\Bar{\bSigma}_{a,t}$ represents as $f^a_t(\bmu) = \norm{\bmu-\bmu_{a}^{\star} }_{\Bar{\bSigma}_{a,t}}^2$. 
    In the following proof, since we will mainly focus on one arm, we will write $\bmu_t$, $\bmu_{\star}$, $\bg_t$, $f_t$, $\widehat{\bSigma}_{t}$, $\Bar{\bSigma}_{t}$ etc instead of $\bmu_{a,t}$, $\bmu_{a}^{\star}$, $\bg_{a,t}$, $f^{a}_t$, $\widehat{\bSigma}_{a,t}$ and $\Bar{\bSigma}_{a,t}$, etc to easy the notation. An argument analog to the proof of Theorem \ref{thm:sparse-est} gives that:

\begin{equation}\label{eq:sparse-est-divcov-itr}
    \begin{aligned}
        \norm{\bmu_t-\bmu_\star}_2^2 \le & \left(1+\frac{3}{2}\sqrt{\varrho } \right) \left( \norm{\bmu_{t-1}-\bmu_{\star}}_2^2-  2\beta_t \left\langle \cP_{\Omega}(\bg_t), \bmu_{t-1} -\bmu_\star \right\rangle +  \beta_t^2 \norm{\cP_{\Omega}(\bg_t) }_2^2\right) \\
         \le &\left(1+\frac{3}{2}\sqrt{\varrho } \right) \left( \norm{\bmu_{t-1}-\bmu_{\star}}_2^2-  2\beta_t \left\langle \nabla f_t(\bmu_{t-1}) , \bmu_{t-1} -\bmu_\star \right\rangle +  2\beta_t^2 \norm{\cP_{\Omega}(\bg_t-\nabla f_t(\bmu_{t-1}) ) }_2^2\right.  \\
         & \left. +2\beta_t^2 \norm{\cP_{\Omega}(\nabla f_t(\bmu_{t-1})) }_2^2 + 2\beta_t \norm{\cP_{\Omega}(\bg_t-\nabla f_t(\bmu_{t-1}) ) }_2\norm{\bmu_{t-1}-\bmu_{\star}}_2  \right),
    \end{aligned}
\end{equation}
where we use the fact that $\left\langle \nabla f_t(\bmu_{t-1}) , \bmu_{t-1} -\bmu_\star \right\rangle=\left\langle \cP_{\Omega}(\nabla f_t(\bmu_{t-1})) , \bmu_{t-1} -\bmu_\star \right\rangle$ by the definition of $\cP_{\Omega}(\cdot)$. Because we are interested in the new objective function $f_t(\bmu) = \norm{\bmu-\bmu_{\star} }_{\Bar{\bSigma}_{t}}^2$, we need to check the sparse eigenvalue of $\Bar{\bSigma}_{t}$. Since for any $\bbeta$ such that $\|\bbeta\|_{0} \leq\lceil 2s\rceil$, we have $ \bbeta^\top \bE \left[\bx_t \bx_t^{\top}\cdot \bI\left\{y_t=a \right\} \middle| \cH_{t-1}\right] \bbeta \le \bbeta^{\top}\bE \left[\bx_t \bx_t^{\top} \middle| \cH_{t-1}\right] \bbeta \le \phi_{\max}(s) \norm{\bbeta}_2^2 $, then it is clear that the $2s$-sparse maximal eigenvalue of $\Bar{\bSigma}_{t} = \sum_{j=1}^{t}\bE \left[\bx_j \bx_j^{\top}\cdot \bI\left\{y_j=a \right\} \middle| \cH_{j-1}\right]/t$ is bounded by $\phi_{\max}(s)$. For the minimum eigenvalue, it follows by Assumption \ref{asm:diverse-cov} that given any unit vector $\bv$,
\begin{equation}\label{eq:sp-eigen-divcov}
  \begin{aligned}
        \bv^\top \bE \left[ \bx_t \bx_t^\top \bI\left\{y_t=a \right\}  \middle| \cH_{t-1}\right]  \bv & \ge  \bE \left[\bv^\top \bx_t \bx_t^\top \bv \bI\left\{y_t=a \right\}  \bI\left\{ \bv^\top \bx_t \bx_t^\top \bv \bI\left\{y_t=a \right\} \ge \gamma(K)  \right\} \middle|  \cH_{t-1}\right]  \\
        & \ge \bE \left[\gamma(K)  \bI\left\{ \bv^\top \bx_t \bx_t^\top \bv \bI\left\{y_t=a \right\} \ge \gamma(K)  \right\} \middle| \cH_{t-1} \right] \\
        & \ge \gamma(K)\zeta(K).
  \end{aligned}
\end{equation}
It is clear that the $2s$-sparse minimum eigenvalue of $\Bar{\bSigma}_{t}$ can be lower bounded by $\gamma(K)\zeta(K)$. We therefore take the condition number of $\Bar{\bSigma}_{t}$ as $\kappa_1 = \frac{\phi_{\max}(s) }{\gamma(K)\zeta(K)}$. The eigenvalues of $\Bar{\bSigma}_{t}$ also imply:
\begin{equation*}
    \begin{gathered}
        \left\langle \nabla f_t(\bmu_{t-1}) , \bmu_{t-1} -\bmu_\star \right\rangle \ge 2 \gamma(K)\zeta(K) \norm{ \bmu_{t-1} -\bmu_\star }_2^2, \\
        \norm{\cP_{\Omega}(\nabla f_t (\bmu_{t-1})) } \le 2 \phi_{\max}(s)\norm{\bmu_{t-1} -\bmu_\star}_2.
    \end{gathered}
\end{equation*}
We can show that 
\begin{equation}\label{eq:divcov-recur}
    \begin{aligned}
        \norm{\bmu_t-\bmu_\star}_2^2  \le &\left(1+\frac{3}{2}\sqrt{\varrho } \right) \left( 1-4\gamma(K)\zeta(K)  \beta_t+8\beta_t^2\phi_{\max}^2(s) \right) \norm{\bmu_{t-1} -\bmu_\star}_2^2 \\
        & + 6 \beta_t^2 \norm{\cP_{\Omega}(\bg_t-\nabla f_t(\bmu_{t-1}) ) }_2^2  + 6 \beta_t \norm{\cP_{\Omega}(\bg_t-\nabla f_t(\bmu_{t-1}) ) }_2\norm{\bmu_{t-1}-\bmu_{\star}}_2\\
        \le & \left(1+\frac{3}{2}\sqrt{\varrho } \right) \left( 1-4\gamma(K)\zeta(K)  \beta_t+8\beta_t^2\phi_{\max}^2(s) \right) \norm{\bmu_{t-1} -\bmu_\star}_2^2  \\
        & + 18 s\beta_t^2  \max_{i\in [d] } \abs{\left\langle \bg_t -\nabla f_t({\bmu}_{t-1}), {\bm e}_i  \right\rangle}^2  + 18 \beta_t \sqrt{s} \max_{i\in [d] } \abs{\left\langle \bg_t -\nabla f_t({\bmu}_{t-1}), {\bm e}_i  \right\rangle}\norm{\bmu_{t-1}-\bmu_{\star}}_2\\
    \end{aligned}
\end{equation}
The following lemma, which echoes with aforementioned Lemma \ref{lemma:max-grad}, quantifies the variation of the averaged stochastic gradient under the diverse covariate condition without $\varepsilon$-greedy strategy:

\begin{lemma}\label{lemma:max-grad-divcov} Define $\{\bm e_i\}_1^{d}$ as the canonical basis of $\bR^d$. Under Assumption \ref{asm:size-of-prob}, \ref{asm:bounded-ftr}, the variance of stochastic gradient $\bg_t$ at the point ${\bmu}_{t-1}$ given in Algorithm \ref{alg:online-iht} can be  bounded by the following inequality:
    \begin{equation}
        \bE \max_{i\in [d] } \abs{\left\langle \bg_t -\nabla f_t({\bmu}_{t-1}), {\bm e}_i  \right\rangle}^2 \le  C \frac{sD^2\log(dt)}{t}\bE \norm{\bmu_{t-1} -\bmu_{\star} }_2^2 + C\frac{\sigma^2 D^2 \log d }{t}.
    \end{equation}
    Moreover, the following inequality also holds with probability at least $1-\epsilon$
\begin{equation*}
    \max_{i\in [d] } \abs{\left\langle \bg_t -\nabla f_t({\bmu}_{t-1}), {\bm e}_i  \right\rangle}^2 \le C s D^2\frac{\log (d/\epsilon)  }{t} \norm{\bmu_{t-1} -\bmu_{\star} }_2^2 +  C\frac{\sigma^2 D^2\log(d/\epsilon) }{t}.
\end{equation*}
\end{lemma}
We defer the proof of Lemma \ref{lemma:max-grad-divcov} to the next section.

We set $\varrho = \frac{1}{36\kappa_1^4}$, and $\beta_t= \frac{1}{4\kappa_1\phi_{\max}(s) } $. Plugging in the expectation bound in Lemma \ref{lemma:max-grad-divcov}, we have
\begin{equation*}
    \begin{aligned}
        \bE\norm{\bmu_t-\bmu_\star}_2^2  
        \le & \left( 1-\frac{1}{4\kappa_1^2} +  C \frac{s_0 D\sqrt{\log(dt)}}{ \gamma(K)\zeta(K)
        \sqrt{t} }  \right)   \bE\norm{\bmu_{t-1} -\bmu_\star}_2^2  \\
      & +  C\frac{s_0\sigma^2 D^2 \log d }{\gamma^2(K)\zeta^2(K) t } + C \sqrt{ \frac{s_0\sigma^2 D^2 \log d }{\gamma^2(K)\zeta^2(K) t}\bE\norm{\bmu_{t-1} -\bmu_\star}_2^2}.
    \end{aligned}
\end{equation*}
Following the proof of  of Theorem \ref{thm:sparse-est} in Section \ref{sec:proof-thm1}, it is clear that when $t$ is larger that $C(\kappa_1^4\frac{s_0^2 D^2 \log(dT) }{ \gamma^2(K)\zeta^2(K)
         } +\kappa_1^2\log s_0)\asymp  \kappa_1^4\frac{s_0^2 D^2 \log(dT) }{ \gamma^2(K)\zeta^2(K)
         } $, essentially we have 

\begin{equation*}
    \begin{aligned}
        \bE\norm{\bmu_t-\bmu_\star}_2^2  
        &\le  \left( 1-\frac{1}{5\kappa_1^2}  \right)   \bE\norm{\bmu_{t-1} -\bmu_\star}_2^2  \\
      & \quad +  C\frac{s_0\sigma^2 D^2 \log d }{\gamma^2(K)\zeta^2(K) t} + C \sqrt{ \frac{s_0\sigma^2 D^2 \log d }{\gamma^2(K)\zeta^2(K) t}\bE\norm{\bmu_{t-1} -\bmu_\star}_2^2} 
      \\
      &\le  \left( 1-\frac{1}{6\kappa_1^2}  \right)   \bE\norm{\bmu_{t-1} -\bmu_\star}_2^2 + C\frac{\kappa_1^2 s_0\sigma^2 D^2 \log d }{\gamma^2(K)\zeta^2(K) t}.
    \end{aligned}
\end{equation*}
This instantly gives us the expectation bound
    \begin{equation*}
        \bE \norm{\bmu_t-\bmu_\star }_2^2 \lesssim \frac{\sigma^2 D^2 s_0 }{\gamma^2(K)\zeta^2(K) } \frac{\log d }{t},
    \end{equation*}
which proves the first claim. Apply  Lemma \ref{lemma:max-grad-divcov} again to the recursive relationship in \eqref{eq:divcov-recur}, we also have the second claim:
  \begin{equation*}
        \norm{\bmu_t-\bmu_\star }_2^2 \lesssim \frac{\sigma^2 D^2 s_0 }{ \gamma^2(K)\zeta^2(K) } \frac{\log (dT/\varepsilon) }{t} 
    \end{equation*}
holds for $t\gtrsim\kappa_1^4\frac{s_0^2 D^2 \log(dT) }{ \gamma^2(K)\zeta^2(K)
         } $ with probability at least $1-\epsilon$.
    
\end{myproof}

\subsection{Dual convergence with sparse estimation}
\begin{theorem}[Dual convergence with sparse estimation]\label{thm:est-dual-convergence}
    Under Assumption \ref{assump:general}, \ref{asm:local-dual-conditions}, \ref{asm:dual-regularity-log-1}, suppose  the primal estimation from time $t-1$ satisfies the error rate 
    $$
\max_{a}\norm{\bmu_{a,t-1}^{ \mathsf{s} } - \bmu_{a}^\star  }_2\vee \norm{\widehat{\bW}_{a,t-1} -\bW_{a}^{\star}}_{2,\max}\le\delta_{\est},
$$
for some $\delta_{\est}\gtrsim \sqrt{s_0mK^2\log^2K\log d/t}$ with probability at least $1-T^{-2}$.  Then at time $t$,  the solution to the dual problem with any $\brho$, i.e.,
\begin{equation*}
    \widehat{\blambda }_t(\brho) = \arg\min_{\blambda \succeq \bm{0}, \norm{\blambda}_1\le Z} \bar{D}_t(\bmu_{a,t}^{\mathsf{s} },\widehat{\bm{W}}_{a},\blambda,\brho) : = \frac{1}{t}\sum_{j=1}^t \max_{a\in[K]\cup\{0\}}\left\langle \bmu_{a,t}^{\mathsf{s} } - (\widehat{\bm{W}}_{a})^{\top}\blambda,\bx_j \right\rangle + \brho^\top \blambda
\end{equation*}
satisfies
\begin{equation*}
    \sup_{\brho\in \cB^+(\brho_0,\delta_\rho')}\norm{\widehat{\blambda}_t (\brho) - \blambda^*(\brho) }_1\le \varepsilon
\end{equation*}
for any  $\varepsilon\ge C B_{\max} M_{\sfm }\sqrt{s_0}DZ\delta_{\est}/\cL$, with probability at least $1-2m \exp \left(-\frac{t \cL^2\varepsilon^2}{C B_{\max}^2}\right)-2T^{-2}$.
\end{theorem}
Notice that here we allow the average constraint $\brho$ to depend on the past data, which is necessary for the analysis. 

\begin{theorem}[primal estimation]\label{thm:est-induction}  Suppose Assumption \ref{assump:general}, \ref{asm:local-dual-conditions}, \ref{asm:dual-regularity-log-1} hold and $\{\brho_{j}\}_{j=1}^{t-1}$ are within $\cB^+(\brho_0,\delta_{\rho}')$. Choosing $T_0=\widetilde{C}Z^2\log(mKdT)$,
We have
        $$
\max_{a}\norm{\bmu_{a,t}^{ \mathsf{s} } - \bmu_{a}^\star  }_2^2\vee \norm{\widehat{\bW}_{a,t} -\bW_{a}^{\star}}_{2,\max}^2\le C\frac{ 
\sigma^2 D^2{s_0\log(mKdT)}}{ \gamma^2(K)\zeta^2(K) t }
$$
for all $t\ge \kappa^4 s_0^2 D^2\log(mKdT)/\phi_{\min}^2(s)$ with probability at least $1-T^{-2}$, provided that the algorithm does not stop at $t$.
\end{theorem}

\subsection{Proof of Theorem \ref{thm:est-dual-convergence}}\label{sec:proof-est-dual-conv}
In this section, we write $\delta_{\rho}$ as as the smaller one between the original $\delta_{\rho}$ used in Assumption \ref{asm:local-dual-conditions}, etc, and $\delta_{\rho}'$ defined in Lemma \ref{lemma:region-B+} to ease the notation.
\subsubsection{Preliminary bound by dual convergence.}
For any $\{(\bmu_a,\bW_a)\}_{a\in[K]}$ and $\blambda$, we write the population and sample versions of dual problems as:
\begin{equation*}
\begin{gathered}
        D(\bmu,\bW,\blambda,\brho) =  \bE_{\bx}\left[\max_{a\in[K]\cup\{0\}}\left\langle \bmu_{a} - (\bm{W}_{a})^{\top}\blambda,\bx_t \right\rangle \right]+ \brho^\top \blambda.  \\
        \bar{D}_t(\bmu,\bW,\blambda,\brho) := \frac{1}{t}\sum_{j=1}^t D_j(\bmu,\bW,\blambda,\brho) = \frac{1}{t}\sum_{j=1}^t \max_{a\in[K]\cup\{0\}}\left\langle \bmu_{a} - (\bm{W}_{a})^{\top}\blambda,\bx_j \right\rangle + \brho^\top \blambda,
\end{gathered}
\end{equation*}
When we take the the first two arguments as $\{(\bmu^\star_a,\bW^\star_a)\}_{a\in[K]}$, we will write $D(\bmu^\star,\bW^\star,\blambda,\brho)=D(\blambda,\brho)$, and $\bar D_t(\bmu^\star,\bW^\star,\blambda,\brho)=\bar D_t(\blambda,\brho)$ in shorthand  in the case when there is no ambiguity.

Now, we first construct a preliminary bound on $\widehat{\blambda}_t$ by studying  the dual convergence when $\{(\bmu^\star_a,\bW^\star_a)\}_{a\in[K]}$ is known.

Define the second-order function of $D$ as 
\begin{equation*}
    S(\blambda,\brho)=D(\blambda,\brho) - D(\blambda^*(\brho),\brho)-\left\langle\nabla_{\blambda}D(\blambda^*(\brho),\brho),\blambda- \blambda^*(\brho)\right\rangle,
\end{equation*}
with its empirical version:
\begin{equation*}
     \bar S_t(\blambda,\brho) := \frac{1}{t}\sum_{j=1}^t S_j(\blambda,\brho) = \frac{1}{t}\sum_{j=1}^t \left[D_j(\blambda,\brho) - D_j(\blambda^*(\brho),\brho)-\left\langle\nabla_{\blambda}D_j(\blambda^*(\brho),\brho),\blambda- \blambda^*(\brho)\right\rangle\right].
\end{equation*}
Here for each $j$, the (sub)gradient is $\nabla_{\blambda}D_j(\blambda,\brho) = \brho-\bW^\star_{a^*_j(\blambda)}\bx_j$.

\begin{proposition}  There exists $ \underline{\cL} = \frac{1}{2}\cL$ and $\overline{\cL} = 2B_{\max}^2 M_{\sfm}$ such that
   \begin{equation}
     \underline{\cL} \norm{\blambda-\blambda^*(\brho)}_1   \le S(\blambda,\brho) \le \overline{\cL}\norm{\blambda-\blambda^*(\brho)}_1
   \end{equation}
\end{proposition}
The justification of this proposition can be found in Proposition 4 of \cite{ma2024optimal} with the help of Lemma \ref{lemma:smoothness}. We present the key technical lemmas for the convergence in Lemma \ref{lemma:concentrate-first-order}, \ref{lemma:lb-second-order}.

\begin{lemma}\label{lemma:concentrate-first-order}  We have the uniform upper bound of the first-order term:
\begin{equation*}
        \sup_{\brho\in \cB^+(\brho_0,\delta_\rho)} \norm{\nabla_{\blambda}\bar{D}_t(\blambda^*(\brho),\brho)- \nabla_{\blambda}{D}(\blambda^*(\brho),\brho)}_{\infty} \le \epsilon_1 + CB_{\max}\sqrt{\frac{m}{t}},
\end{equation*}
will probability at least $1-m\exp \left(-\frac{t \epsilon_1^2}{2 B_{\max}^2}\right)$.
\end{lemma}

\begin{lemma}\label{lemma:lb-second-order}  We have the uniform lower bound of the second-order term:
\begin{equation*}
        \inf_{\brho\in \cB^+(\brho_0,\delta_\rho),\norm{\blambda-\blambda^*(\brho)}_{1}\le\varepsilon}  \bar S_t(\blambda,\brho) \ge  \frac{\cL}{2}\norm{\blambda-\blambda^*(\brho)}_1^2 -C B_{\max}\varepsilon \sqrt{\frac{m}{t}}  -\epsilon_2,
\end{equation*}
will probability at least $1-\exp \left(-\frac{t \epsilon_2^2}{8 B_{\max}^2\varepsilon^2}\right)$.
\end{lemma}
With Lemma \ref{lemma:concentrate-first-order}, \ref{lemma:lb-second-order}, we can establish a preliminary bound on $\widehat{\blambda}_t$ such that it falls into the local smooth region $\norm{\blambda-\blambda^*(\brho)}_{1}\le\delta_{\lambda}, \text{ where }\brho\in \cB(\brho_0,\delta_\rho)$ with high probability. Since 
\begin{equation*}
\begin{aligned}
        & \bar D_t(\blambda,\brho) - \bar D_t(\blambda^*(\brho),\brho)  = \bar S_t(\blambda,\brho)+ \left\langle\nabla_{\blambda} D(\blambda^*(\brho),\brho), \blambda- \blambda^*(\brho)\right\rangle 
         \\
         & \quad\quad \quad\quad \quad\quad \quad\quad \quad\quad \quad\quad+ \left\langle\nabla_{\blambda}\bar D_t(\blambda^*(\brho),\brho)-\nabla_{\blambda} D(\blambda^*(\brho),\brho),\blambda- \blambda^*(\brho)\right\rangle \\
         & \ge \bar S_t(\blambda,\brho) -  \sup_{\brho\in \cB^+(\brho_0,\delta_\rho)} \norm{\nabla_{\blambda}\bar{D}_t(\blambda^*(\brho),\brho)- \nabla_{\blambda}{D}(\blambda^*(\brho),\brho)}_{\infty} \norm{\blambda- \blambda^*(\brho)}_1,
\end{aligned}
\end{equation*}
we can choose $\epsilon_1={\cL/8}\cdot\varepsilon$, $\epsilon_2=\cL\varepsilon^2/64$, and derive that when $\varepsilon> C\frac{B_{\max}}{\cL}\sqrt{\frac{m}{t}}$, we have
\begin{equation}\label{eq:emp-D-growth}
    \bar D_t(\blambda,\brho) - \bar D_t(\blambda^*(\brho),\brho)  \ge \frac{\cL}{2}\left(\norm{\blambda- \blambda^*(\brho)}_1-\frac{\varepsilon}{4}\right)^2 -\cL\frac{\varepsilon^2}{4},
\end{equation}
with probability at least $1-2m \exp \left(-\frac{t \cL^2\varepsilon^2}{C B_{\max}^2}\right)$ for all $\brho\in \cB^+(\brho_0,\delta_\rho),\norm{\blambda-\blambda^*(\brho)}_{1}\le\varepsilon$. 

Now, suppose that the primal estimation achieves the error rate $$
\max_{a}\norm{ \bmu_{a}^\star -\bmu_{a,t-1}^{ \mathsf{s} } }_2\vee \norm{\widehat{\bW}_{a,t-1} -\bW_{a}^{\star}}_{2,\max}\le\delta_{\est},
$$
with probability at least $1-T^{-2}$ for some small $\delta_{\est}$ at time $t$. Then, we have 
\begin{equation*}
    \begin{aligned}
        D_j(\blambda,\brho) &= \max_{a\in[K]\cup\{0\}}\left\langle \bmu_{a}^\star - (\bm{W}_{a}^\star)^{\top}\blambda,\bx_j \right\rangle + \brho^\top \blambda \\
        & \le 3\sqrt{s_0}DZ\delta_{\est} + \left\langle \widehat\bmu^{\mathsf{s}}_{a^*_t} - (\widehat{\bm{W}}_{a^*_t})^{\top}\blambda,\bx_j \right\rangle + \brho^\top \blambda 
        \\
        &\le   3\sqrt{s_0}DZ\delta_{\est} +  \max_{a\in[K]\cup\{0\}} \left\langle \widehat\bmu^{\mathsf{s}}_{a} - (\widehat{\bm{W}}_{a})^{\top}\blambda,\bx_j \right\rangle 
        \\
        &\le  3\sqrt{s_0}DZ\delta_{\est} + D_j(\widehat\bmu^{\mathsf{s}},\widehat{\bm{W}},\blambda,\brho).
    \end{aligned} 
\end{equation*}
Also, the same argument gives that $D_j(\widehat\bmu^{\mathsf{s}},\widehat{\bm{W}},\blambda,\brho)\le 3\sqrt{s_0}DZ\delta_{\est} + D_j(\blambda,\brho) $. This leads to:
\begin{equation}
    \label{eq:dual-sample-est}
    \begin{aligned}
        \abs{\bar D_t(\widehat\bmu^{\mathsf{s}},\widehat{\bm{W}},\blambda,\brho) - \bar  D_t(\blambda,\brho)} \le  3\sqrt{s_0}DZ\delta_{\est}.
    \end{aligned}
\end{equation}
When we take $\blambda$ as $\widehat{\blambda}_t$ that minimize the estimated sample dual problem $\bar D_t(\widehat\bmu^{\mathsf{s}},\widehat{\bm{W}},\blambda,\brho)$, we can derive from \eqref{eq:emp-D-growth} and  \eqref{eq:dual-sample-est} that 
\begin{equation*}
    \begin{aligned}
           0&\ge \bar D_t(\widehat\bmu^{\mathsf{s}},\widehat{\bm{W}},\widehat{\blambda}_t,\brho) - \bar D_t(\widehat\bmu^{\mathsf{s}},\widehat{\bm{W}},{\blambda}^*,\brho)  
            \\
            &\ge\bar D_t(\widehat{\blambda}_t,\brho) - \bar D_t(\blambda^*(\brho),\brho) -6\sqrt{s_0}DZ\delta_{\est}
            \\
            &\ge \frac{\cL}{2}\left(\norm{\widehat{\blambda}_t - \blambda^*(\brho)}_1-\frac{\varepsilon}{4}\right)^2 -\cL\frac{\varepsilon^2}{4} -6\sqrt{s_0}DZ\delta_{\est},
    \end{aligned}
\end{equation*}
if $\widehat{\blambda}_t$ falls into the local region $\norm{\blambda-\blambda^*(\brho)}_{1}\le\varepsilon$.
Taking $\varepsilon =\sqrt{ 32\sqrt{s_0}DZ\delta_{\est}/\cL}$, we are able to ensure that $\norm{\widehat{\blambda}_t-\blambda^*(\brho)}_{1}\le\varepsilon$ must hold (otherwise we can always find a $\blambda'$ within the $\cB_1(\varepsilon)$ ball around $\blambda^*(\brho)$  such that $\frac{\cL}{2}\left(\norm{\blambda' - \blambda^*(\brho)}_1-\frac{\varepsilon}{4}\right)^2 -\cL\frac{\varepsilon^2}{4} -6\sqrt{s_0}DZ\delta_{\est}>0$ due to the convexity of $\bar D_t(\blambda,\brho)$). This step is elaborated in \cite{ma2024optimal} for more details. Now we let
\begin{equation}\label{eq:init-req-dual-preliminary}
\begin{gathered}
       \varepsilon=\sqrt{ 32\sqrt{s_0}DZ\delta_{\est}/\cL}\le \delta_{\lambda} \\
   t\varepsilon^2 =  32\sqrt{s_0}DZ\delta_{\est}t /\cL\ge C B_{\max}^2 \log(mT),
\end{gathered}
\end{equation}
which lead to the probability bound that $\sup_{\cB^+(\brho_0,\delta_\rho)}\norm{\widehat{\blambda}_t-\blambda^*(\brho)}_{1}\le\sqrt{ 32\sqrt{s_0}DZ\delta_{\est}/\cL}\le \delta_{\lambda}$, with probability at least $1-2T^{-2}$ for all $t\ge T_0$ after some initialization period $T_0$ which will be determined later.

\subsubsection{Refined dual convergence with primal estimation.}

Since the preliminary bound established above shows that $\widehat{\blambda}_t$ falls into the smooth region of the dual function with high probability, we can thus give a finer control of $\norm{\widehat{\blambda}_t-\blambda^*(\brho)}_{1}$. We start from the optimality of $\widehat{\blambda}_t$:
\begin{equation}\label{eq:opt-emp-lamstar}
    \begin{aligned}
        &\left\langle \nabla_{\blambda} \bar D_t(\widehat\bmu^{\mathsf{s}},\widehat{\bm{W}}, \blambda^*(\brho),\brho),  \widehat{\blambda}_t - \blambda^*(\brho)\right \rangle 
        \le  \bar D_t(\widehat\bmu^{\mathsf{s}},\widehat{\bm{W}},\widehat{\blambda}_t,\brho) - \bar D_t(\widehat\bmu^{\mathsf{s}},\widehat{\bm{W}},\blambda^*(\brho),\brho) \le 0 \\
        & \left\langle \nabla_{\blambda} \bar D_t(\widehat\bmu^{\mathsf{s}},\widehat{\bm{W}}, \widehat{\blambda}_t,\brho),  \widehat{\blambda}_t - \blambda^*(\brho)\right \rangle 
        \le 0
    \end{aligned}.
\end{equation}
\eqref{eq:opt-emp-lamstar} suggests that
\begin{equation}
    \left\langle \nabla_{\blambda} \bar D_t(\widehat\bmu^{\mathsf{s}},\widehat{\bm{W}}, \blambda^*(\brho)+z \bv,\brho),  \widehat{\blambda}_t - \blambda^*(\brho)\right \rangle 
        \le 0,
\end{equation}
for any $z\in[0,1]$, where $\bv = \widehat{\blambda}_t - \blambda^*(\brho)$. 
We thus have the second-order term satisfying:
\begin{equation}\label{eq:second-order-up-1}
\begin{aligned}
        \bar S_t(\widehat{\blambda}_t,\brho)  
        & = \int_0^1 \left\langle\nabla_{\blambda} \bar D_t(\blambda^*(\brho)+z\bv,\brho), \widehat{\blambda}_t - \blambda^*(\brho) \right\rangle dz 
        \\
         & \le  \int_0^1 \left\langle \nabla_{\blambda} \bar D_t(\blambda^*(\brho)+z\bv,\brho) - \nabla_{\blambda}\bar D_t(\widehat\bmu^{\mathsf{s}},\widehat{\bm{W}}, \blambda^*(\brho)+z \bv,\brho), \widehat{\blambda}_t - \blambda^*(\brho) \right\rangle dz 
         \\
         & \le  \sup_{\norm{\blambda-\blambda^*(\brho)}_1\le \delta_{\lambda},\brho\in\cB^+(\brho_0,\delta_\rho)} \norm{\nabla_{\blambda} \bar D_t(\blambda,\brho) - \nabla_{\blambda}\bar D_t(\widehat\bmu^{\mathsf{s}},\widehat{\bm{W}}, \blambda,\brho)}_{\infty}\norm{\widehat{\blambda}_t - \blambda^*(\brho) }_1,
\end{aligned}
\end{equation}
with probability at least $1-2T^{-2}$. 

To control \eqref{eq:second-order-up-1}, we write the decision variables as
\begin{equation*}
     \begin{gathered}
         a^*_t(\blambda) =  \arg\max_{a\in[K]\cup\{0\}} \left\langle \bmu_{a}^\star - (\bm{W}_{a}^{\star})^{\top}\blambda,\bx_t \right\rangle 
         \\
\widehat{a}_t(\blambda) =  \arg\max_{a\in[K]\cup\{0\}} \left\langle  \bmu_{a,t-1}^{\mathsf{s}} - (\widehat{\bm{W}}_{a,t-1} )^{\top}\blambda,\bx_t \right\rangle,
     \end{gathered}
\end{equation*}
where the estimators $\bmu_{a,t-1}^{\mathsf{s}}$ and $\widehat{\bm{W}}_{a,t-1}$ are known to take values within
\begin{equation*}
    \bOmega_{\est} = \left\{\bmu_a,\bW_a: \max_{a}\norm{ \bmu_{a}^\star -\bmu_{a} }_2\vee \norm{{\bW}_{a} -\bW_{a}^{\star}}_{2,\max}\le\delta_{\est}, \norm{\bmu_a}_0\le s_0, \norm{\bW_{a,i}}_0\le s_0\right\}
\end{equation*}
with probability at least $1-T^{-2}$. We now express the difference between gradients in  \eqref{eq:second-order-up-1} as 
\begin{equation}\label{eq:est-first-order-ub}
\begin{aligned}
        &\sup_{\norm{\blambda-\blambda^*(\brho)}_1\le \delta_{\lambda},\brho\in\cB^+(\brho_0,\delta_\rho)} \norm{\nabla_{\blambda} \bar D_t(\blambda,\brho) - \nabla_{\blambda}\bar D_t(\widehat\bmu^{\mathsf{s}},\widehat{\bm{W}}, \blambda,\brho)}_{\infty}
        \\
    &\le 3\sqrt{s_0}DZ\delta_{\est} +  \sup_{\norm{\blambda-\blambda^*(\brho)}_1\le \delta_{\lambda},\brho\in\cB^+(\brho_0,\delta_\rho)} \frac{2B_{\max}}{t} \sum_{j=1}^t\mathbbm{1}\{a^*_j(\blambda)\neq \widehat{a}_j(\blambda)\} 
    \\
    &\le  3\sqrt{s_0}DZ\delta_{\est} +  \sup_{\norm{\blambda-\blambda^*(\brho)}_1\le \delta_{\lambda},\brho\in\cB^+(\brho_0,\delta_\rho),\bOmega_{\est}} \frac{2B_{\max}}{t} \sum_{j=1}^t\left(\mathbbm{1}\{a^*_j(\blambda)\neq \widehat{a}_j(\blambda)\} -\bp(a^*_j(\blambda)\neq \widehat{a}_j(\blambda))\right) 
    \\
    & \quad \quad \quad + \sup_{\norm{\blambda-\blambda^*(\brho)}_1\le \delta_{\lambda},\brho\in\cB^+(\brho_0,\delta_\rho),\bOmega_{\est}} 2B_{\max} \bp(a^*_j(\blambda)\neq \widehat{a}_j(\blambda)),
\end{aligned}
\end{equation}
with probability at least $1-2T^{-2}$. Here for each $j$, the probability $\bp(a^*_j(\blambda)\neq \widehat{a}_j(\blambda))$ is taken only with respect to $\bx_j$ by fixing other quantities, and the supremum is taken with respect to $\brho$, $\blambda$, and $(\bmu_{a,t-1}^{\mathsf{s}},\widehat{\bm{W}}_{a,t-1})\in\bOmega_{\est}$. Since within $\bOmega_{\est}$, $\{ a^*_j(\blambda)\neq \widehat{a}_j(\blambda) \}$ implies that
\begin{equation*}
\begin{gathered}
    \left\langle \widehat\bmu^{\mathsf{s}}_{ a^*_j(\blambda)  } - (\widehat{\bm{W}}_{ a^*_j(\blambda) })^{\top}\blambda,\bx_j \right\rangle  <    \left\langle \widehat\bmu^{\mathsf{s}}_{ \widehat{a}_j(\blambda)  } - (\widehat{\bm{W}}_{ \widehat{a}_j(\blambda)  })^{\top}\blambda,\bx_j \right\rangle 
    \\
    \left\langle \bmu_{a^*_t(\blambda) }^\star - (\bm{W}_{a^*_t(\blambda)}^{\star})^{\top}\blambda,\bx_t \right\rangle - \left\langle \bmu_{\widehat{a}_j(\blambda)}^\star - (\bm{W}_{\widehat{a}_j(\blambda)}^{\star})^{\top}\blambda,\bx_t \right\rangle < 3\sqrt{s_0}DZ\delta_{\est},
\end{gathered}
\end{equation*}
we can therefore derive that
\begin{equation}\label{eq:est-first-order-ub-2}
\begin{aligned}
       & \sup_{\norm{\blambda-\blambda^*(\brho)}_1\le \delta_{\lambda},\brho\in\cB^+(\brho_0,\delta_\rho),\bOmega_{\est}}\bp( a^*_j(\blambda)\neq \widehat{a}_j(\blambda)) 
       \\
    &\le   \sup_{\norm{\blambda-\blambda^*(\brho)}_1\le \delta_{\lambda},\brho\in\cB^+(\brho_0,\delta_\rho)}  \bp\left(\left\langle \bmu_{a^*_t(\blambda) }^\star - (\bm{W}_{a^*_t(\blambda)}^{\star})^{\top}\blambda,\bx_t \right\rangle - \max_{a\neq a^*_t(\blambda)}\left\langle \bmu_{a}^\star - (\bm{W}_{a}^{\star})^{\top}\blambda,\bx_t \right\rangle \le 3\sqrt{s_0}DZ\delta_{\est}\right) 
    \\
    &\le 3M_{\sfm } \sqrt{s_0}DZ\delta_{\est},
\end{aligned}
\end{equation}

by the margin condition in Assumption \ref{asm:local-dual-conditions}. Here we use the fact that for any $\brho\in\cB^+(\brho_0,\delta_\rho)$, we can always truncate its non-binding dimension and get a $\brho^-$ such that $\brho^-\in \cB(\brho_0,\delta_\rho)$ and $\blambda^*(\brho)=\blambda^*(\brho^-)$.
To ease the expression, we write the set $\bOmega_{\lambda}=\left\{\blambda: \norm{\blambda-\blambda^*(\brho)}_1\le \delta_{\lambda},\forall\brho\in\cB(\brho_0,\delta_\rho)\right\}$, with $\bOmega_{\lambda}\in\bR^{m}$. To control the empirical process in \eqref{eq:est-first-order-ub}, it suffices to bound:
\begin{equation*}
\begin{aligned}
       & \underbrace{\sup_{\bOmega_{\lambda},\bOmega_{\est}} \frac{1}{t} \sum_{j=1}^t\left(\mathbbm{1}\{a^*_j(\blambda)\neq \widehat{a}_j(\blambda)\} -\bp(a^*_j(\blambda)\neq \widehat{a}_j(\blambda))\right) - \bE\left[\sup_{\bOmega_{\lambda},\bOmega_{\est}} \frac{1}{t} \sum_{j=1}^t\left(\mathbbm{1}\{a^*_j(\blambda)\neq \widehat{a}_j(\blambda)\} -\bp(a^*_j(\blambda)\neq \widehat{a}_j(\blambda))\right)\right] }_{\mathfrak{C}_1}\\
   & + \underbrace{\bE\left[\sup_{\bOmega_{\lambda},\bOmega_{\est}} \frac{1}{t} \sum_{j=1}^t\left(\mathbbm{1}\{a^*_j(\blambda)\neq \widehat{a}_j(\blambda)\} -\bp(a^*_j(\blambda)\neq \widehat{a}_j(\blambda))\right)\right] }_{\mathfrak{C}_2}.
\end{aligned}
\end{equation*}
For the term $\mathfrak{C}_1$, we have 
\begin{equation}\label{eq:est-first-order-ub-3}
    \bp({\mathfrak{C}_1}\ge \epsilon_3 )\le \exp(-\frac{t\epsilon^2_3}{4}),
\end{equation}
by the bouned difference condition \citep{koltchinskii2011oracle}.

\noindent For the term $\mathfrak{C}_2$, we can control it by computing the VC dimension of the function class $\{\mathbbm{1}\{a^*_j(\blambda)\neq \widehat{a}_j(\blambda)\}\}$, which we denote by $V_{\mathfrak{C}_2}$. Notice that each set $\{a^*_j(\blambda)\neq \widehat{a}_j(\blambda)\}\}$ is composed of:
\begin{equation*}
    \begin{aligned}
      \{a^*_j(\blambda)\neq \widehat{a}_j(\blambda)\}\}=  &\bigcup_{a=0}^K\left[\bigcap_{k\neq a}\{\left\langle \bmu_{a }^\star - (\bm{W}_{a}^{\star})^{\top}\blambda,\bx_t \right\rangle> \left\langle \bmu_{k }^\star - (\bm{W}_{k }^{\star})^{\top}\blambda,\bx_t \right\rangle\}
         \right.\\
         &\left. \quad \quad\quad 
        \bigcap\left\{\bigcup_{k\neq a}\{ \left\langle \widehat\bmu^{\mathsf{s}}_{ a  } - (\widehat{\bm{W}}_{ a })^{\top}\blambda,\bx_j \right\rangle  <    \left\langle \widehat\bmu^{\mathsf{s}}_{ k  } - (\widehat{\bm{W}}_{ k  })^{\top}\blambda,\bx_j \right\rangle\}   \right\} \right] \\
            & :=\bigcup_{a=0}^K \cC_{a}
    \end{aligned} 
\end{equation*}
Denote the VC dimension of set class consisting of $ \cC_{a}$ as $V_{\cC}$. By the VC theory on the union of sets, \citep{eisenstat2007vc}, we know that $V_{\mathfrak{C}_2}\le CKV_{\cC}\log K$. To control $V_{\cC}$, we consider its growth function $\Pi_{\cC}(V)$ that counts the most dichotomies on any $V$ points.

Define 
$\Pi_{\cD_1}(V)$ as the growth function on set class of $\{\left\langle \bmu_{a }^\star - (\bm{W}_{a}^{\star})^{\top}\blambda,\bx_t \right\rangle> \left\langle \bmu_{k }^\star - (\bm{W}_{k }^{\star})^{\top}\blambda,\bx_t \right\rangle\}$, and $\Pi_{\cD_2}(V)$  as the growth function on set class of $\bigcup_{k\neq a}\left\{ 
 \left\langle \bmu_{ a  } - ({\bm{W}}_{ a })^{\top}\blambda,\bx_j \right\rangle  <   \left\langle \bmu_{ k  } - ({\bm{W}}_{ k  })^{\top}\blambda,\bx_j \right\rangle \right\}$,  where $\bmu_{ a  }$,${\bm{W}}_{ a }$, $\bmu_{ k  }$, ${\bm{W}}_{ k  }$ are sparse.  Notice that $\Pi_{\cD_1}(V)$ can be easily controlled by $O(V^m)$ since $\blambda\in\bR^m$. Then, by the VC dimenion of $s_0$-sparse vectors (Lemma 1, \cite{abramovich2018high}) and Sauer’s Lemma, we have 
\begin{equation*}
    \Pi_{\cD_2}(V) \le \binom{d}{s_0}^K\cdot \sum_{i=0}^{C s_0mK\log K} \binom{V}{i} \le \left(\frac{de}{s_0}\right)^{s_0K}\cdot \left(\frac{Ve}{Cs_0mK\log K}\right)^{Cs_0mK\log K}.
\end{equation*}
Therefore, we have
\begin{equation*}
    \Pi_{\cC} \le \Pi_{\cD_1}(V)^K\cdot  \Pi_{\cD_2}(V) \le \left(\frac{Ve}{m}\right)^{mK} \cdot\left(\frac{de}{s_0}\right)^{s_0K}\cdot \left(\frac{Ve}{Cs_0mK\log K}\right)^{Cs_0mK\log K}
\end{equation*}
We can conclude that $V_{\cC}\le Cs_0m K\log K\log d $, and thus  $V_{\mathfrak{C}_2}\le CKV_{\cC}\log K\le Cs_0 mK^2\log^2 K\log d$. 

Invoking the Rademacher complexity bound by VC dimension
\citep{van1996weak,gine2021mathematical}, we have
\begin{equation}\label{eq:est-first-order-ub-4}
   \mathfrak{C}_2\le C\sqrt{\frac{V_{\mathfrak{C}_2}}{t} }\le C \sqrt{\frac{s_0 mK^2\log^2 K\log d}{t}}
\end{equation}

% \cite{maximov2016tight}, Multi-Class Margin Classifiers \cite{yin2019rademacher}

Associating \eqref{eq:est-first-order-ub}, \eqref{eq:est-first-order-ub-2}, \eqref{eq:est-first-order-ub-3}, \eqref{eq:est-first-order-ub-4} with the lower bound obtained in Lemma \ref{lemma:lb-second-order}, 
\begin{equation*}
\begin{aligned}
          &C\left( B_{\max} M_{\sfm }\sqrt{s_0}DZ\delta_{\est}+   B_{\max} \sqrt{\frac{s_0 mK^2\log^2 K\log d}{t}}\right)\norm{\widehat{\blambda}_t - \blambda^*(\brho) }_1 +\frac{1}{128}\cL\varepsilon  \norm{\widehat{\blambda}_t - \blambda^*(\brho) }_1 
          \\
          &\ge \bar S_t(\widehat{\blambda}_t,\brho) 
          \\
          & \bar S_t(\blambda,\brho) 
          \\
          &\ge  \frac{\cL}{2}\left(\norm{\blambda- \blambda^*(\brho)}_1-\frac{\varepsilon}{4}\right)^2 -\cL\frac{\varepsilon^2}{4},
\end{aligned}
\end{equation*}
with probability at least $1-2m \exp \left(-\frac{t \cL^2\varepsilon^2}{C B_{\max}^2}\right)-2T^{-2}$ for all $\brho\in \cB^+(\brho_0,\delta_\rho),\norm{\blambda-\blambda^*(\brho)}_{1}\le\varepsilon$, where $\varepsilon> C\frac{B_{\max}}{\cL}\sqrt{\frac{m}{t}}$. Here we treat $\delta_{\est}\gtrsim \sqrt{s_0mK^2\log^2K\log d/t}$ since $\delta_{\est}$ can be chosen large. Taking $\varepsilon\ge C B_{\max} M_{\sfm }\sqrt{s_0}DZ\delta_{\est}/\cL$, we have 
\begin{equation}
\begin{aligned}
        &\frac{1}{64}\cL\varepsilon  \norm{\widehat{\blambda}_t - \blambda^*(\brho) }_1 \ge  \bar S_t(\widehat{\blambda}_t,\brho),
        \\
    &\bar S_t(\blambda,\brho)
    \ge  \frac{\cL}{2}\left(\norm{\blambda- \blambda^*(\brho)}_1-\frac{\varepsilon}{4}\right)^2 -\cL\frac{\varepsilon^2}{4}
\end{aligned}
\end{equation}
Now, we claim that $\norm{\widehat{\blambda}_t - \blambda^*(\brho) }_1\le \varepsilon$ must hold. If not, then we have a $\widehat{\blambda}_t$ such that $\norm{\widehat{\blambda}_t - \blambda^*(\brho) }_1 > \varepsilon$ and  $\bar S_t(\widehat{\blambda}_t,\brho)\le \frac{1}{64}\cL\varepsilon  \norm{\widehat{\blambda}_t - \blambda^*(\brho) }_1$. Notice that $\bar S_t(\widehat{\blambda}_t,\brho)$ is a convex function and $\bar S_t(\blambda^*(\brho),\brho)=0$. We can take $\blambda'$ as a point on the line segment between $\blambda^*(\brho)$ and $\widehat{\blambda}_t$ by 
$$
\blambda'=\varepsilon\cdot\left(\widehat{\blambda}_t-\blambda^*(\brho)\right)/\norm{\widehat{\blambda}_t-\blambda^*(\brho)}_1 +\blambda^*(\brho) = \frac{\varepsilon}{\norm{\widehat{\blambda}_t-\blambda^*(\brho)}_1}\widehat{\blambda}_t+\left(1-\frac{\varepsilon}{\norm{\widehat{\blambda}_t-\blambda^*(\brho)}_1}\right)\blambda^*(\brho).
$$
For $\blambda'$, we have $\norm{\blambda'-\blambda^*(\brho)}_1=\varepsilon$, and by the convexity of $S_t(\blambda,\brho)$ it satisfies
\begin{equation}\label{eq:barS-ub-contradict}
    \bar S_t(\blambda',\brho) \le \frac{\varepsilon}{\norm{\widehat{\blambda}_t-\blambda^*(\brho)}_1}\bar S_t(\widehat{\blambda}_t,\brho)  \le \frac{1}{64}\cL\varepsilon^2.
\end{equation}
However, $\blambda'$ is within $\norm{\blambda-\blambda^*(\brho)}_{1}\le\varepsilon$ and $\bar S_t(\blambda',\brho)$ is lower bounded by
\begin{equation}\label{eq:barS-lb-contradict}
    \bar S_t(\blambda',\brho)
    \ge  \frac{\cL}{2}\left(\norm{\blambda'- \blambda^*(\brho)}_1-\frac{\varepsilon}{4}\right)^2 -\cL\frac{\varepsilon^2}{4} \ge \frac{1}{32} \cL \varepsilon^2.
\end{equation}
\eqref{eq:barS-ub-contradict} and \eqref{eq:barS-lb-contradict} contradict each other! Thus, we conclude that $\norm{\widehat{\blambda}_t - \blambda^*(\brho) }_1\le \varepsilon$ with probability at least $1-2m \exp \left(-\frac{t \cL^2\varepsilon^2}{C B_{\max}^2}\right)-2T^{-2}$ for all $\brho\in \cB^+(\brho_0,\delta_\rho)$, where $\varepsilon\ge C B_{\max} M_{\sfm }\sqrt{s_0}DZ\delta_{\est}/\cL$.

\subsection{Proof of Theorem \ref{thm:est-induction}}
Notice that the algorithm in the initial phase always guarantees a convergence of rate $\widetilde{O}(\frac{K}{t})$ by Theorem \ref{thm:sparse-est} (by treating $\underline{p_j}=\frac{1}{K} $). Therefore we can always select a large $T_0$ such that $\delta_{\est}$ is small before $T_0+1$. We study the iteration after time $T_0+1$. To prove a high-probability bound, it suffices to look at the convergence for any fixed $\bmu^\star_a$, and then relax it to all $a\in[K]$ and all the rows of $\bW^\star_a$. Thus, we only focus on one  $\bmu^\star_a$, and adopt the notations in Section \ref{sec:proof-sparse-est-divcov} for simplicity.  We start with \eqref{eq:sparse-est-divcov-itr}, which gives

\begin{equation*}
    \begin{aligned}
        \norm{\bmu_t-\bmu_\star}_2^2 
         \le &\left(1+\frac{3}{2}\sqrt{\varrho } \right) \left( \norm{\bmu_{t-1}-\bmu_{\star}}_2^2-  2\beta_t \left\langle \nabla f_t(\bmu_{t-1}) , \bmu_{t-1} -\bmu_\star \right\rangle +  2\beta_t^2 \norm{\cP_{\Omega}(\bg_t-\nabla f_t(\bmu_{t-1}) ) }_2^2\right.  \\
         & \left. +2\beta_t^2 \norm{\cP_{\Omega}(\nabla f_t(\bmu_{t-1})) }_2^2 + 2\beta_t \norm{\cP_{\Omega}(\bg_t-\nabla f_t(\bmu_{t-1}) ) }_2\norm{\bmu_{t-1}-\bmu_{\star}}_2  \right),
    \end{aligned}
\end{equation*}
where we write $\bar{\bSigma}_{t} = \bar{\bSigma}_{a,t} = \sum_{j=1}^{t}\bE \left[\bx_j \bx_j^{\top}\cdot \bI\left\{y_j=a \right\} \middle| \cH_{j-1}\right]/t$, and the corresponding objective function $f_t(\bmu) = \norm{\bmu-\bmu_{a}^{\star} }_{\Bar{\bSigma}_{a,t}}^2$. We now define a new covariance matrix when both $\bmu^\star_a$, $\bW^\star_a$ and optimal dual solution $\blambda^*(\brho)$ is known:
\begin{equation*}
    {\bSigma}^\star_{t}(\{\brho_j\}_{j=T_0}^{t-1}) = \left(\sum_{j=1}^{T_0}\frac{\bSigma}{K} +\sum_{j=T_0+1}^{t}\bE \left[\bx_j \bx_j^{\top}\cdot \bI\left\{a=a^*_j(\blambda^*(\brho_{j-1}))  \right\} \right]\right)/t,
\end{equation*}
where $a^*_j(\blambda^*(\brho)) =  \arg\max_{a\in[K]\cup\{0\}} \left\langle \bmu_{a}^\star - (\bm{W}_{a}^{\star})^{\top}\blambda^*(\brho),\bx_j \right\rangle$. Notice that $ {\bSigma}^\star_{t}(\{\brho_j\}_{j=T_0}^{t-1}) $ is a matrix that is only determined by $t$ and $\{\brho_j\}_{j=T_0+1}^t$. Denote the corresponding $g_{t}(\bmu)=\norm{\bmu-\bmu^\star_a}_{{\bSigma}^\star_{t}(\{\brho_j\}_{j=T_0}^{t-1})}$. 
Then, we can write the iteration as
\begin{equation*}
    \begin{aligned}
        \norm{\bmu_t-\bmu_\star}_2^2 
         \le &\left(1+\frac{3}{2}\sqrt{\varrho } \right) \left( \norm{\bmu_{t-1}-\bmu_{\star}}_2^2-  2\beta_t \left\langle \nabla f_t(\bmu_{t-1}) , \bmu_{t-1} -\bmu_\star \right\rangle +  2\beta_t^2 \norm{\cP_{\Omega}(\bg_t-\nabla f_t(\bmu_{t-1}) ) }_2^2\right.  \\
         & \left. +2\beta_t^2 \norm{\cP_{\Omega}(\nabla f_t(\bmu_{t-1})) }_2^2 + 2\beta_t \norm{\cP_{\Omega}(\bg_t-\nabla f_t(\bmu_{t-1}) ) }_2\norm{\bmu_{t-1}-\bmu_{\star}}_2  \right) 
         \\
          \le  &\left(1+\frac{3}{2}\sqrt{\varrho } \right) \left( \norm{\bmu_{t-1}-\bmu_{\star}}_2^2-  2\beta_t \left\langle \nabla g_t(\bmu_{t-1} ) , \bmu_{t-1} -\bmu_\star \right\rangle  +  2\beta_t^2 \norm{\cP_{\Omega}(\bg_t-\nabla f_t(\bmu_{t-1}) ) }_2^2\right.  \\
         & \left. +2\beta_t \left\langle \nabla g_t(\bmu_{t-1} )-\nabla f_t(\bmu_{t-1}) , \bmu_{t-1} -\bmu_\star \right\rangle +2\beta_t^2 \norm{\cP_{\Omega}(\nabla f_t(\bmu_{t-1})) }_2^2 \right.
         \\
         &\left.  + 2\beta_t \norm{\cP_{\Omega}(\bg_t-\nabla f_t(\bmu_{t-1}) ) }_2\norm{\bmu_{t-1}-\bmu_{\star}}_2  \right) 
    \end{aligned}
\end{equation*}
When $\brho_j\in\cB^+(\brho_0,\delta_{\rho}')$, we can always truncate  $\brho_j$ into $\brho^-_j\in\cB(\brho_0,\delta_{\rho}')$ by reducing the resources in non-binding dimensions without change the optimal dual solution, i.e., $\blambda^*(\brho_j) = \blambda^*(\brho^-_j)$, and therefore ${\bSigma}^\star_{t}(\{\brho_j\}_{j=T_0}^{t-1})={\bSigma}^\star_{t}(\{\brho_j^-\}_{j=T_0}^{t-1})$. By the diverse covariate condition, we have
\begin{equation}\label{eq:div-cov-eigen-lb}
    \left\langle \nabla g_t(\bmu_{t-1},\{\brho_j\}_{j=T_0}^{t-1}) , \bmu_{t-1} -\bmu_\star \right\rangle = \left\langle \nabla g_t(\bmu_{t-1},\{\brho_j^-\}_{j=T_0}^{t-1}) , \bmu_{t-1} -\bmu_\star \right\rangle \ge  2 \gamma(K)\zeta(K) \norm{ \bmu_{t-1} -\bmu_\star }_2^2,
\end{equation}
where we use the implicitly condition that $\phi_{\min}(s)\ge K\gamma(K)\zeta(K)$.  Otherwise, the minimum sparse eigenvalue of $\bSigma$, $\phi_{\min}(s)$, will be smaller than the lower bounded we derived from diverse covariate condition $K\gamma(K)\zeta(K)$ since 
\begin{equation*}
 \bv^\top \bSigma \bv= \sum_{a=0}^K  \bv^\top\bE\left[\bx_j \bx_j^{\top}\cdot \bI\left\{a=a^*_j(\blambda^*(\brho))  \right\}\right]\bv \succeq K\gamma(K)\zeta(K).
\end{equation*}

Plugging \eqref{eq:div-cov-eigen-lb} and \eqref{eq:divcov-recur} into the iterative relationship above, we have
\begin{equation}\label{eq:div-cov-itr-1}
    \begin{aligned}
          \norm{\bmu_t-\bmu_\star}_2^2    \le  &\left(1+\frac{3}{2}\sqrt{\varrho } \right) \left( 1-4\gamma(K)\zeta(K)  \beta_t+8\beta_t^2\phi_{\max}^2(s) \right) \norm{\bmu_{t-1} -\bmu_\star}_2^2  \\
        & + 18 s\beta_t^2  \max_{i\in [d] } \abs{\left\langle \bg_t -\nabla f_t({\bmu}_{t-1}), {\bm e}_i  \right\rangle}^2  + 18 \beta_t \sqrt{s} \max_{i\in [d] } \abs{\left\langle \bg_t -\nabla f_t({\bmu}_{t-1}), {\bm e}_i  \right\rangle}\norm{\bmu_{t-1}-\bmu_{\star}}_2 \\
        & + 18\beta_t \underbrace{\left\langle \nabla g_t(\bmu_{t-1},\{\brho_j^-\}_{j=T_0}^{t-1})-\nabla f_t(\bmu_{t-1}) , \bmu_{t-1} -\bmu_\star \right\rangle}_{\mathfrak{D}_1}
    \end{aligned}
\end{equation}
We control the additional term $\mathfrak{D}_1$ by noticing that for any $t\le T_0$, we have $\bE\left[\bx_j \bx_j^{\top}\cdot \bI\left\{y_j=a \right\} \middle| \cH_{j-1}\right]=\frac{\bSigma}{K}$. This suggests that
\begin{equation*}
\begin{aligned}
        &\left\langle \nabla g_t(\bmu_{t-1},\{\brho_j^-\}_{j=T_0+1}^t)-\nabla f_t(\bmu_{t-1}) , \bmu_{t-1} -\bmu_\star \right\rangle  
        \\ &= (\bmu_{t-1} -\bmu_\star)^\top \frac{ \sum_{j=T_0+1}^t\bE \left[\bx_j \bx_j^{\top}\cdot(\bI\left\{a=a^*_j(\blambda^*(\brho_{j-1}^-))\right\}-  \bI\left\{y_j=a \right\} )\middle| \cH_{j-1}\right]}{t} (\bmu_{t-1} -\bmu_\star)\\
        & \le 2sD^2 \frac{\sum_{j=T_0+1}^t \bp_{\bx_j}(a^*_j(\blambda^*(\brho_{j-1}^-))\neq y_j |\cH_{j-1} ) }{t} \norm{\bmu_{t-1} -\bmu_\star}_2^2.
\end{aligned}
\end{equation*}
where we use the fact that $\brho_{j-1}$ is independent of the future data $\bx_j$. Now we denote the high-probability bound as         
$$
\max_{a}\norm{\bmu_{a,t}^{ \mathsf{s} } - \bmu_{a}^\star  }_2\vee \norm{\widehat{\bW}_{a,t} -\bW_{a}^{\star}}_{2,\max}\le\delta_{\est}(t),
$$
which holds with probability at least $1-\frac{1}{T^2}$ for each $t$.

When $\brho_{j}$ are all within the $\cB^+(\brho_0,\delta_{\rho}')$, we can bound the probability of wrong selection $\bp_{\bx_j}(a^*_j(\blambda^*(\brho_{j-1}^-)\neq y_j |\cH_{j-1} )$ by 
\begin{equation*}
\begin{aligned}
        &\bp_{\bx_j}\left(a^*_j(\blambda^*(\brho_{j-1}^-))\neq y_j |\cH_{j-1} \right) 
    =
    \bp_{\bx_j}\left(  \left\langle \widehat\bmu^{\mathsf{s}}_{ a^*_j(\blambda)  } - (\widehat{\bm{W}}_{ a^*_j(\blambda) })^{\top}\widehat{\blambda}_{j-1},\bx_j \right\rangle  <    \left\langle \widehat\bmu^{\mathsf{s}}_{ y_j    } - (\widehat{\bm{W}}_{ y_j  })^{\top}\widehat{\blambda}_{j-1},\bx_j \right\rangle \middle|\cH_{j-1}\right) 
    \\
    & \le \frac{1}{T^2} + \bp\left(\left\langle \bmu_{a^*_t(\blambda) }^\star - (\bm{W}_{a^*_j(\widehat{\blambda}_{j-1})}^{\star})^{\top}\widehat{\blambda}_{j-1},\bx_j \right\rangle - \left\langle \bmu_{y_j }^\star - (\bm{W}_{y_j }^{\star})^{\top}\widehat{\blambda}_{j-1},\bx_j \right\rangle < 3\sqrt{s_0}DZ\delta_{\est}(j) \middle|\cH_{j-1}\right) 
    \\
    & \le \frac{1}{T^2} + \bp(\norm{\widehat{\blambda}_{j-1}-\blambda^*(\brho_{j-1})}_1\ge\delta_{\lambda}) + 3\sqrt{s_0}M_{\sfm}DZ\delta_{\est}(j-1) 
    \\
    &  \le \frac{3}{T^2} + 3\sqrt{s_0}M_{\sfm}DZ\delta_{\est}(j-1),
\end{aligned}
\end{equation*}
where by Section \ref{sec:proof-est-dual-conv}, we let  $T_0$ to be large such that the rate $\sqrt{ 32\sqrt{s_0}DZ\delta_{\est}(t)/\cL}\le \delta_{\lambda}$ for all $t\ge T_0$. This gives the bound on $\mathfrak{D}_1$:
\begin{equation}\label{eq:control-D1}
    \mathfrak{D}_1 \le s \sqrt{s_0}  D^3 M_{\sfm}Z\frac{\sum_{j=T_0+1}^{t} \delta_{\est}(j-1)}{t} \norm{\bmu_{t-1} -\bmu_\star}_2^2
\end{equation}

By Lemma \ref{lemma:max-grad-divcov}, we know that with probability at least $1-T^2$,
\begin{equation*}
    \max_{i\in [d] } \abs{\left\langle \bg_t -\nabla f_t({\bmu}_{t-1}), {\bm e}_i  \right\rangle}^2 \le C s D^2\frac{\log (dT)  }{t} \norm{\bmu_{t-1} -\bmu_{\star} }_2^2 +  C\frac{\sigma^2 D^2\log(dT) }{t}.
\end{equation*}
Now, consider estimating $\bmu^\star_a$ for all $a\in[K]$ and all the rows of $\bW^\star_a$ at the same time.  We can derive that 
\begin{equation*}
\begin{gathered}
        \max_{i\in [d],a\in[K] } \abs{\left\langle \bg_{a,t} -\nabla f_t^a({\bmu}_{a,t-1}), {\bm e}_i  \right\rangle}^2 \le C s D^2\frac{\log (dKT)  }{t} \norm{\bmu_{a,t-1} -\bmu^{\star}_a }_2^2 +  C\frac{\sigma^2 D^2\log(dT) }{t} 
        \\
                \max_{i\in [d], a\in[K], l\in [m]} \abs{\left\langle \bh_{a,l,t} -\nabla h_t^a({\bW}_{a,l\cdot,t-1}), {\bm e}_i  \right\rangle}^2 \le C s D^2\frac{\log (mdKT)  }{t} \norm{{\bW}_{a,l\cdot,t-1} -\bW^{\star}_{a,l\cdot} }_2^2 +  C\frac{\sigma^2 D^2\log(mdKT) }{t}
\end{gathered}
\end{equation*}
hold with probability at least $1-T^2$. Here $\bh_{a,l,t}$ is stochastic gradient at time $t$ for estimating the $l$-th row of $\bW^\star_a$. Associating this result with \eqref{eq:control-D1} and plugging them in \eqref{eq:div-cov-itr-1}, we essentially have
\begin{equation*}
\begin{aligned}
        & \max_{a}\norm{\bmu_{a,t}^{ \mathsf{s} } - \bmu_{a}^\star  }_2^2\vee \norm{\widehat{\bW}_{a,t} -\bW_{a}^{\star}}_{2,\max}^2 \\
        & \le \left( 1-\frac{1}{4\kappa_1^2} +  C \frac{s_0 D\sqrt{\log(mKdT)}}{ \gamma(K)\zeta(K)
        \sqrt{t} } +s\beta_t \sqrt{s_0}  D^3 M_{\sfm}Z\frac{\sum_{j=T_0+1}^{t} \delta_{\est}(j-1)}{t} \right)  \delta_{\est}^2(t-1)
        \\
      & +  C\frac{s_0\sigma^2 D^2 \log(mKdT) }{\gamma^2(K)\zeta^2(K) t } + C \sqrt{ \frac{s_0\sigma^2 D^2 \log(mKdT) }{\gamma^2(K)\zeta^2(K) t} \delta_{\est}^2(t-1)} 
      \\
      & \le \left(1-\frac{1}{6\kappa_1^2}+ C \frac{s_0 D\sqrt{\log(mKdT)}}{ \gamma(K)\zeta(K)
        \sqrt{t} } +C\frac{\kappa_1^2 s_0^{\frac{3}{2}} D^3M_{\sfm} Z}{\gamma(K)\zeta(K)}\frac{\sum_{j=T_0+1}^{t} \delta_{\est}(j-1)}{t}  \right) \delta_{\est}^2(t-1)  \\
       & +  C\frac{\kappa_1^2s_0\sigma^2 D^2 \log(mKdT) }{\gamma^2(K)\zeta^2(K) t }
\end{aligned}
\end{equation*}
where we take  $\varrho = \frac{1}{36\kappa_1^4}$, and $\beta_t= \frac{1}{4\kappa_1\phi_{\max}(s) } $.

Notice again that in Section \ref{sec:proof-est-dual-conv}, dual variable satisfies at time $T_0$: 
$\sup_{\cB^+(\brho_0,\delta_\rho)}\norm{\widehat{\blambda}_{T_0}-\blambda^*(\brho)}_{1}\le\sqrt{ 32\sqrt{s_0}DZ\delta_{\est}(T_0)/\cL}\le \delta_{\lambda}$, with probability at least $1-2T^{-2}$.

Now, we take 
\begin{equation}\label{eq:est-induction-rate}
    \delta_{\est}(t) = C_0\frac{\kappa_1^2 
\sigma D\sqrt{s_0\log(mKdT)}}{ \gamma(K)\zeta(K) \sqrt{t}},
\end{equation}
and justify its validity. Clearly, from Theorem \ref{thm:sparse-est}, it is correct for all $t\ge \kappa^4 s_0^2 D^2\log(mKdT)/\phi_{\min}^2(s)$. We now only need to select a large $T_0$ such that the iterative relationship fits $ \delta_{\est}(t)$ for $t\ge T_0$. To see how large $T_0$ is required, we suppose the rate is valid for any $T_0\le j \le t-1$, then at time $t$ with $t\ge C \kappa_1^4 s_0^2D^2\log(mKdT)/\gamma^2(K)\zeta^2(K)$, we have
\begin{equation*}
    \begin{aligned}
               & \max_{a}\norm{\bmu_{a,t}^{ \mathsf{s} } - \bmu_{a}^\star  }_2^2\vee \norm{\widehat{\bW}_{a,t} -\bW_{a}^{\star}}_{2,\max}^2 \\
      & \le \left(1-\frac{1}{7\kappa_1^2} +C\frac{\kappa_1^2 s_0^{\frac{3}{2}} D^3M_{\sfm} Z }{\gamma(K)\zeta(K)  }\frac{\sum_{j=T_0+1}^{t} \delta_{\est}(j-1)}{t}  \right) \delta_{\est}^2(t-1) +  C\frac{\kappa_1^2s_0\sigma^2 D^2 \log(mKdT) }{\gamma^2(K)\zeta^2(K) t }  \\
      &\le \left(1-\frac{1}{7\kappa_1^2} +C\cdot C_0\frac{\kappa_1^2 s_0^{\frac{3}{2}} D^3M_{\sfm} Z }{\gamma(K)\zeta(K)  }\frac{\kappa_1^2 
\sigma D\sqrt{s_0\log(mKdT)}}{ \gamma(K)\zeta(K) }\left(\sqrt{\frac{1}{t}}-\sqrt{\frac{T_0}{t^2}}\right)  \right) \delta_{\est}^2(t-1) 
\\
& \quad +  C\frac{\kappa_1^2s_0\sigma^2 D^2 \log(mKdT) }{\gamma^2(K)\zeta^2(K) t }  \\
& \le \left(1-\frac{1}{7\kappa_1^2} +C\cdot C_0\frac{\kappa_1^2 s_0^{\frac{3}{2}} D^3M_{\sfm} Z }{\gamma(K)\zeta(K)   }\frac{\kappa_1^2 
\sigma D\sqrt{s_0\log(mKdT)}}{ \gamma(K)\zeta(K) \sqrt{T_0} }  \right) \delta_{\est}^2(t-1) 
\\
& \quad +  C\frac{\kappa_1^2s_0\sigma^2 D^2 \log(mKdT) }{\gamma^2(K)\zeta^2(K) t }. 
    \end{aligned}
\end{equation*}
where we use $\sqrt{\frac{1}{t}}-\sqrt{\frac{T_0}{t^2}}\le  \frac{1}{4\sqrt{T_0}}$ for any $t$. Taking 
\begin{equation*}
    T_0\ge C\cdot C_0^2 \frac{\kappa_1^{12}  \sigma^2 s_0^{4} D^8 M^2_{\sfm} Z^2\log(mKdT) }{ \gamma^4(K)\zeta^4(K) },
\end{equation*}
we have for any $t\ge T_0+1$
\begin{equation*}
    \max_{a}\norm{\bmu_{a,t}^{ \mathsf{s} } - \bmu_{a}^\star  }_2^2\vee \norm{\widehat{\bW}_{a,t} -\bW_{a}^{\star}}_{2,\max}^2 \le  \left(1-\frac{1}{10\kappa_1^2}  \right) \delta_{\est}^2(t-1)  +  C\frac{\kappa_1^2s_0\sigma^2 D^2 \log(mKdT) }{\gamma^2(K)\zeta^2(K) t }.
\end{equation*}
When $t\ge 40\kappa_1^2$, we have
\begin{equation*}
    \begin{aligned}
        & \max_{a}\norm{\bmu_{a,t}^{ \mathsf{s} } - \bmu_{a}^\star  }_2^2\vee \norm{\widehat{\bW}_{a,t} -\bW_{a}^{\star}}_{2,\max}^2  \le  \left(1-\frac{1}{10\kappa_1^2}  \right) \delta_{\est}^2(t-1)  +  C\frac{\kappa_1^2s_0\sigma^2 D^2 \log(mKdT) }{\gamma^2(K)\zeta^2(K) t } \\
        & \le \left(1-\frac{1}{20\kappa_1^2}  \right) \delta_{\est}^2(t-1)  +  C\frac{\kappa_1^2s_0\sigma^2 D^2 \log(mKdT) }{\gamma^2(K)\zeta^2(K) t } - \frac{1}{20\kappa_1^2} C_0\cdot \frac{\kappa_1^4s_0\sigma^2 D^2 \log(mKdT) }{\gamma^2(K)\zeta^2(K) t } \\
        &\le \frac{t-1}{t} \delta_{\est}^2(t-1) =  \left(C_0\frac{\kappa_1^2 
\sigma D\sqrt{s_0\log(mKdT)}}{ \gamma(K)\zeta(K) \sqrt{t}}\right)^2 = \delta_{\est}^2(t). 
    \end{aligned}
\end{equation*}
Thus, by induction, we have $\max_{a}\norm{\bmu_{a,t}^{ \mathsf{s} } - \bmu_{a}^\star  }_2\vee \norm{\widehat{\bW}_{a,t} -\bW_{a}^{\star}}_{2,\max}\le \delta_{\est}(t)$ for all $t\ge T_0$ with probability at least $1-T^{-2}$. 

Section \ref{sec:proof-est-dual-conv} shows that we need to choose 
$$T_0\ge C \frac{ \kappa^4 s_0\sigma^2D^2\log(mKdT) \cdot  s_0D^2Z^2 }{\cL^2\delta_{\lambda}^4 } = \frac{\kappa^4 \sigma^2 s_0^2 D^4 Z^2\log(mKdT)}{ \cL^2\delta_{\lambda}^4 } 
$$ 
to ensure the relationship $\sqrt{ 32\sqrt{s_0}DZ\delta_{\est}(T_0)/\cL}\le \delta_{\lambda}$. Thus, our final choice of $T_0$ is
\begin{equation}\label{eq:T0-value}
    T_0 \ge C  \frac{\kappa_1^{12}  \sigma^2 s_0^{4} D^8 M^2_{\sfm} Z^2\log(mKdT) }{ \gamma^4(K)\zeta^4(K) } \bigvee \frac{\kappa^4 \sigma^2 s_0^2 D^4 Z^2\log(mKdT)}{ \cL^2\delta_{\lambda}^4 }. 
\end{equation}
\subsection{Proof of Theorem \ref{thm:resolving-regret}}

Notice that, since our decision is made based on the estimation of parameters from time $t-1$, the reward we collected can not be explicitly expressed as a part of dual function. Therefore, the regret decomposition in \cite{li2022online,balseiro2023best,ma2024optimal,bray2024logarithmic} is not directly applicable to our setting. Contrarily, we propose the following new decomposition of the regret with :

\begin{lemma}\label{lemma:regret-decomp-log}
    Denote $\tau$ as the first time that $\rho_{t}\notin \cB^+(\brho_0,\delta_{\rho}')$ or $\tau=T$ if   $\rho_{t}\in \cB^+(\brho_0,\delta_{\rho}')$ always holds for all $t\in[T]$.
\begin{equation*}
\begin{aligned}
         \operatorname{Regret}(\pi)  &\le  C M_{\sfm} s_0 D^2Z^2 \bE\sum_{t=T_0+1}^{\tau}\max_{a}\norm{\bmu_{a,t-1}^{ \mathsf{s} } - \bmu_{a}^\star  }_2^2\vee \norm{\widehat{\bW}_{a,t-1} -\bW_{a}^{\star}}_{2,\max}^2 
    \\ 
     & +C M_{\sfm} B_{\max}^2\bE\sum_{t=T_0+1}^{\tau}  \norm{\widehat{\blambda}_{t-1}-\blambda^*}_1^2 +C (R_{\max}+2 B_{\max}\frac{\VU}{C_{\min}}) \bE (T-\tau)
     \\
     & + \widetilde{C}(R_{\max}+2 B_{\max}\frac{\VU}{C_{\min}})Z^2\log(mKdT)  
\end{aligned}
\end{equation*}
\end{lemma}
Starting from Lemma \ref{lemma:regret-decomp-log}, we proceed to deal with the regret by noticing that:
\begin{enumerate}
    \item The fist term can be bounded by $O(\log T\log(mKdT))$ due to the primal estimation from Theorem \ref{thm:est-induction}
    \item The second term can be decomposed by two terms by 
    $$\bE\sum_{t=T_0+1}^{\tau}  \norm{\widehat{\blambda}_{t-1}-\blambda^*}_1^2 \le 2 \bE\sum_{t=T_0+1}^{\tau}  \norm{\widehat{\blambda}_{t-1}-\blambda^*(\brho_{t-1})}_1^2 + 2\bE\sum_{t=T_0+1}^{\tau}  \norm{\blambda^*(\brho_{t-1}) - \blambda^*}_1^2, 
    $$
    while the term $\bE\sum_{t=T_0+1}^{\tau}  \norm{\widehat{\blambda}_{t-1}-\blambda^*(\brho_{t-1})}_1^2 $ can been controlled by  Theorem \ref{thm:est-dual-convergence}.
\end{enumerate}
Therefore, we only need to control the term $\bE\sum_{t=T_0+1}^{\tau}  \norm{\blambda^*(\brho_{t-1}) - \blambda^*}_1^2$ and $\bE (T-\tau)$. Denote a new stopped series as $\brho'_t=\brho_{t\wedge\tau}$. We now show that there two terms can all be bounded by $\bE\sum_{t=1}^T\norm{\brho_{t-1}'-\brho_0}_{2,\sfB}^2$. Here the norm $\norm{\cdot}_{2,\sfB}$ means the $\ell_2$ norm only on binding dimensions $I_{\sfB}$. For the first  term, by the proof of Lemma \ref{lemma:region-B+}, we have
\begin{equation}\label{eq:itr-rho-1}
    \bE\sum_{t=T_0+1}^{\tau}  \norm{\blambda^*(\brho_{t-1}) - \blambda^*}_1^2 \le  \frac{1}{\cL}\bE\sum_{t=T_0+1}^{\tau}  \norm{\brho_{t-1}'-\brho_0}_{\infty,\sfB}^2 \le  \frac{1}{\cL}\bE\sum_{t=T_0+1}^{\tau}  \norm{\brho_{t-1}'-\brho_0}_{2,\sfB}^2
\end{equation}

% For any $\bx\in\bR^m$, define $\norm{\bx}_{\infty,+}=\sup_{i\in I_{\sfB}}\abs{x_i}\vee \sup_{i\in I_{\NB}} (x_i)_+ $, which measures how far the vector $\brho_0-\bx$ goes toward the boundary of $\cB^+(\brho_0,\delta'_{\rho})$. We have $\norm{\bx}_{\infty,+}\le \norm{\bx}_2$, and for any $t$,
For all the binding dimensions, denote its minimum stopping time that $\rho_{t,i} $ is not in $\cB^+(\brho_0,\delta_{\rho}')$ for some $i\in I_{\sfB}$ as $\tau_{\sfB}$. Also define the non-binding stopping time analogously as $\tau_{\NB}$. Then we have $\tau=\tau_{\sfB} \wedge\tau_{\NB}$. Therefore,
\begin{equation*}
    \begin{aligned}
       \bp(\tau_{\sfB} \le t)= \bp(\norm{\brho_0-\brho_{t}'}_{\infty,\sfB} \ge \delta'_{\rho})\le \frac{\bE \norm{\brho_0-\brho_{t}'}_{\infty,\sfB}^2 }{(\delta'_{\rho})^2} \le  \frac{\bE \norm{\brho_0-\brho_{t}'}_{2,\sfB}^2 }{(\delta'_{\rho})^2}
    \end{aligned}
\end{equation*}

\begin{equation}\label{eq:itr-rho-2}
    \begin{aligned}
        \bE(T-\tau_{\sfB}) = \sum_{t=1}^T \bp(T-\tau_{\sfB}\ge t) \le  \sum_{t=1}^T \bp(\tau_{\sfB}\le t) \le 1+\sum_{t=1}^T \frac{\bE \norm{\brho_0-\brho_{t-1}'}_{2,\sfB}^2 }{(\delta'_{\rho})^2}
    \end{aligned}
\end{equation}
Since we do the uniform sampling before the initial phase $T_0$, we can turn to study the deviation of $\brho_t$ after $\brho_{T_0}$ since 
\begin{equation*}
\begin{gathered}
        \norm{\brho_{t}'-\brho_{0}}_{2,\sfB}^2\le 2   \norm{\brho_{t}'-\brho_{T_0}}_{2,\sfB}^2 + 2 \norm{\brho_{T_0}-\brho_{0}}_{2,\sfB}^2 \\
    \norm{ \brho_{T_0}-\brho_{0}}_{2,\sfB}^2 =  \norm{\frac{\sum_{t=1}^{T_0}\brho_{0}-\bW^\star_{y_t}\bx_t}{T-T_0}}_2^2 \le  \frac{2 mT_0\widetilde{B}_{\max} }{T}.
\end{gathered}
\end{equation*}
Here we use the fact that $\tau\ge T_0+1$ since the initial phase is only $\widetilde{C}\log(T)$. We study the iterative relationship of $\brho_{t}'-\brho_{T_0}$ after time $T_0$ for each $i\in I_{\sfB}$:
\begin{equation*}
\begin{aligned}
        \bE \left(\brho_{t}'-\brho_{T_0}\right)_i^2 
        & =  \bE\left(\brho_{t-1}'-\brho_{T_0}\right)_i^2 \mathbbm{1}\{t-1< \tau\}  +2\bE\left(\brho_{t-1}'-\brho_{T_0}\right)_i \frac{\left(\brho_{t-1}-\bW^\star_{y_t}\bx_t\right)_i}{T-t} \mathbbm{1}\{t-1< \tau\} 
        \\
        &\quad +  \bE \frac{\left(\brho_{t-1}-\bW^\star_{y_t}\bx_t-\omega_{t,i}\right)_i^2}{(T-t)^2}\mathbbm{1}\{t-1< \tau\} 
        \\
        &\le  \bE\left(\brho_{t-1}'-\brho_{T_0}\right)_i^2 \mathbbm{1}\{t-1< \tau\} + \frac{\widetilde{B}_{\max}^2}{(T-t)^2} 
        \\
        & \quad+ 2\bE\left(\brho_{t-1}'-\brho_{T_0}\right)_i \frac{\left(\brho_{t-1}-\bW^\star_{a^*_t(\blambda^*(\brho_{t-1}'))}\bx_t\right)_i}{T-t} \mathbbm{1}\{t-1< \tau\} \\
        & \quad +  2\bE\left(\brho_{t-1}'-\brho_{T_0}\right)_i \frac{\left(\bW^\star_{a^*_t(\blambda^*(\brho_{t-1}'))}\bx_t- \bW^\star_{y_t}\bx_t\right)_i}{T-t} \mathbbm{1}\{t-1< \tau\}
        \\
        & \le \bE\left(\brho_{t-1}'-\brho_{T_0}\right)_i^2  + \frac{\widetilde{B}_{\max}^2}{(T-t)^2}  +2\bE B_{\max}\frac{\abs{\left(\brho_{t-1}'-\brho_{T_0}\right)_i}\bp\left( y_t\neq a^*_t(\blambda^*(\brho_{t-1}'))\middle| \cH_{t-1} \right)}{T-t}
\end{aligned}
\end{equation*}
Here in the second inequality, we use the fact that $a^*_t(\blambda^*(\brho_{t-1}'))$ is the best primal choice for the dual problem $\min_{\blambda} D(\blambda,\brho_{t-1})$, i.e., $\bE\left(\brho_{t-1}-\bW^\star_{a^*_t(\blambda^*(\brho_{t-1}'))}\bx_t\right)_i =0$ for each $i\in I_{\sfB}$.

We can control the probability of $\bp\left( y_t\neq a^*_t(\blambda^*(\brho_{t-1}'))\middle| \cH_{t-1} \right)$ by the analyses we have conducted in Section \ref{sec:proof-reg-decomp-log}:
\begin{equation*}
\begin{aligned}
        & \bE\bp\left( y_t\neq a^*_t(\blambda^*(\brho_{t-1}'))\middle| \cH_{t-1} \right) \\
    &\le   CM_{\sfm}\sqrt{s_0} DZ \bE\max_{a}\norm{\bmu_{a,t-1}^{ \mathsf{s} } - \bmu_{a}^\star  }_2\vee \norm{\widehat{\bW}_{a,t-1} -\bW_{a}^{\star}}_{2,\max} +  CM_{\sfm} B_{\max} \bE\norm{\widehat{\blambda}_{t-1}-\blambda^*(\brho_{t-1})}_1, \\
    &\le   CM_{\sfm}\sqrt{s_0} DZ  \sqrt{\frac{ 
\sigma^2  D^2{s_0\log(mKdT)}}{ \gamma^2(K)\zeta^2(K) t }} + C B_{\max}^2 M_{\sfm }^2\sqrt{s_0}DZ/\cL  \sqrt{\frac{ 
\sigma^2 m D^2{s_0\log(mKdT)}}{ \gamma^2(K)\zeta^2(K) t }} \\
& \le  C \frac{\kappa_1^2 B_{\max}^2 M_{\sfm }^2\sqrt{s_0}DZ}{\cL}  \sqrt{\frac{ 
\sigma^2 m D^2{s_0\log(mKdT)}}{ \gamma^2(K)\zeta^2(K) t }}. 
\end{aligned}
\end{equation*}
Here, we use Theorem \ref{thm:est-dual-convergence} and Theorem \ref{thm:est-induction} by selecting $\delta_{\est} =C \kappa_1^2  \sqrt{\frac{ 
\sigma^2 m D^2{s_0\log(mKdT)}}{ \gamma^2(K)\zeta^2(K) t }}$ in Theorem \ref{thm:est-dual-convergence}.

This further gives the bound on all the binding dimensions:
\begin{equation*}
    \begin{aligned}
         \bE\norm{\brho_{t}'-\brho_{T_0}}^2_{2,\sfB}&\le \bE \norm{\brho_{t-1}'-\brho_{T_0}}^2_{2,\sfB}  + \frac{m \widetilde{B}_{\max}^2}{(T-t)^2} 
         \\
         &\quad +2\bE B_{\max}\sqrt{m}\frac{\sqrt{\bE \norm{\brho_{t-1}'-\brho_{T_0}}^2_{2,\sfB}}\bp\left( y_t\neq a^*_t(\blambda^*(\brho_{t-1}'))\middle| \cH_{t-1} \right)}{T-t}  
         \\
         &\le \bE \norm{\brho_{t-1}'-\brho_{T_0}}^2_{2,\sfB}  + \frac{m \widetilde{B}_{\max}^2}{(T-t)^2}  
         \\
         & \quad + C \frac{\kappa_1^2 m B_{\max}^3 M_{\sfm }^2\sqrt{s_0}DZ}{\cL}  \sqrt{\frac{ 
\sigma^2  D^2{s_0\log(mKdT)}}{ \gamma^2(K)\zeta^2(K) t }}\frac{ \sqrt{\bE \norm{\brho_{t-1}'-\brho_{T_0}}^2_{2,\sfB}} }{T-t}.
    \end{aligned}
\end{equation*}
By induction, we can derive that:
\begin{equation*}
     \bE\norm{\brho_{t}'-\brho_{T_0}}^2_{2,\sfB} \le  C \frac{\kappa_1^4 m^2 B_{\max}^6 M_{\sfm }^4 s_0^2 D^4Z^2}{\cL^2}  {\frac{ 
\sigma^2  {\log(mKdT)}}{ \gamma^2(K)\zeta^2(K)  }}\frac{t}{T(T-t+1)}
\end{equation*}
Plugging this into \eqref{eq:itr-rho-2}, we have the following bound:
\begin{equation}\label{eq:binding-stp}
     \bE(T-\tau_{\sfB}) \le 1+\sum_{t=1}^T \frac{\bE \norm{\brho_0-\brho_{t-1}'}_{2,\sfB}^2 } {(\delta'_{\rho})^2} \le  C \frac{\kappa_1^4 m^2 B_{\max}^6 M_{\sfm }^4 s_0^2 D^4Z^2 \sigma^2  {\log(mKdT)} \log T}{\cL^2 \delta'^2_{\rho} \gamma^2(K)\zeta^2(K) } 
\end{equation}

However, controlling the non-binding dimension is more challenging if one wants a sharp bound with respect to $m$. Our approach here is different from the non-binding dimension analysis in \cite{li2022online,ma2024optimal} as the regret bound therein are  $O(m^2)$ due using dimension splitting. In our case, using dimension splitting will result in a $O(m^3)$ rate since the parameter space in our problem in quite large. To tackle this issue, we invoke the following $\ell_{\infty}$ version of Doob's maximal inequality to control the stopping time:

\begin{lemma}[Doob's Maximal Inequality] \label{lemma:doobs}Suppose that $\{M_t(\alpha)\}_{\alpha\in\mathcal{A} }$ are the differences of a group of martingales  relative to a  filtration $\{\cH_t\}_{t=0}^T$ with $\bE[M_{t}(\alpha)|\cH_{t-1}]=0 $, where the index set $\mathcal{A}$ is finite. Then we have
\begin{equation*}
    \bp(\max_{ \alpha\in\mathcal{A} }\max_{t\in [T]} \sum_{j=1}^t M_j(\alpha) \ge \delta )\le \frac{\bE \max_{ \alpha\in\mathcal{A}  }\left[ \sum_{j=1}^T M_j(\alpha)\right]^2 }{\delta^2}
\end{equation*}
    
\end{lemma}
We now define an auxiliary series $\brho^{'-}_{t}$ with
$$
\rho^{'-}_{t,i}= \begin{cases} \rho^{'-}_{t,i}, &\text{ if }i\in I_{\textsf{B}} \\ \rho_{0,i}-\delta_{\rho}', &\text{ if }i\in I_{\textsf{NB}}\end{cases}.
$$
Clearly, we have $\blambda^*(\brho_{t-1}')=\blambda^*(\brho^{'-}_{t-1})$. This suggests that when $i\in I_{\NB}$, $\bE\left[\left(\brho_{t-1}^{'-}-\bW^\star_{a^*_t(\blambda^*(\brho_{t-1}'))}\bx_t\right)_i\middle| \cH_{t-1} \right] >0$, i.e., $\bE \left[\left(\bW^\star_{a^*_t(\blambda^*(\brho_{t-1}'))}\bx_t\right)_i\middle|\cH_{t-1}\right] < \rho_{0,i}-\delta_{\rho}'$. We thus have

\begin{equation}\label{eq:nb-stopping}
    \begin{aligned}
        \bp(\tau_{\NB}\le t) &=   \bp\left(\max_{i\in \NB}\max_{t'\in [t]}\sum_{j=1}^{t'}\left[ \left(\bW^\star_{y_j}\bx_j\right)_i - \rho_{0,i}+ \delta_{\rho}'\right] \ge t\delta_{\rho}' \right)  \\
        &  \le \bp\left(\max_{i\in \NB}\max_{t'\in [t]}\sum_{j=1}^{t'}\left[ \left(\bW^\star_{y_j}\bx_j\right)_i - \rho_{0,i}+ \delta_{\rho}'\right] \ge t\delta_{\rho}' \right) 
        \\
        &  \le \bp\left(  \max_{i\in \NB}\max_{t'\in [t]}\sum_{j=1}^{t'} \left(\bW^\star_{y_j}\bx_j\right)_i- \bE \left[\left(\bW^\star_{a^*_j(\blambda^*(\brho_{j-1}'))}\bx_t\right)_i\middle|\cH_{j-1}\right] \ge t\delta_{\rho}' \right)
        \\
        & = \bp\left(  \max_{i\in \NB}\max_{t'\in [t]}\sum_{j=1}^{t'} \left(\bW^\star_{y_j}\bx_j\right)_i - \bE \left[\left(\bW^\star_{y_j}\bx_j\right)_i\middle|\cH_{j-1}\right] \right.
        \\
        & \quad\quad\quad \left.
        +\bE \left[\left(\bW^\star_{y_j}\bx_j\right)_i\middle|\cH_{j-1}\right] - \bE \left[\left(\bW^\star_{a^*_j(\blambda^*(\brho_{j-1}'))}\bx_t\right)_i\middle|\cH_{j-1}\right] \ge t\delta_{\rho}' \right)
        \\
        & \le \underbrace{\bp\left(  \max_{i\in \NB}\max_{t'\in [t]}\sum_{j=1}^{t'} \left(\bW^\star_{y_j}\bx_j\right)_i - \bE \left[\left(\bW^\star_{y_j}\bx_j\right)_i\middle|\cH_{j-1}\right] \ge \frac{t\delta_{\rho}'}{2}\right) }_{\mathfrak{F}_1 }
        \\
        & \quad +  \underbrace{\bp\left(\max_{i\in \NB}\max_{t'\in [t]}\sum_{j=1}^{t'} \bE \left[\left(\bW^\star_{y_j}\bx_j\right)_i\middle|\cH_{j-1}\right] - \bE \left[\left(\bW^\star_{a^*_j(\blambda^*(\brho_{j-1}'))}\bx_t\right)_i\middle|\cH_{j-1}\right] \ge\frac{t\delta_{\rho}'}{2} \right)}_{\mathfrak{F}_2 }.
    \end{aligned}
\end{equation}
For $\mathfrak{F}_1 $, we use Lemma \ref{lemma:doobs}, which gives:
\begin{equation*}
    \mathfrak{F}_1  \le 4 \bE  \frac{\max_{i\in[m]}\left(\sum_{j=1}^{t} \left(\bW^\star_{y_j}\bx_j\right)_i - \bE \left[\left(\bW^\star_{y_j}\bx_j\right)_i\middle|\cH_{j-1}\right]\right)^2}{t^2\delta_{\rho}'^2}.
\end{equation*}
Since $H_j:=\left(\bW^\star_{y_j}\bx_j\right)_i - \bE \left[\left(\bW^\star_{y_j}\bx_j\right)_i\middle|\cH_{j-1}\right]$ is martingale difference bounded by $\abs{H_j}\le 2B_{\max}$, we can use Azuma Hoeffding’s inequality:
\begin{equation*}
    \bp\left(\abs{\sum_{j=1}^t H_j} \ge z\right)\le 2\exp\left(\frac{z^2}{2B_{\max}^2 t}\right),
\end{equation*}
which leads to
\begin{equation*}
    \begin{aligned}
        \bp\left( \max_{i\in[m]}\left(\sum_{j=1}^{t} \left(\bW^\star_{y_j}\bx_j\right)_i - \bE \left[\left(\bW^\star_{y_j}\bx_j\right)_i\middle|\cH_{j-1}\right]\right)^2 \ge z^2\right)\le 2\exp\left(\log m-\frac{z^2}{2B_{\max}^2 t}\right)\wedge 1.
    \end{aligned}
\end{equation*}
Integrating gives 
\begin{equation*}
\begin{aligned}
        &\bE \max_{i\in[m]}\left(\sum_{j=1}^{t} \left(\bW^\star_{y_j}\bx_j\right)_i - \bE \left[\left(\bW^\star_{y_j}\bx_j\right)_i\middle|\cH_{j-1}\right]\right)^2 
    \\
    &\le \int_{0}^{4B_{\max}^2}   \bp\left( \max_{i\in[m]}\left(\sum_{j=1}^{t} \left(\bW^\star_{y_j}\bx_j\right)_i - \bE \left[\left(\bW^\star_{y_j}\bx_j\right)_i\middle|\cH_{j-1}\right]\right)^2 \ge \varepsilon\right)d\varepsilon 
    \\
    &\le 2 B^2_{\max} t\log m +   \int_{ 2 B^2_{\max} t\log m}^{4B_{\max}^2} 2\exp\left(\log m-\frac{\varepsilon}{2B_{\max}^2 t}\right) d\varepsilon
    \\
    &\le  2 B^2_{\max} t\log m  +4 B^2_{\max}t \le 6 B^2_{\max} t\log m,
\end{aligned}
\end{equation*}
i.e.,
\begin{equation*}
    \mathfrak{F}_1  \le \frac{24 B_{\max}^2\log m}{t \delta_{\rho}'^2}.
\end{equation*}
We now look at $\mathfrak{F}_2$:
\begin{equation*}
\begin{aligned}
        \mathfrak{F}_2 & \le \bp\left(\max_{i\in \NB}\max_{t'\in [t]}\sum_{j=1}^{t'} B_{\max}\bp\left(y_j\neq a^*_j(\blambda^*(\brho_{j-1}')) \middle| \cH_{j-1} \right) \ge \frac{t\delta_{\rho}'}{2} \right) 
    \\
    & \le \bp\left(\sum_{j=1}^{t} B_{\max}\bp\left(y_j\neq a^*_j(\blambda^*(\brho_{j-1}')) \middle| \cH_{j-1} \right) \ge \frac{t\delta_{\rho}'}{2} \right) 
    \\
    & \le \bp\left(\sum_{j=1}^{t} B_{\max}M_{\sfm}  \widehat{\gamma}_{j-1} \ge \frac{t\delta_{\rho}'}{2} \right) 
\end{aligned}
\end{equation*}
where we apply the analyses we have conducted in Section \ref{sec:proof-reg-decomp-log}, and write the term
\begin{equation*}
   \widehat{\gamma}_{j-1} := \sqrt{s_0} DZ \max_{a}\norm{\bmu_{a,j-1}^{ \mathsf{s} } - \bmu_{a}^\star  }_2\vee \norm{\widehat{\bW}_{a,j-1} -\bW_{a}^{\star}}_{2,\max} +  B_{\max} \norm{\widehat{\blambda}_{j-1}-\blambda^*(\brho_{j-1})}_1.
\end{equation*}
By Theorem \ref{thm:est-dual-convergence} and \ref{thm:est-induction} and selecting $\delta_{\est} =C \kappa_1^2  \sqrt{\frac{ 
\sigma^2 m D^2{s_0\log(mKdT)}}{ \gamma^2(K)\zeta^2(K) t }}$ in Theorem \ref{thm:est-dual-convergence}, we know that for any $t\ge T_0+1$,
\begin{equation*}
\begin{aligned}
         &\sum_{j=1}^{t} B_{\max}\bp\left(y_j\neq a^*_j(\blambda^*(\brho_{j-1}')) \middle| \cH_{j-1} \right)
         \\
         &\le B_{\max} T_0+\sum_{j=T_0+1}^t B_{\max} M_{\sfm}\widehat{\gamma}_{t} 
\\
&\le B_{\max} \widetilde{C}\log(mKdT)+ C \frac{\kappa_1^2 B_{\max}^2 M_{\sfm }^2\sqrt{s_0}DZ}{\cL}  \sqrt{\frac{ 
\sigma^2 m D^2{s_0\log(mKdT) t}}{ \gamma^2(K)\zeta^2(K)  }}.
\\
\end{aligned}
\end{equation*}
with probability at least $1-T^{-2}$. This further implies that when 
$$t\gtrsim \frac{B_{\max}T_0}{\delta_{\rho}'^2} + \frac{1}{\delta_{\rho}'^4}  \frac{\kappa_1^4 B_{\max}^4 M_{\sfm }^4s_0^2D^4 Z^2 \sigma^2 m \log(mKdT) }{\cL^2 \gamma^2(K)\zeta^2(K)},$$
then with probability at least $1-T^{}$, we have
\begin{equation*}
    \sum_{j=1}^{t} B_{\max}\bp\left(y_j\neq a^*_j(\blambda^*(\brho_{j-1}')) \middle| \cH_{j-1} \right)   <  \frac{t\delta_{\rho}'}{2} .  
\end{equation*}
We therefore conclude that 
\begin{equation}\label{eq:F2-bound}
    \begin{aligned}
          \mathfrak{F}_2 \le \frac{1}{T}, \text{ for } t\gtrsim \frac{B_{\max}T_0}{\delta_{\rho}'^2} + \frac{1}{\delta_{\rho}'^4}  \frac{\kappa_1^4 B_{\max}^4 M_{\sfm }^4s_0^2D^4 Z^2 \sigma^2 m \log(mKdT) }{\cL^2 \gamma^2(K)\zeta^2(K)}.
    \end{aligned}
\end{equation}
With the bounds on $ \mathfrak{F}_1$, $ \mathfrak{F}_2$, we can return to \eqref{eq:nb-stopping} can control the term $ \bp(\tau_{\NB}\le t)$ by
\begin{equation}\label{eq:nonbinding-stp}
     \bp(\tau_{\NB}\le t) \le   \mathfrak{F}_1 + \mathfrak{F}_2 \le \frac{24 B_{\max}^2\log m}{t \delta_{\rho}'^2} + \frac{1}{T}, \text{ for all } t \text{ in }\eqref{eq:F2-bound}.
\end{equation}

Combining \eqref{eq:binding-stp} with \eqref{eq:nonbinding-stp}, we can show that:
\begin{equation}
    \bE(T-\tau)\le \sum_{t=1}^T  \bp(\tau_{\NB}\le t) +\bp(\tau_{\sfB}\le t)\le  C \frac{\kappa_1^4 m^2 B_{\max}^6 M_{\sfm }^4 s_0^2 D^4Z^2 \sigma^2  {\log(mKdT)} \log T}{\cL^2 \delta'^4_{\rho} \gamma^2(K)\zeta^2(K) } 
\end{equation}
Now, plugging this bound into Lemma \ref{lemma:regret-decomp-log}, and combined with Theorem \ref{thm:est-dual-convergence}, \ref{thm:est-induction}, we can prove the claim that:
\begin{equation*}
    \operatorname{Regret}(\pi)\lesssim \frac{\kappa_1^4 m^2 B_{\max}^6 \widetilde{B}_{\max} M_{\sfm }^4 s_0^2 D^4Z^2 \sigma^2  {\log(mKdT)} \log T}{\cL^2 \delta'^4_{\rho} \gamma^2(K)\zeta^2(K) } 
\end{equation*}

\subsection{Proof of Lemma \ref{lemma:doobs}}
Denote the stopping time $\tau=\inf_{t}\{ \max_{ \alpha\in\mathcal{A} } \sum_{j=1}^t M_j(\alpha) \ge \delta \}\wedge T$. Then we have
\begin{equation*}
     \bp(\max_{ \alpha\in\mathcal{A} }\max_{t\in [T]} \sum_{j=1}^t M_j(\alpha) \ge \delta ) =  \bp(\max_{ \alpha\in\mathcal{A} } \sum_{j=1}^\tau M_j(\alpha) \ge \delta ) \le \frac{\bE\left[\max_{ \alpha\in\mathcal{A} } \sum_{j=1}^\tau M_j(\alpha)\right]^2 }{\delta^2}
\end{equation*}
Denote $\alpha_{\tau}=\arg\max_{ \alpha\in\mathcal{A} } \sum_{j=1}^\tau M_j(\alpha)$, where $\{\alpha_{\tau}\in\cD\}\in\cH_{\tau}$ for any $\cD$. We then have
\begin{equation*}
\begin{aligned}
        \bE \max_{ \alpha\in\mathcal{A}  }\left[ \sum_{j=1}^T M_j(\alpha)\right]^2 & \ge \bE \left[ \sum_{j=1}^T M_j(\alpha_{\tau})\right]^2 =  \bE \left[ \sum_{j=1}^\tau  M_j(\alpha_{\tau}) + \sum_{j=\tau+1}^T  M_j(\alpha_{\tau})\right]^2  \\
        & = \bE \left[\sum_{j=1}^\tau  M_j(\alpha_{\tau}) \right]^2 +\bE\bE\left[ \left(\sum_{j=\tau+1}^T  M_j(\alpha_{\tau}) \right)^2\mid\cH_{\tau}\right] \\
        & \ge \bE \left[\sum_{j=1}^\tau  M_j(\alpha_{\tau}) \right]^2 =  \bE\left[\max_{ \alpha\in\mathcal{A} } \sum_{j=1}^\tau M_j(\alpha)\right]^2.
\end{aligned}
\end{equation*}
Therefore, we prove the claim.

\subsection{Proof of Lemma \ref{lemma:regret-decomp-log}}\label{sec:proof-reg-decomp-log}
Since $\VU= T\cdot D(\blambda^*,\brho_0)$, we have
\begin{equation*}
    \begin{aligned}
         \operatorname{Regret}(\pi) & \le \bE\left[\sum_{t=1}^{\tau} D_t(\blambda^*,\brho_0) + \sum_{t=\tau+1}^{T} D_t(\blambda^*,\brho_0) - \sum_{t=1}^{\tau} \left\langle \bmu_{y_t}^\star,\bx_t \right\rangle \right] \\
         & \le  \bE\left[\sum_{t=1}^{\tau} \left\langle \bmu_{a_t^*}^\star-(\bW^\star_{a_t^*})^\top\blambda^* ,\bx_t \right\rangle  +\brho_0^\top \blambda^*  - \left\langle \bmu_{y_t}^\star- (\bW^\star_{y_t})^\top\blambda^* ,\bx_t \right\rangle  - (\blambda^*)^\top\bW^\star_{y_t} \bx_t \right] \\
         & \quad + R_{\max}\bE(T-\tau)  
         \\
         & \le  \underbrace{\bE\left[\sum_{t=1}^{\tau} \left\langle \bmu_{a_t^*}^\star-(\bW^\star_{a_t^*})^\top\blambda^* ,\bx_t \right\rangle  - \left\langle \bmu_{y_t}^\star- (\bW^\star_{y_t})^\top\blambda^* ,\bx_t \right\rangle\right]}_{\mathfrak{E}_1} + \underbrace{\bE\left[\sum_{t=1}^{\tau} \left\langle\blambda^*,\brho_0- \bW^\star_{y_t} \bx_t\right\rangle  \right]}_{\mathfrak{E}_2 }
         \\
         & \quad +R_{\max}\bE(T-\tau) 
    \end{aligned} 
\end{equation*}
We now control $\mathfrak{E}_1$ and $\mathfrak{E}_2$. For $\mathfrak{E}_1$, it is clear that when $\{a^*_t\neq y_t\}$, we can derive that 
\begin{equation*}
    \begin{aligned}
        \left\langle\bmu_{y_t,t-1}^{\mathsf{s}} -\widehat{\bW}_{y_t,t-1}^\top\widehat{\blambda}_{t-1},\bx_t \right\rangle >   \left\langle\bmu_{a_t^*,t-1}^{\mathsf{s}} -\widehat{\bW}_{a_t^*,t-1}^\top\widehat{\blambda}_{t-1},\bx_t \right\rangle
    \end{aligned}.
\end{equation*}
This further leads to 
\begin{equation*}
    \begin{aligned}
        & \left\langle \bmu_{a_t^*}^\star-(\bW^\star_{a_t^*})^\top\blambda^* ,\bx_t \right\rangle  - \left\langle \bmu_{y_t}^\star- (\bW^\star_{y_t})^\top\blambda^* ,\bx_t \right\rangle \\
        &< 3 \sqrt{s_0} DZ \max_{a}\norm{\bmu_{a,t-1}^{ \mathsf{s} } - \bmu_{a}^\star  }_2\vee \norm{\widehat{\bW}_{a,t-1} -\bW_{a}^{\star}}_{2,\max} + 2B_{\max}\norm{\widehat{\blambda}_{t-1}-\blambda^*}_1:=\widehat{\gamma}_{t-1}
    \end{aligned}
\end{equation*}
Using this inequality, we derive that when $t> T_0=\widetilde{C}Z^2\log(mKdT)$, 
\begin{equation*}
\begin{aligned}
        &\bE\left[\left\langle \bmu_{a_t^*}^\star-(\bW^\star_{a_t^*})^\top\blambda^* ,\bx_t \right\rangle  - \left\langle \bmu_{y_t}^\star- (\bW^\star_{y_t})^\top\blambda^* ,\bx_t \right\rangle\right]  \\
    & \le  \bE\widehat{\gamma}_{t-1} \mathbbm{1}\{ \left\langle \bmu_{a_t^*}^\star-(\bW^\star_{a_t^*})^\top\blambda^* ,\bx_t \right\rangle  - \left\langle \bmu_{y_t}^\star- (\bW^\star_{y_t})^\top\blambda^* ,\bx_t \right\rangle<\widehat{\gamma}_{t-1} \} 
    \\
    & \le M_{\sfm}\bE \widehat{\gamma}_{t-1}^2
    \\
    & = C M_{\sfm} s_0 D^2Z^2 \bE\max_{a}\norm{\bmu_{a,t-1}^{ \mathsf{s} } - \bmu_{a}^\star  }_2^2\vee \norm{\widehat{\bW}_{a,t-1} -\bW_{a}^{\star}}_{2,\max}^2 + C M_{\sfm} B_{\max}^2\bE \norm{\widehat{\blambda}_{t-1}-\blambda^*}_1^2
\end{aligned}
\end{equation*}

For $t\le T_0=\widetilde{C}Z^2\log(mKdT)$, we can simply bound the terms in $\mathfrak{E}_1$ by its size $2R_{\max}+2 B_{\max}\frac{\VU}{C_{\min}}$. We thus have:
\begin{equation}\label{eq:regret-decomp-log-1}
\begin{aligned}
        \mathfrak{E}_1 &\le \widetilde{C}(R_{\max}+2 B_{\max}\frac{\VU}{C_{\min}})Z^2\log(mKdT)  
        \\
        & + C M_{\sfm} s_0 D^2Z^2 \bE\sum_{t=T_0+1}^{\tau}\max_{a}\norm{\bmu_{a,t-1}^{ \mathsf{s} } - \bmu_{a}^\star  }_2^2\vee \norm{\widehat{\bW}_{a,t-1} -\bW_{a}^{\star}}_{2,\max}^2 
    \\ 
     & +C M_{\sfm} B_{\max}^2\bE\sum_{t=T_0+1}^{\tau}  \norm{\widehat{\blambda}_{t-1}-\blambda^*}_1^2.
\end{aligned}
\end{equation}
We now turn to control term $\mathfrak{E}_2$ by 
\begin{equation}\label{eq:regret-decomp-log-2}
\begin{aligned}
        \mathfrak{E}_2 &\le  \bE\left[\sum_{t=1}^{\tau} \left\langle\blambda^*,\brho_0- \bW^\star_{y_t} \bx_t\right\rangle  \right] 
        = \bE \left[ \left\langle\blambda^*, \tau\brho_0- \sum_{t=1}^{\tau} \bW^\star_{y_t} \bx_t\right\rangle  \right]  = \bE \left[ \left\langle\blambda^*, (T-\tau)\brho_{t} -  (T-\tau)\brho_0 \right\rangle  \right]  \\
        & = \bE (T-\tau)  \left\langle\blambda^*, \brho_{\tau} -  \brho_0 \right\rangle \le
        \bE (T-\tau) \norm{\blambda^*}_1 \max_{i\in I_{\sfB}} \abs{\brho_{\tau} -  \brho_0}_i \le 2\frac{\VU}{C_{\min}}\delta_{\rho} \bE (T-\tau)
\end{aligned}
\end{equation}
Combining \eqref{eq:regret-decomp-log-1}, \eqref{eq:regret-decomp-log-2}, we conclude that
\begin{equation*}
\begin{aligned}
         \operatorname{Regret}(\pi)  &\le  \mathfrak{E}_1 +\mathfrak{E}_2 + R_{\max}\bE(T-\tau) \le  \widetilde{C}(R_{\max}+2 B_{\max}\frac{\VU}{C_{\min}})Z^2\log(mKdT)  
        \\
        & + C M_{\sfm} s_0 D^2Z^2 \bE\sum_{t=T_0+1}^{\tau}\max_{a}\norm{\bmu_{a,t-1}^{ \mathsf{s} } - \bmu_{a}^\star  }_2^2\vee \norm{\widehat{\bW}_{a,t-1} -\bW_{a}^{\star}}_{2,\max}^2 
    \\ 
     & +C M_{\sfm} B_{\max}^2\bE\sum_{t=T_0+1}^{\tau}  \norm{\widehat{\blambda}_{t-1}-\blambda^*}_1^2 +C (R_{\max}+2 B_{\max}\frac{\VU}{C_{\min}}) \bE (T-\tau)
\end{aligned}
\end{equation*}

\subsection{Proof of Lemma \ref{lemma:max-grad-divcov}}
\begin{myproof}
    The idea essentially follows the proof of Lemma \ref{lemma:max-grad}, with some modifications in the martingale concentration argument. Notice that, in Algorithm \ref{alg:online-iht}, for any arm $a\in [K]$, we have 
\begin{equation*}
    \begin{aligned}
        \bg_{t} & = 2  \widehat{\bSigma}_t \bmu_{t-1} - \frac{2}{t}\sum_{j=1}^{t}  \bI\left\{y_t=a \right\} \bx_j r_j = \frac{2}{t}\sum_{j=1}^{t}\left( {  \bI\left\{y_j=a \right\}  \bx_j \bx_j^\top} \right)\left(\bmu_{t-1}-\bmu_{\star} \right) - \frac{2}{t}\sum_{j=1}^{t}  \bI\left\{y_t=a \right\} \bx_j \xi_j , \\
        & = 2 \widehat{\bSigma}_t (\bmu_{t-1} -\bmu_{\star})  - \frac{2}{t}\sum_{j=1}^{t}  \bI\left\{y_j=a \right\}\bx_j \xi_j.
    \end{aligned}
\end{equation*}
Still, we can write
\begin{equation*}
\begin{aligned}
         \abs{\left\langle \bg_t -\nabla f_t({\bmu}_{t-1}), {\bm e}_i  \right\rangle} & = \abs{ \left\langle 2 \left(\widehat{\bSigma}_t-\Bar{\bSigma}_t\right) (\bmu_{t-1} -\bmu_{\star})  - \frac{2}{t}\sum_{j=1}^{t} y_j\bx_j \xi_j/p_t , {\bm e}_i \right\rangle } \\
        & \le \underbrace{2\abs{ \left\langle \left(\widehat{\bSigma}_t-\Bar{\bSigma}_t \right) (\bmu_{t-1} -\bmu_{\star}) , {\bm e}_i \right\rangle }}_{\text{ Part 1}} + \underbrace{2\abs{\frac{1}{t}\sum_{j=1}^{t} \bI\left\{y_j=a \right\} \bx_{j,i} \xi_j  }}_{\text{ Part 2}} \\
\end{aligned}
\end{equation*}
We consider the two parts separately. 

In Part $1$, for any $i,k\in[d]$, by the martingale structure of $\frac{1}{t}\sum_{j=1}^{t} \bI\left\{y_j=a \right\} \bx_{j,i}\bx_{j,k}  -\Bar{\bSigma}_{t,ik}$:
\begin{equation*}
    \bE \sum_{j=1}^{t}\left[ \bI\left\{y_j=a \right\} \bx_{j,i}\bx_{j,k} \middle| \cH_{j-1} \right] -t\bar{\bSigma}_{t,ik} =0, \ \   \abs{ \bI\left\{y_j=a \right\}\bx_{j,i}\bx_{j,k}  - \bE \left[ \bI\left\{y_j=a \right\} \bx_{j,i}\bx_{j,k}\middle| \cH_{t-1}\right] }\le 2D^2,
\end{equation*}
We can use the Bernstein-type martingale concentration inequality in Lemma \ref{lemma:martingale-con} to derive the following bound: 
\begin{equation*}
    \begin{aligned}
    \bp\left( \abs{\frac{1}{t}\sum_{j=1}^{t} \bI\left\{y_j=a \right\} \bx_{j,i}\bx_{j,k}  -\Bar{\bSigma}_{t,ik} } \ge z \right)\le 2 \exp\left( -\frac{c z^2}{ D^4/t  + 2D^2 z/t  } \right),
    \end{aligned}
\end{equation*}
where we select $v^2 = D^4/t$, and $b=2D^2/t $. This leads to the concentration that with 
 probability at least $1-\epsilon$, 

\begin{equation*}
     \max_{i,k\in[d] }\abs{\frac{1}{t}\sum_{j=1}^{t} \bI\left\{y_j=a \right\} \bx_{j,i}\bx_{j,k}  -\Bar{\bSigma}_{t,ik} }  \le C D^2 \sqrt{\frac{\log(d /\epsilon)}{ t}}.
\end{equation*}

It follows from the process in \eqref{eq:lemma-gradmax-p1} that 
\begin{equation*}
\begin{aligned}
         \bE &\max_{i\in [d] }\abs{ \left\langle \left(\widehat{\bSigma}_t-\bar{\bSigma}_t \right) (\bmu_{t-1} -\bmu_{\star}) , {\bm e}_i \right\rangle }^2 \\
     & \le C s\frac{D^2}{t}\left(\log(dt) +\log(\frac{\Bar{\mu} D^2}{\sigma} )\right) \bE \norm{\bmu_{t-1} -\bmu_{\star} }_2^2  +C\frac{\sigma^2 D^2 }{t}
\end{aligned}     
\end{equation*}

We now proceed to control Part 2 analogously. Invoke Lemma \ref{lemma:martingale-con} again by selecting $v^2 = \sigma^2 D^2 /t$, and $b=\sigma D/t$. We then have the concentration bound:
\begin{equation*}
    \begin{aligned}
    \bp\left( \abs{ \frac{1}{t}\sum_{j=1}^{t} \bI\left\{y_j=a \right\} \bx_{j,i} \xi_j  } \ge z \right) &\le 2 \exp\left( -\frac{c z^2}{ \sigma^2 D^2 /t  + 2 \sigma D z/t } \right) \\
    & \le 4 \exp\left( -\frac{c t z^2}{ 2 \sigma^2 D^2 } \right)+ 4\exp\left( -\frac{c t z}{ 4 \sigma D } \right),
    \end{aligned}
\end{equation*}
which gives the tail bound with probability at least $1-\epsilon$:
\begin{equation*}
    \begin{aligned}
    \max_{i\in[d] }\abs{\frac{1}{t}\sum_{j=1}^{t} \bI\left\{y_j=a \right\}  \bx_{j,i} \xi_j  }^2\le C  \sigma D \sqrt{ \frac{ \log(d /\epsilon) }{t} }.
    \end{aligned}
\end{equation*}

and also the expectation bound for the maxima:
\begin{equation*}
    \begin{aligned}
    \bE \max_{i\in[d] }\abs{\frac{1}{t}\sum_{j=1}^{t} \bI\left\{y_j=a \right\}  \bx_{j,i} \xi_j  }^2\le C\frac{\sigma^2 D^2 \log d }{t}.
    \end{aligned}
\end{equation*}
Combining Part 1 and Part 2 gives us the first claim on the expectation bound:
\begin{equation*}
      \bE \max_{i\in[d] } \abs{\left\langle \bg_t -\nabla f_t({\bmu}_{t-1}), {\bm e}_i  \right\rangle}^2 \le C \frac{sD^2\log(dt)}{t}\bE \norm{\bmu_{t-1} -\bmu_{\star} }_2^2 + C\frac{\sigma^2 D^2 \log d }{t}.
\end{equation*}
The high probability bound in Part 1 and Part 2 directly leads to the probability bound: with a probability at least $1-\epsilon$, the variation can be controlled by 
\begin{equation*}
    \max_{i\in [d] } \abs{\left\langle \bg_t -\nabla f_t({\bmu}_{t-1}), {\bm e}_i  \right\rangle}^2 \le C s D^2\frac{\log (d/\epsilon)  }{t} \norm{\bmu_{t-1} -\bmu_{\star} }_2^2 +  C\frac{\sigma^2 D^2\log(d/\epsilon) }{t}
\end{equation*}

\end{myproof}

\subsection{Proof of Lemma \ref{lemma:smoothness}}
We have 
\begin{equation}
\begin{aligned}
                &\norm{\nabla_{\blambda} D(\blambda,\brho) - \nabla_{\blambda} D(\blambda^*(\brho),\brho)}_{\infty} 
                \\
               \le & 2B_{\max} \bE\mathbbm{1}\{ \arg\max_{a\in[K]\cup\{0\}}\left\langle \bmu_{a}^\star - (\bm{W}_{a}^{\star})^{\top}\blambda,\bx_t \right\rangle \neq \arg\max_{a\in[K]\cup\{0\}}\left\langle \bmu_{a}^\star - (\bm{W}_{a}^{\star})^{\top}\blambda^*(\brho),\bx_t \right\rangle\} 
               \\
           = &  2B_{\max} \bp\left( \left\langle \bmu_{a^*_t(\blambda)}^\star - (\bm{W}_{a^*_t(\blambda)}^{\star})^{\top}\blambda,\bx_t \right\rangle  > \left\langle \bmu_{a^*_t}^\star - (\bm{W}_{a^*_t}^{\star})^{\top}\blambda,\bx_t \right\rangle \right) 
           \\
           = & 2B_{\max} \bp\left( \left\langle \bmu_{a^*_t}^\star - (\bm{W}_{a^*_t}^{\star})^{\top}\blambda^*,\bx_t \right\rangle - \left\langle \bmu_{a^*_t(\blambda)}^\star - (\bm{W}_{a^*_t(\blambda)}^{\star})^{\top}\blambda^*,\bx_t \right\rangle <  \left\langle (\bm{W}_{a^*_t(\blambda)}^{\star}-\bm{W}_{a^*_t}^{\star})^{\top}(\blambda^*-\blambda) ,\bx_t\right\rangle  \right) 
           \\
            \le &  2B_{\max} \bp\left( \left\langle \bmu_{a^*_t}^\star - (\bm{W}_{a^*_t}^{\star})^{\top}\blambda^*,\bx_t \right\rangle - \left\langle \bmu_{a^*_t(\blambda)}^\star - (\bm{W}_{a^*_t(\blambda)}^{\star})^{\top}\blambda^*,\bx_t \right\rangle < 2B_{\max} \norm{\blambda^*-\blambda}_1 \right) 
            \\
            \le & 4B_{\max}^2 M_{\sfm} \norm{\blambda^*-\blambda}_1
\end{aligned}
        \end{equation}

\subsection{Proof of Lemma \ref{lemma:region-B+}}

Following the Lemma 7 of \cite{ma2024high}, we know that dual optimal solution is continuous with respect to $\brho$, i.e.,
\begin{equation*}
\begin{gathered}
        \norm{\blambda^*(\brho_1)-\blambda^*(\brho_2)}_1\le \frac{1}{\cL}\norm{\brho_1-\brho_2}_{\infty}, \ \forall \brho_1,\brho_2\in \cB(\brho_0,\delta_\rho)\\
        \norm{\blambda^*(\brho_1)-\blambda^*(\brho_2)}_1\le \frac{1}{\cL}\norm{\brho_1-\brho_2}_{\infty,\sfB}, \ \forall \brho_1,\brho_2\in \cB(\brho_0,\delta_\rho) \text{ share the same binding dimensions }  I_\sfB.
\end{gathered}
\end{equation*}
Here $\norm{\cdot}_{\infty,\sfB}$ computes the maximum size of entries on the shared binding dimensions $I_\sfB$. 
Therefore, plugging $\brho_0$ in the inequality above, and selecting $\norm{\brho-\brho_0}_{\infty}\le \frac{\cL\delta_{\sfB }}{2} \wedge \delta_{\rho}$, we yield that all the binding dimensions in $\brho_0$ will not change to non-binding. On the other hand, the remaining resource satisfies
\begin{equation*}
    \begin{aligned}
        &\abs{\bE\big(\brho_{0}-\bW_{a^*_t(\blambda^*(\brho_0))}^{\star}\bx_t\big)_i-\bE\big(\brho_{}-\bW_{a^*_t(\blambda^*(\brho))}^{\star}\bx_t\big)_i} 
        \\
        &\le \norm{\brho-\brho_0}_{\infty}+ \norm{\nabla_{\blambda} D(\blambda^*(\brho),\brho_0) - \nabla_{\blambda} D(\blambda^*(\brho_0),\brho_0)}_{\infty} 
        \\
        &\le \norm{\brho-\brho_0}_{\infty} + 4B_{\max}^2 M_{\sfm} \norm{\blambda^*(\brho)-\blambda^*(\brho_0)}_1\le \frac{8B_{\max}^2M_{\sfm}}{\cL}\norm{\brho-\brho_0}_{\infty}.
    \end{aligned}
\end{equation*}
Letting $\norm{\brho-\brho_0}_{\infty}\le \frac{\cL \delta_{\NB}}{16 B_{\max}^2 M_{\sfm}} \wedge \delta_{\rho}$, we yield that all the non-binding dimensions in $\brho_0$ will not change to binding.
Thus, we conclude that $D(\blambda, \brho_0)$ and $D(\blambda, \brho)$ share the same binding and non-binding dimensions when taking $\delta_{\rho}'= \frac{\cL\delta_{\sfB }}{2}\wedge\frac{\cL \delta_{\NB}}{16 B_{\max}^2 M_{\sfm}} \wedge \delta_{\rho}$.

% Here we use the fact that $\widetilde{\bmu}_{t-1}$ and ${\bmu}_{\star }$ are all sparse.
% If further $\norm{\bx_t }_\infty$ is bounded by $D$, we have 
% \begin{equation*}
% \begin{aligned}
%         &\max_{i\in [d] } \abs{\left\langle \bg_t -\nabla f(\widetilde{\bmu}_{t-1} ), {\bm e}_i  \right\rangle}^2 \le 8 D^2 \xi_t^2+ 64s D^4\norm{\widetilde{\bmu}_{t-1}-\bmu_\star }_2^2, \\
%     &\bE \max_{i\in [d] } \abs{\left\langle \bg_t -\nabla f(\widetilde{\bmu}_{t-1} ), {\bm e}_i  \right\rangle}^2 \le 8 D^2 \sigma^2+ 64s D^4\norm{\widetilde{\bmu}_{t-1} -\bmu_\star }_2^2.
% \end{aligned}
% \end{equation*}
% When $\bmu_t$ converges, we have $\bE \max_{i\in [d] } \abs{\left\langle \bg_t -\nabla f(\widetilde{\bmu}_{t-1}), {\bm e}_i  \right\rangle}^2 \lesssim D^2\sigma^2 $

% \begin{equation*}
%     \bE \norm{\Omega(\bg_t -\nabla f(\widetilde{\bmu}_{t-1} )) }_2^2 \le 3s \bE \max_{i\in [d] } \abs{\left\langle \bg_t -\nabla f(\widetilde{\bmu}_{t-1}), {\bm e}_i  \right\rangle}^2 
% \end{equation*}

%This amounts to compute the . $\psi_2$-norm of each 
\endproof

\newpage

\subsection{Proof of Lemma  \ref{lemma:concentrate-first-order}, \ref{lemma:lb-second-order}  }
These two lemmas are the extensions of Theorem 1 in \cite{ma2024optimal}. Lemma  \ref{lemma:concentrate-first-order} can be proved by the bouned difference condition \citep{koltchinskii2011oracle} where $\norm{\nabla_{\blambda}D_j(\blambda,\brho)-\nabla_{\blambda}D_j(\blambda^*(\brho),\brho)}_{\infty}\le 2B_{\max}$, and the VC-dimension of the function class on $\cB(\brho_0,\delta_\rho)$, which is at most $m+2$.
To show Lemma  \ref{lemma:lb-second-order}, we can check that the following bound holds when $\norm{\blambda-\blambda^*(\brho)}_{1}\le\varepsilon$:
\begin{equation}\label{eq:local-size-S}
    \begin{gathered}
        \abs{{S}_j(\blambda,\brho)} = \abs{\int_0^1 \left\langle (\bW^\star_{a^*_t(\blambda^*(\brho)+z\bv)}-\bW^\star_{a^*_j })\bx_j,\blambda-\blambda^*(\brho)\right\rangle dz} \le 2B_{\max} \varepsilon.
    \end{gathered}
\end{equation}
Therefore, the concentration of
\begin{equation*}
     \inf_{\brho\in \cB^+(\brho_0,\delta_\rho),\norm{\blambda-\blambda^*(\brho)}_{1}\le\varepsilon}  \bar S_t(\blambda,\brho)- S(\blambda,\brho ) - \bE \left[\inf_{\brho\in \cB^+(\brho_0,\delta_\rho),\norm{\blambda-\blambda^*(\brho)}_{1}\le\varepsilon}  \bar S_t(\blambda,\brho)- S(\blambda,\brho )\right]
\end{equation*}
can be given by the bouned difference condition \citep{koltchinskii2011oracle}. To control the empirical process $\bE \left[\inf_{\brho\in \cB^+(\brho_0,\delta_\rho),\norm{\blambda-\blambda^*(\brho)}_{1}\le\varepsilon}  \bar S_t(\blambda,\brho)- S(\blambda,\brho )\right]$, we use the Rademacher complexity:
\begin{equation*}
    \begin{aligned}
        &\bE \left[\inf_{\brho\in \cB^+(\brho_0,\delta_\rho),\norm{\blambda-\blambda^*(\brho)}_{1}\le\varepsilon}  \bar S_t(\blambda,\brho)- S(\blambda,\brho )\right]  
        \\
        &\le 2 \mathcal{R}\left(S,\cB(\brho_0,\delta_\rho) \times \cB_1(\varepsilon)\right) := 2 \bE\left[\bE_{\sigma} \sup_{\brho,\Delta\blambda\in \cB_1(\varepsilon) }  \frac{1}{t} \sum_{j=1}^{t} \sigma_j S_j(\blambda^*(\brho)+\Delta\blambda,\brho)\right].
    \end{aligned}
\end{equation*}

% element $\blambda$ in the $\ell_1$ ball $\cB_1(\blambda^*(\brho),\varepsilon)$,  we can focus on

To handle this Rademacher complexity, we notice that for any  $\Delta\blambda=\blambda-\blambda^*(\brho)$, we can always construct a chain with the distance measured by $\ell_1$-norm: $\left\{\bpi^{k}(\Delta\blambda)\right\}_{k\ge0}$ with  $\bpi^{0}(\Delta\blambda)=\bm{0}$, $\bpi^{k}(\Delta\blambda)\to\Delta\blambda$, $\norm{\bpi^{k}(\Delta\blambda)-\bpi^{k-1}(\Delta\blambda)}_{1}\le 2^{-k}\varepsilon$. It is clear that on the chain we have
\begin{equation*}
     \abs{{S}_j(\blambda^*(\brho)+\bpi^k(\Delta\blambda),\brho)- {{S}_j(\blambda^*(\brho)+\bpi^{k-1}(\Delta\blambda),\brho)}}\le 2B_{\max} \cdot 2^{-k}\varepsilon
\end{equation*}
following \eqref{eq:local-size-S}. For each segment on the chain, we have:
\begin{equation}\label{eq:cover-layers-linear}
    \begin{aligned}
        \bE&\left[\bE_{\sigma} \sup_{\bpi^{k-1}(\Delta\blambda), \bpi^{k}(\Delta\blambda)} \sup_{\brho}  \frac{1}{t} \sum_{j=1}^{t} \sigma_j \left( {S}_j(\blambda^*(\brho)+\bpi^k(\Delta\blambda),\brho)- {{S}_j(\blambda^*(\brho)+\bpi^{k-1}(\Delta\blambda),\brho)} \right)\right] 
        \\
        & \le \frac{C B_{\max}  \varepsilon }{2^k\sqrt{t}} \int_{0}^{1} \sqrt{\log \left(N^2(2^{-k}\varepsilon,\cB_1(\varepsilon),L_1)\cdot N\left(2B_{\max}\cdot 2^{-k} \varepsilon \epsilon,\cB(\brho_0,\delta_\rho),L_2( S_j, \bp_t ) \right)\right) }  d\epsilon,
        % \\
        % & \le  \frac{C  \sqrt{n} \bar b D  \varepsilon  \sqrt{\log N(2^{-k}\varepsilon,\cB_1(\varepsilon),L_1)}}{2^k\sqrt{T}} \\
        % &+ \frac{C  \sqrt{n} \bar b D  \varepsilon}{2^k\sqrt{T}}\left( \int_{0}^{1}\sqrt{\log N( \epsilon/2, \Omega_d, \sqrt{n} L_2(\mathbf{G}_1,\bp_T) ) }+ \sqrt{\log N(  \sqrt{n} \bar b D 2^{-k} \varepsilon  \epsilon/2, \Omega_d, \sqrt{n} L_2(\mathbf{G}_2,\bp_T) ) } d\epsilon\right)  \\
        % &\le  \frac{C  \sqrt{n} \bar b D  \varepsilon \left( \sqrt{\log N(2^{-k}\varepsilon,\mathbb{B}(\varepsilon),L_2)} + \sqrt{mn} \right)  }{2^k \sqrt{T}} \\ 
        % &\le \frac{C \sqrt{n} \bar b D  \varepsilon \left(\sqrt{mn} + \sqrt{\log N(2^{-k}\varepsilon,\mathbb{B}(\varepsilon),L_2)}\right) }{2^k \sqrt{T}},
    \end{aligned}
\end{equation}
where the inequality holds because for each $\epsilon>0$, the covering number of the function class consisting of ${S}_j(\blambda^*(\brho)+\bpi^k(\Delta\blambda),\brho)- {{S}_j(\blambda^*(\brho)+\bpi^{k-1}(\Delta\blambda),\brho)} $ is at most $N^2(2^{-k}\varepsilon,\cB_1(\varepsilon),L_1)$ times the covering number on $\brho$. Following \eqref{eq:cover-layers-linear}, we have
\begin{equation*}
    \begin{aligned}
        \bE&\left[\bE_{\sigma} \sup_{\bpi^{k-1}(\Delta\blambda), \bpi^{k}(\Delta\blambda)} \sup_{\brho}  \frac{1}{t} \sum_{j=1}^{t} \sigma_j \left( {S}_j(\blambda^*(\brho)+\bpi^k(\Delta\blambda),\brho)- {{S}_j(\blambda^*(\brho)+\bpi^{k-1}(\Delta\blambda),\brho)} \right)\right] 
        \\
        & \le \frac{C B_{\max}  \varepsilon }{2^k\sqrt{T}} \left(\sqrt{\log N(2^{-k}\varepsilon,\cB_1(\varepsilon),L_1)} + \sqrt{m}\right)
        % \int_{0}^{1}  \sqrt{\log \left(N\left(2B_{\max}\cdot 2^{-k} \varepsilon \epsilon,\cB(\brho_0,\delta_\rho),L_2( S_j, \bp_t ) \right)\right) }  d\epsilon
    \end{aligned}
\end{equation*}
By Dudley's chaining argument, we can sum up $k$ from $k=1$ to $\infty$ and derive the bound on the Rademacher complexity:
\begin{equation}
    \mathcal{R}\left(S,\cB(\brho_0,\delta_\rho) \times \cB_1(\varepsilon)\right) \le \sum_{k=1}^{\infty}\frac{C B_{\max}  \varepsilon }{2^k\sqrt{t}} \left(\sqrt{\log N(2^{-k}\varepsilon,\cB_1(\varepsilon),L_1)} + \sqrt{m}\right)\le CB_{\max}\varepsilon\sqrt{\frac{m}{t}}
\end{equation}
Following the procedures in Theorem 1, \cite{ma2024optimal}, we prove the claims.

\end{document}